\pdfoutput=1
\documentclass[nohyperref]{article}

\usepackage{microtype}
\usepackage{graphicx}
\usepackage{booktabs} %

\usepackage{hyperref}

\usepackage[accepted]{icml2022}

\usepackage{amsmath}
\usepackage{amssymb}
\usepackage{mathtools}
\usepackage{amsthm}

\usepackage[capitalize,noabbrev]{cleveref}

\theoremstyle{plain}
\newtheorem{theorem}{Theorem}[section]

\newtheorem{lemma}[theorem]{Lemma}

\theoremstyle{definition}
\newtheorem{definition}[theorem]{Definition}

\theoremstyle{remark}

\usepackage[T1]{fontenc}
\usepackage[utf8]{inputenc}

\usepackage{caption}
\usepackage{subcaption}

\usepackage{mathtools}
\usepackage{tablefootnote}
\usepackage{colortbl}
\usepackage{environ}
\usepackage{booktabs}
\usepackage{diagbox} 
\usepackage{wrapfig} 
\usepackage{bbding}
\usepackage{pifont}
\usepackage{wasysym}
\usepackage{amssymb}
\usepackage{times}
\usepackage{soul}
\usepackage{url}
\usepackage{caption}
\usepackage{graphicx}
\usepackage{amsmath}
\usepackage{amsthm}
\usepackage{cleveref}
\usepackage{tabularx}
\usepackage{multirow}
\usepackage{fancybox}
\usepackage{mathtools}
\usepackage{bold-extra}
\usepackage{lipsum}
\usepackage{siunitx}
\usepackage[export]{adjustbox}
\usepackage{comment}

\DeclarePairedDelimiterX{\infdivx}[2]{(}{)}{%
  #1\;\delimsize\|\;#2%
}

\makeatletter
\let\MYcaption\@makecaption
\makeatother
\usepackage{subcaption}
\makeatletter
\let\@makecaption\MYcaption
\makeatother

\makeatletter
\newcommand{\printfnsymbol}[1]{%
  \textsuperscript{\@fnsymbol{#1}}%
}
\makeatother
\usepackage{stmaryrd}
\usepackage{trimclip}
\makeatletter
\DeclareRobustCommand{\shortto}{%
  \mathrel{\mathpalette\short@to\relax}%
}

\usepackage{etoolbox}

\makeatletter

\newcommand*{\red}[1]{{\textcolor{red}{#1}}}
\newcommand*{\orange}[1]{{\textcolor{orange}{#1}}}
\newcommand*{\blue}[1]{{\textcolor{blue}{#1}}}

\newcommand{\BoundFuncLoose}{\mathcal{B}}
\newcommand{\BoundFuncTight}{\mathcal{B}^{\normalfont \text{tight}}}

\newcommand{\BoundFuncTightP}{\BoundFuncTight_{p_0}}

\newcommand{\BoundTight}{\BoundFuncTight(\StrengthTriplet)}
\newcommand{\BoundTightWOCombLoss}{\BoundFuncTight(\StrengthDoublet)}
\newcommand{\BoundLoose}{\BoundFuncLoose(\StrengthDoublet)}

\newcommand{\defPercent}{\scalebox{0.93}{[\%]}}

\newcommand{\binaryEntropyWithArgs}{\mathcal{H}_2(p_{\text{\rm err}})}
\newcommand{\binaryEntropyWithArgsP}{\mathcal{H}_2(p)}

\newcommand{\StrengthDoubletWithArgs}{\mathcal{I}(\mathbf{O}, Y)}
\newcommand{\StrengthTripletWithArgs}{\mathcal{E}(\mathbf{O}, Y, \hat{Y})}
\newcommand{\RelevWithArgs}{I_{\normalfont \text{relev}}(\mathbf{O}, Y)}
\newcommand{\RedunWithArgs}{I_{\normalfont \text{redun}}(\mathbf{O}, Y)}
\newcommand{\ComblossWithArgs}{I_{\normalfont \text{combloss}}(\mathbf{O}, Y, \hat{Y})}

\newcommand{\StrengthDoublet}{\mathcal{I}}
\newcommand{\StrengthTriplet}{\mathcal{E}}
\newcommand{\Relev}{I_{\normalfont \text{relev}}}
\newcommand{\Redun}{I_{\normalfont \text{redun}}}
\newcommand{\Combloss}{I_{\normalfont \text{combloss}}}
\newcommand{\RelevHat}{\hat{I}_{\normalfont \text{relev}}}

\newcommand{\relev}{i_{\normalfont \text{relev}}}
\newcommand{\redun}{i_{\normalfont \text{redun}}}
\newcommand{\combloss}{i_{\normalfont \text{combloss}}}

\newcommand{\multiInfo}{I_{\normalfont \text{multi}}}
\newcommand{\perModelMetric}{i}
\newcommand{\perModelMetricDef}{i_{\normalfont \text{\{relev, redun, combloss\}}} = I_{\normalfont \text{\{relev, redun, combloss\}}}/N}

\newcommand{\combfunc}{\mathcal{F}}

\newcommand{\NumTasks}{eight }
\newcommand{\NumSystems}{16 }

\newcommand{\omegaMax}{I(\Omega^{N, \mathrm{max}}_n;Y)}
\newcommand{\omegaMin}{I(\Omega^{N, \mathrm{min}}_n;Y)}
\newcommand{\omegaMaxMin}{I(\Omega^{N, \mathrm{max/min}}_n;Y)}
\Crefformat{figure*}{#2Figure~#1#3}
\Crefmultiformat{figure*}{Figures~#2#1#3}{ and~#2#1#3}{, #2#1#3}{ and~#2#1#3}

\newcommand{\CombLossInformation}{H(Y|\hat{Y}) - H(Y|\mathbf{O})}

\newcommand{\CombLossInformationWithName}{\underbrace{H(Y|\hat{Y}) - H(Y|\mathbf{O})}_{\normalfont \text{combination loss}}}

\newcommand{\YSize}{Y_\text{\rm max}}

\crefname{equation}{}{}

\newcommand*{\cFirst}[0]{\cellcolor[HTML]{f5ffea}}
\newcommand*{\cSecond}[0]{\cellcolor[HTML]{d5efca}}
\newcommand*{\cThird}[0]{\cellcolor[HTML]{b5dfaa}}
\newcommand*{\cFourth}[0]{\cellcolor[HTML]{94c288}}

\newcommand*{\cBase}[0]{\cellcolor[HTML]{f5f5f5}}

\NewEnviron{alignSmall}{%
    \begin{align}
    \BODY
    \end{align}
}
\NewEnviron{gatherSmall}{%
    \begin{gather}
    \BODY
    \end{gather}
}

\icmltitlerunning{Rethinking Fano’s Inequality in Ensemble Learning}

\begin{document}

\twocolumn[
\icmltitle{Rethinking Fano’s Inequality in Ensemble Learning}

\icmlsetsymbol{equal}{*}

\begin{icmlauthorlist}
\icmlauthor{Terufumi Morishita}{comp}
\icmlauthor{Gaku Morio}{equal,comp}
\icmlauthor{Shota Horiguchi}{equal,comp}
\icmlauthor{Hiroaki Ozaki}{comp}
\icmlauthor{Nobuo Nukaga}{comp}
\end{icmlauthorlist}

\icmlaffiliation{comp}{Hitachi, Ltd. Research and Development Group, Kokubunji, Tokyo, Japan}

\icmlcorrespondingauthor{Terufumi Morishita}{terufumi.morishita.wp@hitachi.com}

\icmlkeywords{Machine Learning, ICML}

\vskip 0.3in
]

\printAffiliationsAndNotice{\icmlEqualContribution} %

\begin{abstract}
We propose a fundamental theory on ensemble learning that answers the central question: \textit{what factors make an ensemble system good or bad?}
Previous studies used a variant of Fano's inequality of information theory and derived a lower bound of the classification error rate on the basis of the \textit{accuracy} and \textit{diversity} of models.
We revisit the original Fano's inequality and argue that the studies did not take into account the information lost when multiple model predictions are combined into a final prediction.
To address this issue, we generalize the previous theory to incorporate the information loss, which we name \textit{combination loss}.
Further, we empirically validate and demonstrate the proposed theory through extensive experiments on actual systems.
The theory reveals the strengths and weaknesses of systems on each metric, which will push the theoretical understanding of ensemble learning and give us insights into designing systems.
\end{abstract}

\section{Introduction}  \label{sec:intro}

Ensemble learning has had great success in various fields of machine learning.
Bagging \cite{breiman1996bagging} trains diverse models from artificial datasets built by random sub-sampling on the original one.
It is common to train models with different weight initializations \cite{lakshminarayanan2016simple} or models with different network architectures \cite{qummar2019deep, morishita2020hitachi_task7}.
While models are usually combined by voting on predictions, other methods focus on how to combine them cleverly \cite{omari2015post,morio2020hitachi_task10}.
Stacking \cite{wolpert-1992-stackedgeneralization} trains meta-estimators that make final predictions from model predictions as their inputs.
Mixture of Experts \cite{jacobs1991adaptive, shazeer2017outrageously} focuses more on the models that are best specialized for a given dataset instance.

The central question of ensemble learning has been: \textit{what factors make an ensemble system good or bad?}
It has been widely believed that accurate and diverse models lead to better performance for ensemble systems.
Guided by this intuition, many heuristical metrics have been proposed to measure accuracy and diversity \cite{kohavi1996bias, skalak1996sources, cunningham2000diversity, shipp2002relationships}.
However, these metrics lack theoretical grounding, and indeed, \citet{kuncheva2003measures} empirically showed that there are no connections between the metrics and system performance through a broad range of experiments.
Turning to theoretical viewpoints, \citet{geman1992neural} decomposed the squared error loss used in regression tasks into the bias and covariance of models.
Bias here corresponds to accuracy and covariance diversity.
For classification tasks, \citet{tumer1995theoretical} showed that the error rate reductions obtained by unweighted voting is a decreasing function of models' correlations, indicating that diverse models lead to better performance.

\begin{figure*}[t]
    \centering
    \includegraphics[width=0.8\linewidth]{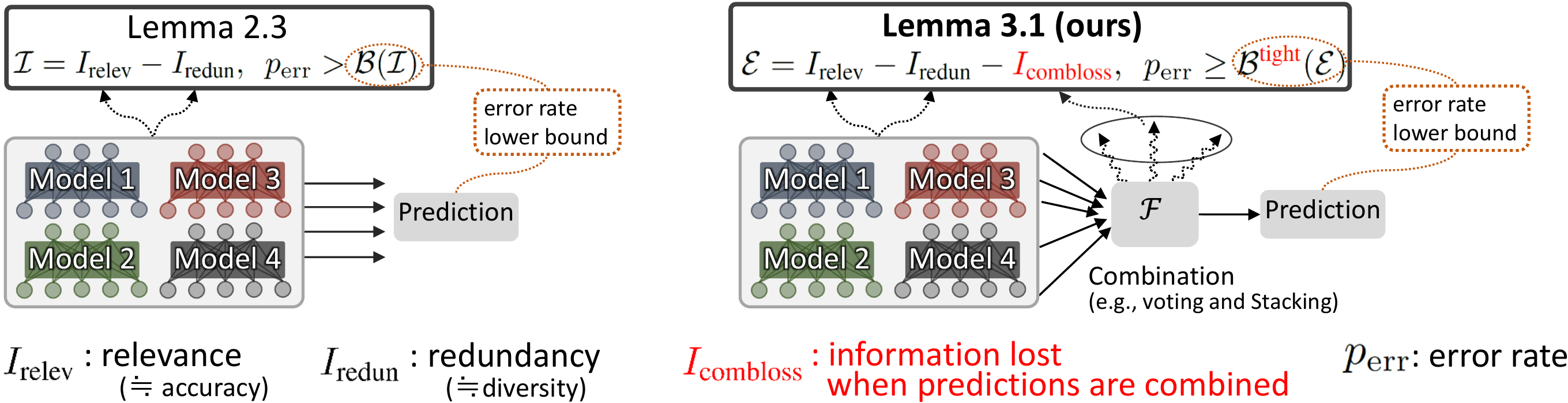}
    \caption{Previous framework \cite{brown_mti,zhou_mti} (left) and ours (right). \label{fig:our_framework}}
\end{figure*}

While the theory of \citet{tumer1995theoretical} deals with classification tasks under a limited setting,
\citet{brown_mti, zhou_mti} first derived accuracy and diversity in a general setting.
Using Fano's inequality of information theory, they derived a lower bound to the error rate of a given system.
Then, they decomposed the lower bound into relevance $\Relev$ and redundancy $\Redun$ (\Cref{lemma:ensemble_bound_loose}, illustrated in  \Cref{fig:our_framework}).
$\Relev$ is the information theoretical version of accuracy and $\Redun$ diversity.
Their framework is promising as a fundamental theory of ensemble learning since it derives well-believed metrics in a general setting.
However, the validity of the framework has not been examined much from both theoretical and empirical perspectives.
Theoretically, we find that the framework rests on implicit assumptions used by a variant of Fano's inequality, which generally do not hold in ensemble learning.
As a result, the framework fails in capturing important aspects of ensemble learning.
Empirically, the experiments of the studies were not extensive enough to justify the framework.
In particular, they did not check whether the framework can predict representative phenomena in ensemble learning.

In this paper, we rethink the theoretical framework from both perspectives.
We first revisit the theory (\Cref{sec:previous_framework,our_framework}).
We argue that the framework does not take into account the information lost when multiple model predictions are combined into a single final prediction.
We call this information loss \textit{combination loss}.
To address the issue, we propose a generalized framework that incorporates the third metric  of combination loss $\Combloss$ based on original Fano's inequality (\Cref{lemma:ensemble_bound_tight}, illustrated in \Cref{fig:our_framework}).
We also solve the issue of the previous framework producing a loose lower bound when the number of classes is small.

Next, we turn to empirical viewpoints.
We first validate the proposed framework in \Cref{sec:setup,sec:validating_framework}.
In contrast to the previous studies,
(i) we directly check whether the framework can predict phenomena in ensemble learning, 
(ii) we use various ensemble systems (\Cref{tb:ensemble_methods}), and
(iii) we use various tasks (\Cref{appendix:tb:tasks}).
Additionally, to be modern and realistic, we use state-of-the-art DNNs such as BERT \cite{devlin-etal-2019-bert}, and the tasks are chosen from widely-used benchmarks such as GLUE \cite{wang-etal-2018-glue}.
Extensive experiments reveal that the previous framework can not predict phenomena such as the performance ranking of ensemble systems (\Cref{fig:ERR_LBR_scatter_plot}) and performance scaling behavior (\Cref{fig:scaling}), ignoring combination loss.
These results refute the previous framework.
In contrast, the proposed framework justifies itself by predicting all these phenomena.
Finally, we demonstrate the proposed framework (\Cref{sec:triplet_decomposition}).
We analyze DNN ensemble systems and answer \textit{why} a system performs well or badly through its strengths and weaknesses in terms of the three metrics (\Cref{tb:ablation}). 
Such analysis pushes the theoretical understanding of ensemble learning and gives us insights into designing systems.
In summary,
\begin{itemize}
    \item We propose a fundamental theory on ensemble learning: which answers the central question, what factors make an ensemble system good or bad. The proposed theory measures a given ensemble system from a well-grounded set of metrics: relevance $\Relev$, redundancy $\Redun$, and combination loss $\Combloss$. The metrics are tied to the bound on the performance of a system. 
    \item We validate the framework through extensive experiments on DNN ensemble systems.
    \item We demonstrate the framework. We analyze the DNN ensemble systems and answer why a system performs well or badly as follows:
    \begin{enumerate}
        \item Systems with models that simply differ in the training seeds perform well because the models are accurate (large $\Relev$) and combinable (small $\Combloss$).
        \item Heterogeneous systems, which use various types of DNNs, also perform well. While some DNNs are inaccurate (small $\Relev$), DNNs are diverse (small $\Redun$). Further, such systems should perform the best among all the systems when DNNs are combined by meta-estimators.
        \item Bagging-based systems do not perform that well. Their models are diverse (small $\Redun$) but inaccurate (small $\Relev$) and uncombinable (large $\Combloss$).
        \item Systems with models with randomly chosen hyperparameters do not perform that well. The models are diverse (small $\Redun$) but inaccurate (small $\Relev$).
        \item Meta-estimators generally push the performance of the systems by combining models smartly to reduce $\Combloss$.
        Further, meta-estimators benefit systems such as 2 and 4 the most since the amount of information of the true label is unevenly distributed on models of varied accuracies and such information is recovered well by meta-estimators.
        Finally, a simple estimator such as logistic regression should be enough on strong DNNs. 
    \end{enumerate}
    \item We release our code as open source.\footnote{Available at: \texttt{\url{https://github.com/hitachi-nlp/ensemble-metrics}}}
\end{itemize}

\section{Conventional Framework Based on Variant of Fano's Inequality}   \label{sec:previous_framework}

\subsection{Fano's Inequality}  \label{sec:fano}
Let $Y \in \{1, 2, \dots, \YSize\}$ be a discrete stochastic variable representing the input
and $\mathbf{O} \in \mathbb{R}^m$ be $m$ stochastic variables representing an observation after a noisy channel.
We want to recover $Y$ from $\mathbf{O}$ by using the reconstruction function $\combfunc:\mathbf{O}\mapsto\hat{Y} \in \left\{1,2,\dots,\YSize\right\}$.
Note that $Y \Rrightarrow \mathbf{O} \Rrightarrow \hat{Y}$ forms a Markov chain.
Fano's inequality relates the information lost in a noisy channel to the error rate when recovering the input as follows.
\begin{lemma}[Fano's inequality \citep{fano1961transmission}]
    \label{lemma:fano}
    For any function $\combfunc$, the following holds:
    {
    \begin{alignSmall}
     \binaryEntropyWithArgs + p_{\text{\rm err}} \log_2 (\YSize - 1) \geq H(Y\mid\hat{Y}), \label{eq:fano}
    \end{alignSmall}
    }%
where $p_{\text{\rm err}} = \mathrm{Pr}[\hat{Y} \neq Y] \in [0, 1]$ is the reconstruction error rate, $\binaryEntropyWithArgsP = - p \log_2 (p) - (1-p) \log_2 (1 - p)$ binary cross entropy, and $H(Y\mid\hat{Y})$ conditional entropy \cref{eq:notation_conditional_entropy}.
\end{lemma}

From the Markovness, the amount of information carried by $\hat{Y}$ is never more than that carried by $\mathbf{O}$; thus, the right-hand side of \Cref{eq:fano} is lower bounded as
{
\begin{align}
    H(Y\mid\hat{Y}) \geq H(Y\mid\mathbf{O}).
    \label{eq:fano_expansion_1}
\end{align}
}%

Since the binary cross entropy never exceeds one, the left side of \Cref{lemma:fano} is upper bounded as
{
\begin{alignSmall}  \label{eq:fano_expansion_2}
    \binaryEntropyWithArgs + p_{\text{\rm err}} \log_2 (\YSize - 1) & \leq 1 + p_{\text{\rm err}} \log_2 (\YSize - 1), \nonumber \\
    & < 1 + p_{\text{\rm err}} \log_2 (\YSize)
\end{alignSmall}
}%
From \cref{eq:fano}--\cref{eq:fano_expansion_2}, we obtain the following well-known variant of Fano's inequality:
{
\begin{lemma}[An error rate lower bound \cite{fano1961transmission}]
    \label{lemma:fano_error_bound_loose}
    \begin{alignSmall}
        p_{\text{\rm err}} > \frac{H\left(Y\mid\mathbf{O}\right) - 1}{\log_2 \YSize}.  \nonumber
    \end{alignSmall}
\end{lemma}
}

\subsection{Error Rate Lower Bound of Ensemble Systems}   \label{sec:ensemble_bound_loose}
In ensemble learning, $Y$ denotes a label on a given instance, and $\mathbf{O} = \{O_1, O_2, \dots, O_N\}$ is the output from $N$ models. Note that the output from $i$-th model $O_i\in\mathbb{R}^{\YSize}$ can be a predicted label ($\YSize=1$) or class probabilities ($\YSize \geq 2$).
$\combfunc$ denotes a model combination method such as voting or Stacking.
\Cref{lemma:fano_error_bound_loose} gives a lower bound of the classification error rate $p_{\text{\rm err}}$ of an ensemble system.

\citet{brown_mti} decomposed the lower bound into relevance and redundancy, the formulation of which was later simplified by \citet{zhou_mti} as follows:
\begin{lemma}[\citeauthor{zhou_mti}, \citeyear{zhou_mti}]
    \label{lemma:ensemble_bound_loose}
    {
    \begin{alignSmall}
        p_{\text{\rm err}} & > \BoundFuncLoose (\StrengthDoubletWithArgs) \coloneqq \frac{H(Y) - \StrengthDoubletWithArgs - 1}{\log_2 \YSize} . \label{eq:ensemble_bound_loose}
    \end{alignSmall}
    }
    $\StrengthDoubletWithArgs$ is defined as follows:
    {
    \begin{alignSmall}
        \StrengthDoubletWithArgs & \coloneqq \RelevWithArgs - \RedunWithArgs, \label{eq:doublet_decompositoin} \\
        \RelevWithArgs & \coloneqq \sum_{i=1}^{N}I(O_i; Y), \nonumber \\
        \RedunWithArgs & \coloneqq  \multiInfo(\mathbf{O}) - \multiInfo(\mathbf{O}|Y), \nonumber
    \end{alignSmall}
    }%
    where $H$ denotes entropy \cref{eq:notation_entropy}, $I$ denotes mutual information \cref{eq:notation_mutual_information}, and $\multiInfo$ denotes \textit{multi-information} \crefrange{eq:notation_multi_information}{eq:notation_conditional_multi_information}, a multivariate generalization of mutual-information.
\end{lemma}
Since $H(Y)$ and $\YSize$ are constants given a machine learning task, the important term in \cref{eq:ensemble_bound_loose} is $\StrengthDoubletWithArgs$ defined in \cref{eq:doublet_decompositoin}, which denotes the amount of unique information on $Y$ carried by $\mathbf{O}$.
The first term $\RelevWithArgs$ is the \textit{relevance}, whose component $I(O_i; Y)$ denotes the amount of information on $Y$ given by $O_i$.
It can be seen as the accuracy of the model $i$ from the information theoretical point of view.
The second term $\RedunWithArgs$ is the \textit{redundancy}, which indicates how strongly the model outputs $\mathbf{O}=\{O_1, O_2, \dots O_N\}$ are correlated with each other.
In other words, it describes the amount of redundant (duplicated) information.
Overall, \Cref{lemma:ensemble_bound_loose} reveals that an ensemble system should include accurate (large $\Relev$) and diverse (small $\Redun$) models to get a small lower bound for the error rate $\BoundFuncLoose(\StrengthDoublet)$.

\section{Proposed Framework Based on Original Fano's Inequality}  \label{our_framework}

\subsection{Error Rate Lower Bound with Better Properties}  \label{sec:ensemble_bound_tight}

\begin{table*}[t!]
    \small
    \caption{
    Extreme toy ensemble systems on imaginary binary classification task for discussing combination loss (\Cref{sec:toy_systems}).
    Each row shows predicted labels on instance from dataset.
    $\mathbf{O}=\{O_1,\dots, O_5\}$: model predictions, $\hat{Y}=\combfunc(\mathbf{O})$: ensemble prediction, and $Y$: ground-truth label.
    Red \red{0}/\red{1} shows wrong ensemble predictions.
    Orange \orange{0}/\orange{1} shows correct but neglected model predictions.
    Blue \blue{0} shows correct prediction recovered by weighted voting.
    \Cref{tb:system-B,tb:system-C} use the same $\mathbf{O}$.
    \label{tb:toy_ensemble_systems}
    }
    \begin{subfigure}{0.21\linewidth}
        \centering
        \subcaption{$\hat{Y}_{\normalfont \text{vote}}$: voting on $\mathbf{O}$.  \label{tb:system-A}}
        \begin{tabular}{lcc}
            $\mathbf{O}$     & $\hat{Y}_{\normalfont \text{vote}}$ & $Y$ \\
            \midrule

            \cBase 11111 & \cSecond 1  & \cThird 1 \\
            \cBase 11111 & \cSecond 1  & \cThird 1 \\
            \cBase 11111 & \cSecond 1  & \cThird 1 \\
            
            \cBase \dots & \cSecond \dots  & \cThird \dots \\           
            
            \cBase 00000 & \cSecond 0  & \cThird 0 \\
            \cBase 11111 & \cSecond \red{1}  & \cThird 0 \\
            \cBase 00000 & \cSecond 0  & \cThird 0 \\
            
            \cBase \dots & \cSecond \dots  & \cThird \dots \\           
            
        \end{tabular}
    \end{subfigure}
    \hfill
    \begin{subfigure}{0.21\linewidth}
        \centering
        \subcaption{$\hat{Y}_{\normalfont \text{vote}}$: voting on $\mathbf{O}$.  \label{tb:system-B}}
        \begin{tabular}{lcc}
            $\mathbf{O}$     & $\hat{Y}_\text{vote}$ & $Y$ \\
            \midrule

            \cBase 11100 & \cSecond 1  & \cThird 1 \\
            \cBase 11111 & \cSecond 1  & \cThird 1 \\
            \cBase 10011 & \cSecond 1  & \cThird 1 \\
            
            \cBase \dots & \cSecond \dots  & \cThird \dots \\           
            
            \cBase \orange{0}11\orange{0}1 & \cSecond \red{1}  & \cThird 0 \\
            \cBase 00000 & \cSecond 0  & \cThird 0 \\           
            \cBase 00011 & \cSecond 0  & \cThird 0 \\           
            
            \cBase \dots & \cSecond \dots  & \cThird \dots \\           
            
        \end{tabular}
    \end{subfigure}
    \hfill
    \begin{subfigure}{0.21\linewidth}
        \centering
        \subcaption{$\hat{Y}_{\normalfont \text{w.vote}}$: just using $O_1$.\label{tb:system-C}}
        \begin{tabular}{lcc}
            $\mathbf{O}$     & $\hat{Y}_\text{w.vote}$ & $Y$ \\
            \midrule

            \cBase 11100 & \cSecond 1  & \cThird 1 \\
            \cBase 11111 & \cSecond 1  & \cThird 1 \\
            \cBase 10011 & \cSecond 1  & \cThird 1 \\
            
            \cBase \dots & \cSecond \dots  & \cThird \dots \\           
            
            \cBase \blue{0}1101 & \cSecond \blue{0}  & \cThird 0 \\
            \cBase 00000 & \cSecond 0  & \cThird 0 \\           
            \cBase 00011 & \cSecond 0  & \cThird 0 \\           
            
            \cBase \dots & \cSecond \dots  & \cThird \dots \\           
             
        \end{tabular}
    \end{subfigure}
    \hfill
    \begin{subfigure}{0.3\linewidth}
        \centering
        \subcaption{$\hat{Y}_{\normalfont \text{vote}}$: voting on $\mathbf{O}$, $\hat{Y}_{\normalfont \text{w.vote}}$: on $O_{3-5}$ \label{tb:system-D}}
        \begin{tabular}{lccc}
            $\mathbf{O}$     & $\hat{Y}_{\normalfont \text{vote}}$ & $\hat{Y}_{\normalfont \text{w.vote}}$ & $Y$ \\
            \midrule

            \cBase 11\orange{1}00 & \cSecond 1 & \cSecond \red{0}  & \cThird 1 \\
            \cBase 11111 & \cSecond 1 & \cSecond 1 & \cThird 1 \\
            \cBase 000\orange{1}\orange{1} & \cSecond \red{0}  & \cSecond 1 & \cThird 1 \\
            
            \cBase \dots & \cSecond \dots & \cSecond \dots & \cThird \dots \\           
            
            \cBase 111\orange{0}\orange{0} & \cSecond \red{1} & \cSecond 0 & \cThird 0 \\
            \cBase 00000 & \cSecond 0  & \cSecond 0 & \cThird 0 \\           
            \cBase 00\orange{0}11 & \cSecond 0  & \cSecond \red{1}  & \cThird 0 \\           
            
            \cBase \dots & \cSecond \dots & \cSecond \dots & \cThird \dots \\           
        \end{tabular}
    \end{subfigure}   
\end{table*}

To derive \Cref{lemma:fano_error_bound_loose}, which is the basis of \Cref{lemma:ensemble_bound_loose}, two bounds, \cref{eq:fano_expansion_1} and \cref{eq:fano_expansion_2}, are used.
However, in a ensemble learning context, both are not tight, so \Cref{lemma:ensemble_bound_loose} would not give a good approximation of the lower bound.

The problem with relying on \cref{eq:fano_expansion_1} is that the existence of a perfect reconstruction function $\combfunc$ is implicitly assumed.
In the information theoretical context, using the noisy-channel coding theorem \cite{shannon1948mathematical}, we can construct a smart reconstruction function $\combfunc$ so that the information lost by $\mathcal{F}$ is zero as $\CombLossInformation \rightarrow 0$.
Thus, the equality in \Cref{eq:fano_expansion_1} holds.
On the other hand, in the ensemble learning context, we usually use a simple function such as voting or a meta-estimator trained on a limited amount of data as $\mathcal{F}$. %
Therefore, the information loss $\CombLossInformation$ caused by combining the outputs from multiple models $\mathbf{O}$ into a single prediction $\hat{Y}$ should also be taken into account.
We refer to this loss as $\textit{combination loss}$.

The problem with relying \cref{eq:fano_expansion_2} is that an exponentially large number of classes is assumed, i.e., $\YSize\gg1$.
In information theory, $Y$ is assumed to be a sequence of symbols (e.g., bits).
Suppose that the sequence length $L \gg 1$ and that there are $C$ types of symbols; $\YSize$ becomes exponentially large as $\YSize = C^L$.
Then, the second term of the left-hand side of \cref{eq:fano_expansion_2} is approximated as $p_{\text{\rm err}} \log_2 (\YSize - 1) \approx p_{\text{\rm err}} L \log_2 C \gg 1$.
Since the first term ($\binaryEntropyWithArgs \leq 1$) becomes negligible, it can be safely replaced with its upper bound (i.e. $1$) without loosening the inequality much.
On the other hand, in the ensemble learning context, the number of classes $\YSize$ can be small; thus, simply neglecting $\binaryEntropyWithArgs$ produces a loose bound.
For example for binary classification problems, the bound by \Cref{lemma:fano_error_bound_loose} is \textit{always negative} as $\frac{H\left(Y\mid\mathbf{O}\right) - 1}{\log_2 \YSize} \leq 0$ because $H(Y\mid\mathbf{O}) \leq 1$ when $\YSize=2$.

To address these two problems, we lower bounded the error rate using the original Fano's inequality (\Cref{lemma:fano}) directly:

\begin{lemma}[Decomposition of error rate lower bound into three metrics]
    \label{lemma:ensemble_bound_tight}
    Let $\mathcal{U}(p) = \mathcal{H}_2(p) + p\log_2 (\YSize - 1)$ and $\mathcal{U}'(p) = \frac{d\mathcal{U}}{dp}(p)$, and let $p_0 \in [0, 1]$ be the approximate error rate.
    Then, for any $p_0$, the error rate $p_{\text{\rm err}}$ is bounded as
    {
    \small
    \begin{alignSmall}
        & p_{\text{\rm err}} \geq \BoundFuncTightP(\StrengthTripletWithArgs) \nonumber \\
        & \coloneqq p_0 + \frac{\mathcal{U}'(p_0)}{4}\left\{1 - \sqrt{1 - 8\frac{H(Y) - \StrengthTripletWithArgs - \mathcal{U}(p_0)}{\mathcal{U}'(p_0)^2} } \right\}, \label{eq:ensemble_bound_tight}
    \end{alignSmall}
    \normalfont
    }%
    where the ensemble strength $\StrengthTripletWithArgs$ is given by
    {
    \small
    \begin{alignSmall}
        \StrengthTripletWithArgs \coloneqq& \RelevWithArgs  - \RedunWithArgs \nonumber \\ &  -  \ComblossWithArgs, \label{eq:triplet_decomposition} \\
        \ComblossWithArgs  \coloneqq&  H(Y|\hat{Y}) - H(Y|\mathbf{O}). \nonumber
    \end{alignSmall}
    \normalfont
    }
\end{lemma}
\textit{Proof.} In \Cref{lemma:fano}, we expand $\binaryEntropyWithArgs$ by using strong convexity and solve for $p_{\text{\rm err}}$.
\Cref{appendix:true_proof} shows the proof.

\Cref{lemma:ensemble_bound_tight} differs from \Cref{lemma:ensemble_bound_loose} in that 
(i) the ensemble strength $\StrengthTriplet$ \cref{eq:triplet_decomposition} includes the third metric of combination loss, and 
(ii) the bound function is tighter\footnote{If $p_0$ is not far from the lower bound values (\Cref{appendix:tightness})}: $\BoundFuncTightP(E) \geq \BoundFuncLoose(E)$, which is the result of removing the large $\YSize$ assumption.

Since $\StrengthTriplet= \StrengthDoublet - \Combloss$ holds, $\StrengthTriplet$ denotes the amount of unique information on $Y$ carried by $\mathbf{O}$ \textit{that can be extracted when a combination $\combfunc$ is applied to $\mathbf{O}$}.
$\BoundFuncTightP$ is still a decreasing function of $\StrengthTriplet$ when $p_0 \in [0, \frac{\YSize - 1}{\YSize}]$, where $\frac{\YSize - 1}{\YSize}$ denotes the error rate of a random-guessing system on a balanced label dataset.
Thus, \Cref{lemma:ensemble_bound_tight} reveals that an ensemble system should include accurate (large $\Relev$) and diverse (small $\Redun$) models and keep $\Combloss$ small in order to have a small lower bound.

\subsection{What Kind of Systems Produce Combination Loss?}  \label{sec:toy_systems}

To clarify in what kind of ensemble systems combination loss becomes apparent, four toy ensemble systems on an imaginary binary classification task are shown in \Cref{tb:toy_ensemble_systems}.
The systems differ in terms of models $\mathbf{O}=\{O_1,O_2,O_3,O_4,O_5\}$ or combination function $\combfunc$.
Although the systems examined here are extremely simplified and the claims here are hypothetical, they can illustrate certain aspects of empirical behaviors of ensemble systems as discussed in \Cref{sec:triplet_decomposition}.

\Cref{tb:system-A} shows the case where the outputs from each model in $\mathbf{O}$ are perfectly correlated, i.e., there is no diversity between models.
Information theoretically, the system has large redundancy $\Redun$.
In this case, simple voting $\hat{Y}_{\normalfont \text{vote}}$ does not lose any information carried by $\mathbf{O}$, so the combination loss is trivially zero.

\begin{table*}[t!]
    \centering
    \small
    \caption{
    Ensemble methods used in this study.
    We built \NumSystems ensemble systems using all combinations of model generation and combination methods.
    Note that Stacking has three variations (i.e., LogR, SVM and RForest).
    All generation methods train $N$ ($\leq 30$) models using different seed for each model.
    Seed affects random aspects of training, e.g., weight initialization or hidden units dropped when using dropout.
    See \Cref{sec:ensemble_systems} for details. \label{tb:ensemble_methods}}
    \begin{tabularx}{1.0\linewidth}{@{}p{17mm} p{27mm} X@{}}
    \toprule
    Type & Method & Description \\
    \midrule
    \multirow{7}{\hsize}{Model \mbox{Generation}} & Random-HyP  & Train models with different hyperparameters randomly sampled around the best value. \\
    \cmidrule(l){2-3}
     & Bagging  & Train models using different dataset instance sets. Each set contains instances randomly sampled from the original dataset.\\
    \cmidrule(l){2-3}
    {} & Random-Seed & Train models that differ only in the seed of fine-tuning.\\
    \cmidrule(l){2-3}
    & Hetero-DNNs & Train models from $L$(=5) types of DNNs. $M$ models from each type so that $L \times M = N$. \\
    \midrule
    \multirow{6}{\hsize}{Model \mbox{Combination}} & Voting & Take a majority vote on labels predicted by models. \\
    \cmidrule(l){2-3}
    &  Stacking \par\noindent(LogR$|$SVM$|$RForest)\label{sec:voting_and_stacking}  &  Use meta-estimators that make prediction from outputs of models as inputs. We used two-layered stacking with a single meta-estimator, which takes predicted labels as inputs. We trained logistic regression (LogR), Support Vector Machine \cite{Platt99probabilisticoutputs} with RBF kernel (SVM) and Random Forest \cite{breiman2001random} (RForest) as meta-estimators.\\
    \bottomrule
    \end{tabularx}
\end{table*}

\Cref{tb:system-B,tb:system-C} show the cases where the models differ in accuracy, among which $O_1$ performs best.
Information theoretically, the amount of information on Y given by the models is unevenly distributed on the models, and especially concentrated on $O_1$.
Note that the same model set is shown in both tables.
If naive voting is used for model combination (\Cref{tb:system-B}), it produces a prediction error \red{1} even though some of the models ($O_1$ and $O_4$ in this case) give correct predictions \orange{0}.
These correct but neglected minorities are the source of combination loss.
On the other hand, if weighted voting that focuses more on the best model (i.e., $O_1$) is used (\Cref{tb:system-C}), it will succeed in recovering the correct prediction, \blue{0}.

\Cref{tb:system-D} shows the case where the models' outputs are diverse but have the same accuracy.
Information theoretically, information on $Y$ given by $\mathbf{O}$ is uniformly distributed on all the models.
In this case, weighted voting will not help much in recovering the correct predictions compared with simple voting, since there are no better models to be focused on.

From the discussion above, it is expected that (i) models' redundancy decreases combination loss, and (ii) smart combination functions help reduce combination loss, especially when the accuracies of models are varied.

\section{Experiments}   \label{sec:setup}

We empirically validate and demonstrate \Cref{lemma:ensemble_bound_tight}.
To this end, we built various ensemble systems and measured their error rates, error rate lower bounds, and the three metric values.
To be modern and realistic, we built ensemble systems on top of state-of-the-art DNNs, specifically pre-trained language models such as BERT \cite{devlin-etal-2019-bert}.
We used various tasks from the GLUE and SuperGLUE benchmarks \cite{wang-etal-2018-glue, NEURIPS2019_4496bf24}.
These benchmarks include challenging tasks from different domains of NLP and are commonly used to compare state-of-the-art models.

Below, we briefly describe these setups.
For reproducibility, we show the details in \Cref{appendix:setup} and release the code.

\subsection{Models}  \label{sec:models}
We fine-tuned the following five types of language models on downstream tasks: BERT \cite{devlin-etal-2019-bert}, RoBERTa \cite{liu-et-al-2019-roberta}, ELECTRA \cite{clark2020electra}, ALBERT \cite{lan2019albert}, and BART \cite{lewis2019bart}.

\subsection{Ensemble Systems}  \label{sec:ensemble_systems}
To build an ensemble system, we must specify a model generation method (i.e., how to train models that produce $\mathbf{O}$) and a combination method (i.e., $\combfunc$).
We used well-established methods that can be used with DNNs (\Cref{tb:ensemble_methods}).
These methods are commonly used with DNNs in a wide range of domains \cite{kumar2016ensemble,liu2017ensemble,qummar2019deep,ma2019ensemble}, especially in competitions where the highest performance is required \cite{szegedy2015going,7299146,atwood2020inclusive,morishita2020hitachi_task8,morio2020hitachi_task11}.
We built \NumSystems systems using all the combinations of generation and combination methods.

For later convenience, we define the baseline system $s_0$ in each task, which is a single DNN (i.e., no-ensemble) that performs the best among DNNs: ELECTRA for the MRPC/Boolq/SST and RoBERTa for the other tasks.

Random-Seed, Random-HyP and Bagging used a single DNN type the same as $s_0$.
Hetero-DNNs used $L$=5 DNN types.

\begin{figure*}[t!]
    \begin{subfigure}[t]{0.32\linewidth}
        \vskip 0pt
        \centering
        \includegraphics[width=0.65\linewidth]{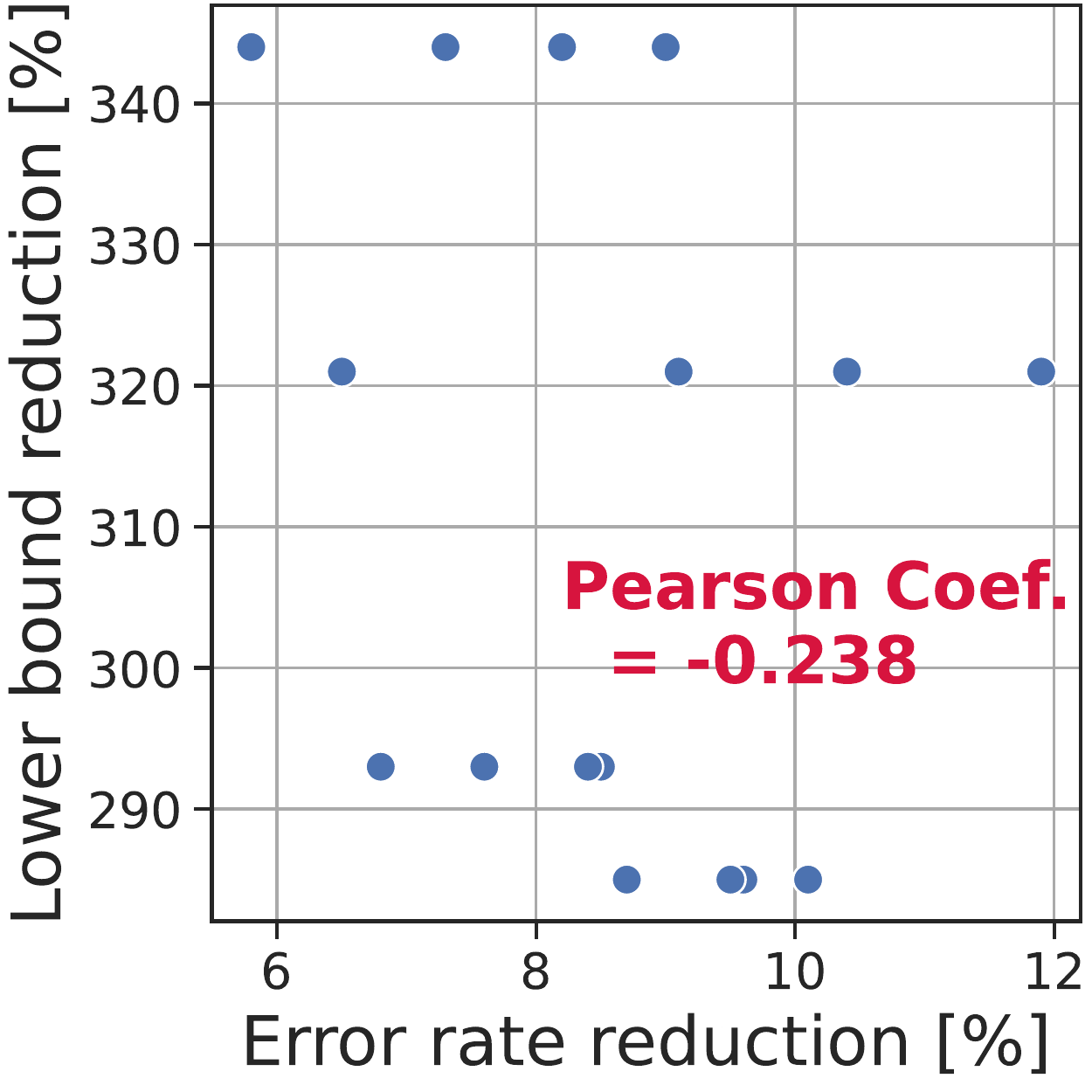}
        \caption{
        \Cref{lemma:ensemble_bound_loose} $\BoundLoose$.
        \label{fig:ERR_LBR_scatter_plot_loose}
        }
    \end{subfigure}
    \hfill
    \begin{subfigure}[t]{0.32\linewidth}
        \vskip 0pt
        \centering
        \includegraphics[width=0.65\linewidth]{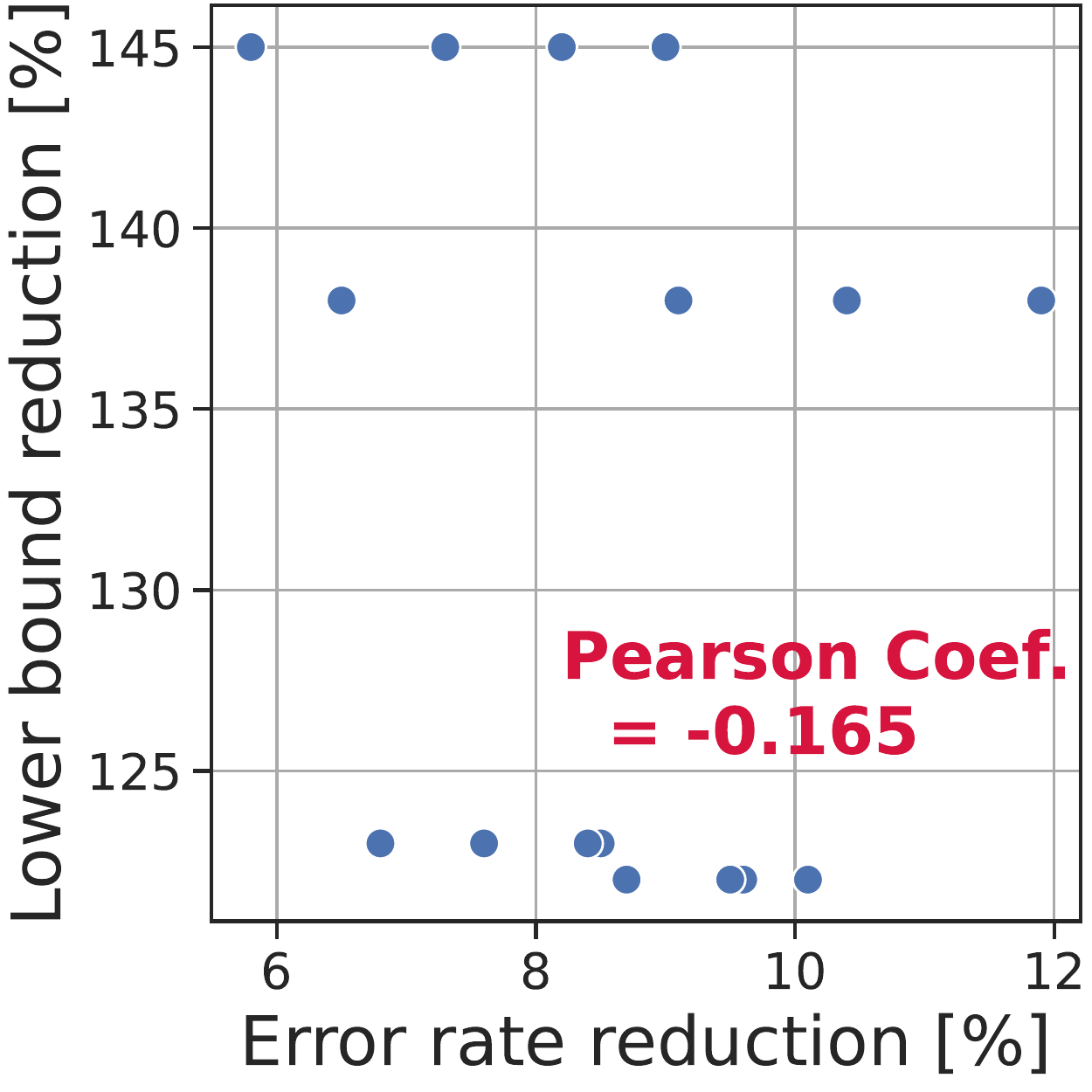}
        \caption{
            $\BoundTightWOCombLoss$.
            \label{fig:ERR_LBR_scatter_plot_tight_wo_combloss}
        }
    \end{subfigure}
    \hfill
    \begin{subfigure}[t]{0.32\linewidth}
        \vskip 0pt
        \centering
        \includegraphics[width=0.65\linewidth]{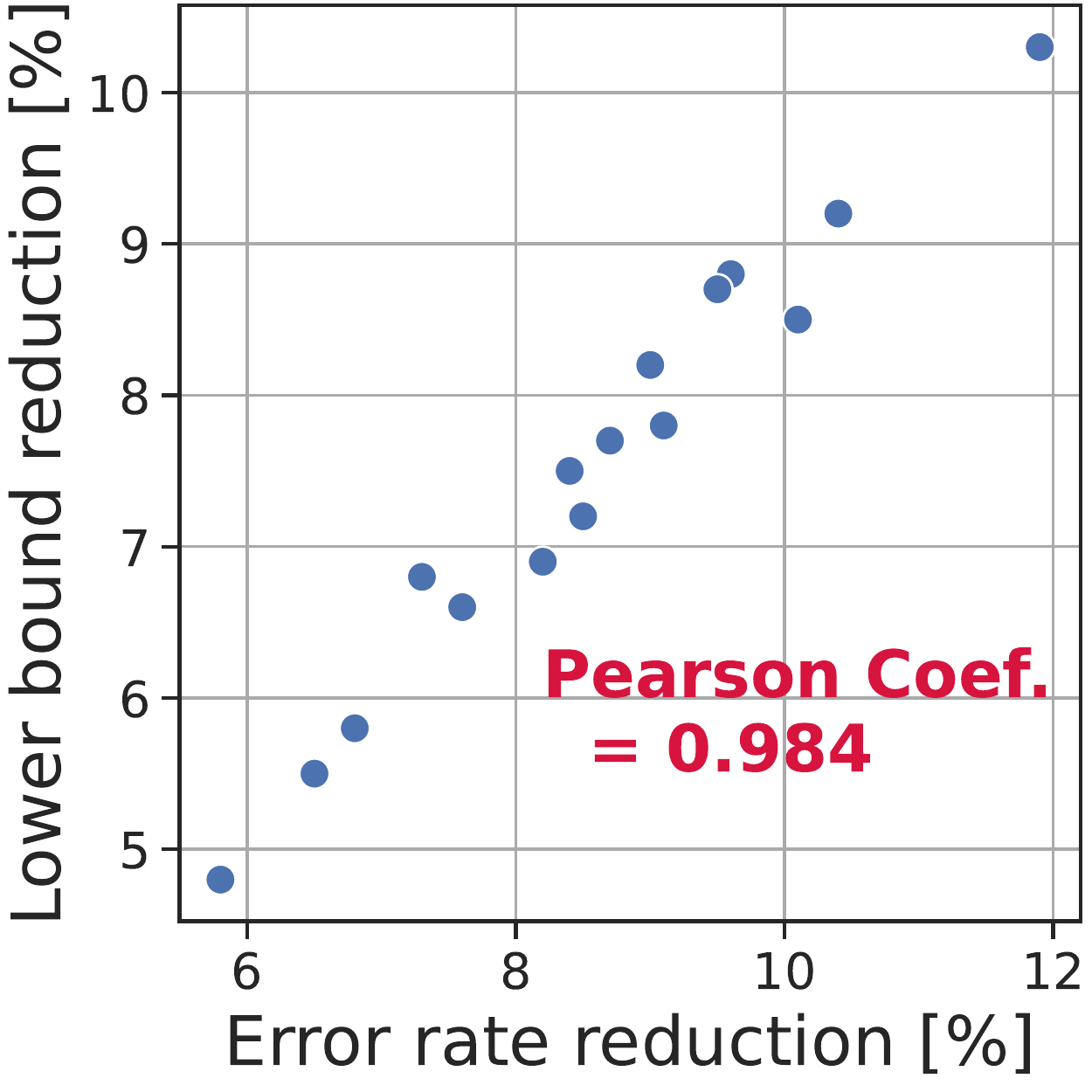}
        \caption{
            \textbf{\Cref{lemma:ensemble_bound_tight}} $\BoundTight$.
            \label{fig:ERR_LBR_scatter_plot_tight}
        }
    \end{subfigure}
    \hfill

    \caption{
    Correlations between error rate reductions and lower bound reductions.
    Each figure uses different type of lower bound.
    Each point in figures shows quantity of specific ensemble system $s$, and quantity is average over \NumTasks tasks.
    See \Cref{tb:ablation_errors} for real value of each point.
    We used \NumSystems ensemble systems described in \Cref{sec:ensemble_systems}.
    Each system $s$ used $N=15$ models.
    Baseline values in \Cref{eq:error_reduction,eq:lower_bound_reduction} are:
    ER($s_0$): \SI{15.5}{\percent},
    LB($s_0$) by $\BoundFuncTight(\StrengthTriplet)$: \SI{2.8}{\percent},
    LB($s_0$) by $\BoundFuncTight(\StrengthDoublet)$: \SI{2.8}{\percent}, and
    LB($s_0$) by $\BoundFuncLoose(\StrengthDoublet)$: \SI{-2.0}{\percent}.
    \label{fig:ERR_LBR_scatter_plot}
    }
\end{figure*}

\subsection{Estimation of Metric Values and Lower Bound} \label{sec:estimation}
We estimated the three metric values ($\Relev$, $\Redun$, and $\Combloss$) and the other quantities appearing in \Cref{lemma:ensemble_bound_loose,lemma:ensemble_bound_tight} on the basis of the observed frequency distribution of the labels $(\mathbf{O}, \hat{Y}, Y)$.
Then, we computed the lower bounds by \Cref{lemma:ensemble_bound_loose,lemma:ensemble_bound_tight}.
All such operations were done on \textit{test sets}\footnote{In order to eliminate from our discussion the statistical fluctuation caused by dataset splitting. Such counfounding factor is undesirable for verifying the theory.}.

To tackle the count sparsity of high-dimensional variables
$\mathbf{O}=\{O_i\mid 1\leq i\leq N, O_i\in\{1, 2, \dots, \YSize\}\}$
, we used the trick of $\mathrm{MTI}_{k=3}$ introduced by \citet{zhou_mti}.

We set the approximate error rate $p_0$ in \cref{eq:ensemble_bound_tight} as the error rate of the baseline $s_0$.
Below, we simply denote $\BoundFuncTight_{p_0}$ as $\BoundFuncTight$.

\subsection{Tasks}  \label{sec:tasks}
We used \NumTasks classification tasks with moderately-sized datasets for computational reasons: Boolq \cite{clark2019boolq}, CoLA \cite{dolan2005automatically}, Cosmos QA \cite{khot2018scitail}, MNLI \cite{MNLI}, MRPC \cite{dolan2005automatically}, SciTail \cite{khot2018scitail},  SST \cite{socher-etal-2013-recursive}, and QQP.

\subsection{Computational Resources / Experimental Runs}  \label{sec:others}
A single run of experiments required about 200 GPUs (V100) $\times$ 1 day.
We ran the experiments three times.

\section{Validation of Framework Through its Predictive Power to Ensemble Phenomena}  \label{sec:validating_framework}
We show that we can predict various phenomena observed on actual ensemble systems using \Cref{lemma:ensemble_bound_tight}.
We show the results aggregated over the \NumTasks tasks here and those for each task in \Cref{appendix:results_for_each_task}.
The discussions here are valid for all tasks, showing their significance.

\Cref{lemma:ensemble_bound_tight} $\BoundFuncTight(\StrengthTriplet)$ differs from \Cref{lemma:ensemble_bound_loose} $\BoundFuncLoose(\StrengthDoublet)$ in two ways, i.e., it has a tightened bound function $\BoundFuncTight$ and ensemble strength with combination loss $\StrengthTriplet$.
To separate contribution of each, we analyze three types of lower bounds hereafter: $\BoundFuncLoose(\StrengthDoublet)$, $\BoundFuncTight(\StrengthDoublet)$, and $\BoundFuncTight(\StrengthTriplet)$.

\subsection{Effect of Bound Function $\BoundFuncTight$}  \label{sec:bound_func_tight}

First, as theoretically expected, the lower bound $\BoundTightWOCombLoss$  was tighter than \Cref{lemma:ensemble_bound_loose} $\BoundLoose$, for example for the baseline system $s_0$, $\BoundFuncTight(\StrengthDoublet_{s_0}) = 2.8\%$ and $\BoundFuncLoose(\StrengthDoublet_{s_0}) = -2.0\%$ (average of \NumTasks tasks).
The captions of \crefrange{tb:ablation_Boolq}{tb:ablation_SST} show the error rates and the error rate lower bounds for \NumTasks tasks.

\subsection{Correlation between Error Rate and Lower Bound}  \label{sec:effectiveness_of_bound}

The error rate lower bound denotes the best-case error rate.
Thus, a system with a smaller lower bound has higher chance of having a smaller error rate \cite{brown_mti, zhou_mti}.
Guided by this intuition, we measured the correlation between the error rates and lower bounds of the ensemble systems.

\Cref{fig:ERR_LBR_scatter_plot} plots the following normalized versions of the error rate and lower bound for each ensemble system $s$:
{
\small
\setlength{\abovedisplayskip}{2pt}
\setlength{\belowdisplayskip}{2pt}
\begin{alignSmall}
    \mathrm{ErrorRateReduction}(s) & = \frac{\mathrm{ER}(s_0) - \mathrm{ER}(s)}{\mathrm{ER}(s_0)} \times 100\ \defPercent, \label{eq:error_reduction} \\
    \mathrm{LowerBoundReduction}(s) & = \frac{\mathrm{LB}(s_0) - \mathrm{LB}(s)}{|\mathrm{LB}(s_0)|} \times 100\ \defPercent.  \label{eq:lower_bound_reduction}
\end{alignSmall}
\normalfont
}%
$s_0$ denotes the single DNN baseline defined in \Cref{sec:ensemble_systems}.
$\mathrm{ER}(s)$ denotes the error rate (i.e., $100\% - \mathrm{accuracy}$) and LB(s) the lower bound.
Note that Pearson correlation coefficient is invariant under this transformation.

\begin{table}[t]
    \centering
    \caption{
    Pearson correlation coefficients between error rate reduction and lower bound reduction.
    In each task, we used the \NumSystems ensemble systems described in \Cref{sec:ensemble_systems}, and each system used $N=15$ models.
    \label{tb:Pearson}
    }
    \resizebox{\linewidth}{!}{%
    \begin{tabular}{@{}lrrr@{}}
        \toprule

        & \multicolumn{3}{c}{Lower bound type} \\

        \cmidrule(l){2-4}
        Task &\Cref{lemma:ensemble_bound_loose}~$\BoundLoose$& $\BoundTightWOCombLoss$ &\textbf{\Cref{lemma:ensemble_bound_tight}}~$\BoundTight$\\
        
        \midrule

Boolq    &  0.341  &    0.330    &    \textbf{0.910}        \\
CoLA    &  -0.211  &    -0.210    &    \textbf{0.991}        \\
CosmosQA    &  -0.324  &    -0.320    &    \textbf{1.000}        \\
MNLI    &  0.226  &    0.216    &    \textbf{0.961}        \\
MRPC    &  0.332  &    0.252    &    \textbf{0.989}        \\
QQP    &  -0.131  &    -0.076    &    \textbf{0.998}        \\
SciTail    &  -0.237  &    -0.191    &    \textbf{0.966}        \\
SST    &  -0.242  &    -0.252    &    \textbf{0.998}        \\
\midrule
average\tablefootnote{The correlation coefficient between the averaged error rate reductions and lower bound reductions. The average is taken over the \NumTasks tasks.}    &  -0.238  &    -0.165    &    \textbf{0.984}        \\

    \bottomrule
    \end{tabular}
}%

\end{table}

\begin{figure*}[t!]
    \begin{subfigure}[t]{0.24\linewidth}
        \vskip 0pt
        \includegraphics[width=\linewidth]{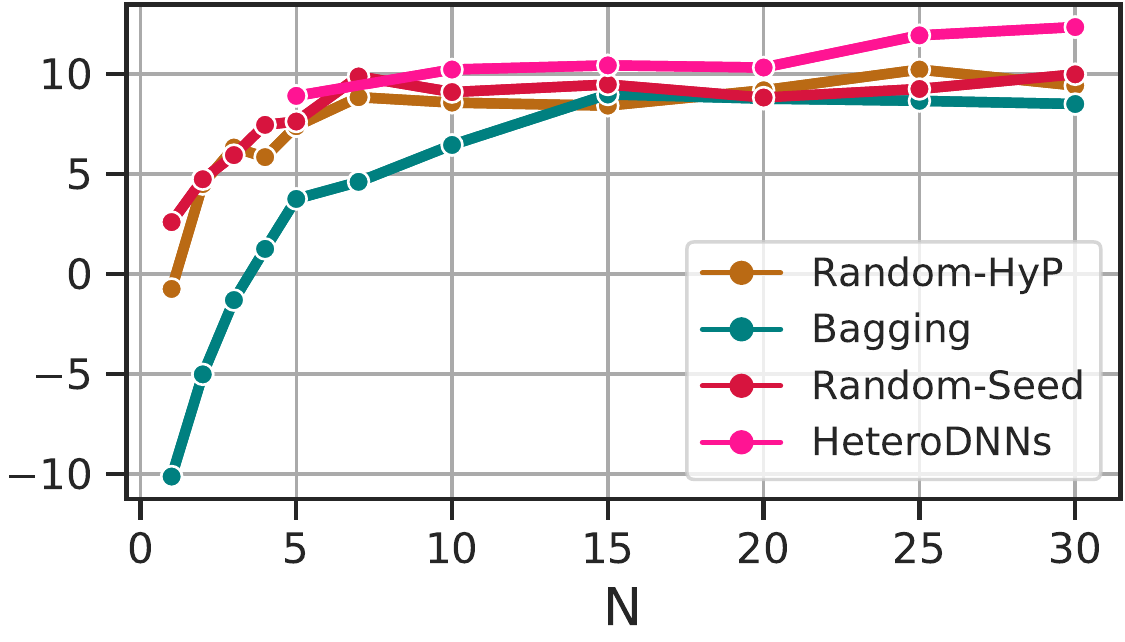}
        \subcaption{Error rate reduction $\defPercent$.\label{fig:scaling.stacking}}
    \end{subfigure}
    \hfill
    \begin{subfigure}[t]{0.24\linewidth}
        \vskip 0pt
        \includegraphics[width=\linewidth]{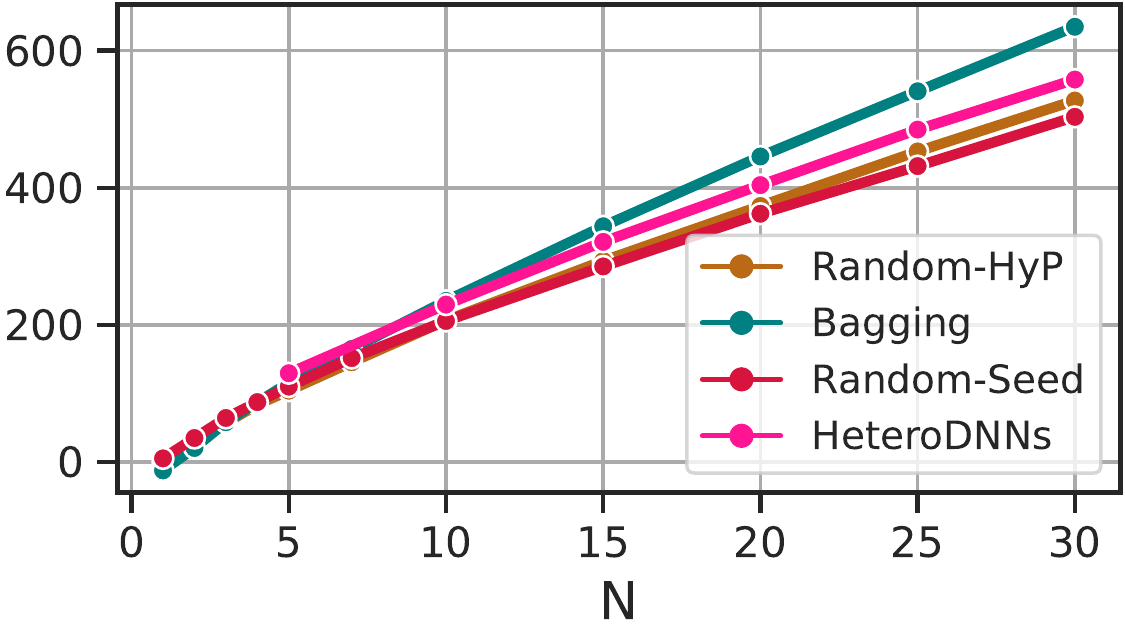}
        \subcaption{
        Lower bound reduction $\defPercent$
        \newline
        \rightline{by \Cref{lemma:ensemble_bound_loose} $\BoundLoose$.}
        \label{fig:scaling.bound.voting.previous_research}
        }
    \end{subfigure}
    \begin{subfigure}[t]{0.24\linewidth}
        \vskip 0pt
        \includegraphics[width=\linewidth]{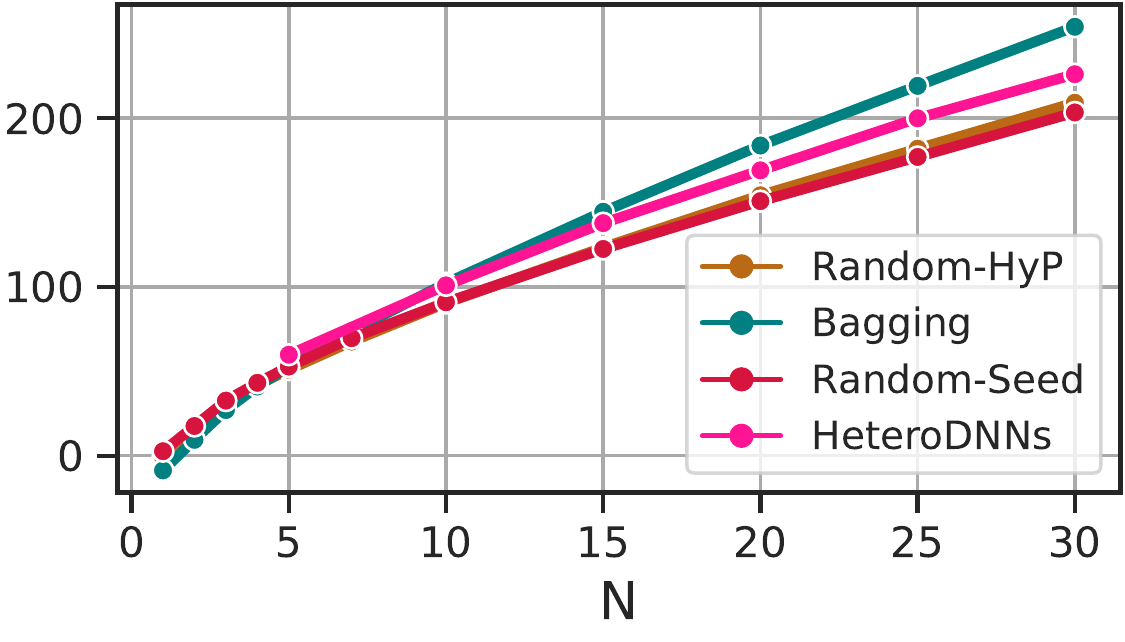}
        \subcaption{
        Lower bound reduction $\defPercent$
        \newline
        \rightline{by $\BoundTightWOCombLoss$.}
        \label{fig:scaling.bound.voting.ours_wo_combloss}
        }
    \end{subfigure}
    \hfill
     \begin{subfigure}[t]{0.24\linewidth}
        \vskip 0pt
        \includegraphics[width=\linewidth]{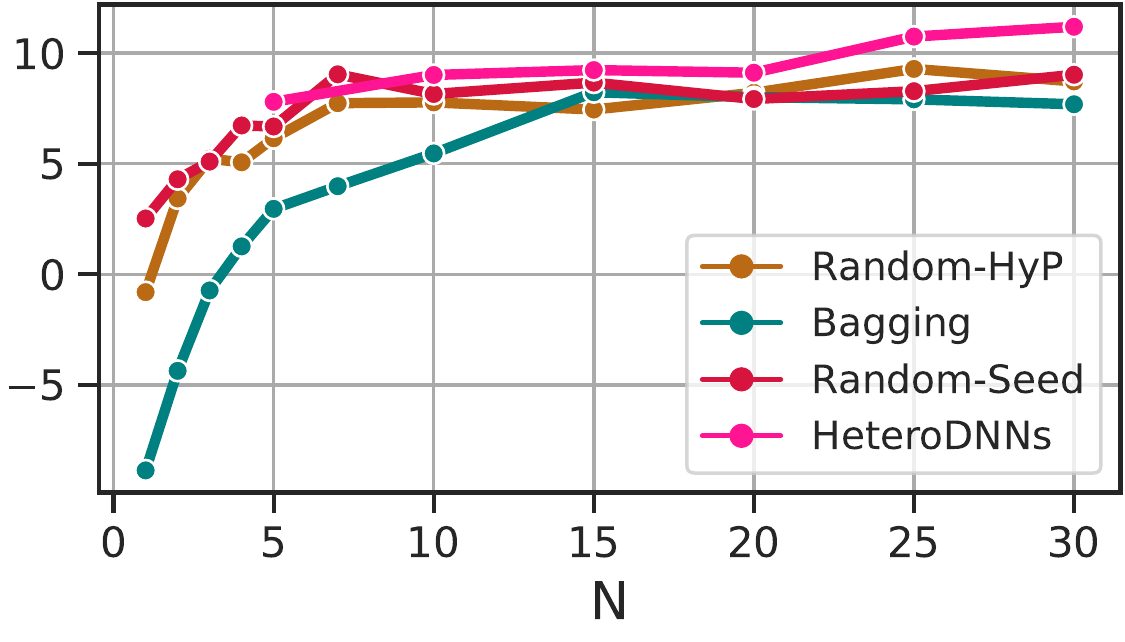}
        \subcaption{
        Lower bound reduction $\defPercent$
        \newline
        \rightline{by \textbf{\Cref{lemma:ensemble_bound_tight}} $\BoundTight$.}
        \label{fig:scaling.stacking.bound}
        }
    \end{subfigure}
    \hfill   
\caption{
Change in error rate reduction and lower bound reduction when number of models $N$ was changed.
Each value is an average of \NumTasks tasks.
Ensemble systems used SVM model combination.
\label{fig:scaling}
}
\end{figure*}

Neither the lower bound reduction by \Cref{lemma:ensemble_bound_loose} $\BoundLoose$ nor that by $\BoundTightWOCombLoss$ correlated with the error rate reduction, as shown in \Cref{fig:ERR_LBR_scatter_plot_tight_wo_combloss,fig:ERR_LBR_scatter_plot_loose}.
In addition, \Cref{lemma:ensemble_bound_loose} $\BoundLoose$ predicted the same lower bound reduction value for different systems that share the same model generation method.
This behavior of the lower bounds can be seen from the points on the same horizontal lines in \Cref{fig:ERR_LBR_scatter_plot_loose}.
This behavior is theoretically expected: since \Cref{lemma:ensemble_bound_loose} $\BoundLoose$ does not include $\Combloss$, it does not consider model combination methods.
This behavior was also observed on $\BoundTightWOCombLoss$ for the same reason.

By contrast, the lower bound reduction by \Cref{lemma:ensemble_bound_tight} $\BoundFuncTight(\StrengthTriplet)$ was very strongly correlated with the error rate reduction, as shown in \Cref{fig:ERR_LBR_scatter_plot_tight}.
Strong correlations were observed for all \NumTasks tasks (\Cref{tb:Pearson}) and also for different $N$s (\crefrange{appendix:tb:Pearson_10}{appendix:tb:Pearson_30}).
These results justify \Cref{lemma:ensemble_bound_tight} and show that $\BoundFuncTight(\StrengthTriplet)$ can be used for comparing systems.
These results also show the importance of combination loss given that the only difference between $\BoundFuncTight(\StrengthTriplet)$ and $\BoundFuncTight(\StrengthDoublet)$ is combination loss.

\subsection{Predicting Error Rate Scaling Curve} \label{sec:scalability}

\Cref{fig:scaling} shows the change in error rate reduction and lower bound reductions when the number of models $N$ was changed.

Both \Cref{lemma:ensemble_bound_loose} $\BoundLoose$ (\Cref{fig:scaling.bound.voting.previous_research}) and $\BoundTightWOCombLoss$ (\Cref{fig:scaling.bound.voting.ours_wo_combloss}) could not predict the shape the of error rate reduction curve (\Cref{fig:scaling.stacking}), especially the saturation over $N \gtrapprox 15$.
By contrast, \Cref{lemma:ensemble_bound_tight} $\BoundTight$ (\Cref{fig:scaling.stacking.bound}) could predict such phenomena.
The results again justify \Cref{lemma:ensemble_bound_tight} and show the importance of combination loss.

Refer to \Cref{appendix:sec:scalability} for more detailed discussions, where we examine the scaling property of each metric values.

\section{Analysis of Ensemble Systems by Framework}  \label{sec:triplet_decomposition}

We demonstrate how we can reveal the strengths and weaknesses of the systems on the basis of the metrics in \Cref{lemma:ensemble_bound_tight}.
The results here are summarized in \Cref{sec:intro}.
We show the results aggregated over the \NumTasks tasks here and those for each task in \Cref{appendix:results_for_each_task}.
The discussions here are valid for all tasks, showing their significance.

\subsection{Justification of Three Metrics for Ensemble System Analysis}

\Cref{tb:ablation} shows the statistics of the ensemble systems.
First, the ranking of the lower bound reduction by $\BoundTight$ in \Cref{tb:ablation_errors} matches the ranking of $\StrengthTriplet$ in \Cref{tb:ablation_triple}.
This is theoretically expected because $\BoundFuncTight$ is a decreasing function.
Thus, $\StrengthTriplet$ can be used for comparing systems, instead of $\BoundTight$.
Furthermore, since $\StrengthTriplet$ is decomposed into the three metrics $(\Relev, \Redun, \Combloss)$ as in \cref{eq:triplet_decomposition}, the three metrics can be used to analyze ensemble systems.

Below, we use \textit{per-model} metrics $\perModelMetricDef$ for intuitive understanding.

\begin{table*}[t!]
    \centering
    \caption{
        Statistics of ensemble systems described in \Cref{sec:ensemble_systems}.
        Rows and columns list model generation and combination methods of \Cref{tb:ensemble_methods}, respectively.
        Each cell shows quantity of specific system $s$.
        Each quantity is average over \NumTasks tasks.
        Each system contains $N=15$ models.
        Color shows rank within \textit{each column} (brighter is better).
        \label{tb:ablation}
    }  
    \begin{subfigure}{\linewidth}
        \centering
        \small
        \tabcolsep 3.0pt
    \subcaption{
        Error rate and lower bound reductions.
        Baseline values used in \Cref{eq:error_reduction,eq:lower_bound_reduction} were
        ER($s_0$): \SI{15.5}{\percent},
        LB($s_0$) by $\BoundFuncTight(\StrengthTriplet)$: \SI{2.8}{\percent},
        LB($s_0$) by $\BoundFuncTight(\StrengthDoublet)$: \SI{2.8}{\percent}, and
        LB($s_0$) by $\BoundFuncLoose(\StrengthDoublet)$: \SI{-2.0}{\percent}.
        \label{tb:ablation_errors}
    }

\begin{tabular}{lccccccccc}
\toprule

& \multicolumn{4}{c}{Error rate reductions $\defPercent$} & &   \multicolumn{4}{c}{Lower bound reductions by $\BoundTight$ $\defPercent$ } \\

\cmidrule(l{\tabcolsep}r{\tabcolsep}){2-5} \cmidrule(l{\tabcolsep}){7-10}

& Voting & LogR & SVM & RForest & & Voting & LogR & SVM & RForest \\

\midrule

Random-HyP    &   \cThird $6.8_{\pm{\mbox{\tiny 1.4}}}$ &    \cThird $8.5_{\pm{\mbox{\tiny 0.9}}}$ &  \cFourth $8.4_{\pm{\mbox{\tiny 1.2}}}$ &   \cThird $7.6_{\pm{\mbox{\tiny 0.7}}}$ &        &   \cThird $5.8_{\pm{\mbox{\tiny 1.4}}}$ &   \cThird $7.2_{\pm{\mbox{\tiny 1.0}}}$ &  \cFourth $7.5_{\pm{\mbox{\tiny 1.1}}}$ &   \cThird $6.6_{\pm{\mbox{\tiny 0.7}}}$ \\
Bagging       &  \cSecond $7.3_{\pm{\mbox{\tiny 2.0}}}$ &   \cFourth $8.2_{\pm{\mbox{\tiny 1.9}}}$ &   \cThird $9.0_{\pm{\mbox{\tiny 1.9}}}$ &  \cFourth $5.8_{\pm{\mbox{\tiny 2.0}}}$ &        &  \cSecond $6.8_{\pm{\mbox{\tiny 2.1}}}$ &  \cFourth $6.9_{\pm{\mbox{\tiny 2.0}}}$ &   \cThird $8.2_{\pm{\mbox{\tiny 2.1}}}$ &  \cFourth $4.8_{\pm{\mbox{\tiny 2.0}}}$ \\ 
Random-Seed   &   \cFirst $9.6_{\pm{\mbox{\tiny 1.2}}}$ &  \cSecond $10.1_{\pm{\mbox{\tiny 0.7}}}$ &  \cSecond $9.5_{\pm{\mbox{\tiny 0.7}}}$ &  \cSecond $8.7_{\pm{\mbox{\tiny 0.2}}}$ &        &   \cFirst $8.8_{\pm{\mbox{\tiny 1.2}}}$ &  \cSecond $8.5_{\pm{\mbox{\tiny 0.7}}}$ &  \cSecond $8.7_{\pm{\mbox{\tiny 0.8}}}$ &  \cSecond $7.7_{\pm{\mbox{\tiny 0.1}}}$ \\
Hetero-DNNs &  \cFourth $6.5_{\pm{\mbox{\tiny 1.4}}}$ &   \cFirst $11.9_{\pm{\mbox{\tiny 0.8}}}$ &  \cFirst $10.4_{\pm{\mbox{\tiny 1.5}}}$ &   \cFirst $9.1_{\pm{\mbox{\tiny 1.9}}}$ &        &  \cFourth $5.5_{\pm{\mbox{\tiny 1.4}}}$ &  \cFirst $10.3_{\pm{\mbox{\tiny 0.8}}}$ &   \cFirst $9.2_{\pm{\mbox{\tiny 1.5}}}$ &   \cFirst $7.8_{\pm{\mbox{\tiny 1.9}}}$  \\

\bottomrule

\end{tabular}

\end{subfigure}

\vspace{2mm}
\begin{subfigure}{\linewidth}
\centering
\small
\tabcolsep 2.0pt
\subcaption{
Breakdown of ensemble strength defined in \cref{eq:triplet_decomposition}.
We show per-model metric values defined as $\perModelMetricDef$. Thus, $\StrengthTriplet = (\relev - \redun - \combloss) \ \times N $ holds.
For intuitive understanding, all values are normalized by ensemble strength of baseline $\StrengthTriplet_{s_0}$, for example, $\Relev = \RelevHat / \StrengthTriplet_{s_0} \times 100$ where $\RelevHat$ is raw value.
\label{tb:ablation_triple}
}

\begin{tabular}{lccccccccccccc}
\toprule

{} & \multicolumn{4}{c}{\multirow{1}{*}{Ensemble strength $\StrengthTriplet$}} & & \multicolumn{6}{c}{Per-model metric values} & \\
\cmidrule(l{\tabcolsep}r{\tabcolsep}){2-5} \cmidrule(l{\tabcolsep}r{\tabcolsep}){6-12}
{} &  & &  &  & & \multirow{2}{*}{$\relev$} & \multirow{2}{*}{$\redun$} & \multicolumn{4}{c}{ $\combloss$} & & \multirow{2}{*}{$\relev - \redun$} \\\cmidrule(lr){9-12}
{} & \multicolumn{1}{c}{Voting} & \multicolumn{1}{c}{LogR} & \multicolumn{1}{c}{SVM} & \multicolumn{1}{c}{RForest} & &  {} &  {} &  \multicolumn{1}{c}{Voting} & \multicolumn{1}{c}{LogR} &  \multicolumn{1}{c}{SVM} & \multicolumn{1}{c}{RForest} & &  {} \\

\midrule
Baseline ($s_0$)                    &    \multicolumn{4}{c}{\cBase 100 (the raw value is 0.478)} &        &   \cBase 100 &                                 \cBase 0 &                                \cBase 0 &                                \cBase 0 &                                \cBase 0 &                                 \cBase 0 &        &   \cBase 100 \\
\midrule

Random-HyP                   &  \cThird $105.0_{\pm{\mbox{\tiny 1.4}}}$ &   \cThird $107.4_{\pm{\mbox{\tiny 1.0}}}$ &   \cFourth $107.5_{\pm{\mbox{\tiny 1.3}}}$ &   \cThird $105.2_{\pm{\mbox{\tiny 0.7}}}$ &        &   \cThird $89.4_{\pm{\mbox{\tiny 0.9}}}$ &   \cThird $74.5_{\pm{\mbox{\tiny 0.9}}}$ &                      \cSecond $7.96_{\pm{\mbox{\tiny 0.36}}}$ &                       \cFirst $7.80_{\pm{\mbox{\tiny 0.34}}}$ &                       \cSecond $8.00_{\pm{\mbox{\tiny 0.37}}}$ &                       \cSecond $7.94_{\pm{\mbox{\tiny 0.29}}}$ &        &  \cFourth $15.0_{\pm{\mbox{\tiny 1.3}}}$ \\
Bagging                      &   \cSecond $105.3_{\pm{\mbox{\tiny 1.9}}}$ &  \cFourth $105.7_{\pm{\mbox{\tiny 1.8}}}$ &  \cThird $108.0_{\pm{\mbox{\tiny 1.1}}}$ &  \cFourth $103.1_{\pm{\mbox{\tiny 1.4}}}$ &        &  \cSecond $90.1_{\pm{\mbox{\tiny 0.3}}}$ &  \cSecond $73.5_{\pm{\mbox{\tiny 0.3}}}$ &                      \cFourth $9.56_{\pm{\mbox{\tiny 0.08}}}$ &                      \cFourth $9.54_{\pm{\mbox{\tiny 0.03}}}$ &                      \cFourth $9.40_{\pm{\mbox{\tiny 0.05}}}$ &                      \cFourth $9.71_{\pm{\mbox{\tiny 0.05}}}$ &        &   \cFirst $16.6_{\pm{\mbox{\tiny 0.4}}}$ \\
Random-Seed                  &   \cFirst $109.2_{\pm{\mbox{\tiny 1.1}}}$ &  \cSecond $108.5_{\pm{\mbox{\tiny 0.9}}}$ &  \cSecond $108.8_{\pm{\mbox{\tiny 1.3}}}$ &  \cSecond $107.7_{\pm{\mbox{\tiny 1.0}}}$ &        &  \cFirst $100.0_{\pm{\mbox{\tiny 0.0}}}$ &  \cFourth $84.9_{\pm{\mbox{\tiny 0.3}}}$ &                       \cFirst $7.79_{\pm{\mbox{\tiny 0.21}}}$ &                      \cSecond $7.84_{\pm{\mbox{\tiny 0.26}}}$ &                      \cFirst $7.82_{\pm{\mbox{\tiny 0.19}}}$ &                      \cFirst $7.89_{\pm{\mbox{\tiny 0.23}}}$ &        &   \cThird $15.1_{\pm{\mbox{\tiny 0.3}}}$ \\
Hetero-DNNs                &  \cFourth $104.5_{\pm{\mbox{\tiny 1.1}}}$ &   \cFirst $110.9_{\pm{\mbox{\tiny 0.8}}}$ &   \cFirst $110.6_{\pm{\mbox{\tiny 1.0}}}$ &   \cFirst $107.8_{\pm{\mbox{\tiny 1.7}}}$ &        &  \cFourth $86.0_{\pm{\mbox{\tiny 0.4}}}$ &   \cFirst $69.9_{\pm{\mbox{\tiny 0.2}}}$ &                       \cThird $9.16_{\pm{\mbox{\tiny 0.24}}}$ &                       \cThird $8.73_{\pm{\mbox{\tiny 0.26}}}$ &                       \cThird $8.75_{\pm{\mbox{\tiny 0.29}}}$ &                       \cThird $8.94_{\pm{\mbox{\tiny 0.19}}}$ &        &  \cSecond $16.1_{\pm{\mbox{\tiny 0.4}}}$ \\

\bottomrule
\end{tabular}

\end{subfigure}

\end{table*}

\subsection{Analysis of Model Generation Methods}

\textbf{Random-Seed and Hetero-DNNs} systems performed the best or second best in each column of \Cref{tb:ablation_errors} (i.e. among the systems with the same combination method).
Looking into the per-model relevance $\relev$ in \Cref{tb:ablation_triple},
Random-Seed had the largest $\relev$ in each column.
$\relev$ denotes the average accuracy of the models.
Indeed, the ranking of $\relev$ coincided with the ranking of the average error rate shown as ``avg'' in \Cref{tb:ablation_omega_statistics}.
Random-Seed had the most accurate models because it used only the best DNN type (cf. Hetero-DNNs), all the dataset instances (cf. Bagging), and only the best hyperparameter (cf. Random-HyP).

On the per-model redundancy $\redun$, Hetero-DNNs had a value smaller than that of Random-Seed (i.e., it had more diverse models), benefitting from the diverse DNN types.

For per-model combination loss $\combloss$
\footnote{
The magnitude of $\combloss$ is smaller than those of $\relev$ and $\redun$. However, $\relev$ and $\redun$ are strongly correlated, and thus, $\combloss$ is not negligible compared with $\relev - \redun$, as shown. Thus, combination loss is significant.},
Random-Seed had the smallest value in the voting column.
We attribute this to it having lowest diversity (i.e., the largest $\redun$), similarly to \Cref{tb:system-A}.
However, the meta-estimators (LogR, SVM, and RForest) reduced $\combloss$ more on Hetero-DNNs than on Random-Seed.
This pushed the performance of Hetero-DNNs to the highest among all the systems.
Regarding this phenomenon,  Hetero-DNNs can be analogous to \crefrange{tb:system-B}{tb:system-C} and Random-Seed to \Cref{tb:system-D}: since Hetero-DNNs uses various DNN types of varied accuracies, the amount of information on $Y$ is concentrated more on better models compared with Random-Seed.
Thus, Hetero-DNNs benefitted more from the meta-estimators, which focused on these models and recovered the information to reduce $\combloss$, similarly to the transition from \crefrange{tb:system-B}{tb:system-C}.
This phenomenon did not occur in Random-Seeed since it uses models of similar accuracies, similarly to \Cref{tb:system-D}.

Indeed, we can see the information concentration and how the meta-estimator handled such information more directly.
To this end, we propose an auxiliary metric of \textit{$n$-model concentration} $Conc^N_n$ (\Cref{appendix:concentration}) which measures the degree to which the amount of information given by $N$ models $\mathbf{O} = \{O_1,\dots, O_N\}$ is concentrated on the top-$n$ models $\Omega^{N, \mathrm{max}}_n$:
{
\small
\setlength{\abovedisplayskip}{0pt}
\setlength{\belowdisplayskip}{0pt}
\begin{alignSmall}
    Conc^N_n(\mathbf{O}, Y) &= \frac{\omegaMax - \omegaMin}{I(\mathbf{O};Y)} \in [0, 1], \label{eq:concentration} \\ 
    I(\Omega^{N, \mathrm{max/min}}_n;Y) & = \underset{\{i_1, i_2, \dots, i_n\} \in \Omega^N_n}{\mathrm{max/min}} I(\{O_{i_1}, O_{i_2}, \dots, O_{i_n}\};Y).  \nonumber
\end{alignSmall}
\normalfont
}

\Cref{tb:ablation_omega_statistics} shows $Conc^{N=15}_{n=3}$ for each model generation method.
Intuitively, $Conc^N_n$ and the standard deviation of model error rates, which denotes the variety in accuracies, were strongly correlated. 
Hetero-DNNs had a larger $Conc^N_n$ and Random-Seed a smaller one, as expected.
\Cref{tb:base_model_type_performances} shows that the meta-estimator for Hetero-DNNs distributed weight $W_t$ to each DNN type $t$ in accordance with its error rate.
Overall, we can see a clear analogy of Hetero-DNNs to \Cref{tb:system-B,tb:system-C}, and of Random-Seed to \Cref{tb:system-D}.

\begin{figure}[t!]
\small
\tabcolsep 3.0pt
    \captionof{table}{
    The information concentration metric $Conc^{N=15}_{n=3}$. See \Cref{eq:concentration}.
    Color shows rank (brighter is better) in each column.
    Values are averages over \NumTasks tasks.
    \label{tb:ablation_omega_statistics}}
    \centering

\begin{tabular}{lccc}
\toprule
{} & \multicolumn{1}{c}{\multirow{2}{*}{$Conc^{N=15}_{n=3}$}} & \multicolumn{2}{c}{Error rates of models $\defPercent$} \\
\cmidrule(lr){3-4} 
{Model generation} & {} & \multicolumn{1}{c}{ avg.} & \multicolumn{1}{c}{std.} \\

\midrule
Baseline ($s_0$)                    &                                  \cBase 0 &    \cBase $16.1_{\pm{\mbox{\tiny 0.9}}}$ &    \cBase - \\
\midrule
Random-HyP                   &   \cFirst $0.28_{\pm{\mbox{\tiny 0.02}}}$ &   \cThird $17.3_{\pm{\mbox{\tiny 0.1}}}$ &   \cFirst $3.4_{\pm{\mbox{\tiny 0.2}}}$ \\
Bagging                      &   \cThird $0.08_{\pm{\mbox{\tiny 0.00}}}$ &  \cSecond $17.1_{\pm{\mbox{\tiny 0.0}}}$ &   \cThird $0.8_{\pm{\mbox{\tiny 0.0}}}$ \\
Random-Seed                  &   \cThird $0.08_{\pm{\mbox{\tiny 0.00}}}$ &   \cFirst $15.5_{\pm{\mbox{\tiny 0.1}}}$ &  \cFourth $0.7_{\pm{\mbox{\tiny 0.1}}}$ \\
Hetero-DNNs                &  \cSecond $0.20_{\pm{\mbox{\tiny 0.00}}}$ &  \cFourth $18.1_{\pm{\mbox{\tiny 0.0}}}$ &  \cSecond $2.3_{\pm{\mbox{\tiny 0.0}}}$ \\

\bottomrule
\end{tabular}

    \tabcolsep 3.0pt
    \small
    \centering
    \caption{
    Logistic regression meta-estimator weight $W_t$ distributed to each DNN type $t$.
    $N$=15 models are generated by Hetero-DNNs (i.e., 3 models per DNN type).
    Values are averages over \NumTasks tasks.
    See \Cref{appendix:stacking_weight_distribution} for details.
    \label{tb:base_model_type_performances}
    }
    \vspace{2mm}
    \begin{tabular}{@{}lcc@{}}
        \toprule
        DNN $t$ & Average error rate of models $\defPercent$ & $W_{t}$\\
        \midrule
        
        RoBERTa &  $15.1 {{\scriptstyle \pm0.3}}$ & 0.49\\
        ELECTRA &  $17.0 {{\scriptstyle \pm0.1}}$ & 0.40\\
        BART &  $17.9 {{\scriptstyle \pm0.1}}$ & 0.25\\
        BERT &  $18.7 {{\scriptstyle \pm0.1}}$ & 0.24\\
        ALBERT &  $20.4 {{\scriptstyle \pm0.1}}$ & 0.21\\
        \bottomrule
        
        \end{tabular}
\end{figure}

\textbf{Bagging and Random-HyP} performed the third or fourth best in each column of \Cref{tb:ablation_errors}.
Similarly to Hetero-DNNs, they had a smaller $\relev$ and $\redun$ compared with Random-Seed (\Cref{tb:ablation_triple}).
The smaller $\relev$ is attributed to Bagging using smaller subsets of training instances and Random-HyP using randomly sampled non-optimal hyperparameters, which degraded model accuracies.
The smaller $\redun$ is due to the diverse instance sets of Bagging and the diverse hyperparameters of Random-HyP.

Bagging had the largest $\combloss$ in each column, and more importantly, the meta-estimators (LogR, SVM and RForest) could not reduce $\combloss$ as much as they could on Hetero-DNNs and Random-HyP.
This phenomenon should be due to the Bagging's smaller $Conc^N_n$ (\Cref{tb:ablation_omega_statistics}), which is the result of models of similar accuracies, similarly to \Cref{tb:system-D}.
Such models were generated because Bagging used dataset sub-sets of the same size.

\subsection{Analysis of Model Combination Methods}

Stacking (i.e., LogR, SVM, and RForest) generally outperformed voting in each row of \Cref{tb:ablation_errors}.
This is due to the smaller $\combloss$ since $\relev$ and $\redun$ are the same in each row.
Simply, the meta-estimators combined the models better to reduce $\combloss$.

Interestingly, the simple meta-estimator of LogR performed on par with or better than the complex ones of SVM and RForest.
We estimate that the DNN's predictions were so good that simple combinations were enough, and complex ones were superfluous.

\section{Other Considerations}  \label{sec:other_considerations}
The limitations of the study are listed in \Cref{appendix:limitations}.
Ethical matters and social impacts are discussed in \Cref{appendix:social_impacts}.

\section{Conclusion}  \label{sec:conclusion}
We proposed a novel and fundamental theoretical framework that measures a given ensemble system on the basis of a well-grounded set of metrics.
We also validated and demonstrated the framework through experiments on DNN ensemble systems.
In the future, we will analyze a broader range of systems, including rec/ent DNN ensemble systems optimized in an end-to-end manner.
We will also incorporate combination loss into ensemble systems as an optimization target (i.e., as a loss-term) for better performance.

\section*{Acknowledgement}
We thank the three anonymous reviewers and the meta-reviewer, who gave us insightful comments and suggestions.
We thank Dr. Masaaki Shimizu and Yasunori Kaneda at Hitachi for the convenience of additional computational resources.
We thank Osamu Imaichi and Masashi Egi at Hitachi for insightful comments.
We thank Dr. Naoaki Okazaki, professor at Tokyo Institute of Technology, for the keen comments.
Computational resources of AI Bridging Cloud Infrastructure (ABCI) provided by the National Institute of Advanced Industrial Science and Technology (AIST) were used.

\bibliography{icml2022}
\bibliographystyle{icml2022}

\appendix

\numberwithin{equation}{section}
\renewcommand{\thefigure}{\Alph{section}.\arabic{figure}}
\renewcommand{\thetable}{\Alph{section}.\arabic{table}}

\newpage

\section{Limitations}  \label{appendix:limitations}
The study has the following limitations:
\begin{itemize}
    \item As stated in \Cref{sec:intro}, the framework \Cref{lemma:ensemble_bound_tight} deals with classification tasks.
    \item As stated in \Cref{sec:toy_systems}, the claims made on the toy ensemble systems of \Cref{tb:toy_ensemble_systems} are hypothetical rather than theoretically driven, although they have explained certain aspects of the experimental results as discussed in \Cref{sec:triplet_decomposition}.
\end{itemize}

\section{Ethics and Social Impacts}  \label{appendix:social_impacts}
Ensemble learning is a generic technology to boost the performance of machine learning models. 
This study provides a theoretical framework on ensemble learning for evaluating a given ensemble system by a set of specific metrics.
The framework enables us to reveal the strengths and weaknesses of ensemble systems on each metric, which will give us insights into the designing of ensemble systems.
Thus, this study should ultimately lead to the better performance of machine learning models.

While it is possible that inappropriate use of improved machine learning models poses negative effects on society, we believe that this study does not directly pose negative effects on society.

\section{Definitions}  \label{appendix:notations}
We show the definitions of information theoretical quantities used in this study.
In the below, we assume that $\mathbf{S}$ and $\mathbf{T}$ denote sets of discrete stochastic variables:
\begin{alignSmall}
\mathbf{S} & = \{S_1, S_2, \dots, S_L\},\ L \in \mathbb{N}, \nonumber\\
\mathbf{T} & = \{T_1, T_2, \dots, T_M\},\ M \in \mathbb{N}, \nonumber
\end{alignSmall}
where $S_i$ and $T_i$ are discrete stochastic variables.
We denote $s_i, t_i$ as the values of $S_i, T_i$, and $p$ as the probability distribution function.

\begin{definition}[Entropy of $\mathbf{S}$]
\begin{gatherSmall}
H(\mathbf{S}) = - \sum_{s_1, \dots, s_L} p(s_1, \dots, s_L) \log_2 p(s_1, \dots, s_L).  \label{eq:notation_entropy}
\end{gatherSmall}
\end{definition}

\begin{definition}[Conditional entropy of $\mathbf{T}$ given $\mathbf{S}$]
\begin{alignSmall}
H(\mathbf{T}|\mathbf{S}) = - \sum_{s_1, \dots, s_L} \sum_{t_1, \dots, t_M} & p(s_1, \dots, s_L, t_1, \dots, t_M) \nonumber \\
 & \times \log_2 p(t_1, \dots t_M|s_1, \dots, s_L). \label{eq:notation_conditional_entropy}
\end{alignSmall}
\end{definition}

\begin{definition}[Mutual-information between $\mathbf{S}$ and $\mathbf{T}$]
\begin{gatherSmall}
I(\mathbf{S};\mathbf{T}) = H(\mathbf{T}) - H(\mathbf{T}|\mathbf{S}).  \label{eq:notation_mutual_information}
\end{gatherSmall}
\end{definition}

\begin{definition}[Multi-information of $\mathbf{S}$]
\begin{alignSmall}
    \multiInfo(\mathbf{S}) = \sum_{s_1, \dots, s_L} p(s_1, \dots, s_L) \log_2 \frac{p(s_1, \dots, s_L)}{p(s_1)\dots.p(s_L)}. \label{eq:notation_multi_information}
\end{alignSmall}
\end{definition}

\begin{definition}[Conditional multi-information of $\mathbf{T}$ given $\mathbf{S}$]
\begin{alignSmall}
    \multiInfo(\mathbf{T}|\mathbf{S}) = \sum_{s_1, \dots, s_L} & \sum_{t_1, \dots, t_M} p(s_1, \dots, s_L, t_1, \dots, t_M) \nonumber \\
    \times \log_2 & \frac{p(t_1, \dots, t_M|s_1, \dots, s_L)}{p(t_1|s_1, \dots, s_L)\dots p(t_M|s_1, \dots, s_L)}. \label{eq:notation_conditional_multi_information}
\end{alignSmall}
\end{definition}

For the interpretation of \crefrange{eq:notation_multi_information}{eq:notation_conditional_multi_information}, see \cite{brown_mti, zhou_mti}.

\section{About \Cref{lemma:ensemble_bound_tight}}\label{appendix:proof}

\subsection{Full Proof}\label{appendix:true_proof}

\newcommand{\QuadraticWithArgs}{\hat{\mathcal{H}}_2(p_{\text{\rm err}})}
\newcommand{\Quadratic}{\hat{\mathcal{H}}_2}
\newcommand{\PBound}{\bar{p}_{p_0}}
\newcommand{\BinaryEntropyWithArgs}{\binaryEntropyWithArgs}
\newcommand{\BinaryEntropy}{\mathcal{H}_2}
\newcommand{\TightRangePPlusMinus}{\mathcal{R}^{\pm}_{p0}}
\newcommand{\TightRangePPlus}{\mathcal{R}^{+}_{p0}}
\newcommand{\TightRangePMinus}{\mathcal{R}^{-}_{p0}}
\newcommand{\UpperBoundFuncTight}[2]{\hat{\mathcal{U}}^{\normalfont \text{tight}}_{m#1, p_0#2}}
\newcommand{\UpperBoundFuncLoose}[2]{\hat{\mathcal{U}}_{m#1, p_0#2}}
\newcommand{\DeltaPlus}{\Delta^{+}_{p_0}}
\newcommand{\DeltaMinus}{\Delta^{-}_{p_0}}

\begin{alignSmall}
    H(Y|\mathbf{O}) & + \CombLossInformationWithName =  H(Y|\hat{Y}), \nonumber \\
    & \leq \underbrace{\binaryEntropyWithArgs} + p_{\text{\rm err}} \log_2 (\YSize - 1) \eqqcolon \mathcal{U}(p_{\text{\rm err}}), \nonumber \\
    & \leq \underbrace{\mathcal{H}_2(p_0) + \mathcal{H}_2'(p_0)(p_{\text{\rm err}} - p_0) -\frac{m}{2}(p_{\text{\rm err}} - p_0)^2}_{\eqqcolon\QuadraticWithArgs} \nonumber \\ 
    & \qquad \qquad \qquad \qquad \qquad \qquad + p_{\text{\rm err}} \log_2 (\YSize - 1), \nonumber \\ 
    & \eqqcolon \UpperBoundFuncTight{}{}(p_{\text{\rm err}}).  \label{appendix:eq:concavity}
\end{alignSmall}
The first inequality follows from Fano's inequality \Cref{lemma:fano}.
In the second inequality, we used strong concavity of binary cross entropy function $\underbrace{\BinaryEntropyWithArgs}$ to upper bound it by another quadratic function $\QuadraticWithArgs$ tangent to ${\BinaryEntropyWithArgs}$ at $p_{\text{\rm err}} = p_0$.
\cref{appendix:eq:concavity} holds for any $p_0 \in [0, 1]$ and $m \leq 4$.

$m$ represents the curvature of $\QuadraticWithArgs$.
Setting $m=4$ produces the most curved quadratic function $\QuadraticWithArgs$, and hence the tightest upper bound of $\BinaryEntropyWithArgs$.
Then, decomposing $H(Y|\mathbf{O})$ of the left-hand side as \Cref{lemma:ensemble_bound_loose} and solving \cref{appendix:eq:concavity} for $p_{\text{\rm err}}$ derives \Cref{lemma:ensemble_bound_tight}.

The choice of $p_0$ of \Cref{lemma:ensemble_bound_tight} is discussed in \Cref{appendix:choice_of_p0}.

\subsection{Which Choice of $p_0$ is Preferable for Ensemble System Comparison}  \label{appendix:choice_of_p0}

\Cref{lemma:ensemble_bound_tight} discloses lower bounds that depend on $p_0$.
For fair comparisons of ensemble systems, we must first choose and fix a specific value of $p_0$ from $[0, 1]$.
Any choice of $p_0$ is ok since it does not change the ranking of lower bounds.
In our experiments, we chose the baseline error rate as our $p_0$ due to the following reason.

As stated in \cref{appendix:true_proof}, we approximated the binary cross entropy function $\BinaryEntropyWithArgs$ as a quadratic function $\QuadraticWithArgs$ tangent to $\BinaryEntropyWithArgs$ at $p_{\text{\rm err}} = p_0$.
Thus, the approximation error $\UpperBoundFuncTight{=4}{}(p_{\text{\rm err}}) - \mathcal{U}(p_{\text{\rm err}})$ is the smallest when $p_0 \sim p_{\text{\rm err}}$, where $p_{\text{\rm err}} = \BoundFuncTightP(E)$ is the actual lower bound obtained by ensemble strength $E$ of each of the ensemble systems.
This means that we should choose a value of $p_0$ that is similar to the error rate lower bounds of the target ensemble systems due to the following reason.

Since we do not know the error lower bounds of the systems before we choose $p_0$ and solve $p_{\text{\rm err}} = \BoundFuncTightP(E)$, it is a bit complicated to tune the value of $p_0$, although it is possible.
Thus, in the experiments of this study, we chose the baseline error rate as our $p_0$ rather than tuning $p_0$.
The baseline error rate is expected to be similar to the error rates of the ensemble systems, and hence it should not be much different from the lower bounds of the systems.

\subsection{Comparison between Tightness of \Cref{lemma:ensemble_bound_tight} and \Cref{lemma:ensemble_bound_loose}}  \label{appendix:tightness}

\Cref{lemma:ensemble_bound_tight} differs from \Cref{lemma:ensemble_bound_loose} in the lower bound functions.
That is, \Cref{lemma:ensemble_bound_tight} uses $\BoundFuncTightP(E)$ while \Cref{lemma:ensemble_bound_loose} uses $\BoundFuncLoose(E)$.
In this section, we show that the bound function $\BoundFuncTightP(E)$ is tighter (i.e. larger) than $\BoundFuncLoose(E)$ if $E$ is in a specific range in which $\BoundFuncTightP(E)$ is not much different from $p_0$.
Hereafter we assume $p_0 \leq \frac{\YSize-1}{\YSize}$, $\BoundFuncTightP(E) \leq \frac{\YSize-1}{\YSize}$, and $\BoundFuncLoose(E) \leq \frac{\YSize-1}{\YSize}$, where $\frac{\YSize - 1}{\YSize}$ means an error rate of a random guessing system on a balanced label dataset.

Firstly, we show how the two lemmas are derived.

\textbf{\Cref{lemma:ensemble_bound_tight}} is derived using \cref{appendix:eq:concavity} as:.
\begin{enumerate}
    \item Set $m=4$. That is, we use $\UpperBoundFuncTight{=4}{}(p_{\text{\rm err}})$ for the upper bound function.
    \item Solving for $p_{\text{\rm err}}$ derives \Cref{lemma:ensemble_bound_tight} $p_{\text{\rm err}} \geq \BoundFuncTightP(\StrengthTriplet)$.
\end{enumerate}

\textbf{\Cref{lemma:ensemble_bound_loose}} is derived in a similar way as:
\begin{alignSmall}
    H(Y|\mathbf{O}) + & \CombLossInformationWithName \leq \UpperBoundFuncTight{}{}(p_{\text{\rm err}}), \nonumber \\
    & \leq \UpperBoundFuncTight{}{}(p_{\text{\rm err}}) + p_{\text{\rm err}} \log_2\frac{\YSize}{\YSize - 1} \eqqcolon \UpperBoundFuncLoose{}{}(p_{\text{\rm err}}). 
    \label{appendix:eq:concavity_loose}
\end{alignSmall}
\begin{enumerate}
    \item Set $m=0, p_0=\frac{1}{2}$. That is, we use $\UpperBoundFuncLoose{=0}{=\frac{1}{2}}(p_{\text{\rm err}})$ for the upper bound function.
    \item Loosen the left-hand side as $H(Y|\mathbf{O}) + \CombLossInformationWithName \geq H(Y|\mathbf{O})$, that is, ignore the combination loss.
    \item Then, solving for $p_{\text{\rm err}}$ derives \Cref{lemma:ensemble_bound_loose} $p_{\text{\rm err}} \geq \BoundFuncLoose(\StrengthDoublet)$.
\end{enumerate}

Viewing these, we can immediately show that if we use $p_0 = \frac{1}{2}$ for \Cref{lemma:ensemble_bound_tight}, it's bound function is tighter as $\forall E, \mathcal{B}^{\normalfont \text{tight}}_{p_0=\frac{1}{2}}(E) \geq \BoundFuncLoose(E)$.
This follows from $\UpperBoundFuncTight{=4}{=\frac{1}{2}}(p_{\text{\rm err}}) \leq  \UpperBoundFuncTight{=0}{=\frac{1}{2}}(p_{\text{\rm err}}) + p_{\text{\rm err}} \log_2 \frac{\YSize}{\YSize - 1} = \UpperBoundFuncLoose{=0}{=\frac{1}{2}}(p_{\text{\rm err}})$.
We also point out that \Cref{lemma:ensemble_bound_loose} poses the following assumptions which may lead to loose bound; (i) $m=0$. This means that the upper bound function $\UpperBoundFuncLoose{=0}{=\frac{1}{2}}(p_{\text{\rm err}})$ is a line. (ii) The existence of positive term $p_{\text{\rm err}} \log_2 \frac{\YSize}{\YSize - 1}$.

When we use $p_0 \neq \frac{1}{2}$, that is more general, the tightness $\BoundFuncTightP(E) \geq \BoundFuncLoose(E)$ holds in limited ranges of $E$.
As stated in \Cref{appendix:choice_of_p0}, the approximation error produced by $\UpperBoundFuncTight{}{}(E)$ is the smallest if $\BoundFuncTightP(E) \sim p_0$. 
Thus, roughly speaking, $\BoundFuncTightP(E) \geq \BoundFuncLoose(E)$ holds if $\BoundFuncTightP(E) \sim p_0$ and $\BoundFuncTightP(E) \leq \BoundFuncLoose(E)$ holds if $\BoundFuncTightP(E)$ and $p_0$ differ much in their values.

We can discuss the details as follows.
The tightness condition on ensemble strength $E$ is given by:
\begin{alignSmall}
    \BoundFuncTightP(E) \geq \BoundFuncLoose(E). \label{appendix:eq:tightness_on_E}
\end{alignSmall}
Let lower bound function $\PBound(E) = \BoundFuncTightP(E)$.
Solving \cref{appendix:eq:tightness_on_E} for $E$ can derive the range of $\PBound(E)$ where the tightness holds:
\begin{gatherSmall}
    \PBound(E) \leq \min(p_0 + \DeltaPlus, p_0 + \DeltaMinus), \label{appendix:eq:tight_range_p_upper}\\
    \max(p_0 + \DeltaPlus, p_0 + \DeltaMinus) \leq \PBound(E), \label{appendix:eq:tight_range_p_lower}
\end{gatherSmall}
where
\begin{alignSmall}
    \Delta^{\pm}_{p_0} & \coloneqq \tau(p_0) \left[1 \sqrt{1 - \frac{1}{2}\frac{1 - \BinaryEntropy(p_0) - \log \frac{\YSize-1}{\YSize}}{\tau(p_0)^2}} \right], \nonumber \\
    \tau(p_0) & \coloneqq \frac{1}{4}\left[\frac{d\BinaryEntropy}{dp}(p_0) - \log \frac{\YSize}{\YSize - 1}\right]. \nonumber
\end{alignSmall}
We assumed $1 - \frac{1}{2}\frac{1 - \BinaryEntropy(p_0) - \log \frac{\YSize-1}{\YSize}}{\tau(p_0)^2} \geq 0$.
Otherwise, the tightness \cref{appendix:eq:tightness_on_E} always holds.

We proceed by specifying $p_0$.
Firstly, suppose $p_0$ is mildly small: $p_0 \leq \frac{\YSize-1}{2\YSize - 1}$.
Then, we can show $\tau(p_0) \geq 0$.
Thus, $\DeltaMinus \leq \DeltaPlus$ holds, and \cref{appendix:eq:tight_range_p_upper} becomes:
\begin{gatherSmall}
\PBound(E) \leq p_0 + \DeltaMinus. \label{appendix:eq:tight_range_p_upper_simple}
\end{gatherSmall}
Additionally, we can show that $\DeltaMinus \geq 0$.
Thus, \cref{appendix:eq:tight_range_p_upper_simple} discloses that if the lower bound $\PBound(E)$ is not much larger than $p_0$, the tightness \cref{appendix:eq:tightness_on_E} holds.
Especially, if $\PBound(E) \leq p_0$ the tightness holds.
This condition applies to the experiments of this study.
We have also directly shown that $\BoundFuncTightP(\StrengthDoublet) > \BoundFuncLoose(\StrengthDoublet)$ in \Cref{tb:ablation_errors}.

If $p_0$ is large $p_0 \geq \frac{\YSize-1}{2\YSize - 1}$, we can show $\tau(p_0) \leq 0$.
Thus, $\DeltaMinus \geq \DeltaPlus$ holds, and \cref{appendix:eq:tight_range_p_lower} becomes:
\begin{gatherSmall}
\PBound(E) \geq p_0 + \DeltaMinus. \label{appendix:eq:tight_range_p_lower_simple}
\end{gatherSmall}
Additionally, we can show that $\DeltaMinus \leq 0$.
Thus, \cref{appendix:eq:tight_range_p_lower_simple} discloses that if the lower bound $\PBound(E)$ is not much smaller than $p_0$, the tightness \cref{appendix:eq:tightness_on_E} holds.

\clearpage

\section{Details of Experimental Setup}   \label{appendix:setup}

\subsection{Models} \label{appendix:DNNs}

\begin{table*}[ht!]
    \centering
    \small
    \tabcolsep 0.9pt
    \caption{DNNs used in study and their error rates for each task. Convention of ``variant'' follows Huggingface's transformer library \cite{wolf-et-al-2019-huggingface}. \textbf{Bold} shows best model in each task, which is used as baseline $s_0$ stated in \Cref{sec:ensemble_systems}. \label{appendix:tb:PTDNNs_in_detail}}

\begin{tabular}{@{}lllllllllll@{}}

\toprule
                            DNN type &                                       variant &                       avg. &                     Boolq &                      CoLA &                  CosmosQA &                      MNLI &                      MRPC &                       QQP &                  SciTail &                       SST \\

\midrule
RoBERTa \cite{liu-et-al-2019-roberta} & base & $\mathbf{15.5}_{\pm \mbox{\tiny 0.3}}$ &   $24.1_{\pm \mbox{\tiny 0.6}}$ &  $\mathbf{15.6}_{\pm \mbox{\tiny 0.2}}$ &  $\mathbf{28.2}_{\pm \mbox{\tiny 0.5}}$ &  $\mathbf{18.7}_{\pm \mbox{\tiny 1.2}}$ &  $13.6_{\pm \mbox{\tiny 0.5}}$ &   $\mathbf{14.1}_{\pm \mbox{\tiny 0.8}}$ &  $\mathbf{4.2}_{\pm \mbox{\tiny 0.2}}$ &  $5.8_{\pm \mbox{\tiny 0.7}}$ \\
ELECTRA \cite{clark2020electra} &  base-discriminator  & $17.3_{\pm \mbox{\tiny 0.3}}$ &  $\mathbf{23.1}_{\pm \mbox{\tiny 1.3}}$ &  $17.0_{\pm \mbox{\tiny 0.5}}$ &  $29.8_{\pm \mbox{\tiny 0.7}}$ &  $22.6_{\pm \mbox{\tiny 1.0}}$ &  $\mathbf{13.3}_{\pm \mbox{\tiny 0.7}}$ &   $18.5_{\pm \mbox{\tiny 0.9}}$ &  $7.1_{\pm \mbox{\tiny 0.2}}$ &  $\mathbf{5.7}_{\pm \mbox{\tiny 0.5}}$ \\
 BART \cite{lewis2019bart} & base & $17.9_{\pm \mbox{\tiny 0.2}}$ &  $25.5_{\pm \mbox{\tiny 1.3}}$ &  $20.9_{\pm \mbox{\tiny 0.5}}$ &  $30.1_{\pm \mbox{\tiny 0.5}}$ &   $22.3_{\pm \mbox{\tiny 0.8}}$ &  $15.8_{\pm \mbox{\tiny 0.7}}$ &  $15.9_{\pm \mbox{\tiny 1.2}}$ &  $4.7_{\pm \mbox{\tiny 0.3}}$ &  $8.3_{\pm \mbox{\tiny 0.5}}$ \\
 BERT \cite{devlin-etal-2019-bert} & base-uncased & $18.7_{\pm \mbox{\tiny 0.1}}$ &   $26.0_{\pm \mbox{\tiny 0.8}}$ &  $17.2_{\pm \mbox{\tiny 0.4}}$ &  $34.1_{\pm \mbox{\tiny 0.6}}$ &   $26.3_{\pm \mbox{\tiny 0.2}}$ &  $17.1_{\pm \mbox{\tiny 0.7}}$ &   $16.5_{\pm \mbox{\tiny 0.7}}$ &  $4.5_{\pm \mbox{\tiny 0.2}}$ &  $8.0_{\pm \mbox{\tiny 0.7}}$ \\
 ALBERT \cite{lan2019albert} & base-v1 & $20.4_{\pm \mbox{\tiny 0.1}}$ &  $25.3_{\pm \mbox{\tiny 2.2}}$ &  $19.5_{\pm \mbox{\tiny 0.2}}$ &  $43.2_{\pm \mbox{\tiny 0.1}}$ &   $27.1_{\pm \mbox{\tiny 0.3}}$ &  $14.5_{\pm \mbox{\tiny 0.4}}$ &   $18.8_{\pm \mbox{\tiny 0.8}}$ &  $4.9_{\pm \mbox{\tiny 0.1}}$ &  $9.6_{\pm \mbox{\tiny 0.3}}$ \\
\bottomrule

\end{tabular}
\end{table*}

\begin{table*}[h!]
    \centering
    \small
    \caption{Hyperparameters used for fine-tuning of DNNs. \label{appendix:tb:PTDNN_hyperparameters}}
    \begin{tabularx}{0.75\linewidth}{lX}
    \toprule
    hyperparameter & value \\
    \midrule
    learning rate & 3e-5 \ ([1e-5, 1e-4] for the random sampling of Random-HyP) \\
    optimizer & Adam \cite{kingma2014adam} ($\epsilon=1e-8$) with linear warmup (data size proportion=0.1), described in \cite{devlin-etal-2019-bert}. \\
    gradient clipping & 1.0 \\
    gradient accumulation steps & 1 \\
    epochs & 5 \\
    dropout & DNN specific values (follows jiant \cite{phang2020jiant}) \\
    training batch size & 16 \\
    inference batch size & 32 \\
    number of softmax layer & 1 \\
    \bottomrule
    \end{tabularx}
\end{table*}

\begin{table*}[h!]
    \centering
    \small
    \caption{
    Meta-estimator hyperparameters.
    Hyperparameter names follow scikit-learn.
    Most of the hyperparameters are set as default values of scikit-learn (version 0.22.2). \label{appendix:tb:PTDNN_hyperparameters_of_meta_estimator}}
    \begin{tabularx}{0.75\linewidth}{llX}
    \toprule
    meta-estimator & hyperparameter & value / search range \\
    \midrule
    logistic regression & C & [1e-2, 3e-2, 1e-1, 3e-1, 1e0] \\
    & penalty & L2 \\
    & solver & liblinear \\
    & max\_iter & 1000 \\
    & multi\_class & auto \\
    & random\_state & 0 \\
    
    SVM & C & [1e-2, 3e-2, 1e-1, 3e-1, 1e0] \\
    & max\_iter & -1 \\
    & decision\_function\_shape & ovr \\
    & random\_state & 0 \\

    Random Forest & ccp\_alpha & [0.0, 0.03, 0.1, 0.3] \\
    & random\_state & 0 \\
    & criterion & gini \\
    & max\_depth & None \\

    \bottomrule
    \end{tabularx}
\end{table*}

\paragraph{DNN types:}
\Cref{appendix:tb:PTDNNs_in_detail} shows the five types of pre-trained language models used in this study. 
Pre-trained language models are essentially large neural networks with self-attention layers that are trained on huge text corpora in an unsupervised manner.
These models are shown to obtain state-of-the-art performance when fine-tuned on downstream tasks \cite{liu-et-al-2019-roberta, lan2019albert, clark2020electra, devlin-etal-2019-bert, lewis2019bart}.
In addition, since they differ in terms of model architecture and pre-training method, they should produce strong diversity, and hence, are suitable for ensembles.

\paragraph{Fine-tuning procedures:}
We trained each DNN on each downstream task following the standard practice of language model fine-tuning (see \citet{devlin-etal-2019-bert} for example) as follows.

We added a new softmax layer on top of the embedding layers of the DNNs.
We preprocessed the input text by the following steps: (i) we tokenized the input text with a DNN-type-specific tokenizer, (ii) if the text included more than two sentences, we added DNN-type-specific ``separator'' tokens between sentences, (iii) we tensorized each token into a one-hot vector using DNN-type-specific vocabulary.

We trained these models on the training sets of the tasks.
TValidation sets were used only during the preliminary experiments to adjust some hyperparameters (shown below).
Please refer to \Cref{appendix:dataset_splitting} for the details of the datset splitting strategy.

We used the hyperparameters shown in \Cref{appendix:tb:PTDNN_hyperparameters} to fine-tune all of the DNN types.
The values were chosen on the basis of the original papers \cite{liu-et-al-2019-roberta, lan2019albert, clark2020electra, devlin-etal-2019-bert, lewis2019bart} and our preliminary experiments.
Note that language models require only a few epochs for convergence.

Some of the ensemble methods in \Cref{tb:ensemble_methods} use different seeds for fine-tuning to produce diverse DNN models.
In our study, seeds affect (i) the initial weights of the softmax layer, (ii) the hidden units dropped by dropout, and (iii) the shuffling order of the training instances.

\paragraph{Implementations:}
We implemented the fine-tuning of DNNs described here using the jiant library \cite{phang2020jiant} (v2.2.0\footnote{github hash: 961bd577f736449956ddb2c15dcfce68bbb75e59}), which in turn utilizes Hugging Face’s Transformers library \cite{wolf-et-al-2019-huggingface}.
Jiant enables us to fine-tune various types of pre-trained language models on various NLP tasks.
See our code for details.

\subsection{Ensemble Systems}  \label{appendix:ensemble_systems}
For the random hyperparameter sampling of Random-HyP, we sampled the fine-tuning learning rate since it affect the resulting model the most.
We sampled the learning rate around the best value of 3e-5, i.e., from [1e-5, 1e-4], as shown in \Cref{appendix:tb:PTDNN_hyperparameters}.

The baseline system $s_0$ was single DNN (i.e. no-ensemble) that performed the best among DNNs.
These baselines are shown as bold in \Cref{appendix:tb:PTDNNs_in_detail}

We implemented the model generation methods in \Cref{tb:ensemble_methods} by ourselves.

We implemented the model combination methods in \Cref{tb:ensemble_methods} using scikit-learn \footnote{\url{https://scikit-learn.org/stable/}}.
For the training of Stacking meta-estimators, we used the hyperparameters shown in \Cref{appendix:tb:PTDNN_hyperparameters_of_meta_estimator}.
We tuned some of the hyperparameters using scikit-learn's GridSearchCV with 5-fold cross validation.
\Cref{appendix:stacking} gives other details on Stacking ensemble used in this study.

\subsection{Estimation of metric values and lower bound}   \label{appendix:mti}

\noindent \textbf{Trick of MTI}

In our experiments, we estimated the three metric values on the basis of the frequency distribution observed for the datasets.
We used the trick of MTI introduced by \cite{zhou_mti}, which approximates quantities appearing in the three metrics which depend on high-dimensional stochastic variables $\mathbf{O}$.
Please refer to \cite{zhou_mti} for more details.

We repeat the three terms of \Cref{lemma:ensemble_bound_tight} below:
\begin{gatherSmall}
    \RelevWithArgs = \sum_{i=1}^{N}I(O_i, Y), \nonumber \\
    \RedunWithArgs =  \multiInfo(\mathbf{O}) - \multiInfo(\mathbf{O}|Y), \nonumber \\
    \ComblossWithArgs =  H(Y|\hat{Y}) - H(Y|\mathbf{O}). \nonumber
\end{gatherSmall}
Looking at above, it can be seen that some terms (i.e. $\multiInfo(\mathbf{O})$, $\multiInfo(\mathbf{O}|Y)$ and $H(Y|\mathbf{O})$) depend on high-dimensional variable $\mathbf{O} = \{O_1,\dots, O_N\}$, where $N$ is the number of models.
Since $N$ can be as large as 30 in our experiments, these terms might not be estimated reliably due to the count sparsity for the limited amount of dataset instances.

Thus, we use the trick of MTI introduced by \citet{zhou_mti}, which approximates the quantities by replacing $\mathbf{O}$ with its smaller subset $\Omega$ as follows:
\begin{gatherSmall}
    \multiInfo(\mathbf{O}) = \sum_{i=1}^{N} I(O_i;O_{1:i-1}) \geq \sum_{i=1}^{N} \max_{\Omega^{i-1}_k}I(O_i|\Omega^{i-1}_k), \label{appendix:eq:mti_redun} \\
    \multiInfo(\mathbf{O}|Y) = \sum_{i=1}^{N} I(O_i;O_{1:i-1}|Y) \geq \sum_{i=1}^{N} \max_{\Omega^{i-1}_k} I(O_i;\Omega^{i-1}_k| Y), \label{appendix:eq:mti_conditional_redun}\\
    H(Y|\mathbf{O}) \leq \min_{\Omega^{N}_{k}} H(Y|\Omega^{N}_k), \label{appendix:eq:mti_entropy}
\end{gatherSmall}
where $\Omega^{i}=\{X_1,\dots, X_{i}\}$, and $\Omega^{i-1}_k$ is a subset of size k.

The first equalities of \cref{appendix:eq:mti_redun,appendix:eq:mti_conditional_redun} were proved by \cite{zhou_mti}.
The last inequality in each equation is understood as follows.
By replacing $\mathbf{O}$ with its subset $\Omega^{i}_{k}$, we lose some amount of information carried by $\mathbf{O}$.
Thus, this transformation might make mutual information in \cref{appendix:eq:mti_redun,appendix:eq:mti_conditional_redun} smaller than the original value and the entropy in \cref{appendix:eq:mti_entropy} larger.
However, if we find $\Omega^{i}_{k}$, which contains the largest amount of information (corresponding to $\max$ and $\min$ operations in each equation), the difference from the original value (i.e,. approximation error) is the smallest.

\citet{zhou_mti} empirically showed that the method works well to produce almost an exact value.
In our experiments, we used $k=3$ ($\mathrm{MTI}_{k=3}$).

\noindent \textbf{On the choice of $p_0$}

We set the approximate error rate $p_0$ in \cref{eq:ensemble_bound_tight} as the error rate of the baseline $s_0$ defined in \Cref{sec:ensemble_systems}.
We state the reason for this in \Cref{appendix:choice_of_p0}.

\subsection{Tasks} \label{appendix:tasks}

\begin{table*}[h!]
    \centering
    \small
    \tabcolsep 4.0pt
    \caption{Tasks used in this study.Mmajority of tasks are from GLUE benchmark \cite{wang-etal-2018-glue} (shown as $\ast$) and SuperGLUE benchmark \cite{NEURIPS2019_4496bf24} (shown as $\star$). All datasets are publicly available. \label{appendix:tb:tasks}}
    \begin{tabularx}{1.0\linewidth}{p{60mm}Xp{8mm}p{80mm}}
    \toprule
    task & dataset size & \# classes ($\YSize$) & description\\
    \midrule
    Boolq$\star$ \ (Boolean Question) \cite{clark2019boolq} & 9.5k & 2 & We are required to choose yes or no about a given question on a given passage. The questions are the ones naturally occurring in Google search engine, rather than the ones artificially built. Answering the questions often requires query for complex, non-factoid information, and difficult entailment-like inference. \\
    \\
    CoLA$\ast$ \ (Corpus of Linguistic Acceptability) \cite{dolan2005automatically} & 8.5k & 2      & We are required to judge linguistic acceptability (i.e., grammatical or non grammatical) of given text such as ``What did Bill buy potatoes?''. The text are drawn from books and journal articles on linguistic theory. Answering the questions requires the rich grammatical knowledge from the local word dependencies such as subject-verb-object order to the non-local dependencies. \\
    \\
    Cosmos QA \ \cite{khot2018scitail}    & 25k     & 4       & After reading a short narrative passage, we are required to answer a question about the passage (such as ``What's a possible reason the writer needed someone to dress him every morning?'') by choosing one answer from four possible candidates. The passages are taken from blogs on the web and personal narratives. Understanding the narrative requires common sense such as inference on causes and effects of events, even when they are not mentioned explicitly in the texts.   \\
    \\
    MNLI$\ast$ \ (Multi-Genre Natural Language Inference) \cite{MNLI}    & 400k (10k used)     & 3       & Given two pieces of text, we answer the relationship of the one piece to the other piece from three choices: ``entails'', ``neutral'', ``contradicts''. The dataset is composed of texts from various distinct genres of written English. The pairs are such as ``At 8:34, the Boston Center controller received a third transmission from American 11'' and ``The Boston Center controller got a third transmission from American 11.'' Answering the question requires total ability of natural language understanding, e.g., handling lexical entailment, quantification, coreference, tense, belief, modality, and lexical and syntactic ambiguity.\\
    \\
    MRPC$\ast$ \  (Microsoft Research Paraphrase Corpus) \cite{dolan2005automatically} & 3k  & 2     & We are required to judge whether given two sentences are semantically equivalent. The sentences are automatically extracted from online news sources and twitter. Pairs are such as: ``Charles O. Prince, 53, was named as Mr. Weill’s successor.'' ``Mr. Weill’s longtime confidant, Charles O. Prince, 53, was named as his successor.''. Recognizing such paraphrase is a fundamental skill needed for various tasks in NLP.\\
    \\
    QQP$\ast$ \ (Quora Question Pairs) \footnote{https://www.quora.com/profile/Ricky-Riche-2/First-Quora-Dataset-Release-Question-Pairs}   & 300k (10k used)         & 2   & We are required to determine whether a pair of questions are semantically equivalent. The questions are taken from the social Q\&A website Quora. The skill is used by question-answering system to recognize the semantically same questions of different linguistic expressions. \\
    \\
    SciTail \ \cite{khot2018scitail}    & 23k         & 4   & We are required to answer a given scientific question such as ``Which of the following best explains how stems transport water to other parts of the plant?'' by choosing one answer from four candidates. We have access to the additional relevant text. The questions are the ones naturally arising in the web rather than ones artificially created. \\
    \\
    SST$\ast$ \ (Stanford Sentiment Treebank) \cite{socher-etal-2013-recursive}  & 50k (10k used)         & 2   & We predict a sentiment label (i.e., positive or negative) of a given sentence. The sentences are taken from movie reviews. The task requires the understanding of compositoinality of langeuage. \\
    \bottomrule
    \end{tabularx}
\end{table*}

\Cref{appendix:tb:tasks} details the \NumTasks tasks used in this study.

\clearpage

\section{Stacking Ensemble}  \label{appendix:stacking}

\begin{figure}[h!]
\centering
\small
    \centering
    \includegraphics[width=0.5\linewidth]{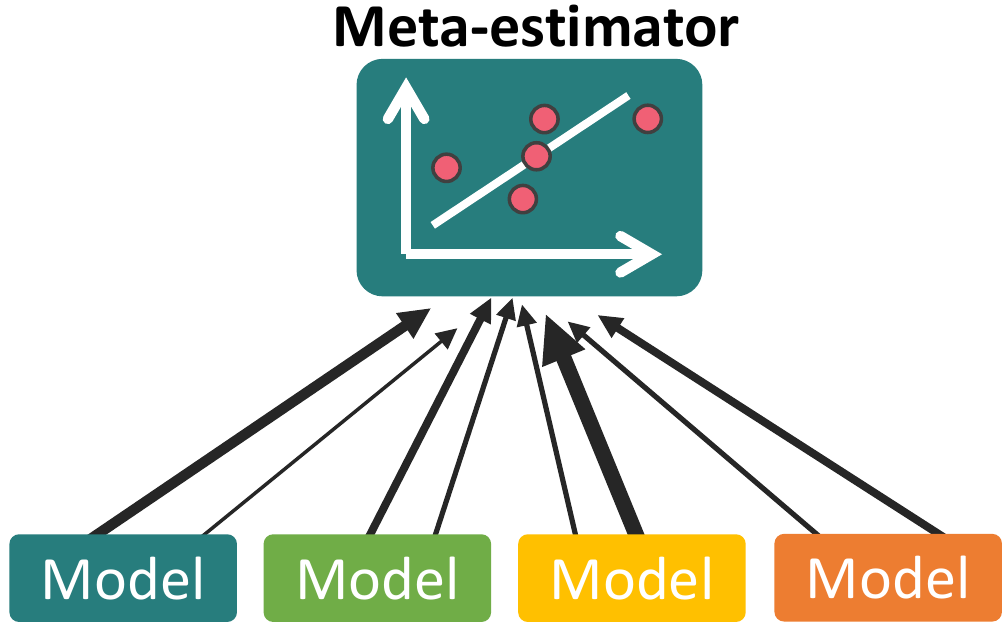}
    \caption{Stacking ensemble used in this study.\label{appendix:fig:stacking}}
\end{figure}

Here, we give destails of the stacking ensemble used in this study.

\subsection{Architecture:}  \label{appendix:stacking_architecture}
\Cref{appendix:fig:stacking} illustrates the stacking ensemble used in this study.
We used the two-layered stacking ensemble where the first-layer models are fine-tuned DNNs, and the second-layer model (i.e. the meta-estimator) is another classification model.
For the meta-estimator, we used logistic regression, Support Vector Machine \cite{Platt99probabilisticoutputs} with RBF kernel, and Random Forest \cite{breiman2001random}.
For the inputs of the meta-estimator, we used class labels predicted by the models.

In the below, we show the details of the logistic regression meta-estimator case.
The meta-estimator estimates the probability for a given instance $i$ belonging to class c $p_{i, c} \in [0, 1]$ from class labels predicted by $N$ models $\mathbf{\hat{y}_i} = \{\hat{y}^1_i, \hat{y}^2_i, \dots, \hat{y}^N_i\}, \ \hat{y}^n_i \in \{0, 1\}$ as:
\begin{alignSmall}
    p_{i, c} = \frac{1}{1 + \exp (- l_{i, c})}, \nonumber \\
    l_{i, c} =  w^0_c + \sum_{m=1}^N w^m_c \hat{y}^m_i. \nonumber
\end{alignSmall}
The class with the largest $p_{i, c}$ is chosen as the final answer.
The meta-estimator is trained using ``meta-feature dataset'' $D_{\text{meta}} = \{(\mathbf{\hat{y}}_1, y_1), (\mathbf{\hat{y}}_2, y_2), \dots, (\mathbf{\hat{y}}_{|D|}, y_{|D|})\}$, where $y_i$ denotes the groundtruth label. 
Details of the meta-estimator training are shown in \ref{appendix:ensemble_systems} and \ref{appendix:dataset_splitting}.

\subsection{Weight Distribution of \Cref{tb:base_model_type_performances}} \label{appendix:stacking_weight_distribution}
The DNN-type-wise weight sum mentioned in \Cref{tb:base_model_type_performances} is calculated as follows:
\begin{gatherSmall}
W_{t} = \sum _{m \in M_t}|w^{m}_{c=1}|, \nonumber
\end{gatherSmall}
where $m$ denote the index of model, $t$ a specific DNN type and $M_t$ the set of indexes of models from DNN type $t$.
Note that since our study used binary classification tasks, it suffices to look $c=1$.

\subsection{Meta-estimator training}

\paragraph{Hyperparameters}
The hyperparameters of the meta-estimators (i.e., logistic regression and the SVM used by Stacking ensemble) are shown in \Cref{appendix:tb:PTDNN_hyperparameters_of_meta_estimator}.

\paragraph{Implementation:}
We implemented the model combination methods in \Cref{tb:ensemble_methods} using scikit-learn \footnote{\url{https://scikit-learn.org/stable/}}.

\subsection{Dataset splitting}  \label{appendix:dataset_splitting}

\begin{figure*}[t!]
    \centering
    \includegraphics[width=1.0\linewidth]{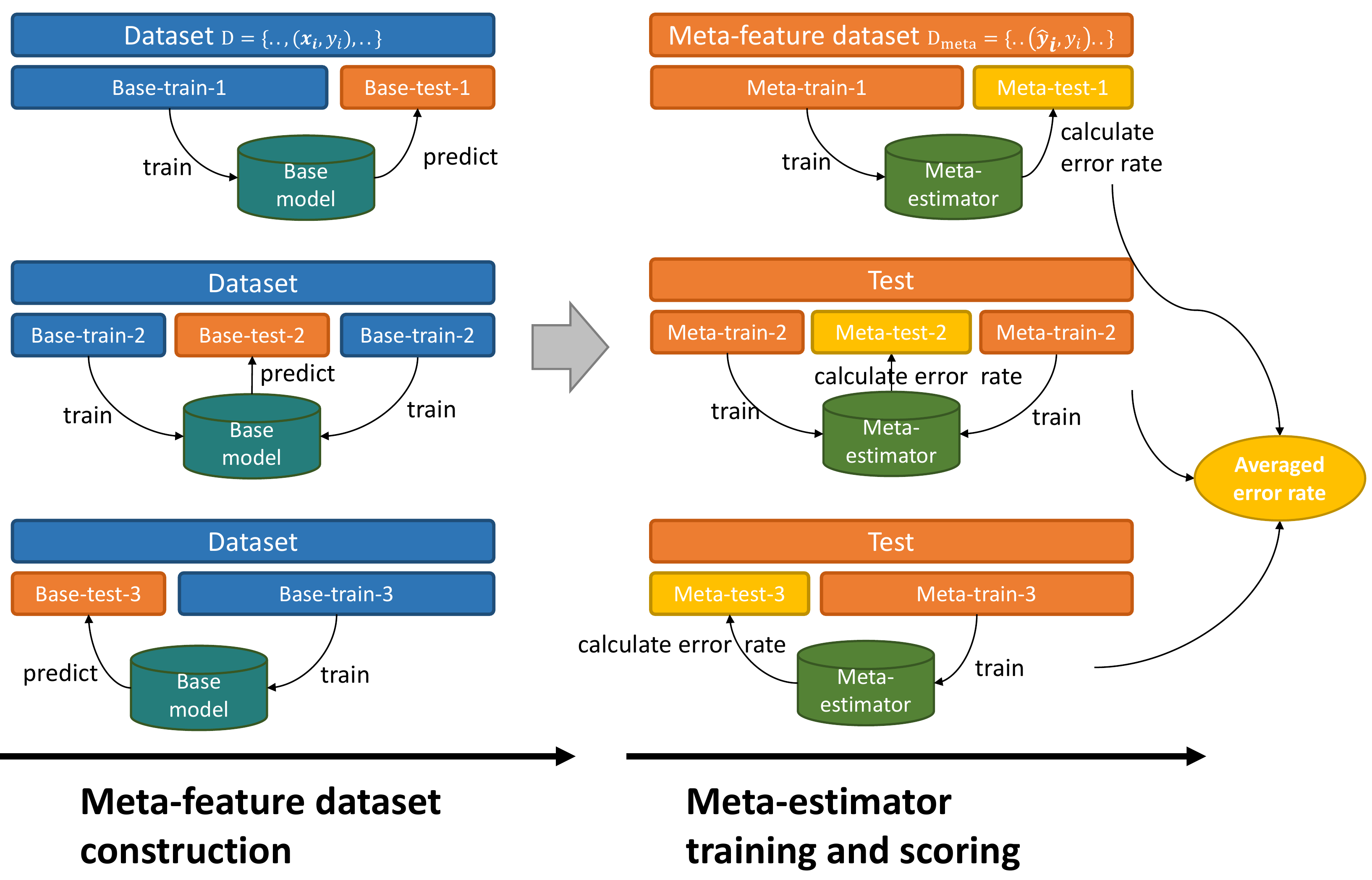}
    \caption{Our dataset splitting strategy (\Cref{appendix:dataset_splitting}) with 3-split case. \label{appendix:fig:dataset_splitting}}
\end{figure*}

In order to train meta-estimator of Stacking, we must take cross-validation based dataset splitting strategy \cite{wolpert-1992-stackedgeneralization}.
In the below, we describe the data splitting strategy, which is illustrated in \Cref{appendix:fig:dataset_splitting}.
Note that the same data splitting strategy was used for voting-based systems for fair comparisons.

Training of stacking meta-estimators requires ``meta-feature dataset'' $D_{\text{meta}} = \{(\mathbf{\hat{y}}_1, y_1), (\mathbf{\hat{y}}_2, y_2), \dots, \mathbf{\hat{y}}_{|D|}, y_{|D|})\}$., as stated in \Cref{appendix:stacking}.
Here, $\mathbf{\hat{y}_i} = \{\hat{y}^1_i, \hat{y}^2_i, \dots, \hat{y}^N_i\}$ where $\ \hat{y}^m_i \in \{0, 1\}$ denotes the label predicted by model $m$ on instance $i$.
$y_i$ denotes the groundtruth label of the same instance $i$.
To prevent overfitting of meta-estimators, the model predictions $\{\mathbf{\hat{y}}_1, \mathbf{\hat{y}}_2, \dots\}$ must be \textit{label-leak-free}.
Thus, the model predictions are usually obtained using n-fold cross-validation as follows.

\paragraph{Meta-feature dataset construction}

For each model $m$, we use $n$-fold cross-validation to obtain its label-leak-free predictions.
Specifically:
\begin{enumerate}
    \item Choose model $m$.
    \item Divide the dataset $D = \{(\mathbf{x}_1, y_1), \dots, (\mathbf{x}_{|D|}, y_{|D|})\}$ into $n$ sets.
    \item One of them (i.e. base-test-$i$) is set aside for testing later.
    \item Train the model $m$ on the rest sets (i.e. base-train-$i$).
    \item Apply the trained model $m$ to the test set (i.e. base-test-$i$) to get label-leak-free predictions.
    \item Repeat 3-5 for $i$ to collect label-leak-free predictions on whole the dataset $\{\hat{y}^m_1, \hat{y}^m_2, \dots, \hat{y}^m_{|D|}\}$ where $\hat{y}^m_i$ denotes a label prediction by model $m$ on the instance $i$, as stated in  \ref{appendix:stacking}.
    \item Repeat 1-6 for $m$ to collect label-leak-free predictions by all the models: $\{\mathbf{\hat{y}}_1, \mathbf{\hat{y}}_2, \dots, \mathbf{\hat{y}}_{|D|}\}$. Then, we concatenate the predictions   $D_{\normalfont \text{meta}}$.
    Then, we merge the predicted labels $\{\mathbf{\hat{y}}_1, \mathbf{\hat{y}}_2, \dots, \mathbf{\hat{y}}_{|D|}\}$ and the groundtruth labels $\{y_1, \dots, y_{|D|}\}$ into the meta-feature dataset $D_{\text{meta}} = \{(\mathbf{\hat{y}}_1, y_1), (\mathbf{\hat{y}}_2, y_2), \dots, (\mathbf{\hat{y}}_{|D|}, y_{|D|})\}$.
    
\end{enumerate}
Note that, since the test set (i.e. base-test-$i$) is never used by model training, the predictions on the test set are label-leak-free.

In this study, we used $n=5$.

\paragraph{Meta-estimator training and scoring}:

Some of the datasets used in this study are small, as shown in \Cref{appendix:tb:tasks}.
The official test-sets of the datasets are also small.
For example, the test sets of RTE dataset includes only 277 instances.
We supposed that the performance measurements conducted on such small test-sets might not be so reliable.
Thus, we conducted the following $l$-fold cross-validation to train and score the meta-estimators:
\begin{enumerate}
    \item Divide $D_{\mathrm{meta}}$ into $l$ sets.
    \item One of them (i.e. meta-test-$i$) is set aside for testing.
    \item Train a meta-estimator on the rest sets (i.e. meta-train-$i$).
    \item Apply the meta-estimator to the test sets (i.e. meta-test-$i$) and calculate its error rate.
    \item Repeat 2-4 for $i$ to get error rates on the test-sets, then calculate the average of them.
\end{enumerate}
In this study, we used $l=4$.

\clearpage

\section{Pearson Correlation Coefficients between Error Rate Reductions and Lower Bound Reductions for Various Number of Models $N$}   \label{appendix:correlation_coefficients}

\crefrange{appendix:tb:Pearson_10}{appendix:tb:Pearson_30} show the Pearson correlation coefficients between the error reductions and lower bound reductions of the ensemble systems in each task.
Each table shows the results of different $N$, which is the number of models used by the ensemble systems.

See \Cref{sec:effectiveness_of_bound} for the discussion of such correlations.

\begin{table}[h!]
    \centering
    \caption{
    $N=10$.
    Pearson correlation coefficients between error rate reduction and lower bound reduction.
    In each task we used the \NumSystems ensemble systems described in \Cref{sec:ensemble_systems}.
    \label{appendix:tb:Pearson_10}
    }
    \resizebox{\linewidth}{!}{%
    \begin{tabular}{@{}lrrr@{}}
        \toprule
        
        & \multicolumn{3}{c}{Lower bound type} \\

        \cmidrule(l){2-4}
        Task &\Cref{lemma:ensemble_bound_loose}~$\BoundLoose$& $\BoundTightWOCombLoss$ &\textbf{\Cref{lemma:ensemble_bound_tight}}~$\BoundTight$\\
        
        \midrule

Boolq        &   0.413        &   0.377        &   \textbf{0.869}    \\
CoLA         &   -0.259       &   -0.245       &   \textbf{0.993}    \\
CosmosQA     &   -0.188       &   -0.174       &   \textbf{1.000}    \\
MNLI         &   -0.275       &   -0.385       &   \textbf{0.955}    \\
MRPC         &   0.218        &   0.218        &   \textbf{0.983}    \\
QQP          &   -0.359       &   -0.330       &   \textbf{0.999}    \\
SciTail      &   -0.076       &   -0.092       &   \textbf{0.944}    \\
SST          &   0.286        &   0.357        &   \textbf{0.998}    \\

\midrule
average\tablefootnote{The correlation coefficient between the averaged error rate reductions and lower bound reductions. The average is taken over the \NumTasks tasks.}    &   -0.482       &   -0.431       &   \textbf{0.975}    \\

    \bottomrule
    \end{tabular}
}%

\end{table}

\begin{table}[h!]
    \centering
    \caption{
    $N=15$.
    Pearson correlation coefficients between error rate reduction and lower bound reduction.
    In each task we used the \NumSystems ensemble systems described in \Cref{sec:ensemble_systems}.
    \label{appendix:tb:Pearson_15}
    }
    \resizebox{\linewidth}{!}{%
    \begin{tabular}{@{}lrrr@{}}
        \toprule
        
        & \multicolumn{3}{c}{Lower bound type} \\

        \cmidrule(l){2-4}
        Task &\Cref{lemma:ensemble_bound_loose}~$\BoundLoose$& $\BoundTightWOCombLoss$ &\textbf{\Cref{lemma:ensemble_bound_tight}}~$\BoundTight$\\
        
        \midrule

Boolq        &   0.341        &   0.330        &   \textbf{0.910}    \\
CoLA         &   -0.211       &   -0.210       &   \textbf{0.991}    \\
CosmosQA     &   -0.324       &   -0.320       &   \textbf{1.000}    \\
MNLI         &   0.226        &   0.216        &   \textbf{0.961}    \\
MRPC         &   0.332        &   0.252        &   \textbf{0.989}    \\
QQP          &   -0.131       &   -0.076       &   \textbf{0.998}    \\
SciTail      &   -0.237       &   -0.191       &   \textbf{0.966}    \\
SST          &   -0.242       &   -0.252       &   \textbf{0.998}    \\

\midrule
average\footnotemark[10]    &   -0.238       &   -0.165       &   \textbf{0.984}    \\

    \bottomrule
    \end{tabular}
}%

\end{table}

\begin{table}[h!]
    \centering
    \caption{
    $N=20$.
    Pearson correlation coefficients between error rate reduction and lower bound reduction.
    In each task we used the \NumSystems ensemble systems described in \Cref{sec:ensemble_systems}.
    \label{appendix:tb:Pearson_20}
    }
    \resizebox{\linewidth}{!}{%
    \begin{tabular}{@{}lrrr@{}}
        \toprule
        
        & \multicolumn{3}{c}{Lower bound type} \\

        \cmidrule(l){2-4}
        Task &\Cref{lemma:ensemble_bound_loose}~$\BoundLoose$& $\BoundTightWOCombLoss$ &\textbf{\Cref{lemma:ensemble_bound_tight}}~$\BoundTight$\\
        
        \midrule

Boolq        &   0.323        &   0.311        &   \textbf{0.915}    \\
CoLA         &   -0.324       &   -0.320       &   \textbf{0.995}    \\
CosmosQA     &   -0.510       &   -0.512       &   \textbf{1.000}    \\
MNLI         &   -0.190       &   -0.192       &   \textbf{0.976}    \\
MRPC         &   -0.235       &   -0.199       &   \textbf{0.964}    \\
QQP          &   0.411        &   0.390        &   \textbf{0.999}    \\
SciTail      &   -0.286       &   -0.307       &   \textbf{0.958}    \\
SST          &   0.032        &   0.024        &   \textbf{0.997}    \\

\midrule
average\footnotemark[10]     &   -0.452       &   -0.425       &   \textbf{0.985}    \\

    \bottomrule
    \end{tabular}
}%

\end{table}

\begin{table}[h!]
    \centering
    \caption{
    $N=30$.
    Pearson correlation coefficients between error rate reduction and lower bound reduction.
    In each task we used the \NumSystems ensemble systems described in \Cref{sec:ensemble_systems}.
    \label{appendix:tb:Pearson_30}
    }
    \resizebox{\linewidth}{!}{%
    \begin{tabular}{@{}lrrr@{}}
        \toprule
        
        & \multicolumn{3}{c}{Lower bound type} \\

        \cmidrule(l){2-4}
        Task &\Cref{lemma:ensemble_bound_loose}~$\BoundLoose$& $\BoundTightWOCombLoss$ &\textbf{\Cref{lemma:ensemble_bound_tight}}~$\BoundTight$\\
        
        \midrule

Boolq        &   0.158        &   0.146        &   \textbf{0.940}    \\
CoLA         &   -0.215       &   -0.213       &   \textbf{0.994}    \\
CosmosQA     &   -0.592       &   -0.588       &   \textbf{1.000}    \\
MNLI         &   -0.048       &   -0.050       &   \textbf{0.976}    \\
MRPC         &   -0.471       &   -0.498       &   \textbf{0.974}    \\
QQP          &   0.187        &   0.231        &   \textbf{0.999}    \\
SciTail      &   -0.379       &   -0.377       &   \textbf{0.954}    \\
SST          &   0.213        &   0.208        &   \textbf{0.996}    \\

\midrule
average\footnotemark[10]     &   -0.330       &   -0.288       &   \textbf{0.990}    \\

    \bottomrule
    \end{tabular}
}%

\end{table}

\clearpage

\begin{figure*}[h!]
    \begin{subfigure}[t]{0.19\linewidth}
        \vskip 0pt
        \includegraphics[width=\linewidth]{figures/jupyter/jiant/average/scaling.stacking.with_legend.pdf}
        \subcaption{Error rate reduction.\label{appendix:fig:scaling.stacking}}
    \end{subfigure}
    \hfill
    \begin{subfigure}[t]{0.19\linewidth}
        \vskip 0pt
        \includegraphics[width=\linewidth]{figures/jupyter/jiant/average/scaling.bound.voting.previous_research.reduction.with_legend.pdf}
        \subcaption{Lower bound reduction by \Cref{lemma:ensemble_bound_loose} $\BoundLoose$.\label{appendix:fig:scaling.bound.voting.previous_research}}
    \end{subfigure}
    \begin{subfigure}[t]{0.19\linewidth}
        \vskip 0pt
        \includegraphics[width=\linewidth]{figures/jupyter/jiant/average/scaling.bound.voting.ours_wo_combloss.reduction.with_legend.pdf}
        \subcaption{Lower bound reduction by $\BoundTightWOCombLoss$. \label{appendix:fig:scaling.bound.voting.ours_wo_combloss}}
    \end{subfigure}
    \hfill
     \begin{subfigure}[t]{0.19\linewidth}
        \vskip 0pt
        \includegraphics[width=\linewidth]{figures/jupyter/jiant/average/scaling.stacking.bound.reduction.with_legend.pdf}
        \subcaption{Lower bound reduction by \textbf{\Cref{lemma:ensemble_bound_tight}} $\BoundTight$.\label{appendix:fig:scaling.stacking.bound}}
    \end{subfigure}
    \hfill
    \begin{subfigure}[t]{0.19\linewidth}
        \vskip 0pt
        \includegraphics[width=\linewidth]{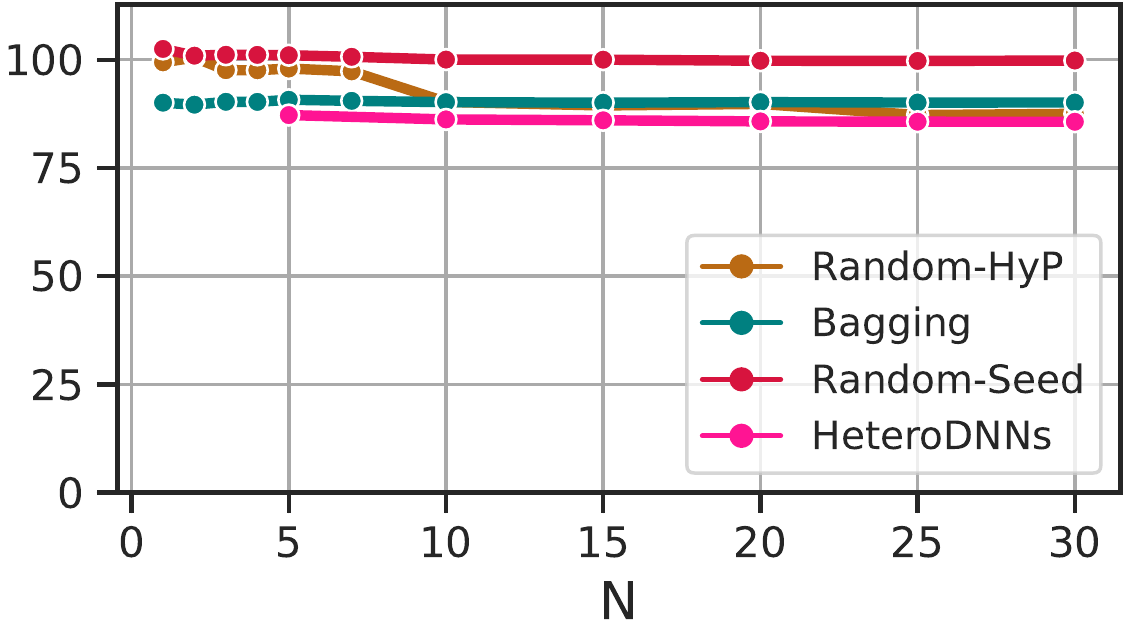}
        \subcaption{$\relev$ \label{appendix:fig:scaling.relevance.per_model}}   
    \end{subfigure}   
    \hfill
    \vfill
    \begin{subfigure}[t]{0.19\linewidth}
        \vskip 0pt
        \includegraphics[width=\linewidth]{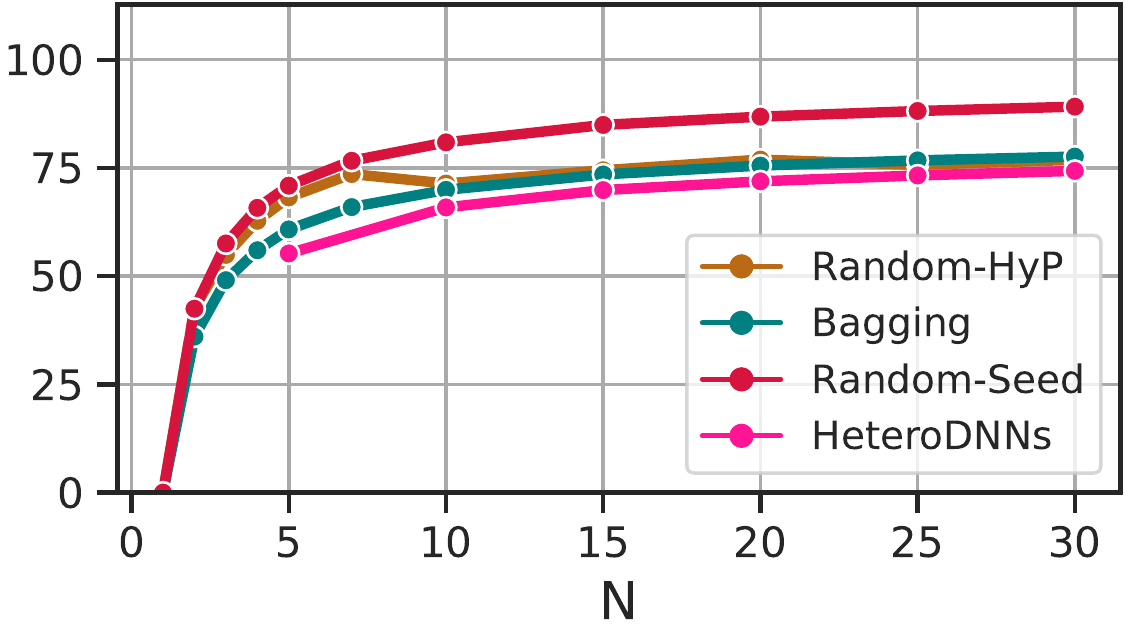}
        \subcaption{$\redun$ \label{appendix:fig:scaling.redundancy.per_model}}   
    \end{subfigure}   
    \hfill
    \begin{subfigure}[t]{0.19\linewidth}
        \vskip 0pt
        \includegraphics[width=\linewidth]{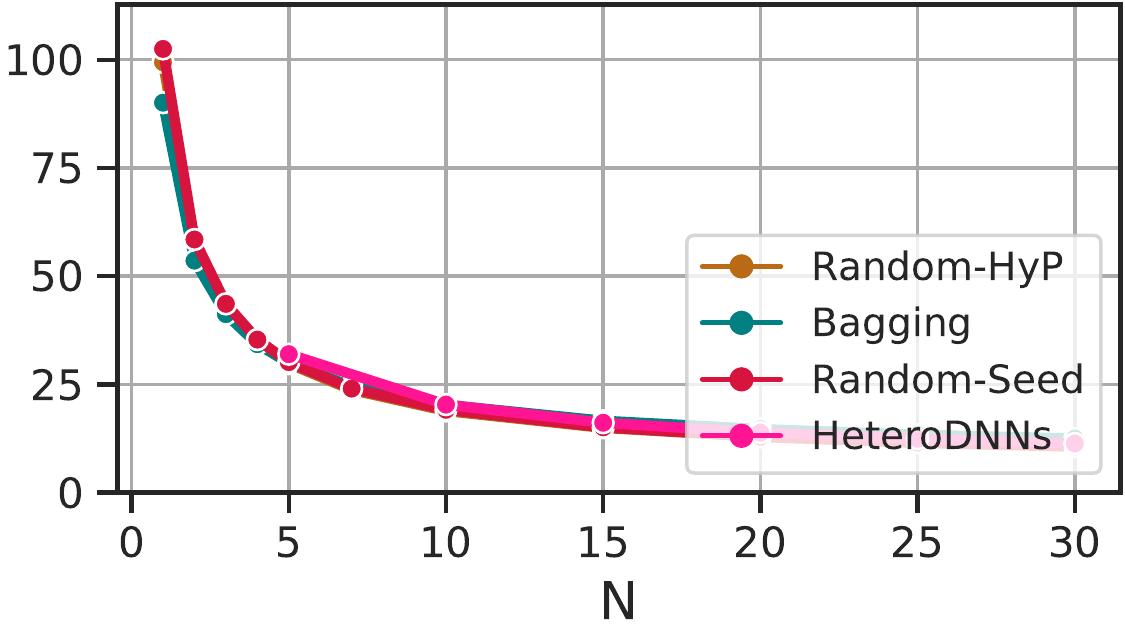}
        \subcaption{$\relev$ $- \redun$ \label{appendix:fig:scaling.novelty.per_model}}
    \end{subfigure}   
    \hfill
    \begin{subfigure}[t]{0.19\linewidth}
        \vskip 0pt
        \includegraphics[width=\linewidth]{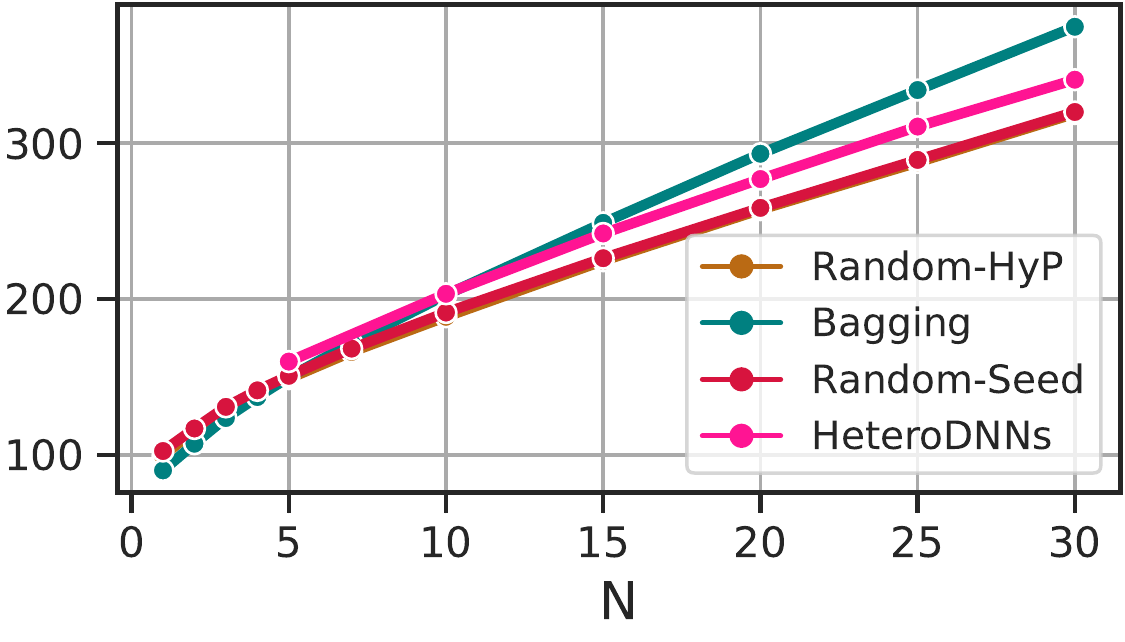}
        \captionsetup{justification=centering}
        \subcaption{$\StrengthDoublet $\newline $= N (\relev - \redun)$ \label{appendix:fig:scaling.E_without_combination_loss}}
    \end{subfigure}
    \hfill
    \begin{subfigure}[t]{0.19\linewidth}
        \vskip 0pt
        \includegraphics[width=\linewidth]{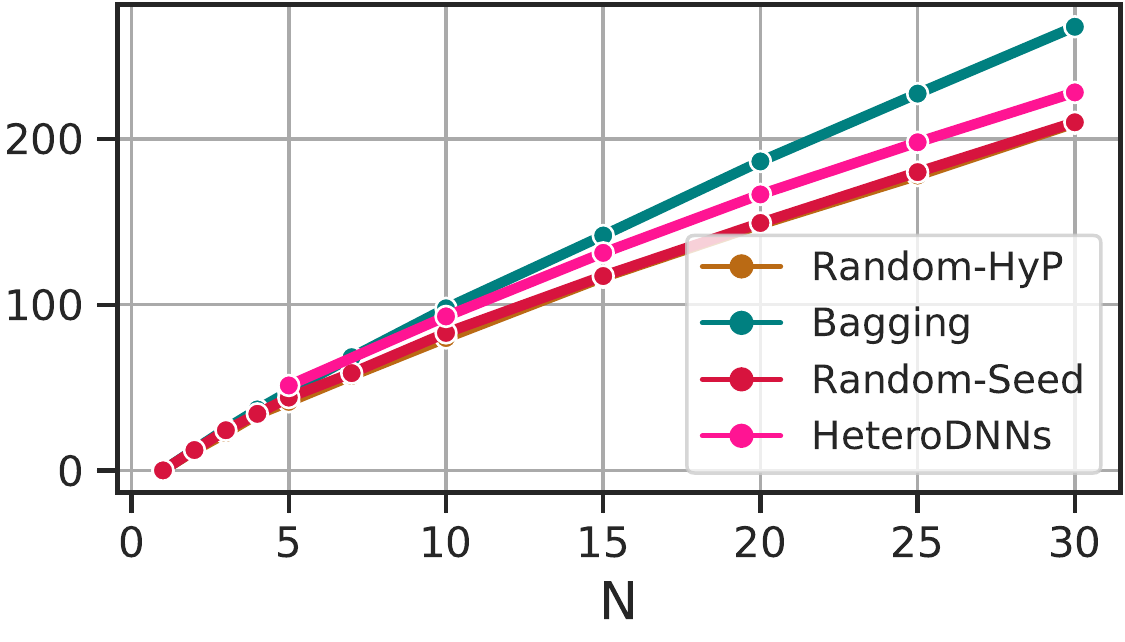}
        \subcaption{$\Combloss$\label{appendix:fig:scaling.combination_loss}}
    \end{subfigure}
    \hfill
    \begin{subfigure}[t]{0.19\linewidth}
        \vskip 0pt
        \includegraphics[width=\linewidth]{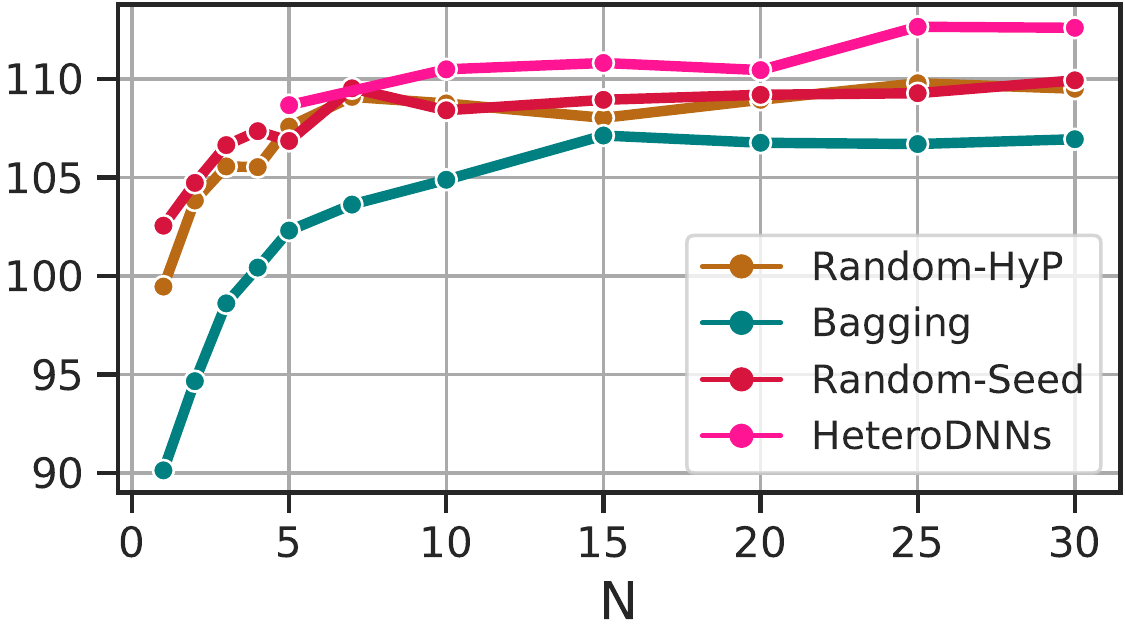}
        \captionsetup{justification=centering}
        \subcaption{$\StrengthTriplet = N (\relev - \redun) - \Combloss $\label{appendix:fig:scaling.E}}
    \end{subfigure}
\caption{
The change in ensemble quantities when the number of models $N$ is changed.
Each figure shows a specific quantity.
The ensemble systems used the SVM model combination.
Each value is an averages of the \NumTasks tasks.
$\perModelMetric$ denotes per-model metric values defined as $\perModelMetricDef$.
\label{appendix:fig:scaling}
}
\end{figure*}

\section{Behavior of Ensemble Quantities When Number of Models $N$ is Changed} \label{appendix:sec:scalability}

In this section, we examine the behavior of the ensemble quantities when the number of models is changed (\Cref{appendix:fig:scaling}).
Most importantly:
(i) both \Cref{lemma:ensemble_bound_loose} $\BoundLoose$ (\Cref{appendix:fig:scaling.bound.voting.previous_research}) and $\BoundTight$ (\Cref{appendix:fig:scaling.bound.voting.ours_wo_combloss}) could not predict the shape of error rate reduction curve (\Cref{appendix:fig:scaling.stacking}), especially the saturation over $N \gtrapprox 15$.
(ii) by contrast \Cref{lemma:ensemble_bound_tight} (\Cref{appendix:fig:scaling.stacking.bound}) could predict the phenomena. This success is attributed to the ensemble strength which consider combination loss (\Cref{appendix:fig:scaling.E}).

\Cref{appendix:fig:scaling.relevance.per_model} shows the per-model relevance $\relev = \Relev / N$, that denotes the average amount of information on $Y$ conveyed by a single model or average accuracy of the models.
All the systems kept it nearly constant, since their model training procedures do not change with respect to $N$.

\Cref{appendix:fig:scaling.redundancy.per_model} shows the per-model redundancy $\redun = \Redun / N$, which denotes the average amount of information on $Y$ conveyed by a single model that is redundant to the other models.
In all of the systems, it increased to about the same as $\relev$.
It increased because as more models come into an ensemble system, it becomes more difficult for a new model to output a ``novel'' prediction distribution compared with those of the existing models.
As a result, new models eventually become totally redundant as $\redun \sim \relev$.

$\relev - \redun$ (\Cref{appendix:fig:scaling.novelty.per_model}), the average amount of \textit{unique} information conveyed by a single model, converged to nearly zero.
Because of this diversity saturation, the increase in the $\StrengthDoublet = N \times (\relev - \redun)$ slowed at large scale (\Cref{appendix:fig:scaling.E_without_combination_loss}).
However, their saturation speed was smaller than the observed one (\Cref{appendix:fig:scaling.stacking}).
As a result, both lower bound reductions of \Cref{lemma:ensemble_bound_loose} $\BoundLoose$ (\Cref{appendix:fig:scaling.bound.voting.previous_research}) and $\BoundTight$ (\Cref{appendix:fig:scaling.bound.voting.ours_wo_combloss}) could not predict the saturation behavior.

\Cref{appendix:fig:scaling.combination_loss} shows the combination loss $\Combloss$.
$\Combloss$ increased in proportion to the increase of $\StrengthDoublet$, since $\Combloss$ represents the amount of information lost from $\StrengthDoublet$ (\Cref{appendix:increasing_combination_loss} gives the intuition behind this increase).
Overall, $\StrengthTriplet = \StrengthDoublet - \Combloss$ saturated at the large scale (\Cref{appendix:fig:scaling.E}).
Thus, the lower bound reduction by \Cref{lemma:ensemble_bound_tight} (\Cref{appendix:fig:scaling.stacking.bound}) produced by $\StrengthTriplet$ succeeded in detecting the observed saturation behavior (\Cref{appendix:fig:scaling.stacking}).

\clearpage

\section{On Increase in Combination Loss with respect to $N$}   \label{appendix:increasing_combination_loss}
We describe the reason for the increase in combination loss $\Combloss$ with respect to number of models $N$ observed in \Cref{appendix:fig:scaling.combination_loss}.

In particular, we discuss that the fact that $\Combloss$ increases with $N$ does not contradict the fact that a larger $N$ leads to better performance.

\subsection{Information theoretical view}
From information theoretical viewpoints:
\begin{enumerate}
    \item Since incoming models bring us more information, $\StrengthDoublet$, which denotes the total amount of information carried by the models, increases with $N$, as shown in \Cref{appendix:fig:scaling.E_without_combination_loss}.
    \item Since $\Combloss$ represents the amount of information lost \textit{from} $\StrengthDoublet$ when a combination function $\mathcal{F}$ is applied, $\Combloss$ generally increases as $\StrengthDoublet$ increases. This is shown in \Cref{appendix:fig:scaling.combination_loss}. This fact is not counter-intuitive, since, for example, if the information loss ``rate'' is constant as $c$, $\Combloss = c \times \StrengthDoublet$ increases at the same speed as $\StrengthDoublet$.
    \item Since the growth of $\StrengthDoublet$ is faster than that of $\Combloss$, $\StrengthTriplet = \StrengthDoublet - \Combloss$, which denotes the total amount of information remaining after the combination, also increases, as shown in \Cref{appendix:fig:scaling.E}.
    \item Since $\StrengthTriplet$ represents the performance of an ensemble system, increasing $\StrengthTriplet$ leads to better performance.
\end{enumerate}
As seen, the fact 2 that $\Combloss$ increases with $N$ does not contradict the fact 3-4 that a larger $N$ leads to better performance.

\subsection{Viewing through neglected minority model predictions}
In \cref{sec:toy_systems}, we discussed that the source of $\Combloss$ is neglected but correct model predictions.
We can also discuss $\Combloss$ from this view as follows:
\begin{enumerate}
    \item The number of neglected minority predictions on a misclassified dataset instance increases as the number of total predictions on the instance increases. Since the latter is rougly proportional to $N$, the former is also roughly proportional to $N$.
    \item The total number of misclassified dataset instances, which denotes error rate, decreases \textit{more slowly} than linearly with $N$. This is empirically known, for example as shown in \Cref{appendix:fig:scaling.stacking}.
    \item The total number of neglected minority predictions in a dataset, which is the source of $\Combloss$, is roughly estimated as [the number of neglected minority predictions on a misclassified dataset instance] $\times$ [the total number of misclassified dataset instances]. From 1 and 2, this quantity increases roughly linearly with $N$.
\end{enumerate}
As seen, the fact 3 that $\Combloss$ increase with $N$ does not contradict with the fact 2 that error rate decrease with $N$.

\section{Measurements of Information Concentration}  \label{appendix:concentration}
To observe this directly, we defined \textit{$n$-model concentration} ($Conc^N_n$) which measures the degree of concentration on top-$n$ models as a value in [0, 1]:
{
\small   %
\begin{align}
    Conc^N_n(\mathbf{O}, Y) &= \frac{\omegaMax - \omegaMin}{I(\mathbf{O};Y)} \in [0, 1], \nonumber \\ 
    I(\Omega^{N, \mathrm{max/min}}_n;Y) & = \underset{\{i_1, i_2, \dots, i_n\} \in \Omega^N_n}{\mathrm{max/min}} I(\{O_{i_1}, O_{i_2}, \dots, O_{i_n}\};Y),  \nonumber
\end{align}
\normalsize
}
where $I$ is mutual information defined by \cref{eq:notation_mutual_information} and $\Omega^N_n$ is all possible combinations of n integers from $[1, N]$.
Since the amount of information on $Y$ carried by a subset $\{O_{i_1},\dots, O_{i_n}\}$ can never be more than that of a full set $\mathbf{O}$, $\omegaMaxMin \leq I(\mathbf{O};Y)$.
This leads to $Conc^N_n(\mathbf{O}, Y) \in [0, 1]$.
The $Conc^N_n$ takes $1$ when all the information carried by $\mathbf{O}$ can be reconstructed by top-n $O_i$ and bottom-n $O_i$s having no information (i.e. $\omegaMax = I(\mathbf{O};Y)$ and $ \omegaMin = 0$).
The $Conc^N_n$ are small when the amount of information on top-n $O_i$ is similar to that of bottom-n $O_i$ (i.e. $\omegaMax \sim \omegaMin$).

\clearpage

\section{Results of each task}   \label{appendix:results_for_each_task}
Below, we show the experimental results of the \NumTasks tasks.
The discussion in \Cref{sec:validating_framework,sec:triplet_decomposition} holds in each task, that is:
\begin{itemize}
    \item $\BoundFuncTight$ generate lower bound tighter than $\BoundFuncLoose$. This is discussed in \Cref{sec:bound_func_tight}.
    \item The lower bound reduction by \Cref{lemma:ensemble_bound_tight} $\BoundTight$ is strongly correlated to the error rate reductions, while those of \Cref{lemma:ensemble_bound_loose} $\BoundLoose$ and $\BoundTightWOCombLoss$ are not. This is discussed in \Cref{sec:effectiveness_of_bound}.
    \item The lower bound reduction by \Cref{lemma:ensemble_bound_tight} $\BoundTight$ successfully predicts the shape of error rate reduction curve when the number of models $N$ is changed, while those of \Cref{lemma:ensemble_bound_loose} $\BoundLoose$ and $\BoundTightWOCombLoss$ do not. This is discussed in \Cref{sec:scalability}.
    \item The strengths and weaknesses of ensemble systems in terms of the three metrics. This is discussed in \Cref{sec:triplet_decomposition}.
\end{itemize}

\begin{figure*}[t!]
    \begin{subfigure}[t]{0.32\linewidth}
        \vskip 0pt
        \centering
        \includegraphics[width=0.65\linewidth]{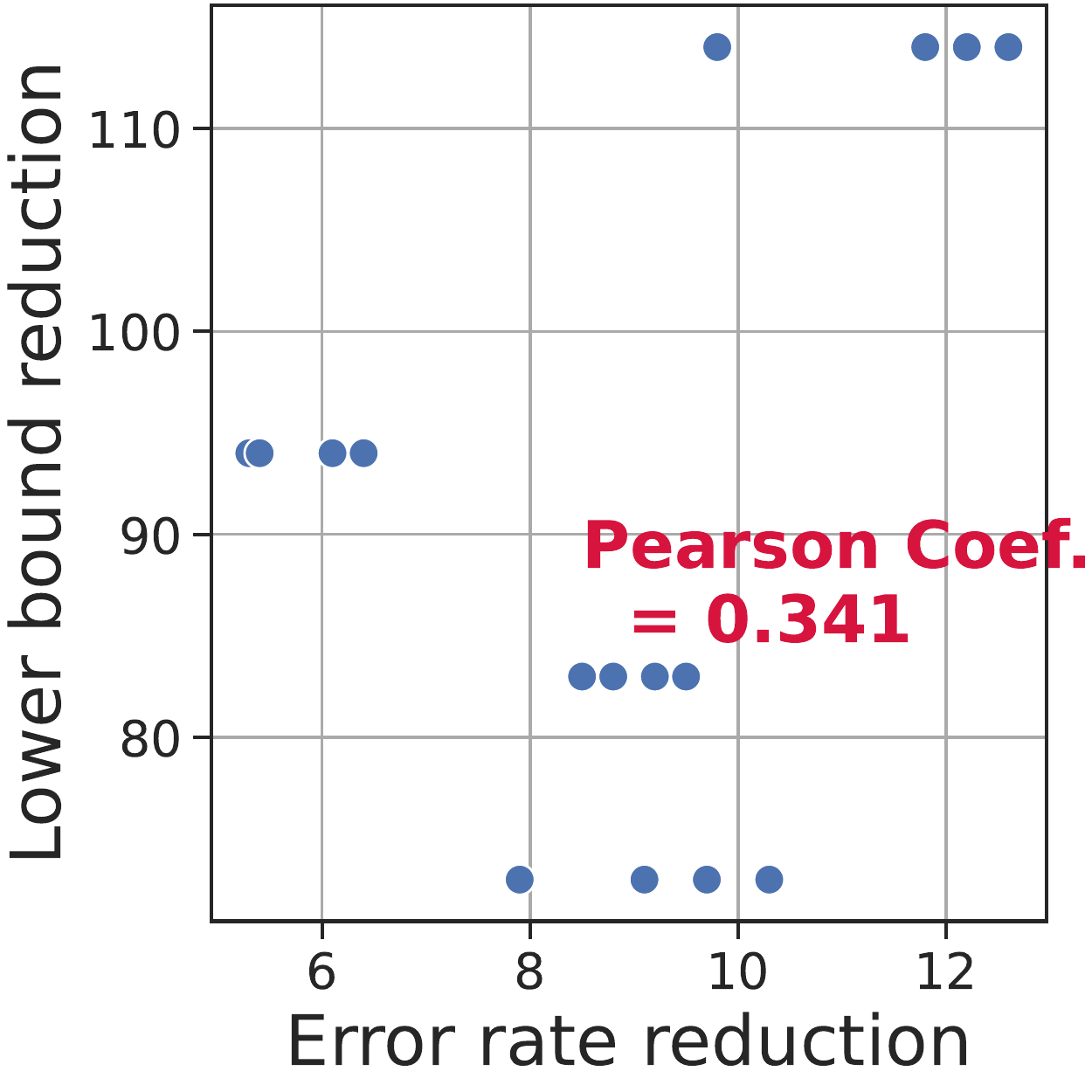}
        \caption{
        \Cref{lemma:ensemble_bound_loose} $\BoundLoose$.
        \label{fig:ERR_LBR_scatter_plot_loose_Boolq}
        }
    \end{subfigure}
    \hfill
    \begin{subfigure}[t]{0.32\linewidth}
        \vskip 0pt
        \centering
        \includegraphics[width=0.65\linewidth]{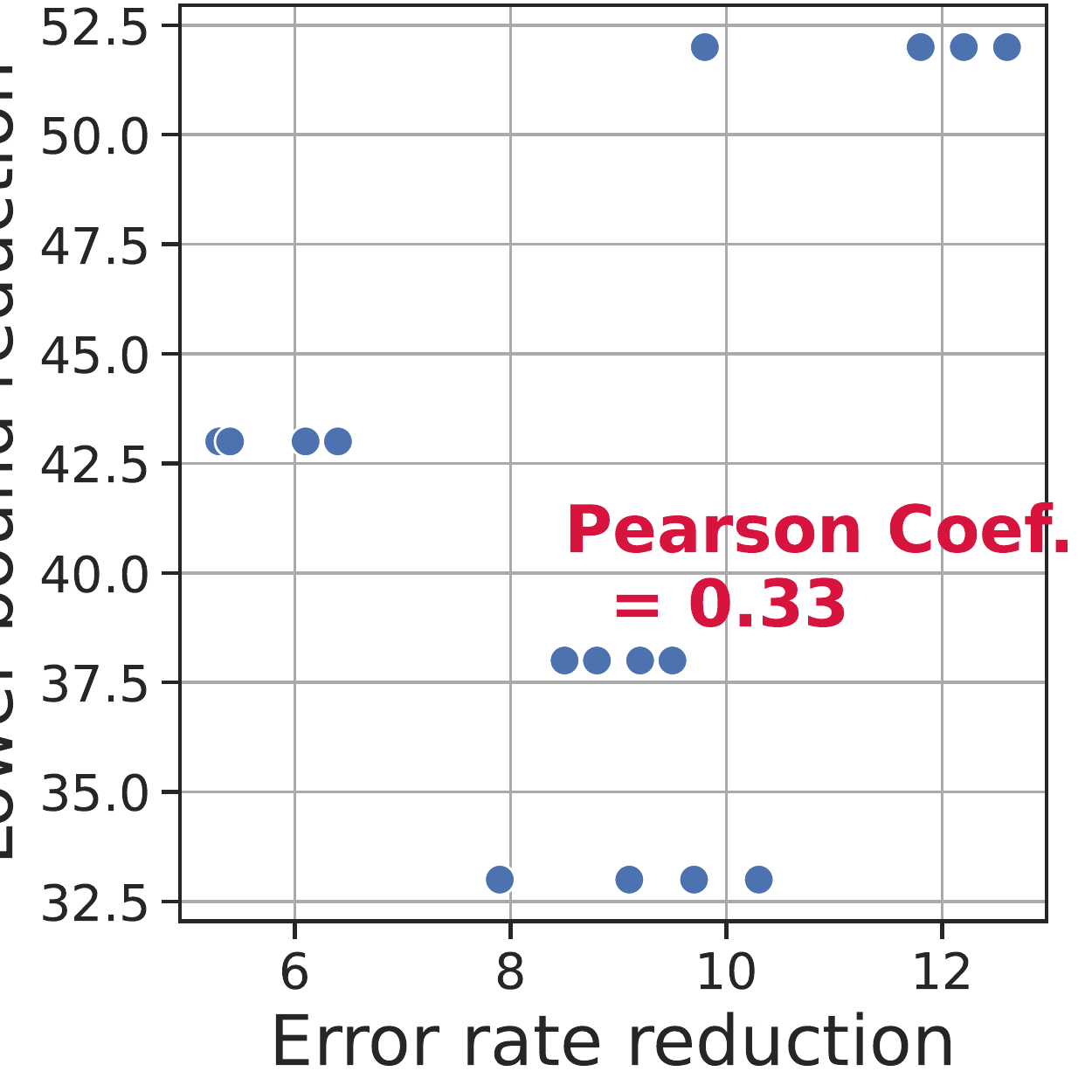}
        \caption{
            $\BoundTightWOCombLoss$.
            \label{fig:ERR_LBR_scatter_plot_tight_wo_combloss_Boolq}
        }
    \end{subfigure}
    \hfill
    \begin{subfigure}[t]{0.32\linewidth}
        \vskip 0pt
        \centering
        \includegraphics[width=0.65\linewidth]{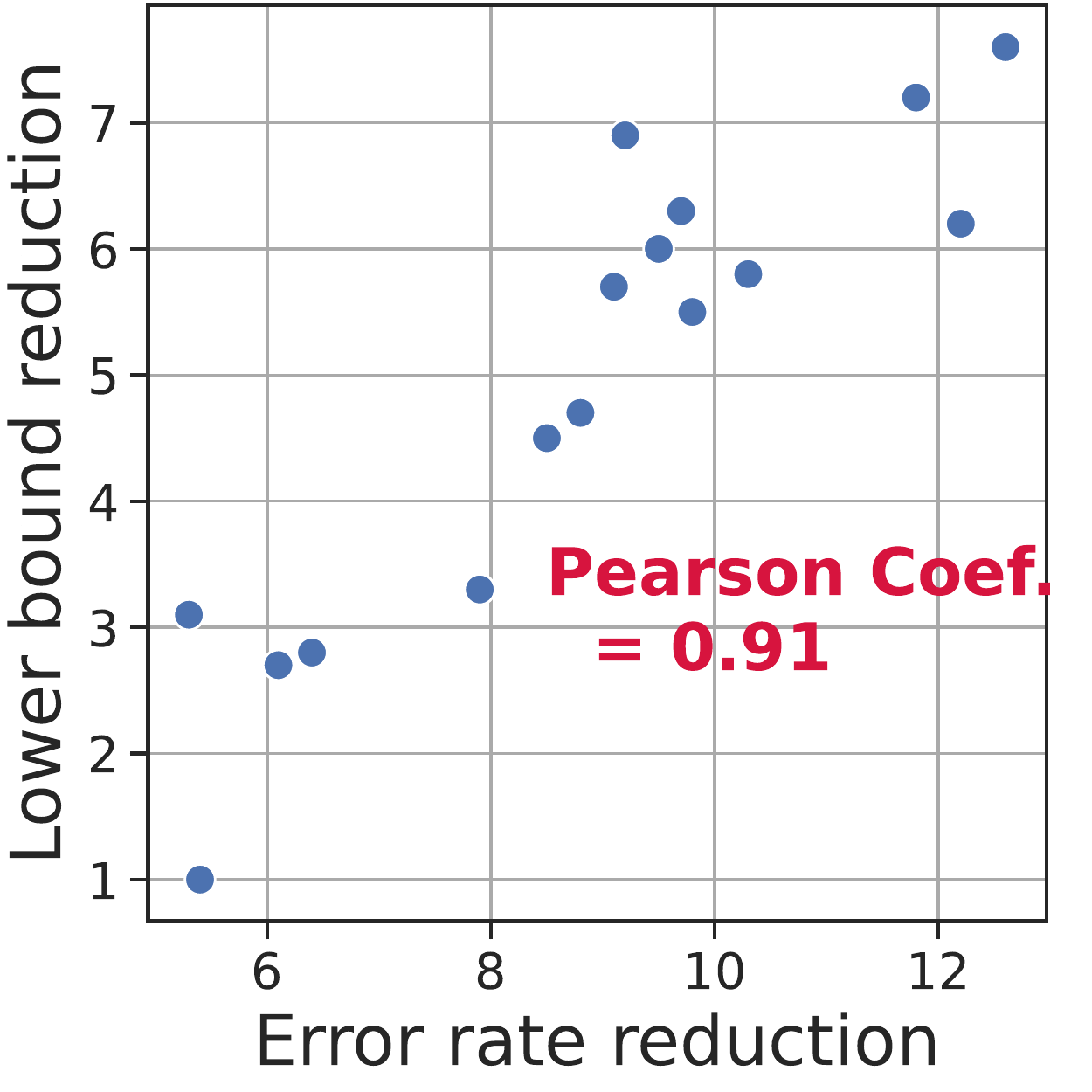}
        \caption{
            \textbf{\Cref{lemma:ensemble_bound_tight}} $\BoundTight$.
            \label{fig:ERR_LBR_scatter_plot_tight_Boolq}
        }
    \end{subfigure}
    \hfill

    \caption{
    \textbf{Boolq task.}
    Correlations between error rate reductions and lower bound reductions.
    Each figure uses different type of lower bound.
    Each point in the figures shows a quantity of a specific ensemble system $s$ and the quantity is the average over the \NumTasks tasks.
    See \Cref{tb:ablation_Boolq} for the real value of each point.
    We used the \NumSystems ensemble systems described in \Cref{sec:ensemble_systems}.
    Each system $s$ used $N=15$ models.
    The baseline values in \Cref{eq:error_reduction,eq:lower_bound_reduction} were the followings:
    ER($s_0$): \SI{24.1}{\percent}.
    LB($s_0$) by $\BoundFuncTight(\StrengthTriplet)$: \SI{3.1}{\percent}.
    LB($s_0$) by $\BoundFuncTight(\StrengthDoublet)$: \SI{3.1}{\percent}.
    LB($s_0$) by $\BoundFuncLoose(\StrengthDoublet)$: \SI{-2.0}{\percent}.
    \label{fig:ERR_LBR_scatter_plot_Boolq}
    }
\end{figure*}

\begin{figure*}[h!]
    \begin{subfigure}[t]{0.19\linewidth}
        \vskip 0pt
        \includegraphics[width=\linewidth]{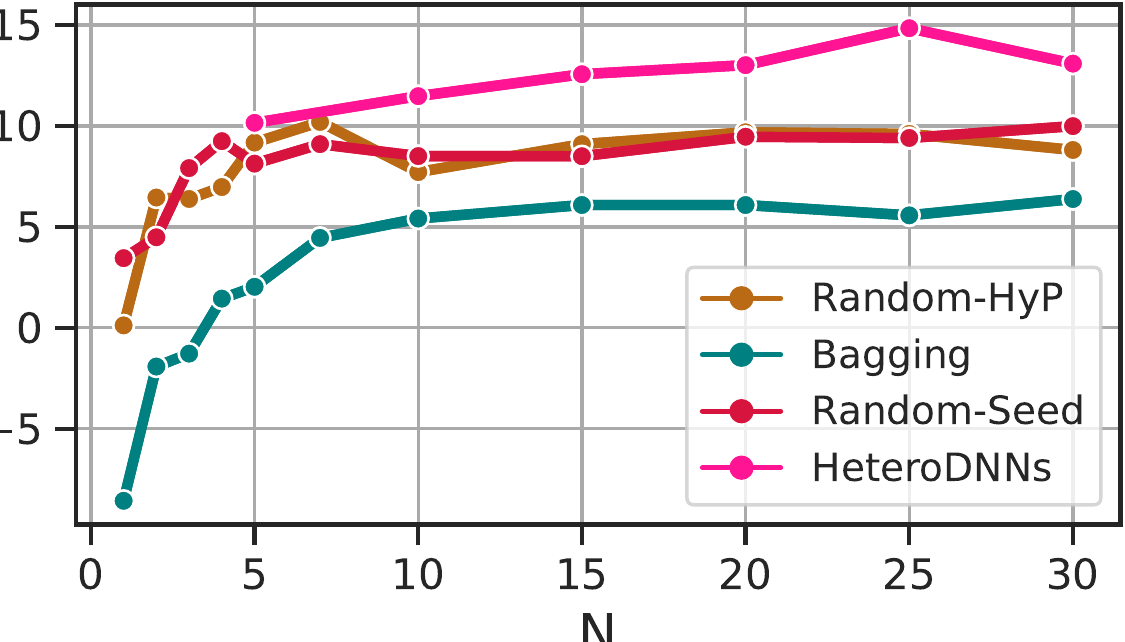}
        \subcaption{Error rate reduction.\label{appendix:fig:scaling.stacking_Boolq}}
    \end{subfigure}
    \hfill
    \begin{subfigure}[t]{0.19\linewidth}
        \vskip 0pt
        \includegraphics[width=\linewidth]{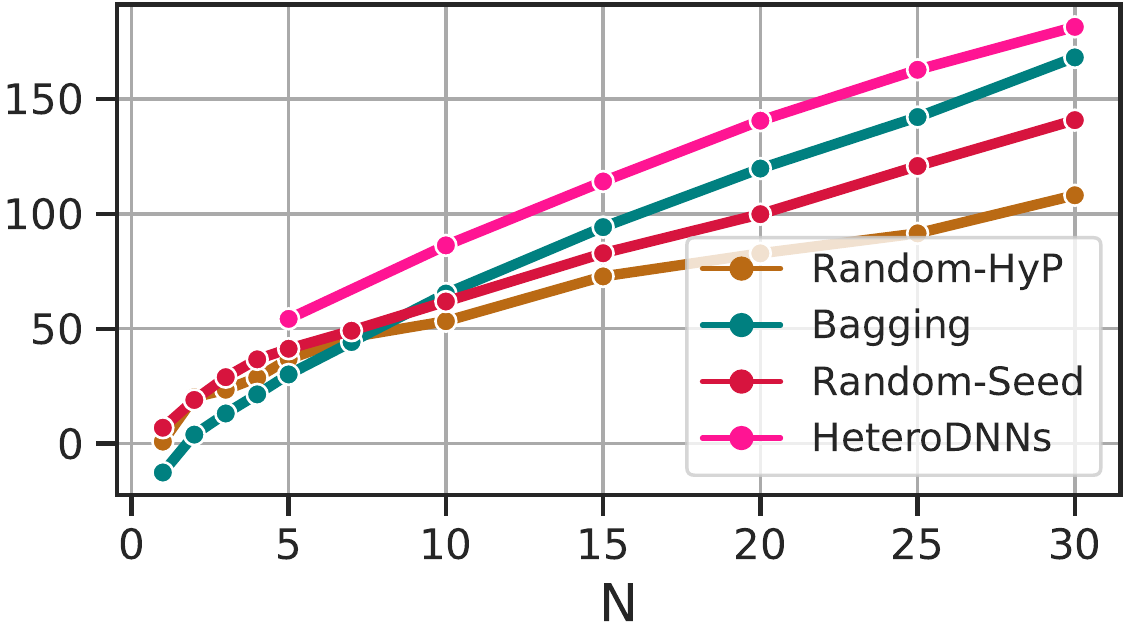}
        \subcaption{Lower bound reduction by \Cref{lemma:ensemble_bound_loose} $\BoundLoose$.\label{appendix:fig:scaling.bound.voting.previous_research_Boolq}}
    \end{subfigure}
    \begin{subfigure}[t]{0.19\linewidth}
        \vskip 0pt
        \includegraphics[width=\linewidth]{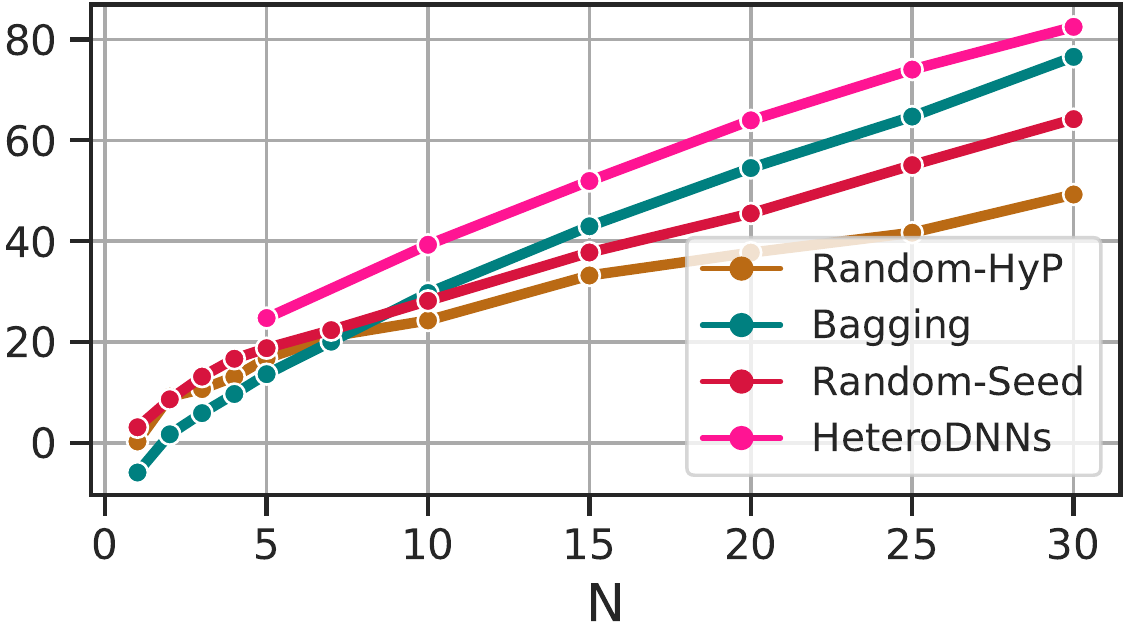}
        \subcaption{Lower bound reduction by $\BoundTightWOCombLoss$. \label{appendix:fig:scaling.bound.voting.ours_wo_combloss_Boolq}}
    \end{subfigure}
    \hfill
     \begin{subfigure}[t]{0.19\linewidth}
        \vskip 0pt
        \includegraphics[width=\linewidth]{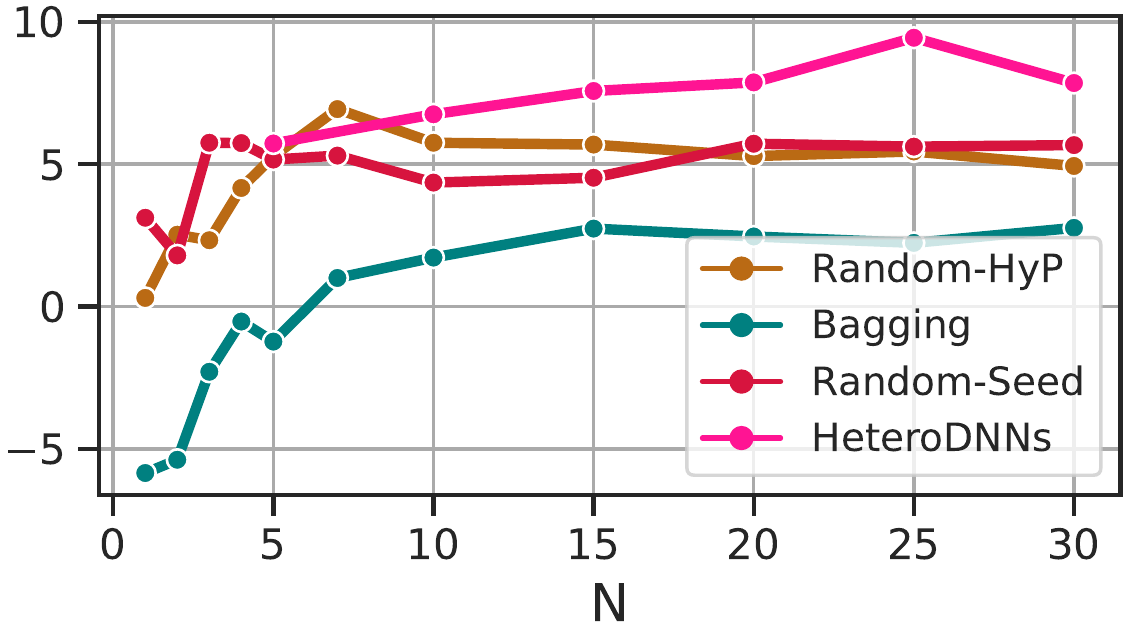}
        \subcaption{Lower bound reduction by \textbf{\Cref{lemma:ensemble_bound_tight}} $\BoundTight$.\label{appendix:fig:scaling.stacking.bound_Boolq}}
    \end{subfigure}
    \hfill
    \begin{subfigure}[t]{0.19\linewidth}
        \vskip 0pt
        \includegraphics[width=\linewidth]{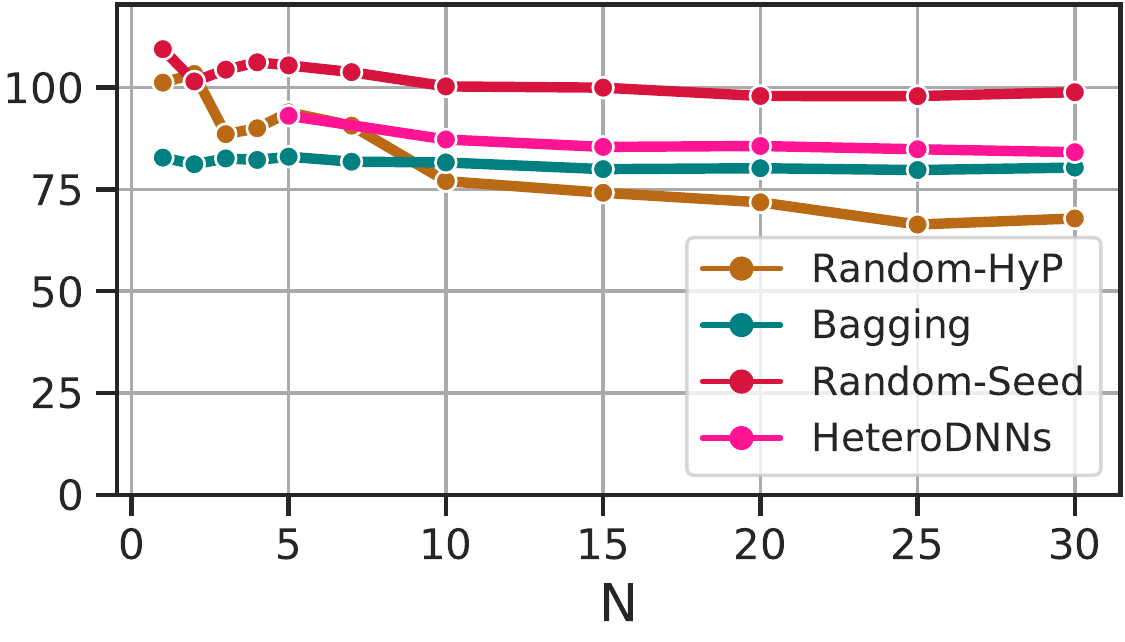}
        \subcaption{$\relev$ \label{appendix:fig:scaling.relevance.per_model_Boolq}}   
    \end{subfigure}   
    \hfill
    \vfill
    \begin{subfigure}[t]{0.19\linewidth}
        \vskip 0pt
        \includegraphics[width=\linewidth]{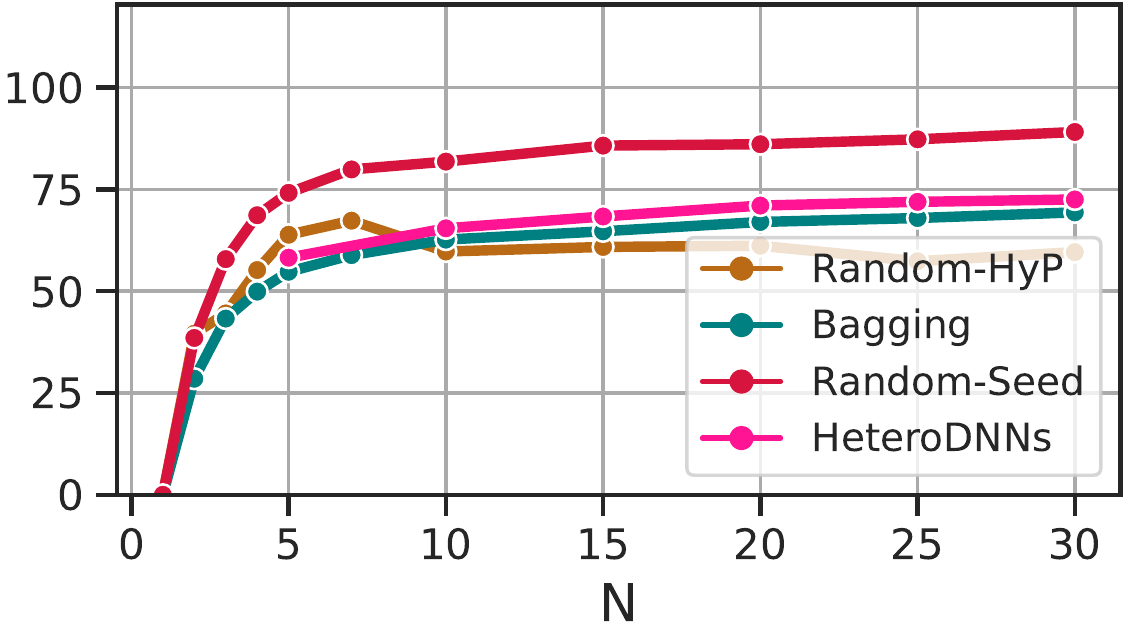}
        \subcaption{$\redun$ \label{appendix:fig:scaling.redundancy.per_model_Boolq}}   
    \end{subfigure}   
    \hfill
    \begin{subfigure}[t]{0.19\linewidth}
        \vskip 0pt
        \includegraphics[width=\linewidth]{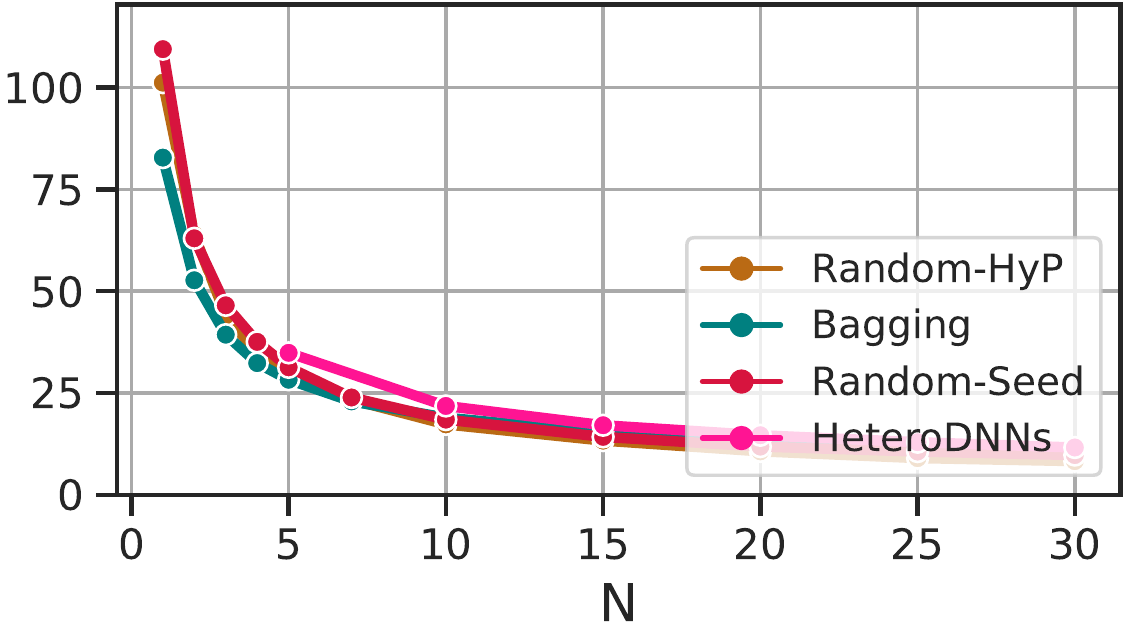}
        \subcaption{$\relev$ $- \redun$ \label{appendix:fig:scaling.novelty.per_model_Boolq}}
    \end{subfigure}   
    \hfill
    \begin{subfigure}[t]{0.19\linewidth}
        \vskip 0pt
        \includegraphics[width=\linewidth]{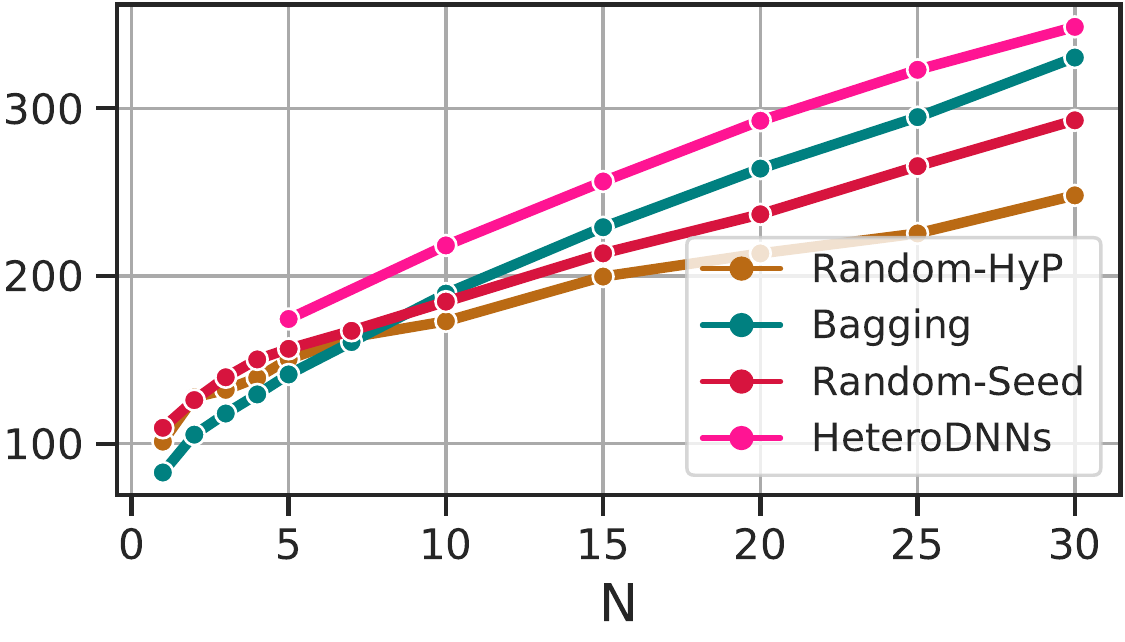}
        \captionsetup{justification=centering}
        \subcaption{$\StrengthDoublet $\newline $= N (\relev - \redun)$ \label{appendix:fig:scaling.E_without_combination_loss_Boolq}}
    \end{subfigure}
    \hfill
    \begin{subfigure}[t]{0.19\linewidth}
        \vskip 0pt
        \includegraphics[width=\linewidth]{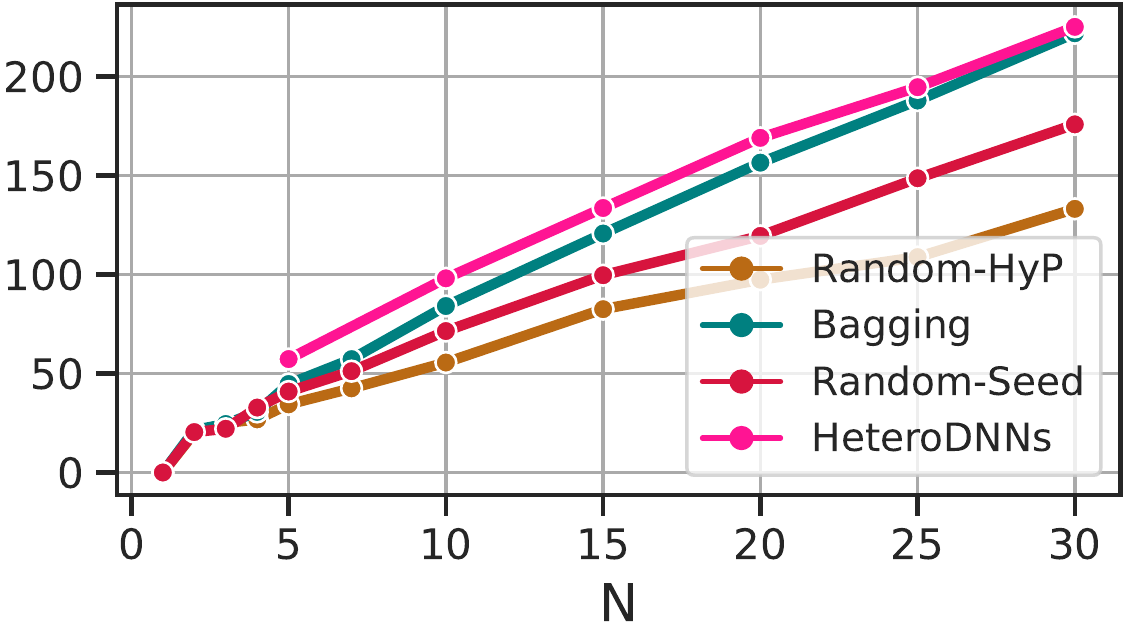}
        \subcaption{$\Combloss$\label{appendix:fig:scaling.combination_loss_Boolq}}
    \end{subfigure}
    \hfill
    \begin{subfigure}[t]{0.19\linewidth}
        \vskip 0pt
        \includegraphics[width=\linewidth]{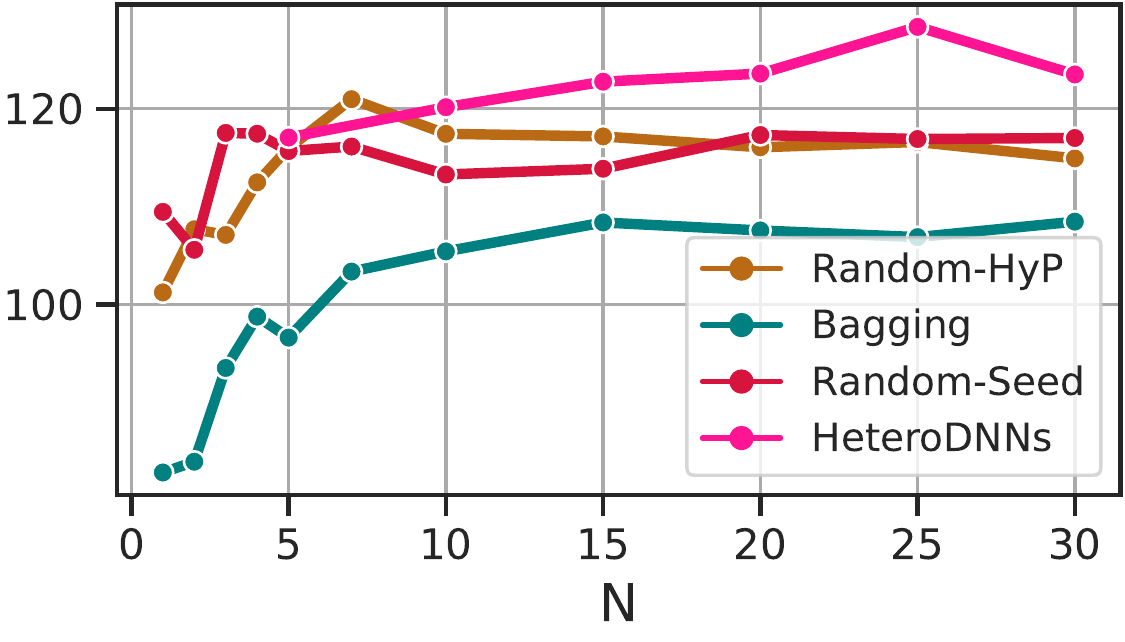}
        \captionsetup{justification=centering}
        \subcaption{$\StrengthTriplet = N (\relev - \redun) - \Combloss $\label{appendix:fig:scaling.E_Boolq}}
    \end{subfigure}
\caption{
\textbf{Boolq task.}
The change in ensemble quantities when the number of models $N$ is changed.
Each figure shows a specific quantity.
The ensemble systems used the SVM model combination.
Each value is an averages of the \NumTasks tasks.
$\perModelMetric$ denotes per-model metric values defined as $\perModelMetricDef$.
\label{appendix:fig:scaling_Boolq}
}
\end{figure*}

\begin{table*}[h!]
    \centering
    \caption{
        \textbf{Boolq task.}
        Statistics of ensemble systems described in \Cref{sec:ensemble_systems}.
        The rows and columns list the model generation and combination methods of \Cref{tb:ensemble_methods}, respectively.
        Each cell shows a quantity of a specific system $s$.
        Each quantity is the average over the \NumTasks tasks.
        Each system contains $N=15$ models.
        Color shows the rank within \textit{each column} (brighter is better).
        \label{tb:ablation_Boolq}
    }  
    \begin{subfigure}{\linewidth}
        \centering
        \small
        \tabcolsep 3.0pt
    \subcaption{
        Error rate reductions and lower bound reductions.
        The baseline values used in \Cref{eq:error_reduction,eq:lower_bound_reduction} were the followings.
        ER($s_0$): \SI{24.1}{\percent}.
        LB($s_0$) by $\BoundFuncTight(\StrengthTriplet)$: \SI{3.1}{\percent}.
        LB($s_0$) by $\BoundFuncTight(\StrengthDoublet)$: \SI{3.1}{\percent}.
        LB($s_0$) by $\BoundFuncLoose(\StrengthDoublet)$: \SI{-2.0}{\percent}.
    }

\begin{tabular}{lcccccccccccc}
\toprule
& \multicolumn{4}{c}{Error rate reductions \cref{eq:error_reduction}} & &   \multicolumn{7}{c}{Lower bound reductions \Cref{eq:lower_bound_reduction}} \\
\cmidrule(l{\tabcolsep}r{\tabcolsep}){2-5} \cmidrule(l{\tabcolsep}){7-13}
 & \multirow{2}{*}{Voting} & \multirow{2}{*}{LogR} & \multirow{2}{*}{SVM} & \multirow{2}{*}{RForest} & & \multicolumn{4}{c}{\textbf{\Cref{lemma:ensemble_bound_tight}} $\BoundFuncTight(\StrengthTriplet)$} & & \multirow{2}{*}{$\BoundFuncTight(\StrengthDoublet)$} & \multirow{2}{*}{\begin{tabular}{c}\Cref{lemma:ensemble_bound_loose} \\ $\BoundFuncLoose(\StrengthDoublet)$\end{tabular}} \\
 &  &  &  &  & & \multicolumn{1}{c}{Voting} & \multicolumn{1}{c}{LogR} & \multicolumn{1}{c}{SVM} & \multicolumn{1}{c}{RForest} & & &  \\

\midrule
Random-HyP    &   \cThird $7.9_{\pm{\mbox{\tiny 1.2}}}$ &  \cSecond $9.6_{\pm{\mbox{\tiny 2.0}}}$ &  \cSecond $9.1_{\pm{\mbox{\tiny 2.0}}}$ &  \cSecond $10.3_{\pm{\mbox{\tiny 1.4}}}$ &        &   \cThird $3.3_{\pm{\mbox{\tiny 1.2}}}$ &  \cSecond $6.3_{\pm{\mbox{\tiny 1.8}}}$ &  \cSecond $5.7_{\pm{\mbox{\tiny 1.0}}}$ &  \cSecond $5.8_{\pm{\mbox{\tiny 1.7}}}$ &        &  \cFourth $33_{\pm{\mbox{\tiny 2}}}$ &  \cFourth $73_{\pm{\mbox{\tiny 5}}}$ \\
Bagging       &  \cFourth $5.3_{\pm{\mbox{\tiny 2.0}}}$ &  \cFourth $6.4_{\pm{\mbox{\tiny 3.0}}}$ &  \cFourth $6.2_{\pm{\mbox{\tiny 2.4}}}$ &   \cFourth $5.4_{\pm{\mbox{\tiny 1.4}}}$ &        &  \cFourth $3.1_{\pm{\mbox{\tiny 1.7}}}$ &  \cFourth $2.8_{\pm{\mbox{\tiny 2.7}}}$ &  \cFourth $2.7_{\pm{\mbox{\tiny 1.8}}}$ &  \cFourth $1.0_{\pm{\mbox{\tiny 1.3}}}$ &        &  \cSecond $43_{\pm{\mbox{\tiny 2}}}$ &  \cSecond $94_{\pm{\mbox{\tiny 3}}}$ \\
Random-Seed   &  \cSecond $9.2_{\pm{\mbox{\tiny 1.9}}}$ &   \cThird $9.5_{\pm{\mbox{\tiny 1.8}}}$ &   \cThird $8.5_{\pm{\mbox{\tiny 3.1}}}$ &    \cThird $8.8_{\pm{\mbox{\tiny 2.9}}}$ &        &   \cFirst $6.9_{\pm{\mbox{\tiny 1.7}}}$ &   \cThird $6.0_{\pm{\mbox{\tiny 1.7}}}$ &   \cThird $4.5_{\pm{\mbox{\tiny 3.1}}}$ &   \cThird $4.7_{\pm{\mbox{\tiny 2.5}}}$ &        &   \cThird $38_{\pm{\mbox{\tiny 1}}}$ &   \cThird $83_{\pm{\mbox{\tiny 3}}}$ \\
Hetero-DNNs &   \cFirst $9.8_{\pm{\mbox{\tiny 0.3}}}$ &  \cFirst $11.8_{\pm{\mbox{\tiny 1.6}}}$ &  \cFirst $12.6_{\pm{\mbox{\tiny 0.4}}}$ &   \cFirst $12.2_{\pm{\mbox{\tiny 1.5}}}$ &        &  \cSecond $5.5_{\pm{\mbox{\tiny 0.4}}}$ &   \cFirst $7.2_{\pm{\mbox{\tiny 1.3}}}$ &   \cFirst $7.6_{\pm{\mbox{\tiny 0.4}}}$ &   \cFirst $6.2_{\pm{\mbox{\tiny 1.4}}}$ &        &   \cFirst $52_{\pm{\mbox{\tiny 2}}}$ &  \cFirst $114_{\pm{\mbox{\tiny 7}}}$ \\
\bottomrule

\end{tabular}

    \end{subfigure}
    \vfill
    \begin{subfigure}{\linewidth}
    \centering
    \small
    \tabcolsep 2.0pt
    \subcaption{
    Breakdown of ensemble strength defined in \cref{eq:triplet_decomposition}.
    We show per-model metric values defined as $\perModelMetricDef$. Thus, $\StrengthTriplet = (\relev - \redun - \combloss) \ \times N $ holds.
    For intuitive understanding, all the values are normalized by the ensemble strength of baseline $\StrengthTriplet_{s_0}$, for example, $\Relev = \RelevHat / \StrengthTriplet_{s_0} \times 100$ where $\RelevHat$ is the raw value.
    \label{tb:ablation_triple_Boolq}
    }

\begin{tabular}{lccccccccccccc}
\toprule

{} & \multicolumn{4}{c}{\multirow{1}{*}{$\StrengthTripletWithArgs$}} & & \multicolumn{6}{c}{Per-model metric values} & \\
\cmidrule(l{\tabcolsep}r{\tabcolsep}){2-5} \cmidrule(l{\tabcolsep}r{\tabcolsep}){6-12}
{} &  & &  &  & & \multirow{2}{*}{$\perModelMetric_{\normalfont \text{relev}}$} & \multirow{2}{*}{$\perModelMetric_{\normalfont \text{redun}}$} & \multicolumn{4}{c}{ $\perModelMetric_{\normalfont \text{combloss}}$} & & \multirow{2}{*}{$\perModelMetric_{\normalfont \text{relev}} - \perModelMetric_{\normalfont \text{redun}}$} \\
{} & \multicolumn{1}{c}{Voting} & \multicolumn{1}{c}{LogR} & \multicolumn{1}{c}{SVM} & \multicolumn{1}{c}{RForest} & &  {} &  {} &  \multicolumn{1}{c}{Voting} & \multicolumn{1}{c}{LogR} &  \multicolumn{1}{c}{SVM} & \multicolumn{1}{c}{RForest} & &  {} \\

\midrule
Baseline ($s_0$)                    &    \multicolumn{4}{c}{\cBase 100 (the raw value is 0.182)} &        &   \cBase 100 &                                 \cBase 0 &                                \cBase 0 &                                \cBase 0 &                                \cBase 0 &                                 \cBase 0 &        &   \cBase 100 \\

\midrule

Random-HyP                   &   \cThird $110.1_{\pm{\mbox{\tiny 3.8}}}$ &  \cSecond $119.2_{\pm{\mbox{\tiny 6.0}}}$ &  \cSecond $117.2_{\pm{\mbox{\tiny 3.5}}}$ &  \cSecond $117.7_{\pm{\mbox{\tiny 5.6}}}$ &        &  \cFourth $74.2_{\pm{\mbox{\tiny 3.2}}}$ &   \cFirst $60.9_{\pm{\mbox{\tiny 2.8}}}$ &   \cFirst $5.97_{\pm{\mbox{\tiny 0.32}}}$ &   \cFirst $5.36_{\pm{\mbox{\tiny 0.47}}}$ &   \cFirst $5.50_{\pm{\mbox{\tiny 0.48}}}$ &   \cFirst $5.46_{\pm{\mbox{\tiny 0.55}}}$ &        &  \cFourth $13.3_{\pm{\mbox{\tiny 4.3}}}$ \\
Bagging                      &  \cFourth $109.5_{\pm{\mbox{\tiny 5.3}}}$ &  \cFourth $108.8_{\pm{\mbox{\tiny 8.5}}}$ &  \cFourth $108.4_{\pm{\mbox{\tiny 5.7}}}$ &  \cFourth $103.2_{\pm{\mbox{\tiny 3.9}}}$ &        &   \cThird $80.0_{\pm{\mbox{\tiny 2.4}}}$ &  \cSecond $64.7_{\pm{\mbox{\tiny 2.5}}}$ &   \cThird $7.97_{\pm{\mbox{\tiny 0.12}}}$ &   \cThird $8.02_{\pm{\mbox{\tiny 0.41}}}$ &   \cThird $8.04_{\pm{\mbox{\tiny 0.28}}}$ &   \cThird $8.39_{\pm{\mbox{\tiny 0.07}}}$ &        &  \cSecond $15.3_{\pm{\mbox{\tiny 3.4}}}$ \\
Random-Seed                  &   \cFirst $120.8_{\pm{\mbox{\tiny 5.6}}}$ &   \cThird $118.1_{\pm{\mbox{\tiny 5.8}}}$ &   \cThird $113.9_{\pm{\mbox{\tiny 9.8}}}$ &   \cThird $114.3_{\pm{\mbox{\tiny 8.0}}}$ &        &  \cFirst $100.0_{\pm{\mbox{\tiny 0.0}}}$ &  \cFourth $85.8_{\pm{\mbox{\tiny 0.3}}}$ &  \cSecond $6.18_{\pm{\mbox{\tiny 0.27}}}$ &  \cSecond $6.36_{\pm{\mbox{\tiny 0.17}}}$ &  \cSecond $6.64_{\pm{\mbox{\tiny 0.41}}}$ &  \cSecond $6.61_{\pm{\mbox{\tiny 0.27}}}$ &        &   \cThird $14.2_{\pm{\mbox{\tiny 0.3}}}$ \\
Hetero-DNNs                &  \cSecond $116.4_{\pm{\mbox{\tiny 1.6}}}$ &   \cFirst $121.4_{\pm{\mbox{\tiny 3.1}}}$ &   \cFirst $122.8_{\pm{\mbox{\tiny 1.8}}}$ &   \cFirst $118.7_{\pm{\mbox{\tiny 4.9}}}$ &        &  \cSecond $85.4_{\pm{\mbox{\tiny 2.0}}}$ &   \cThird $68.4_{\pm{\mbox{\tiny 1.5}}}$ &  \cFourth $9.33_{\pm{\mbox{\tiny 0.65}}}$ &  \cFourth $8.99_{\pm{\mbox{\tiny 0.92}}}$ &  \cFourth $8.90_{\pm{\mbox{\tiny 0.63}}}$ &  \cFourth $9.18_{\pm{\mbox{\tiny 0.41}}}$ &        &   \cFirst $17.1_{\pm{\mbox{\tiny 2.5}}}$ \\

\bottomrule
\end{tabular}

\end{subfigure}

\end{table*}

\clearpage

\begin{figure*}[t!]
    \begin{subfigure}[t]{0.32\linewidth}
        \vskip 0pt
        \centering
        \includegraphics[width=0.65\linewidth]{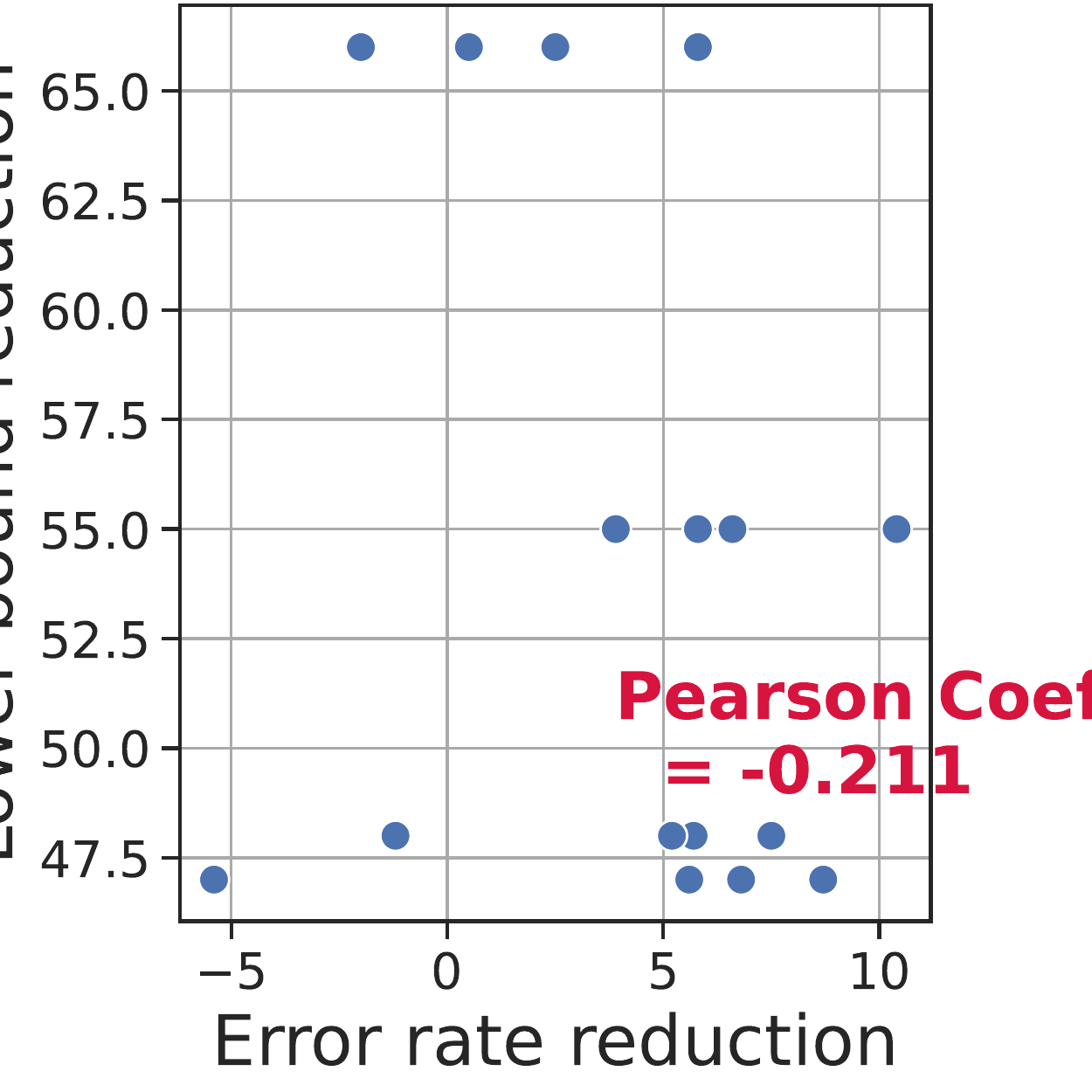}
        \caption{
        \Cref{lemma:ensemble_bound_loose} $\BoundLoose$.
        \label{fig:ERR_LBR_scatter_plot_loose_CoLA}
        }
    \end{subfigure}
    \hfill
    \begin{subfigure}[t]{0.32\linewidth}
        \vskip 0pt
        \centering
        \includegraphics[width=0.65\linewidth]{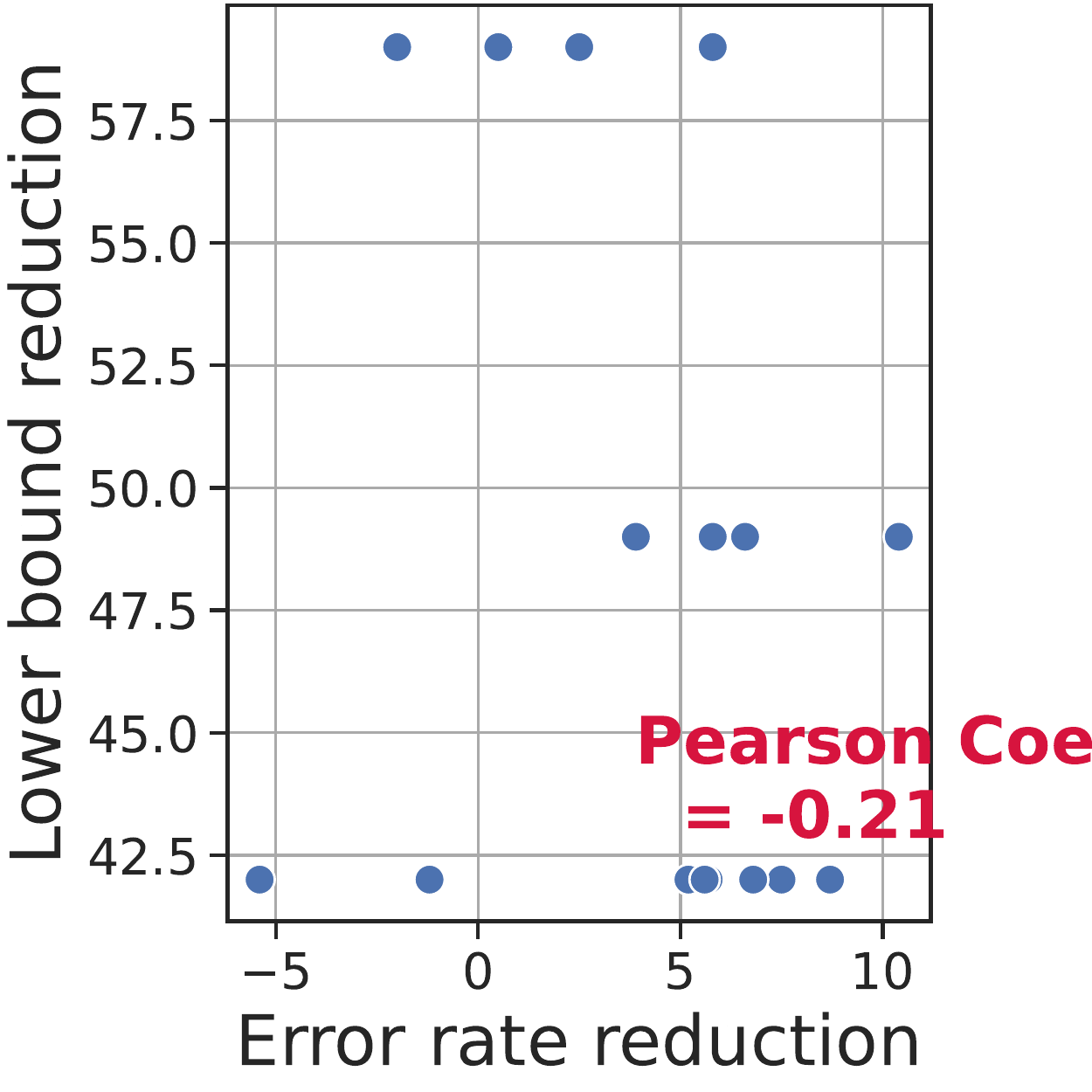}
        \caption{
            $\BoundTightWOCombLoss$.
            \label{fig:ERR_LBR_scatter_plot_tight_wo_combloss_CoLA}
        }
    \end{subfigure}
    \hfill
    \begin{subfigure}[t]{0.32\linewidth}
        \vskip 0pt
        \centering
        \includegraphics[width=0.65\linewidth]{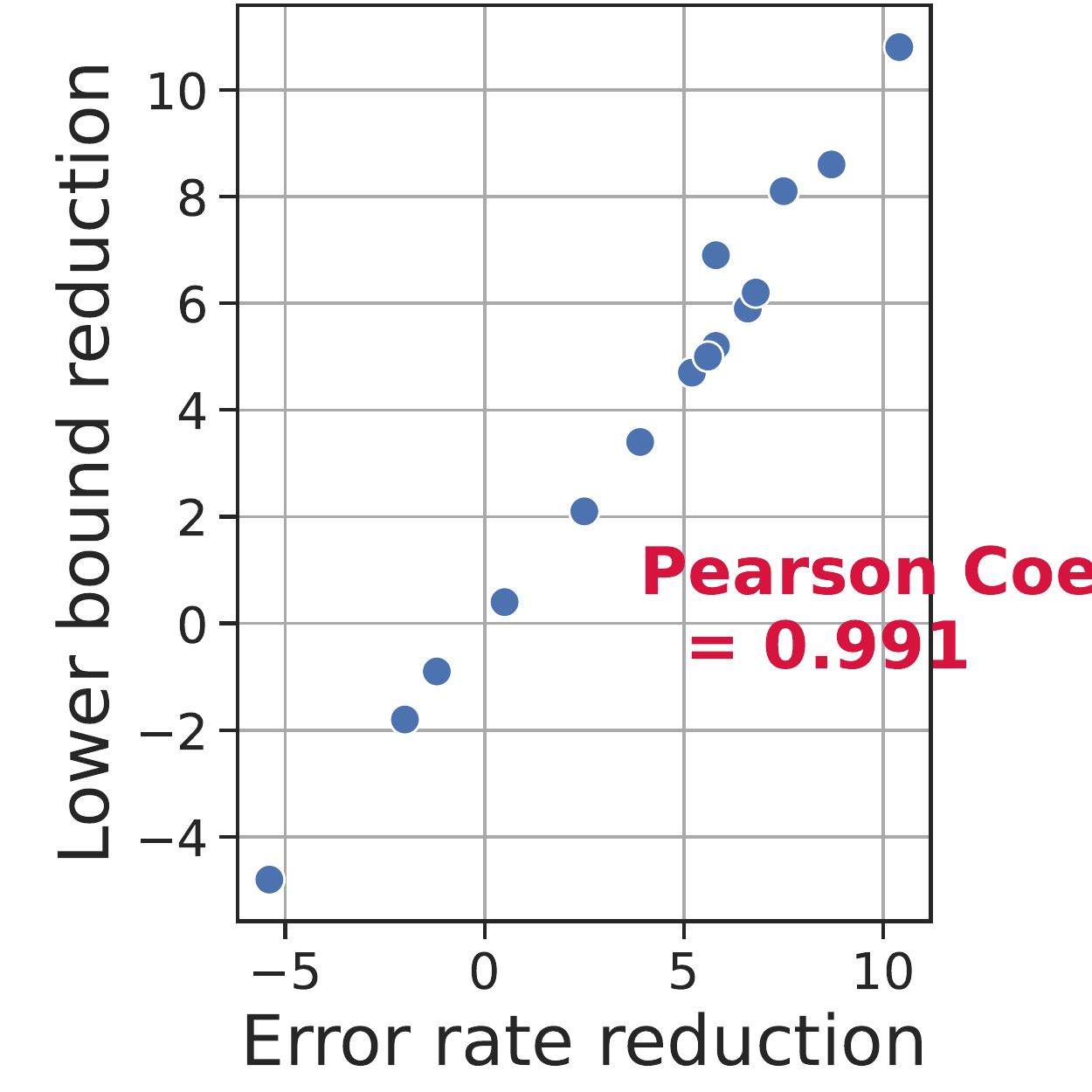}
        \caption{
            \textbf{\Cref{lemma:ensemble_bound_tight}} $\BoundTight$.
            \label{fig:ERR_LBR_scatter_plot_tight_CoLA}
        }
    \end{subfigure}
    \hfill

    \caption{
    \textbf{CoLA task.}
    Correlations between error rate reductions and lower bound reductions.
    Each figure uses different type of lower bound.
    Each point in the figures shows a quantity of a specific ensemble system $s$ and the quantity is the average over the \NumTasks tasks.
    See \Cref{tb:ablation_CoLA} for the real value of each point.
    We used the \NumSystems ensemble systems described in \Cref{sec:ensemble_systems}.
    Each system $s$ used $N=15$ models.
    The baseline values in \Cref{eq:error_reduction,eq:lower_bound_reduction} were the followings:
    ER($s_0$): \SI{15.6}{\percent}.
    LB($s_0$) by $\BoundFuncTight(\StrengthTriplet)$: \SI{2.6}{\percent}.
    LB($s_0$) by $\BoundFuncTight(\StrengthDoublet)$: \SI{2.6}{\percent}.
    LB($s_0$) by $\BoundFuncLoose(\StrengthDoublet)$: \SI{-2.1}{\percent}.
    \label{fig:ERR_LBR_scatter_plot_CoLA}
    }
\end{figure*}

\begin{figure*}[h!]
    \begin{subfigure}[t]{0.19\linewidth}
        \vskip 0pt
        \includegraphics[width=\linewidth]{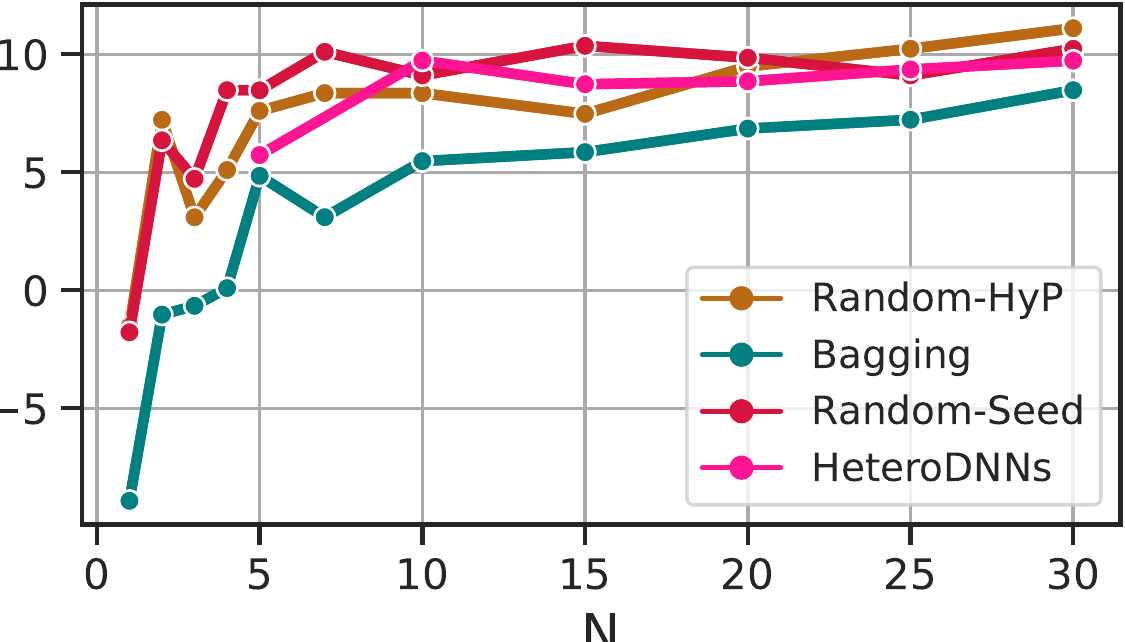}
        \subcaption{Error rate reduction.\label{appendix:fig:scaling.stacking_CoLA}}
    \end{subfigure}
    \hfill
    \begin{subfigure}[t]{0.19\linewidth}
        \vskip 0pt
        \includegraphics[width=\linewidth]{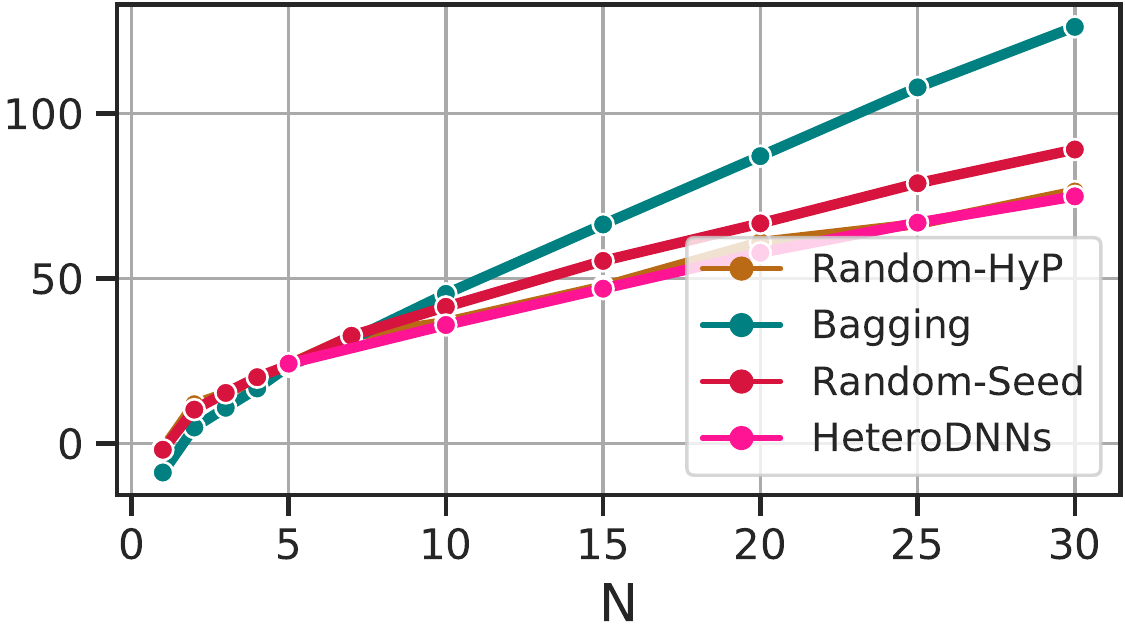}
        \subcaption{Lower bound reduction by \Cref{lemma:ensemble_bound_loose} $\BoundLoose$.\label{appendix:fig:scaling.bound.voting.previous_research_CoLA}}
    \end{subfigure}
    \begin{subfigure}[t]{0.19\linewidth}
        \vskip 0pt
        \includegraphics[width=\linewidth]{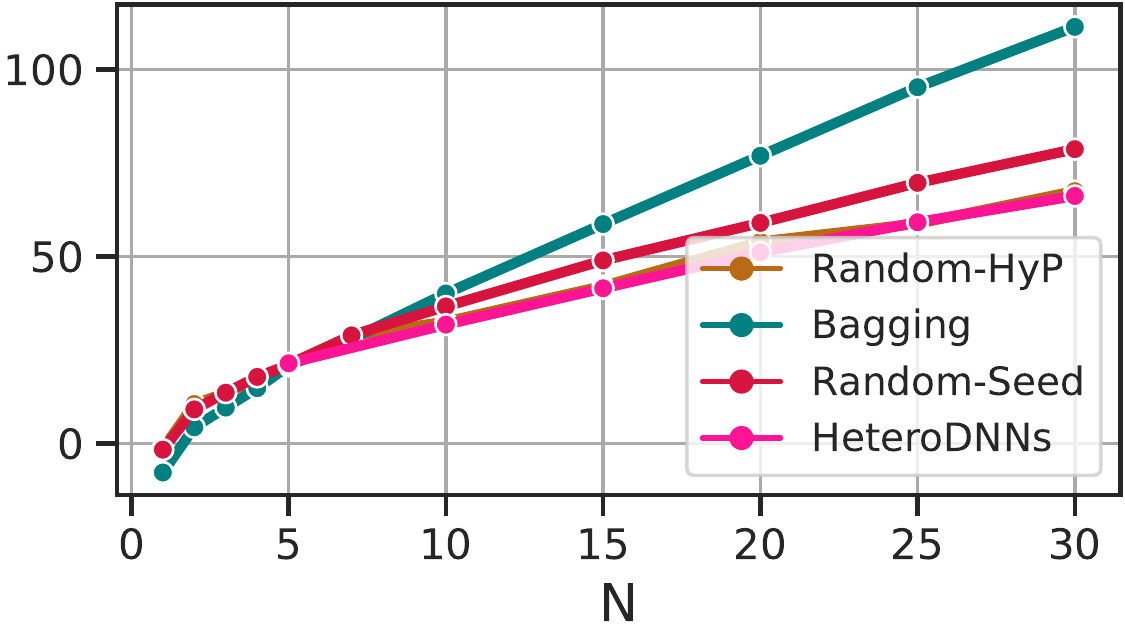}
        \subcaption{Lower bound reduction by $\BoundTightWOCombLoss$. \label{appendix:fig:scaling.bound.voting.ours_wo_combloss_CoLA}}
    \end{subfigure}
    \hfill
     \begin{subfigure}[t]{0.19\linewidth}
        \vskip 0pt
        \includegraphics[width=\linewidth]{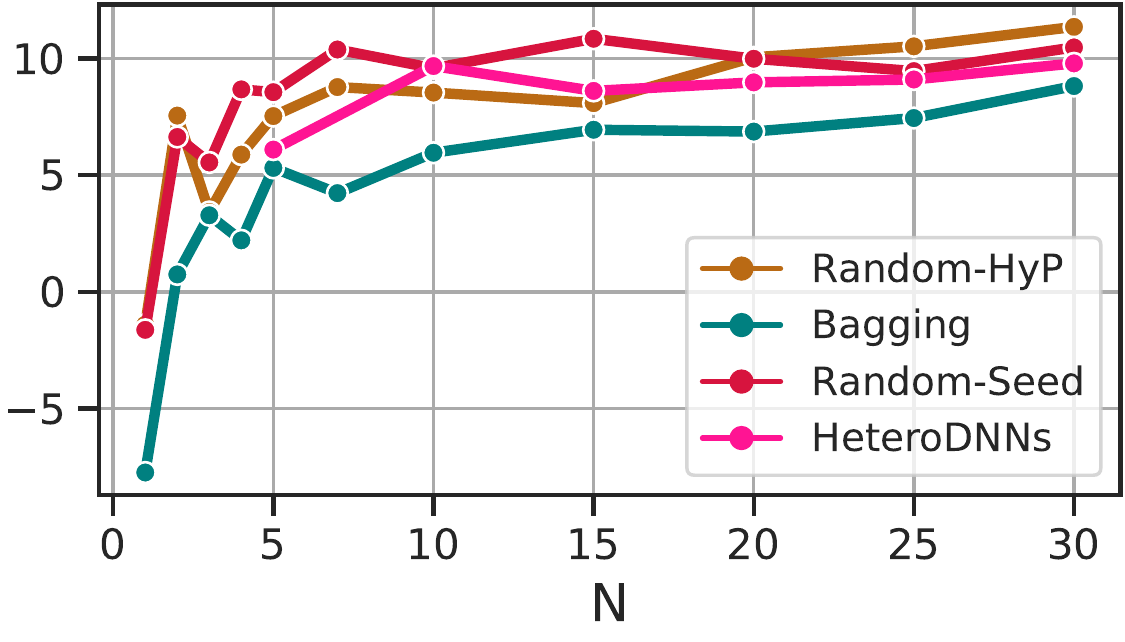}
        \subcaption{Lower bound reduction by \textbf{\Cref{lemma:ensemble_bound_tight}} $\BoundTight$.\label{appendix:fig:scaling.stacking.bound_CoLA}}
    \end{subfigure}
    \hfill
    \begin{subfigure}[t]{0.19\linewidth}
        \vskip 0pt
        \includegraphics[width=\linewidth]{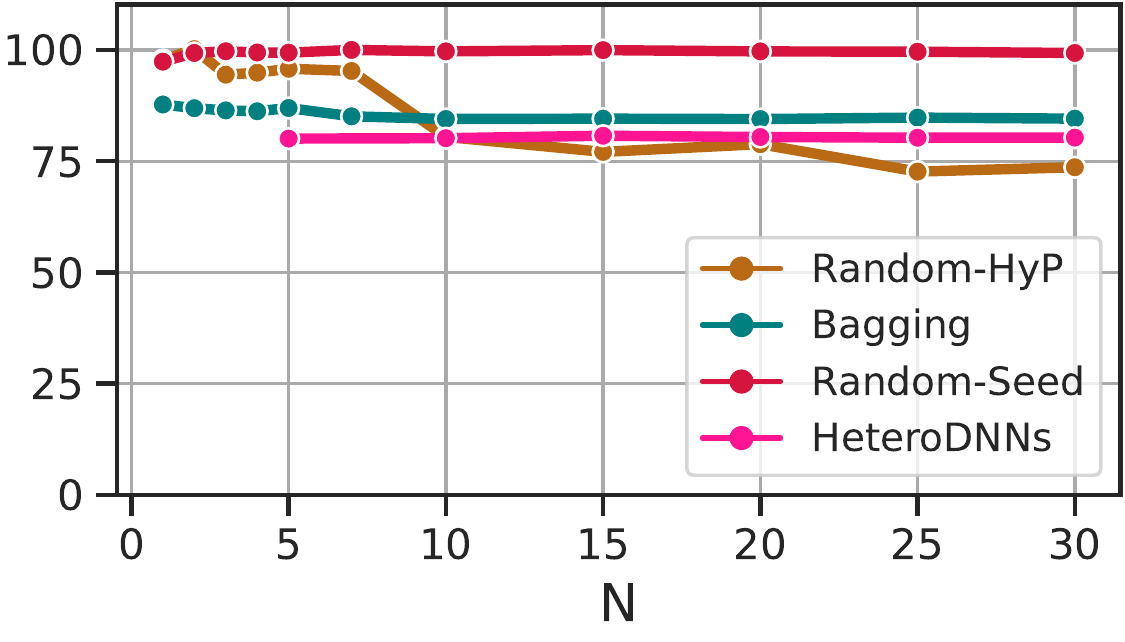}
        \subcaption{$\relev$ \label{appendix:fig:scaling.relevance.per_model_CoLA}}   
    \end{subfigure}   
    \hfill
    \vfill
    \begin{subfigure}[t]{0.19\linewidth}
        \vskip 0pt
        \includegraphics[width=\linewidth]{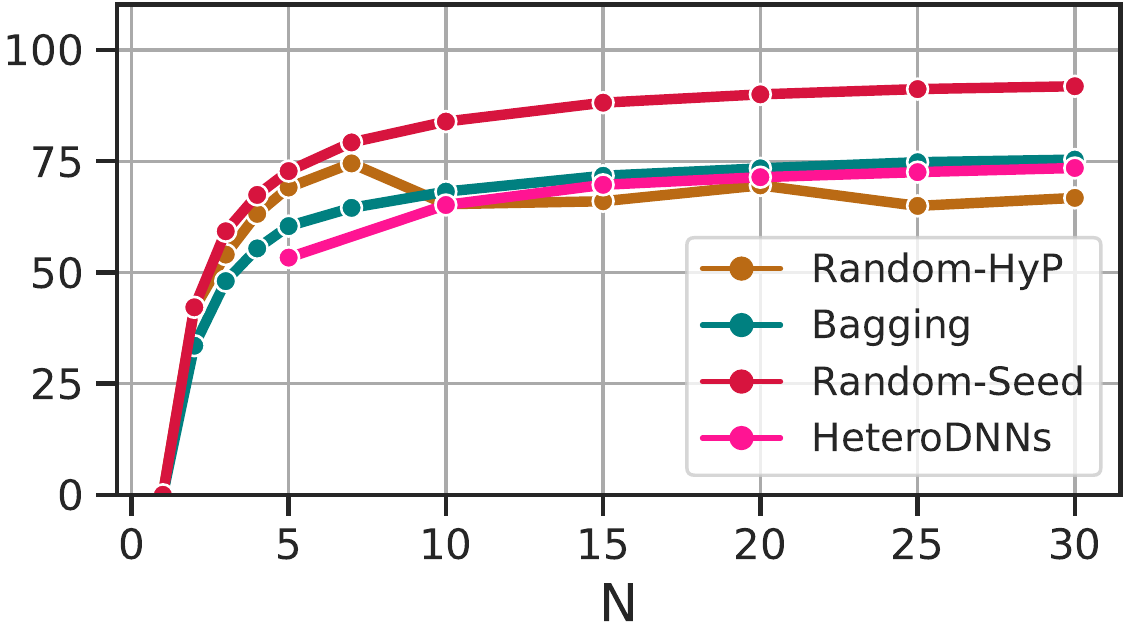}
        \subcaption{$\redun$ \label{appendix:fig:scaling.redundancy.per_model_CoLA}}   
    \end{subfigure}   
    \hfill
    \begin{subfigure}[t]{0.19\linewidth}
        \vskip 0pt
        \includegraphics[width=\linewidth]{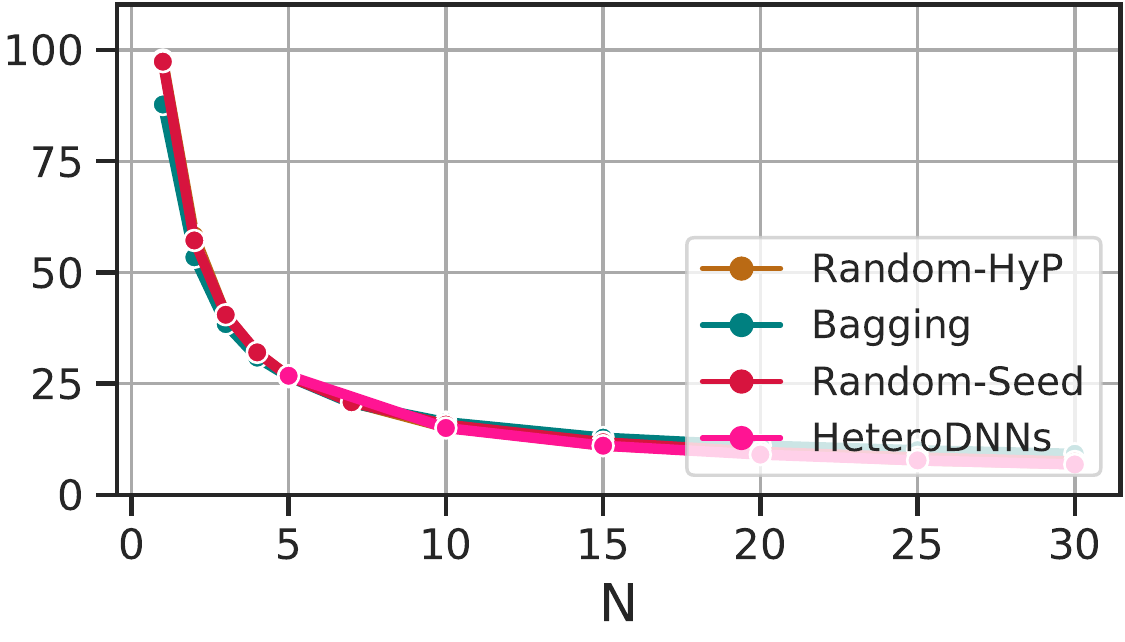}
        \subcaption{$\relev$ $- \redun$ \label{appendix:fig:scaling.novelty.per_model_CoLA}}
    \end{subfigure}   
    \hfill
    \begin{subfigure}[t]{0.19\linewidth}
        \vskip 0pt
        \includegraphics[width=\linewidth]{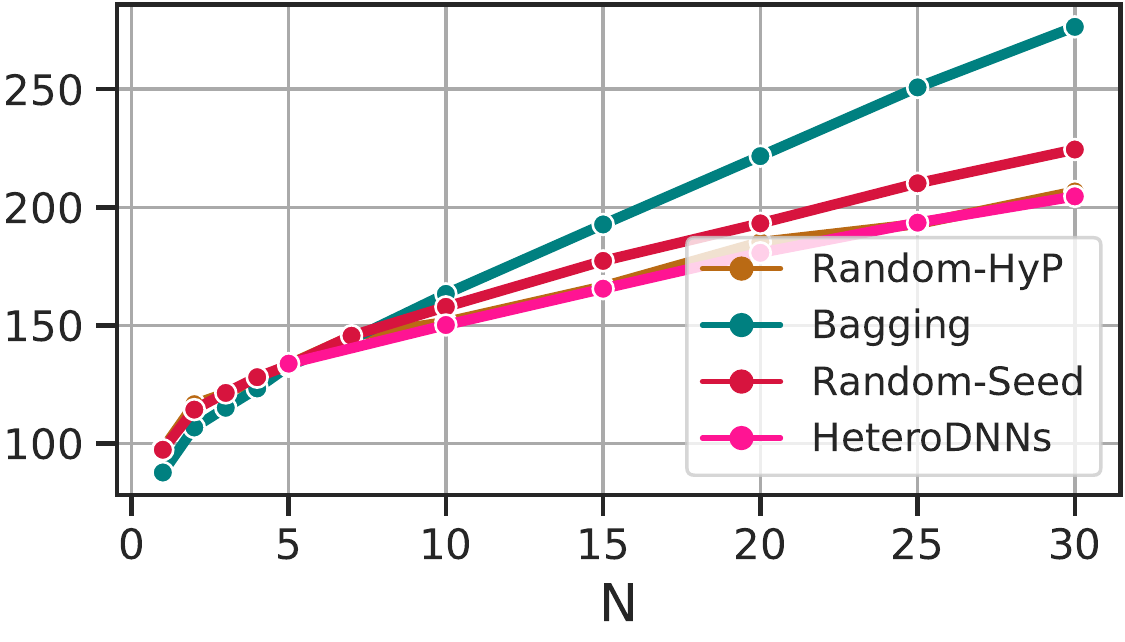}
        \captionsetup{justification=centering}
        \subcaption{$\StrengthDoublet $\newline $= N (\relev - \redun)$ \label{appendix:fig:scaling.E_without_combination_loss_CoLA}}
    \end{subfigure}
    \hfill
    \begin{subfigure}[t]{0.19\linewidth}
        \vskip 0pt
        \includegraphics[width=\linewidth]{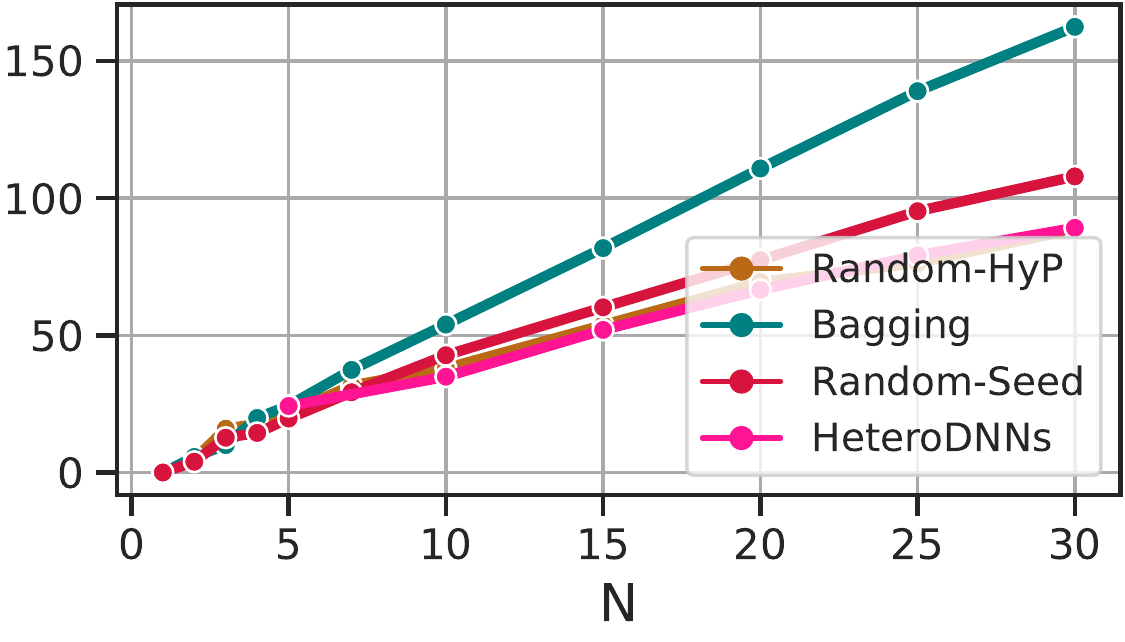}
        \subcaption{$\Combloss$\label{appendix:fig:scaling.combination_loss_CoLA}}
    \end{subfigure}
    \hfill
    \begin{subfigure}[t]{0.19\linewidth}
        \vskip 0pt
        \includegraphics[width=\linewidth]{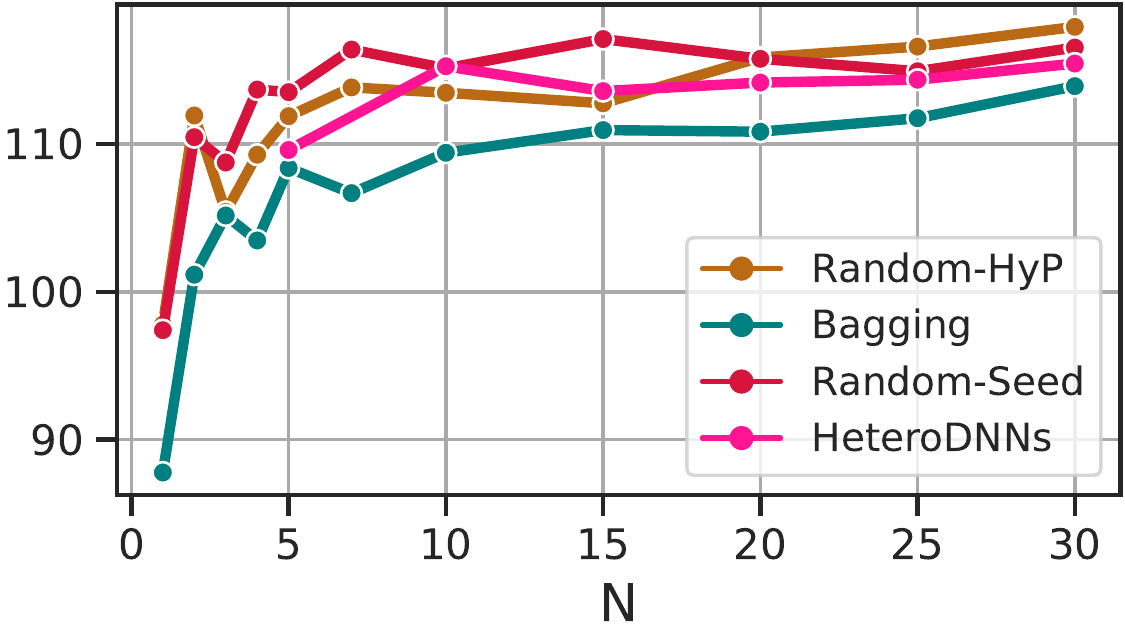}
        \captionsetup{justification=centering}
        \subcaption{$\StrengthTriplet = N (\relev - \redun) - \Combloss $\label{appendix:fig:scaling.E_CoLA}}
    \end{subfigure}
\caption{
\textbf{CoLA task.}
The change in ensemble quantities when the number of models $N$ is changed.
Each figure shows a specific quantity.
The ensemble systems used the SVM model combination.
Each value is an averages of the \NumTasks tasks.
$\perModelMetric$ denotes per-model metric values defined as $\perModelMetricDef$.
\label{appendix:fig:scaling_CoLA}
}
\end{figure*}

\begin{table*}[t]
    \centering
    \caption{
        \textbf{CoLA task}.
        Statistics of ensemble systems described in \Cref{sec:ensemble_systems}.
        The rows and columns list the model generation and combination methods of \Cref{tb:ensemble_methods}, respectively.
        Each cell shows a quantity of a specific system $s$.
        Each quantity is the average over the \NumTasks tasks.
        Each system contains $N=15$ models.
        Color shows the rank within \textit{each column} (brighter is better).
        \label{tb:ablation_CoLA}
    }  
    \begin{subfigure}{\linewidth}
        \centering
        \small
        \tabcolsep 3.0pt
    \subcaption{
        Error rate reductions and lower bound reductions.
        The baseline values used in \Cref{eq:error_reduction,eq:lower_bound_reduction} were the followings.
        ER($s_0$): \SI{15.6}{\percent}.
        LB($s_0$) by $\BoundFuncTight(\StrengthTriplet)$: \SI{2.6}{\percent}.
        LB($s_0$) by $\BoundFuncTight(\StrengthDoublet)$: \SI{2.6}{\percent}.
        LB($s_0$) by $\BoundFuncLoose(\StrengthDoublet)$: \SI{-2.1}{\percent}.
        \label{tb:ablation_errors_CoLA}
    }

\begin{tabular}{lcccccccccccc}
\toprule
& \multicolumn{4}{c}{Error rate reductions \cref{eq:error_reduction}} & &   \multicolumn{7}{c}{Lower bound reductions \Cref{eq:lower_bound_reduction}} \\
\cmidrule(l{\tabcolsep}r{\tabcolsep}){2-5} \cmidrule(l{\tabcolsep}){7-13}
 & \multirow{2}{*}{Voting} & \multirow{2}{*}{LogR} & \multirow{2}{*}{SVM} & \multirow{2}{*}{RForest} & & \multicolumn{4}{c}{\textbf{\Cref{lemma:ensemble_bound_tight}} $\BoundFuncTight(\StrengthTriplet)$} & & \multirow{2}{*}{$\BoundFuncTight(\StrengthDoublet)$} & \multirow{2}{*}{\begin{tabular}{c}\Cref{lemma:ensemble_bound_loose} \\ $\BoundFuncLoose(\StrengthDoublet)$\end{tabular}} \\
 &  &  &  &  & & \multicolumn{1}{c}{Voting} & \multicolumn{1}{c}{LogR} & \multicolumn{1}{c}{SVM} & \multicolumn{1}{c}{RForest} & & &  \\

\midrule
Random-HyP    &   \cThird $\scalebox{1.5}[1.0]{-}1.2_{\pm{\mbox{\tiny 1.3}}}$ &   \cThird $5.6_{\pm{\mbox{\tiny 2.2}}}$ &   \cThird $7.5_{\pm{\mbox{\tiny 1.6}}}$ &                         \cThird $5.2_{\pm{\mbox{\tiny 2.5}}}$ &        &   \cThird $\scalebox{1.5}[1.0]{-}0.9_{\pm{\mbox{\tiny 1.2}}}$ &   \cThird $5.1_{\pm{\mbox{\tiny 2.1}}}$ &   \cThird $8.1_{\pm{\mbox{\tiny 1.6}}}$ &                         \cThird $4.7_{\pm{\mbox{\tiny 2.3}}}$ &        &   \cThird $42_{\pm{\mbox{\tiny 3}}}$ &   \cThird $48_{\pm{\mbox{\tiny 3}}}$ \\
Bagging       &                        \cSecond $0.5_{\pm{\mbox{\tiny 1.5}}}$ &  \cFourth $2.5_{\pm{\mbox{\tiny 1.5}}}$ &  \cFourth $5.9_{\pm{\mbox{\tiny 1.2}}}$ &  \cFourth $\scalebox{1.5}[1.0]{-}2.0_{\pm{\mbox{\tiny 2.3}}}$ &        &                        \cSecond $0.4_{\pm{\mbox{\tiny 1.3}}}$ &  \cFourth $2.1_{\pm{\mbox{\tiny 1.4}}}$ &  \cFourth $6.9_{\pm{\mbox{\tiny 1.6}}}$ &  \cFourth $\scalebox{1.5}[1.0]{-}1.8_{\pm{\mbox{\tiny 2.1}}}$ &        &   \cFirst $59_{\pm{\mbox{\tiny 3}}}$ &   \cFirst $66_{\pm{\mbox{\tiny 3}}}$ \\
Random-Seed   &                         \cFirst $3.9_{\pm{\mbox{\tiny 0.5}}}$ &  \cSecond $6.6_{\pm{\mbox{\tiny 1.7}}}$ &  \cFirst $10.4_{\pm{\mbox{\tiny 0.5}}}$ &                         \cFirst $5.8_{\pm{\mbox{\tiny 1.0}}}$ &        &                         \cFirst $3.4_{\pm{\mbox{\tiny 0.4}}}$ &  \cSecond $5.9_{\pm{\mbox{\tiny 1.6}}}$ &  \cFirst $10.8_{\pm{\mbox{\tiny 1.1}}}$ &                         \cFirst $5.2_{\pm{\mbox{\tiny 0.9}}}$ &        &  \cSecond $49_{\pm{\mbox{\tiny 3}}}$ &  \cSecond $55_{\pm{\mbox{\tiny 3}}}$ \\
Hetero-DNNs &  \cFourth $\scalebox{1.5}[1.0]{-}5.4_{\pm{\mbox{\tiny 3.0}}}$ &   \cFirst $6.8_{\pm{\mbox{\tiny 0.9}}}$ &  \cSecond $8.7_{\pm{\mbox{\tiny 1.0}}}$ &                        \cSecond $5.6_{\pm{\mbox{\tiny 2.0}}}$ &        &  \cFourth $\scalebox{1.5}[1.0]{-}4.8_{\pm{\mbox{\tiny 2.5}}}$ &   \cFirst $6.2_{\pm{\mbox{\tiny 0.9}}}$ &  \cSecond $8.6_{\pm{\mbox{\tiny 0.7}}}$ &                        \cSecond $5.0_{\pm{\mbox{\tiny 1.9}}}$ &        &   \cThird $42_{\pm{\mbox{\tiny 1}}}$ &  \cFourth $47_{\pm{\mbox{\tiny 1}}}$ \\
\bottomrule

\end{tabular}

    \end{subfigure}
    \vfill
    \begin{subfigure}{\linewidth}
    \centering
    \small
    \tabcolsep 2.0pt
    \subcaption{
    Breakdown of ensemble strength defined in \cref{eq:triplet_decomposition}.
    We show per-model metric values defined as $\perModelMetricDef$. Thus, $\StrengthTriplet = (\relev - \redun - \combloss) \ \times N $ holds.
    For intuitive understanding, all the values are normalized by the ensemble strength of baseline $\StrengthTriplet_{s_0}$, for example, $\Relev = \RelevHat / \StrengthTriplet_{s_0} \times 100$ where $\RelevHat$ is the raw value.
    \label{tb:ablation_triple_CoLA}
    }

\begin{tabular}{lccccccccccccc}
\toprule

{} & \multicolumn{4}{c}{\multirow{1}{*}{$\StrengthTripletWithArgs$}} & & \multicolumn{6}{c}{Per-model metric values} & \\
\cmidrule(l{\tabcolsep}r{\tabcolsep}){2-5} \cmidrule(l{\tabcolsep}r{\tabcolsep}){6-12}
{} &  & &  &  & & \multirow{2}{*}{$\perModelMetric_{\normalfont \text{relev}}$} & \multirow{2}{*}{$\perModelMetric_{\normalfont \text{redun}}$} & \multicolumn{4}{c}{ $\perModelMetric_{\normalfont \text{combloss}}$} & & \multirow{2}{*}{$\perModelMetric_{\normalfont \text{relev}} - \perModelMetric_{\normalfont \text{redun}}$} \\
{} & \multicolumn{1}{c}{Voting} & \multicolumn{1}{c}{LogR} & \multicolumn{1}{c}{SVM} & \multicolumn{1}{c}{RForest} & &  {} &  {} &  \multicolumn{1}{c}{Voting} & \multicolumn{1}{c}{LogR} &  \multicolumn{1}{c}{SVM} & \multicolumn{1}{c}{RForest} & &  {} \\

\midrule
Baseline ($s_0$)                    &    \multicolumn{4}{c}{\cBase 100 (the raw value is 0.252)} &        &   \cBase 100 &                                 \cBase 0 &                                \cBase 0 &                                \cBase 0 &                                \cBase 0 &                                 \cBase 0 &        &   \cBase 100 \\

\midrule

Random-HyP                   &    \cThird $98.6_{\pm{\mbox{\tiny 1.8}}}$ &   \cThird $108.1_{\pm{\mbox{\tiny 3.2}}}$ &   \cThird $112.8_{\pm{\mbox{\tiny 2.6}}}$ &   \cThird $107.4_{\pm{\mbox{\tiny 3.6}}}$ &        &  \cFourth $77.1_{\pm{\mbox{\tiny 2.6}}}$ &   \cFirst $66.0_{\pm{\mbox{\tiny 2.3}}}$ &   \cFirst $4.53_{\pm{\mbox{\tiny 0.38}}}$ &  \cSecond $3.89_{\pm{\mbox{\tiny 0.52}}}$ &  \cSecond $3.58_{\pm{\mbox{\tiny 0.25}}}$ &  \cSecond $3.94_{\pm{\mbox{\tiny 0.55}}}$ &        &   \cThird $11.1_{\pm{\mbox{\tiny 3.5}}}$ \\
Bagging                      &  \cSecond $100.6_{\pm{\mbox{\tiny 2.0}}}$ &  \cFourth $103.4_{\pm{\mbox{\tiny 2.2}}}$ &  \cFourth $110.9_{\pm{\mbox{\tiny 2.5}}}$ &   \cFourth $97.1_{\pm{\mbox{\tiny 3.2}}}$ &        &  \cSecond $84.6_{\pm{\mbox{\tiny 0.5}}}$ &   \cThird $71.7_{\pm{\mbox{\tiny 0.2}}}$ &  \cFourth $6.14_{\pm{\mbox{\tiny 0.16}}}$ &  \cFourth $5.96_{\pm{\mbox{\tiny 0.29}}}$ &  \cFourth $5.45_{\pm{\mbox{\tiny 0.44}}}$ &  \cFourth $6.37_{\pm{\mbox{\tiny 0.12}}}$ &        &   \cFirst $12.9_{\pm{\mbox{\tiny 0.5}}}$ \\
Random-Seed                  &   \cFirst $105.3_{\pm{\mbox{\tiny 0.6}}}$ &  \cSecond $109.3_{\pm{\mbox{\tiny 2.5}}}$ &   \cFirst $117.1_{\pm{\mbox{\tiny 1.6}}}$ &   \cFirst $108.3_{\pm{\mbox{\tiny 1.5}}}$ &        &  \cFirst $100.0_{\pm{\mbox{\tiny 0.0}}}$ &  \cFourth $88.2_{\pm{\mbox{\tiny 0.3}}}$ &  \cSecond $4.80_{\pm{\mbox{\tiny 0.22}}}$ &   \cThird $4.53_{\pm{\mbox{\tiny 0.32}}}$ &   \cThird $4.01_{\pm{\mbox{\tiny 0.17}}}$ &   \cThird $4.61_{\pm{\mbox{\tiny 0.33}}}$ &        &  \cSecond $11.8_{\pm{\mbox{\tiny 0.3}}}$ \\
Hetero-DNNs                &   \cFourth $92.4_{\pm{\mbox{\tiny 4.0}}}$ &   \cFirst $109.8_{\pm{\mbox{\tiny 1.4}}}$ &  \cSecond $113.6_{\pm{\mbox{\tiny 1.1}}}$ &  \cSecond $107.9_{\pm{\mbox{\tiny 3.0}}}$ &        &   \cThird $80.7_{\pm{\mbox{\tiny 0.4}}}$ &  \cSecond $69.7_{\pm{\mbox{\tiny 0.4}}}$ &   \cThird $4.88_{\pm{\mbox{\tiny 0.23}}}$ &   \cFirst $3.72_{\pm{\mbox{\tiny 0.08}}}$ &   \cFirst $3.47_{\pm{\mbox{\tiny 0.05}}}$ &   \cFirst $3.84_{\pm{\mbox{\tiny 0.18}}}$ &        &  \cFourth $11.0_{\pm{\mbox{\tiny 0.6}}}$ \\

\bottomrule
\end{tabular}

\end{subfigure}

\end{table*}

\clearpage

\begin{figure*}[t!]
    \begin{subfigure}[t]{0.32\linewidth}
        \vskip 0pt
        \centering
        \includegraphics[width=0.65\linewidth]{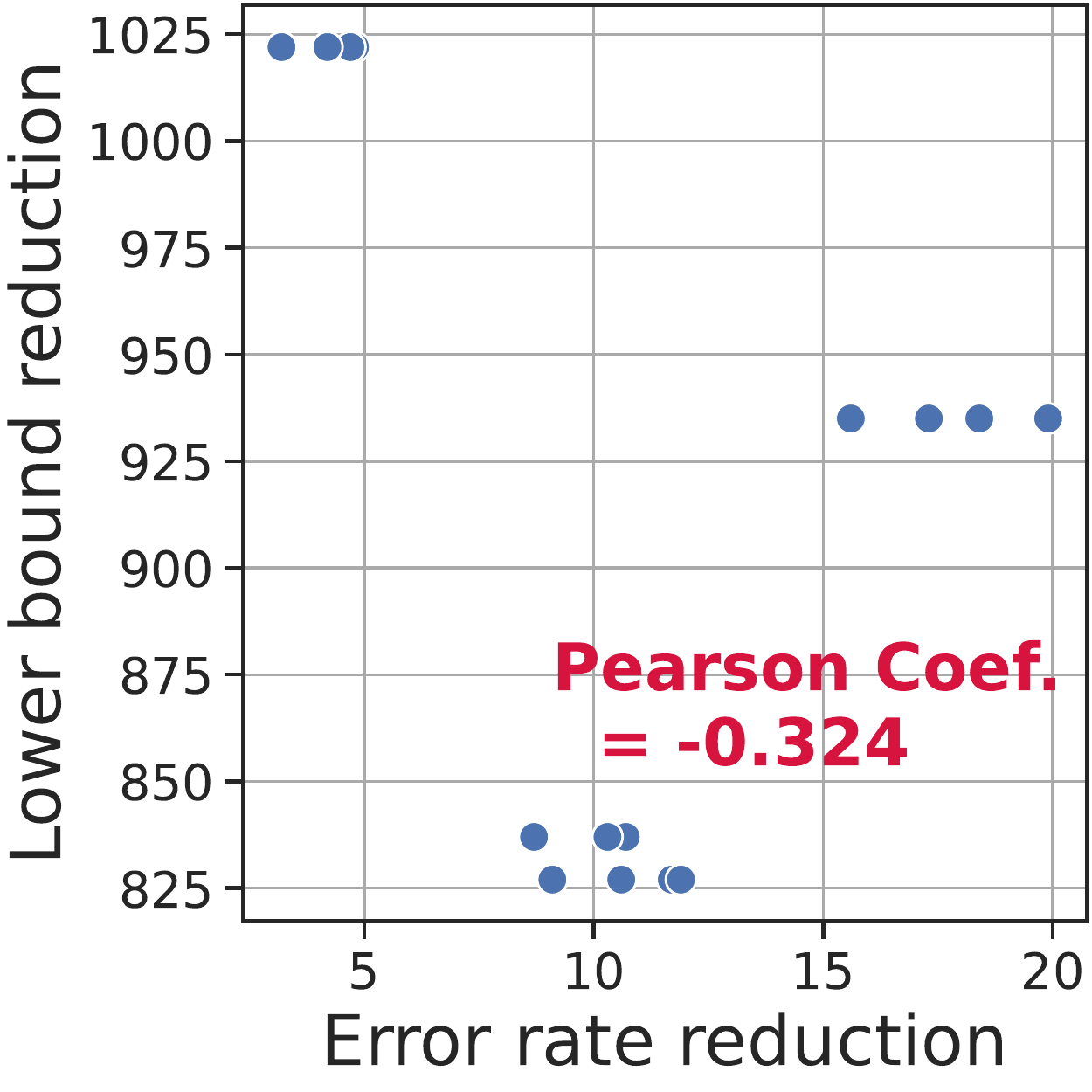}
        \caption{
        \Cref{lemma:ensemble_bound_loose} $\BoundLoose$.
        \label{fig:ERR_LBR_scatter_plot_loose_CosmosQA}
        }
    \end{subfigure}
    \hfill
    \begin{subfigure}[t]{0.32\linewidth}
        \vskip 0pt
        \centering
        \includegraphics[width=0.65\linewidth]{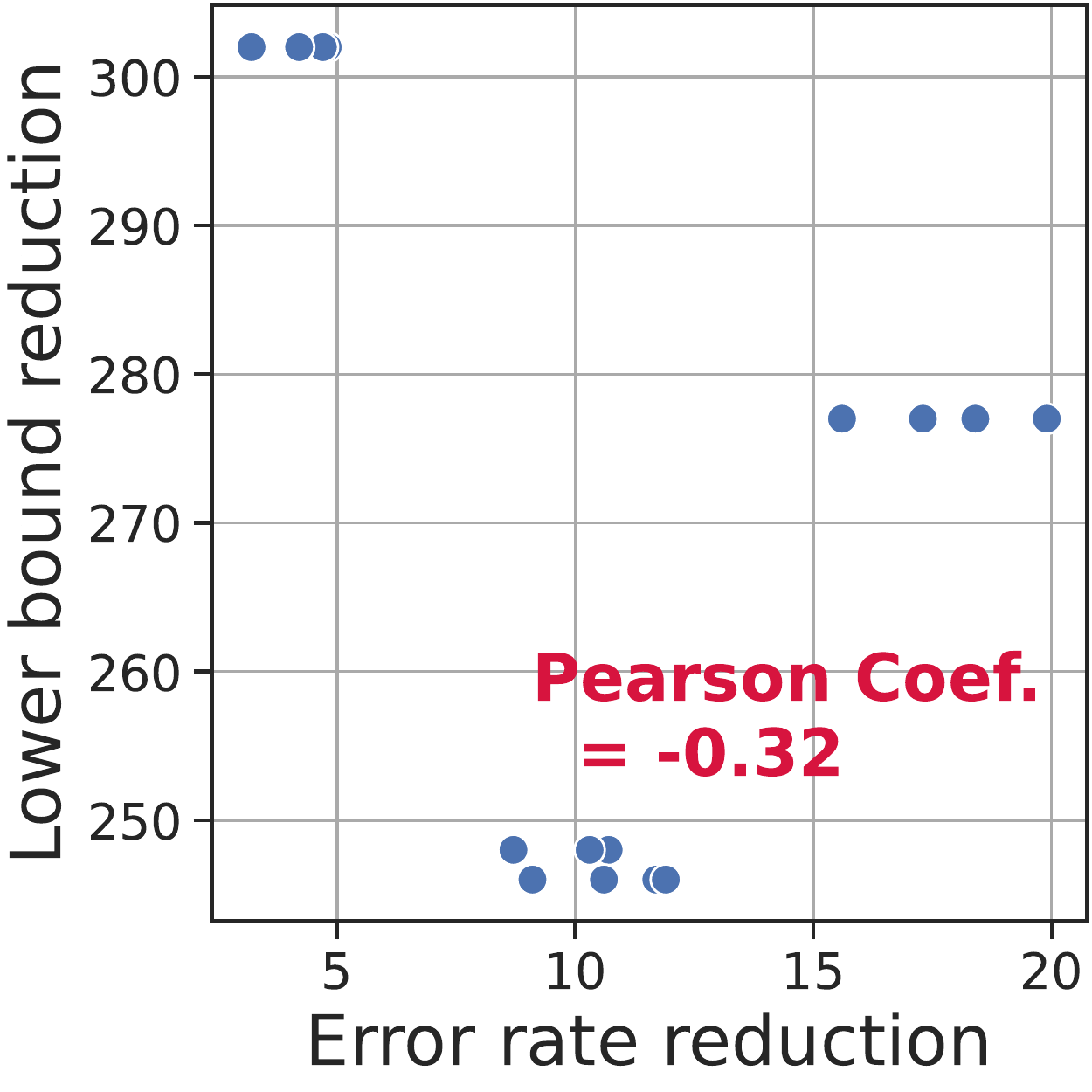}
        \caption{
            $\BoundTightWOCombLoss$.
            \label{fig:ERR_LBR_scatter_plot_tight_wo_combloss_CosmosQA}
        }
    \end{subfigure}
    \hfill
    \begin{subfigure}[t]{0.32\linewidth}
        \vskip 0pt
        \centering
        \includegraphics[width=0.65\linewidth]{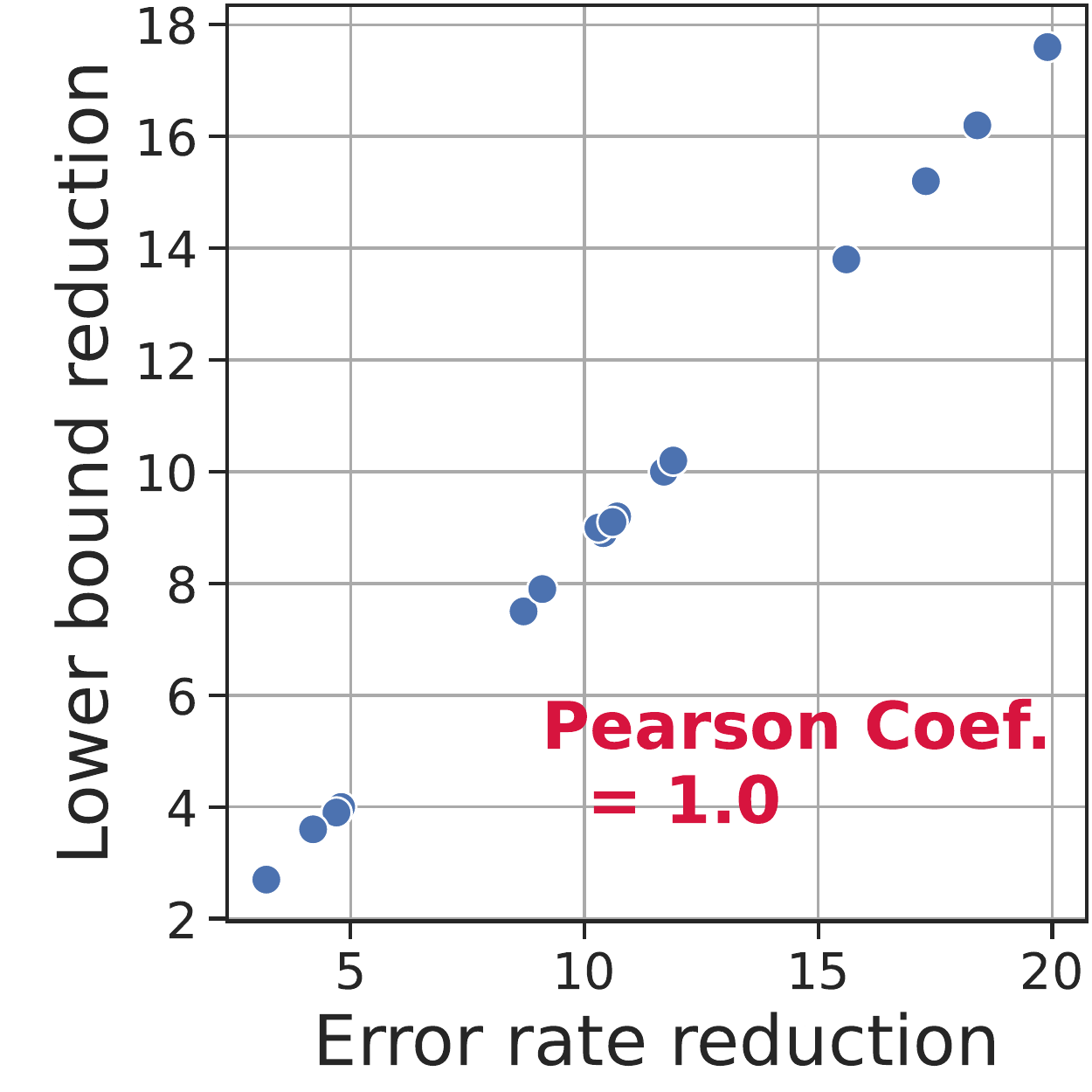}
        \caption{
            \textbf{\Cref{lemma:ensemble_bound_tight}} $\BoundTight$.
            \label{fig:ERR_LBR_scatter_plot_tight_CosmosQA}
        }
    \end{subfigure}
    \hfill

    \caption{
    \textbf{CosmosQA task.}
    Correlations between error rate reductions and lower bound reductions.
    Each figure uses different type of lower bound.
    Each point in the figures shows a quantity of a specific ensemble system $s$ and the quantity is the average over the \NumTasks tasks.
    See \Cref{tb:ablation_CosmosQA} for the real value of each point.
    We used the \NumSystems ensemble systems described in \Cref{sec:ensemble_systems}.
    Each system $s$ used $N=15$ models.
    The baseline values in \Cref{eq:error_reduction,eq:lower_bound_reduction} were the followings:
    ER($s_0$): \SI{28.2}{\percent}.
    LB($s_0$) by $\BoundFuncTight(\StrengthTriplet)$: \SI{6.2}{\percent}.
    LB($s_0$) by $\BoundFuncTight(\StrengthDoublet)$: \SI{6.2}{\percent}.
    LB($s_0$) by $\BoundFuncLoose(\StrengthDoublet)$: \SI{2.0}{\percent}.
    \label{fig:ERR_LBR_scatter_plot_CosmosQA}
    }
\end{figure*}

\begin{figure*}[h!]
    \begin{subfigure}[t]{0.19\linewidth}
        \vskip 0pt
        \includegraphics[width=\linewidth]{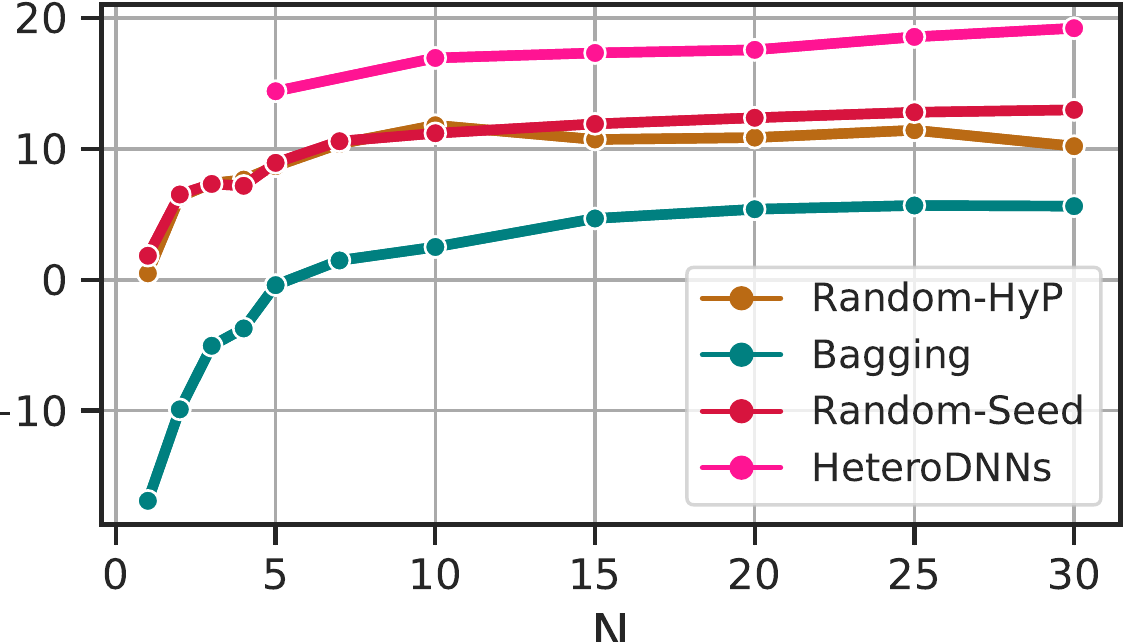}
        \subcaption{Error rate reduction.\label{appendix:fig:scaling.stacking_CosmosQA}}
    \end{subfigure}
    \hfill
    \begin{subfigure}[t]{0.19\linewidth}
        \vskip 0pt
        \includegraphics[width=\linewidth]{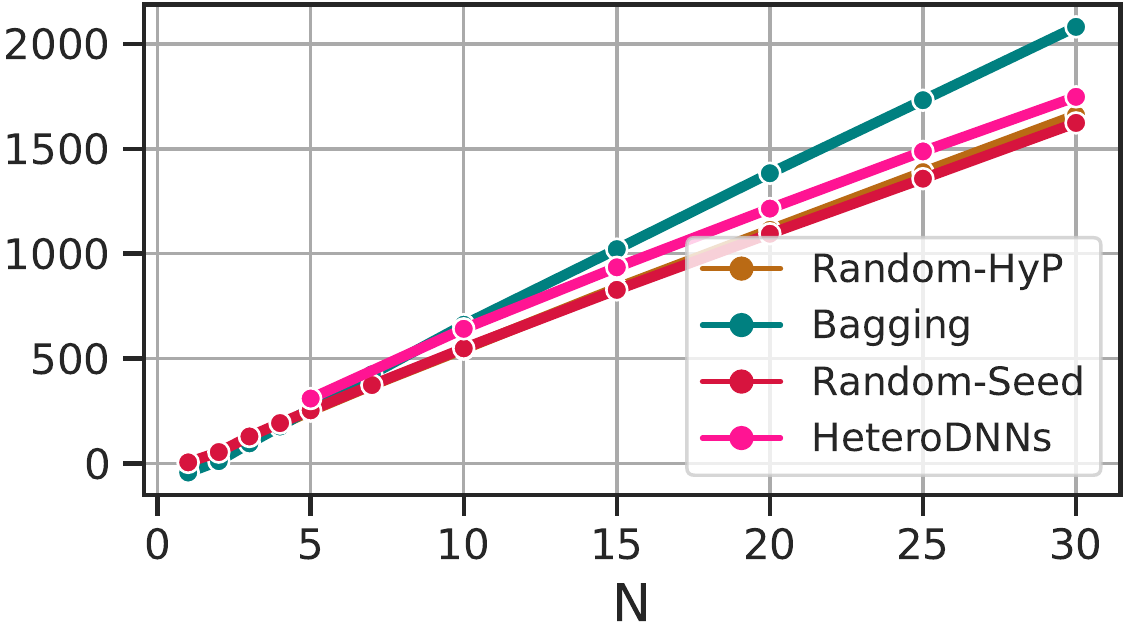}
        \subcaption{Lower bound reduction by \Cref{lemma:ensemble_bound_loose} $\BoundLoose$.\label{appendix:fig:scaling.bound.voting.previous_research_CosmosQA}}
    \end{subfigure}
    \begin{subfigure}[t]{0.19\linewidth}
        \vskip 0pt
        \includegraphics[width=\linewidth]{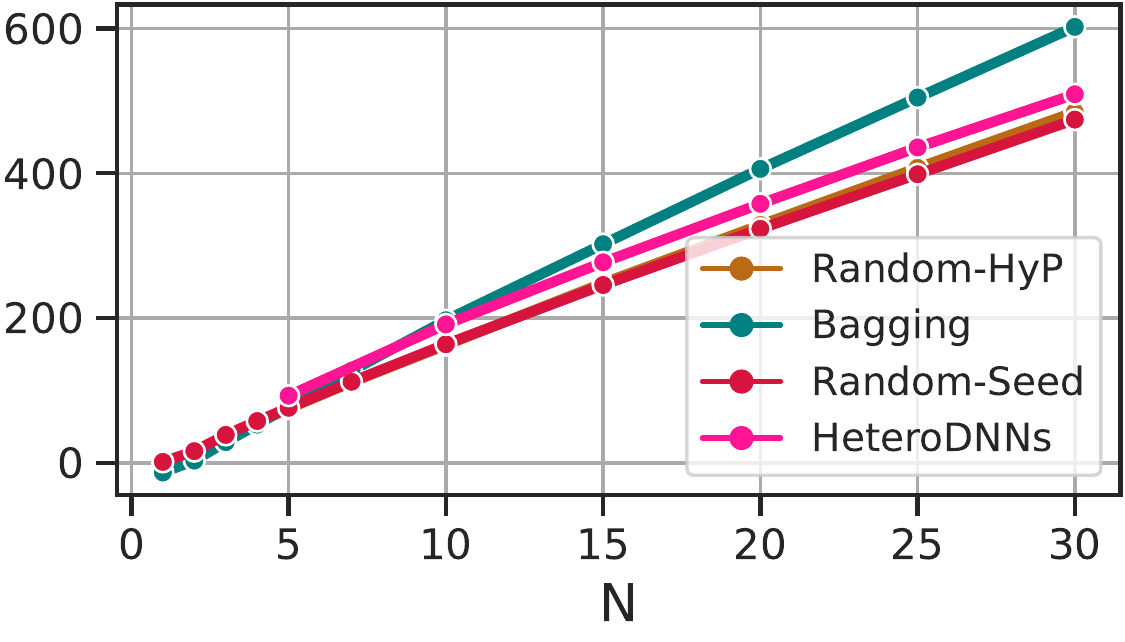}
        \subcaption{Lower bound reduction by $\BoundTightWOCombLoss$. \label{appendix:fig:scaling.bound.voting.ours_wo_combloss_CosmosQA}}
    \end{subfigure}
    \hfill
     \begin{subfigure}[t]{0.19\linewidth}
        \vskip 0pt
        \includegraphics[width=\linewidth]{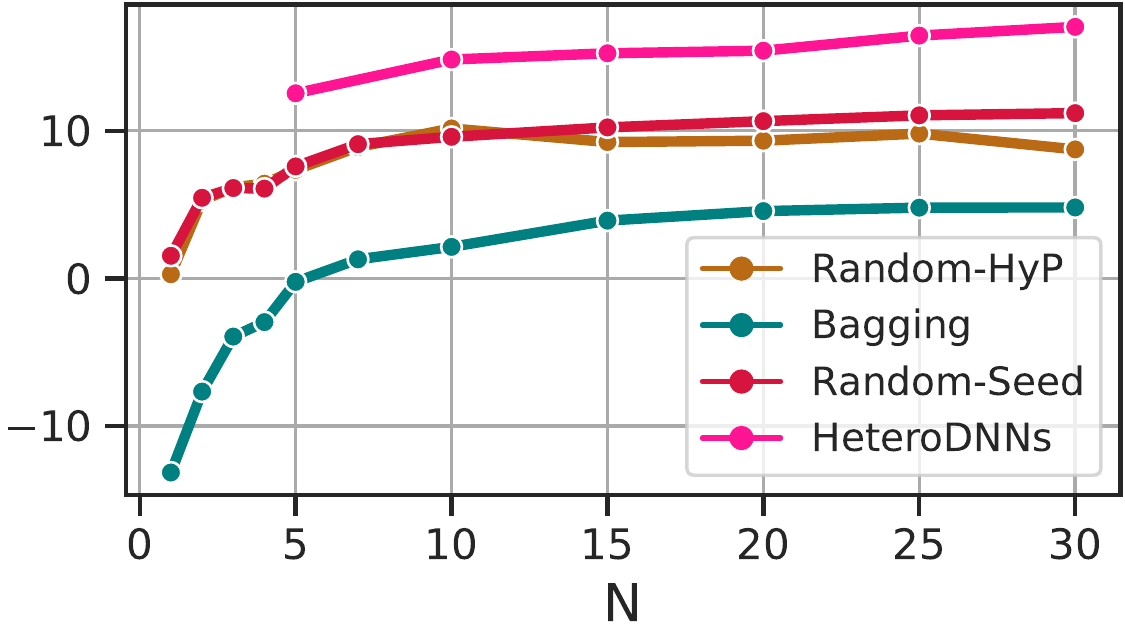}
        \subcaption{Lower bound reduction by \textbf{\Cref{lemma:ensemble_bound_tight}} $\BoundTight$.\label{appendix:fig:scaling.stacking.bound_CosmosQA}}
    \end{subfigure}
    \hfill
    \begin{subfigure}[t]{0.19\linewidth}
        \vskip 0pt
        \includegraphics[width=\linewidth]{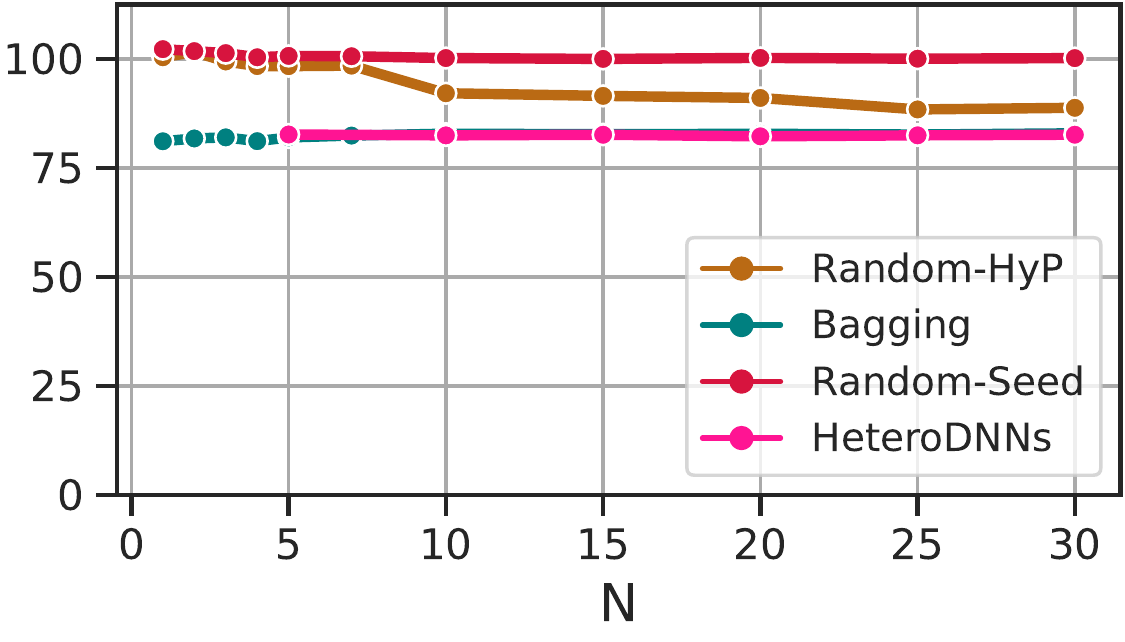}
        \subcaption{$\relev$ \label{appendix:fig:scaling.relevance.per_model_CosmosQA}}   
    \end{subfigure}   
    \hfill
    \vfill
    \begin{subfigure}[t]{0.19\linewidth}
        \vskip 0pt
        \includegraphics[width=\linewidth]{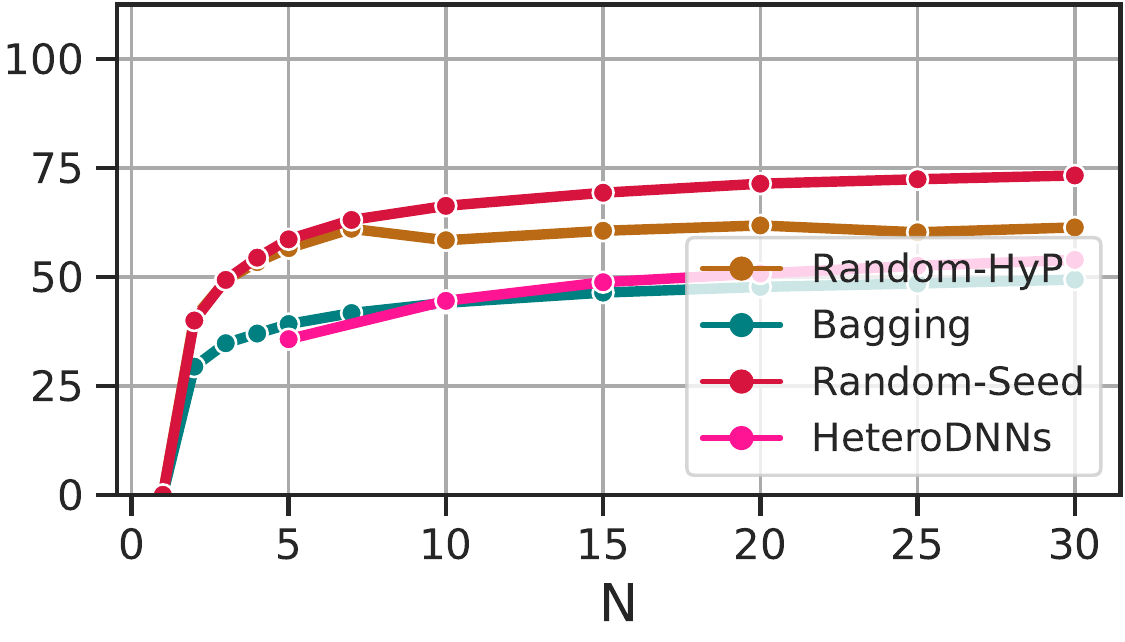}
        \subcaption{$\redun$ \label{appendix:fig:scaling.redundancy.per_model_CosmosQA}}   
    \end{subfigure}   
    \hfill
    \begin{subfigure}[t]{0.19\linewidth}
        \vskip 0pt
        \includegraphics[width=\linewidth]{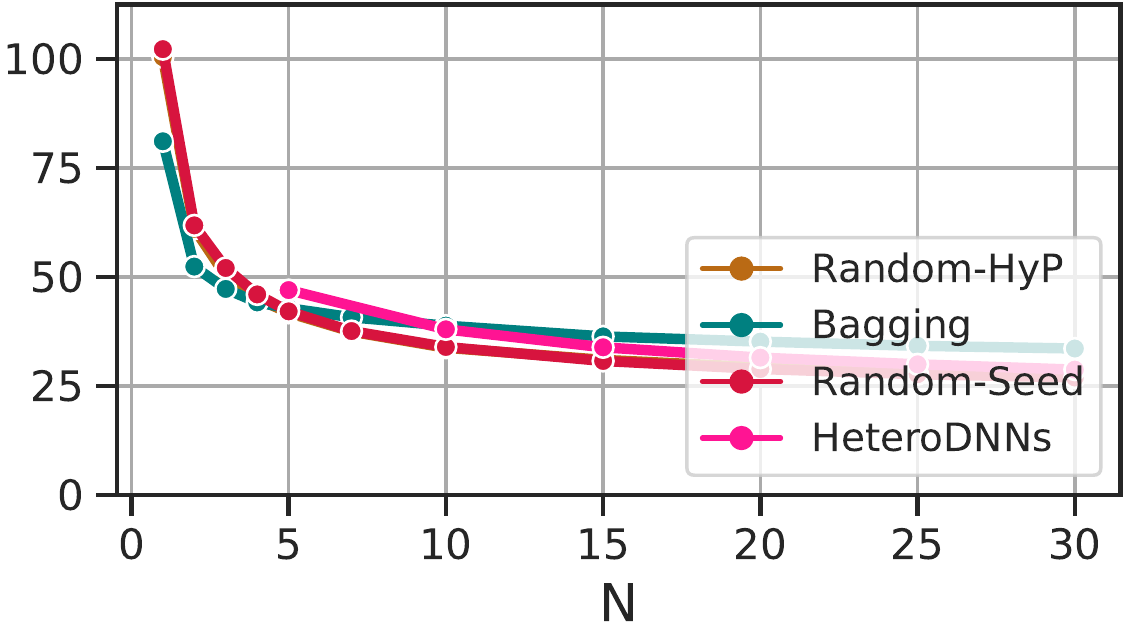}
        \subcaption{$\relev$ $- \redun$ \label{appendix:fig:scaling.novelty.per_model_CosmosQA}}
    \end{subfigure}   
    \hfill
    \begin{subfigure}[t]{0.19\linewidth}
        \vskip 0pt
        \includegraphics[width=\linewidth]{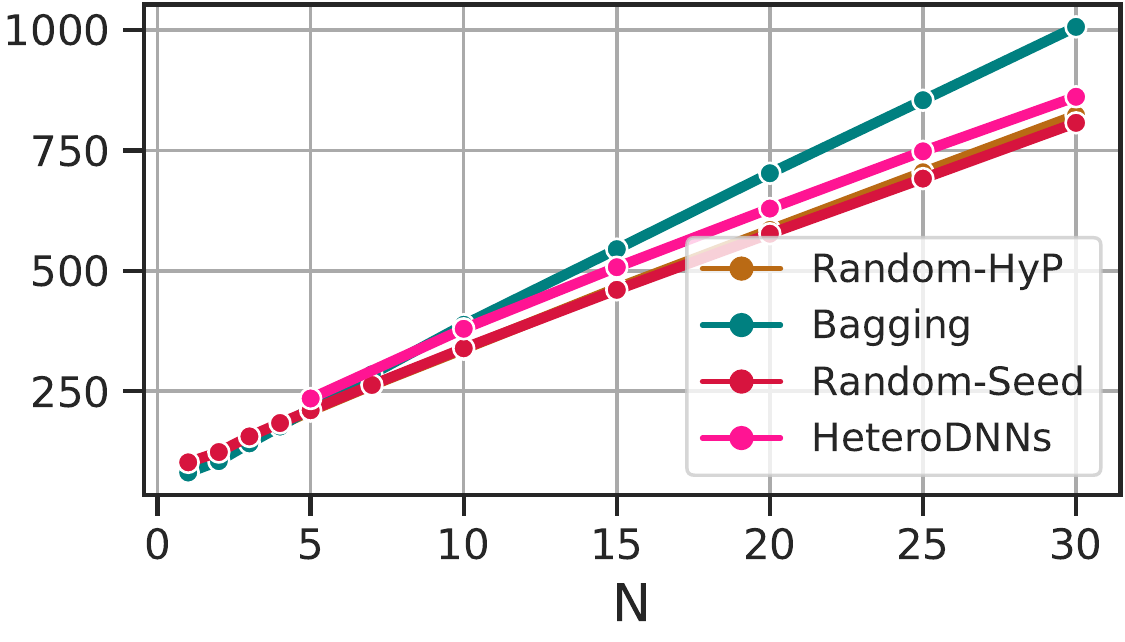}
        \captionsetup{justification=centering}
        \subcaption{$\StrengthDoublet $\newline $= N (\relev - \redun)$ \label{appendix:fig:scaling.E_without_combination_loss_CosmosQA}}
    \end{subfigure}
    \hfill
    \begin{subfigure}[t]{0.19\linewidth}
        \vskip 0pt
        \includegraphics[width=\linewidth]{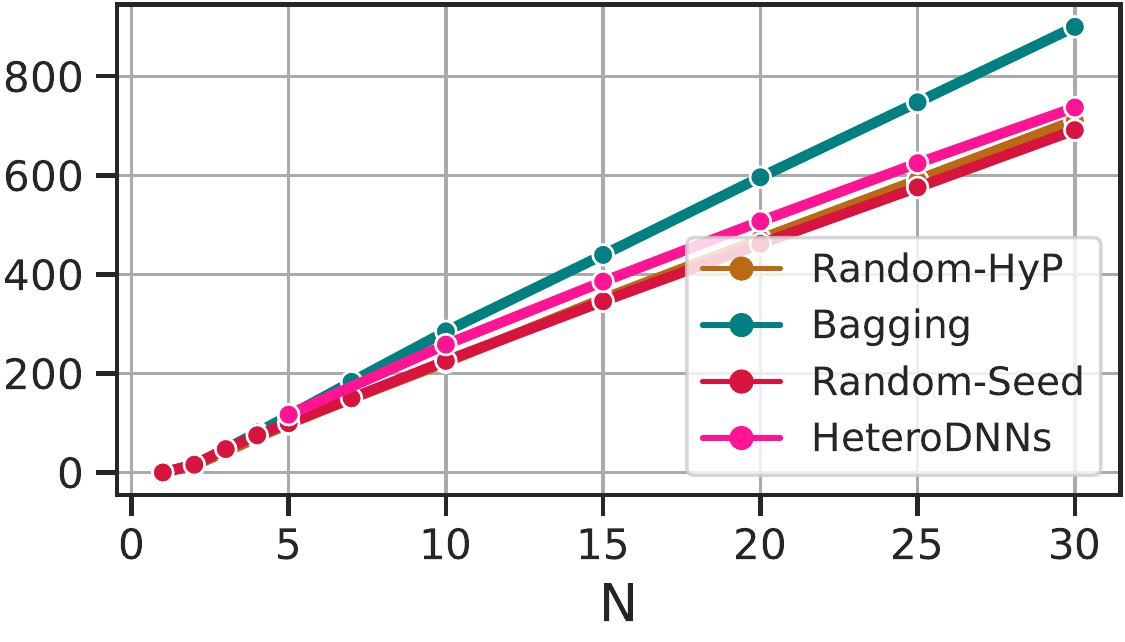}
        \subcaption{$\Combloss$\label{appendix:fig:scaling.combination_loss_CosmosQA}}
    \end{subfigure}
    \hfill
    \begin{subfigure}[t]{0.19\linewidth}
        \vskip 0pt
        \includegraphics[width=\linewidth]{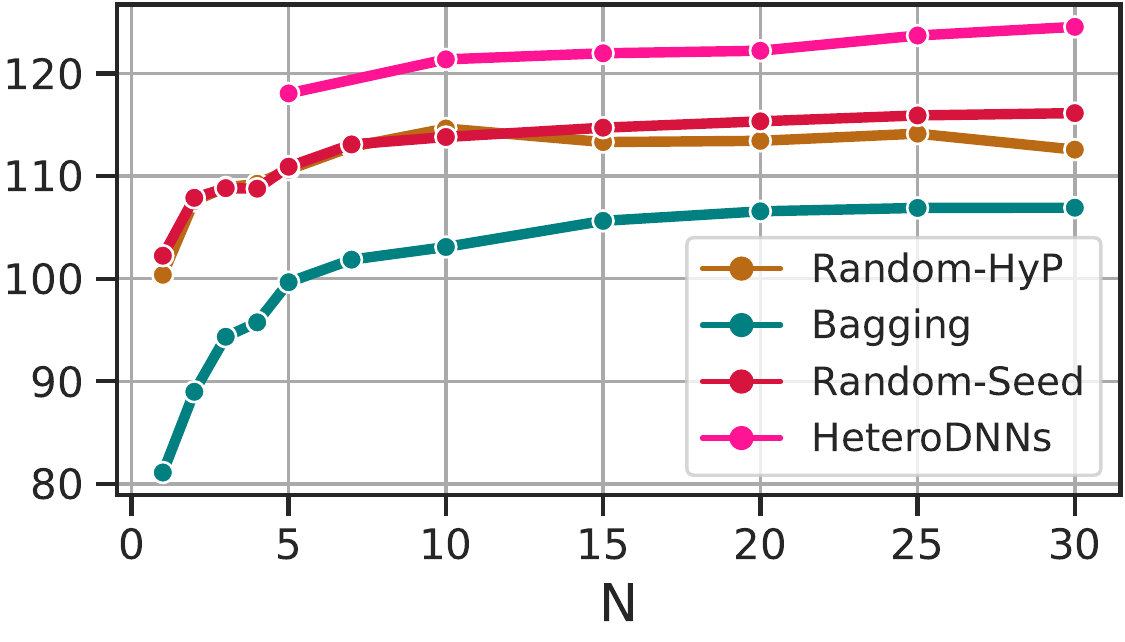}
        \captionsetup{justification=centering}
        \subcaption{$\StrengthTriplet = N (\relev - \redun) - \Combloss $\label{appendix:fig:scaling.E_CosmosQA}}
    \end{subfigure}
\caption{
\textbf{CosmosQA task.}
The change in ensemble quantities when the number of models $N$ is changed.
Each figure shows a specific quantity.
The ensemble systems used the SVM model combination.
Each value is an averages of the \NumTasks tasks.
$\perModelMetric$ denotes per-model metric values defined as: $\perModelMetricDef$.
\label{appendix:fig:scaling_CosmosQA}
}
\end{figure*}

\begin{table*}[h]
    \centering
    \caption{
        \textbf{CosmosQA task}.
        Statistics of ensemble systems described in \Cref{sec:ensemble_systems}.
        The rows and columns list the model generation and combination methods of \Cref{tb:ensemble_methods}, respectively.
        Each cell shows a quantity of a specific system $s$.
        Each quantity is the average over the \NumTasks tasks.
        Each system contains $N=15$ models.
        Color shows the rank within \textit{each column} (brighter is better).
        \label{tb:ablation_CosmosQA}
    }  
    \begin{subfigure}{\linewidth}
        \centering
        \small
        \tabcolsep 3.0pt
    \subcaption{
        Error rate reductions and lower bound reductions.
        The baseline values used in \Cref{eq:error_reduction,eq:lower_bound_reduction} were the followings.
        ER($s_0$): \SI{28.2}{\percent}.
        LB($s_0$) by $\BoundFuncTight(\StrengthTriplet)$: \SI{6.2}{\percent}.
        LB($s_0$) by $\BoundFuncTight(\StrengthDoublet)$: \SI{6.2}{\percent}.
        LB($s_0$) by $\BoundFuncLoose(\StrengthDoublet)$: \SI{2.0}{\percent}.
        \label{tb:ablation_errors_CosmosQA}
    }

\begin{tabular}{lcccccccccccc}
\toprule
& \multicolumn{4}{c}{Error rate reductions \cref{eq:error_reduction}} & &   \multicolumn{7}{c}{Lower bound reductions \Cref{eq:lower_bound_reduction}} \\
\cmidrule(l{\tabcolsep}r{\tabcolsep}){2-5} \cmidrule(l{\tabcolsep}){7-13}
 & \multirow{2}{*}{Voting} & \multirow{2}{*}{LogR} & \multirow{2}{*}{SVM} & \multirow{2}{*}{RForest} & & \multicolumn{4}{c}{\textbf{\Cref{lemma:ensemble_bound_tight}} $\BoundFuncTight(\StrengthTriplet)$} & & \multirow{2}{*}{$\BoundFuncTight(\StrengthDoublet)$} & \multirow{2}{*}{\begin{tabular}{c}\Cref{lemma:ensemble_bound_loose} \\ $\BoundFuncLoose(\StrengthDoublet)$\end{tabular}} \\
 &  &  &  &  & & \multicolumn{1}{c}{Voting} & \multicolumn{1}{c}{LogR} & \multicolumn{1}{c}{SVM} & \multicolumn{1}{c}{RForest} & & &  \\

\midrule
Random-HyP    &    \cThird $8.7_{\pm{\mbox{\tiny 1.0}}}$ &   \cThird $10.4_{\pm{\mbox{\tiny 1.0}}}$ &   \cThird $10.7_{\pm{\mbox{\tiny 1.4}}}$ &  \cSecond $10.3_{\pm{\mbox{\tiny 0.6}}}$ &        &   \cThird $7.5_{\pm{\mbox{\tiny 0.9}}}$ &    \cThird $8.9_{\pm{\mbox{\tiny 0.8}}}$ &    \cThird $9.2_{\pm{\mbox{\tiny 1.1}}}$ &  \cSecond $9.0_{\pm{\mbox{\tiny 0.5}}}$ &        &   \cThird $248_{\pm{\mbox{\tiny 7}}}$ &   \cThird $837_{\pm{\mbox{\tiny 34}}}$ \\
Bagging       &   \cFourth $3.2_{\pm{\mbox{\tiny 0.8}}}$ &   \cFourth $4.8_{\pm{\mbox{\tiny 0.2}}}$ &   \cFourth $4.7_{\pm{\mbox{\tiny 0.3}}}$ &   \cFourth $4.2_{\pm{\mbox{\tiny 0.7}}}$ &        &  \cFourth $2.7_{\pm{\mbox{\tiny 0.7}}}$ &   \cFourth $4.0_{\pm{\mbox{\tiny 0.2}}}$ &   \cFourth $3.9_{\pm{\mbox{\tiny 0.2}}}$ &  \cFourth $3.6_{\pm{\mbox{\tiny 0.6}}}$ &        &   \cFirst $302_{\pm{\mbox{\tiny 3}}}$ &  \cFirst $1022_{\pm{\mbox{\tiny 20}}}$ \\
Random-Seed   &  \cSecond $10.6_{\pm{\mbox{\tiny 1.1}}}$ &  \cSecond $11.7_{\pm{\mbox{\tiny 0.3}}}$ &  \cSecond $11.9_{\pm{\mbox{\tiny 0.6}}}$ &    \cThird $9.1_{\pm{\mbox{\tiny 2.8}}}$ &        &  \cSecond $9.1_{\pm{\mbox{\tiny 1.0}}}$ &  \cSecond $10.0_{\pm{\mbox{\tiny 0.3}}}$ &  \cSecond $10.2_{\pm{\mbox{\tiny 0.6}}}$ &   \cThird $7.9_{\pm{\mbox{\tiny 2.2}}}$ &        &  \cFourth $246_{\pm{\mbox{\tiny 1}}}$ &   \cFourth $827_{\pm{\mbox{\tiny 7}}}$ \\
Hetero-DNNs &   \cFirst $15.6_{\pm{\mbox{\tiny 1.3}}}$ &   \cFirst $18.4_{\pm{\mbox{\tiny 0.7}}}$ &   \cFirst $17.3_{\pm{\mbox{\tiny 0.7}}}$ &   \cFirst $19.9_{\pm{\mbox{\tiny 1.2}}}$ &        &  \cFirst $13.8_{\pm{\mbox{\tiny 1.1}}}$ &   \cFirst $16.2_{\pm{\mbox{\tiny 0.6}}}$ &   \cFirst $15.2_{\pm{\mbox{\tiny 0.6}}}$ &  \cFirst $17.6_{\pm{\mbox{\tiny 1.1}}}$ &        &  \cSecond $277_{\pm{\mbox{\tiny 5}}}$ &   \cSecond $935_{\pm{\mbox{\tiny 9}}}$ \\
\bottomrule

\end{tabular}

    \end{subfigure}
    \vfill
    \begin{subfigure}{\linewidth}
    \centering
    \small
    \tabcolsep 1.0pt
    \subcaption{
    Breakdown of ensemble strength defined in \cref{eq:triplet_decomposition}.
    We show per-model metric values defined as: $\perModelMetricDef$. Thus, $\StrengthTriplet = (\relev - \redun - \combloss) \ \times N $ holds.
    For intuitive understanding, all the values are normalized by the ensemble strength of baseline $\StrengthTriplet_{s_0}$, for example, $\Relev = \RelevHat / \StrengthTriplet_{s_0} \times 100$ where $\RelevHat$ is the raw value.
    \label{tb:ablation_triple_CosmosQA}
    }

\begin{tabular}{lccccccccccccc}
\toprule

{} & \multicolumn{4}{c}{\multirow{1}{*}{$\StrengthTripletWithArgs$}} & & \multicolumn{6}{c}{Per-model metric values} & \\
\cmidrule(l{\tabcolsep}r{\tabcolsep}){2-5} \cmidrule(l{\tabcolsep}r{\tabcolsep}){6-12}
{} &  & &  &  & & \multirow{2}{*}{$\perModelMetric_{\normalfont \text{relev}}$} & \multirow{2}{*}{$\perModelMetric_{\normalfont \text{redun}}$} & \multicolumn{4}{c}{ $\perModelMetric_{\normalfont \text{combloss}}$} & & \multirow{2}{*}{$\perModelMetric_{\normalfont \text{relev}} - \perModelMetric_{\normalfont \text{redun}}$} \\
{} & \multicolumn{1}{c}{Voting} & \multicolumn{1}{c}{LogR} & \multicolumn{1}{c}{SVM} & \multicolumn{1}{c}{RForest} & &  {} &  {} &  \multicolumn{1}{c}{Voting} & \multicolumn{1}{c}{LogR} &  \multicolumn{1}{c}{SVM} & \multicolumn{1}{c}{RForest} & &  {} \\

\midrule
Baseline ($s_0$)                    &    \multicolumn{4}{c}{\cBase 100 (the raw value is 0.683)} &        &   \cBase 100 &                                 \cBase 0 &                                \cBase 0 &                                \cBase 0 &                                \cBase 0 &                                 \cBase 0 &        &   \cBase 100 \\

\midrule

Random-HyP                   &   \cThird $110.8_{\pm{\mbox{\tiny 1.2}}}$ &   \cThird $112.9_{\pm{\mbox{\tiny 1.3}}}$ &   \cThird $113.3_{\pm{\mbox{\tiny 1.8}}}$ &  \cSecond $112.9_{\pm{\mbox{\tiny0.9}}}$ &        &  \cSecond $91.5_{\pm{\mbox{\tiny 1.9}}}$ &   \cThird $60.6_{\pm{\mbox{\tiny 1.6}}}$ &  \cSecond $23.57_{\pm{\mbox{\tiny0.47}}}$ &  \cSecond $23.43_{\pm{\mbox{\tiny0.54}}}$ &  \cSecond $23.40_{\pm{\mbox{\tiny0.56}}}$ &  \cSecond $23.43_{\pm{\mbox{\tiny0.51}}}$ &        &   \cThird $31.0_{\pm{\mbox{\tiny 2.5}}}$ \\
Bagging                      &  \cFourth $103.9_{\pm{\mbox{\tiny 1.0}}}$ &  \cFourth $105.8_{\pm{\mbox{\tiny0.3}}}$ &  \cFourth $105.6_{\pm{\mbox{\tiny0.4}}}$ &  \cFourth $105.1_{\pm{\mbox{\tiny0.9}}}$ &        &   \cThird $82.7_{\pm{\mbox{\tiny0.4}}}$ &   \cFirst $46.4_{\pm{\mbox{\tiny0.3}}}$ &  \cFourth $29.41_{\pm{\mbox{\tiny0.27}}}$ &  \cFourth $29.28_{\pm{\mbox{\tiny0.33}}}$ &  \cFourth $29.29_{\pm{\mbox{\tiny0.30}}}$ &  \cFourth $29.33_{\pm{\mbox{\tiny0.36}}}$ &        &   \cFirst $36.3_{\pm{\mbox{\tiny0.5}}}$ \\
Random-Seed                  &  \cSecond $113.1_{\pm{\mbox{\tiny 1.3}}}$ &  \cSecond $114.5_{\pm{\mbox{\tiny0.3}}}$ &  \cSecond $114.7_{\pm{\mbox{\tiny0.7}}}$ &   \cThird $111.4_{\pm{\mbox{\tiny 3.3}}}$ &        &  \cFirst $100.0_{\pm{\mbox{\tiny0.0}}}$ &  \cFourth $69.3_{\pm{\mbox{\tiny0.4}}}$ &   \cFirst $23.16_{\pm{\mbox{\tiny0.46}}}$ &   \cFirst $23.07_{\pm{\mbox{\tiny0.39}}}$ &   \cFirst $23.06_{\pm{\mbox{\tiny0.42}}}$ &   \cFirst $23.28_{\pm{\mbox{\tiny0.25}}}$ &        &  \cFourth $30.7_{\pm{\mbox{\tiny0.4}}}$ \\
Hetero-DNNs                &   \cFirst $119.9_{\pm{\mbox{\tiny 1.8}}}$ &   \cFirst $123.3_{\pm{\mbox{\tiny 1.1}}}$ &   \cFirst $122.0_{\pm{\mbox{\tiny 1.0}}}$ &   \cFirst $125.5_{\pm{\mbox{\tiny 1.8}}}$ &        &  \cFourth $82.6_{\pm{\mbox{\tiny0.7}}}$ &  \cSecond $48.8_{\pm{\mbox{\tiny0.0}}}$ &   \cThird $25.84_{\pm{\mbox{\tiny0.66}}}$ &   \cThird $25.61_{\pm{\mbox{\tiny0.76}}}$ &   \cThird $25.70_{\pm{\mbox{\tiny0.77}}}$ &   \cThird $25.47_{\pm{\mbox{\tiny0.70}}}$ &        &  \cSecond $33.8_{\pm{\mbox{\tiny0.8}}}$ \\

\bottomrule
\end{tabular}

\end{subfigure}

\end{table*}

\clearpage

\begin{figure*}[t!]
    \begin{subfigure}[t]{0.32\linewidth}
        \vskip 0pt
        \centering
        \includegraphics[width=0.65\linewidth]{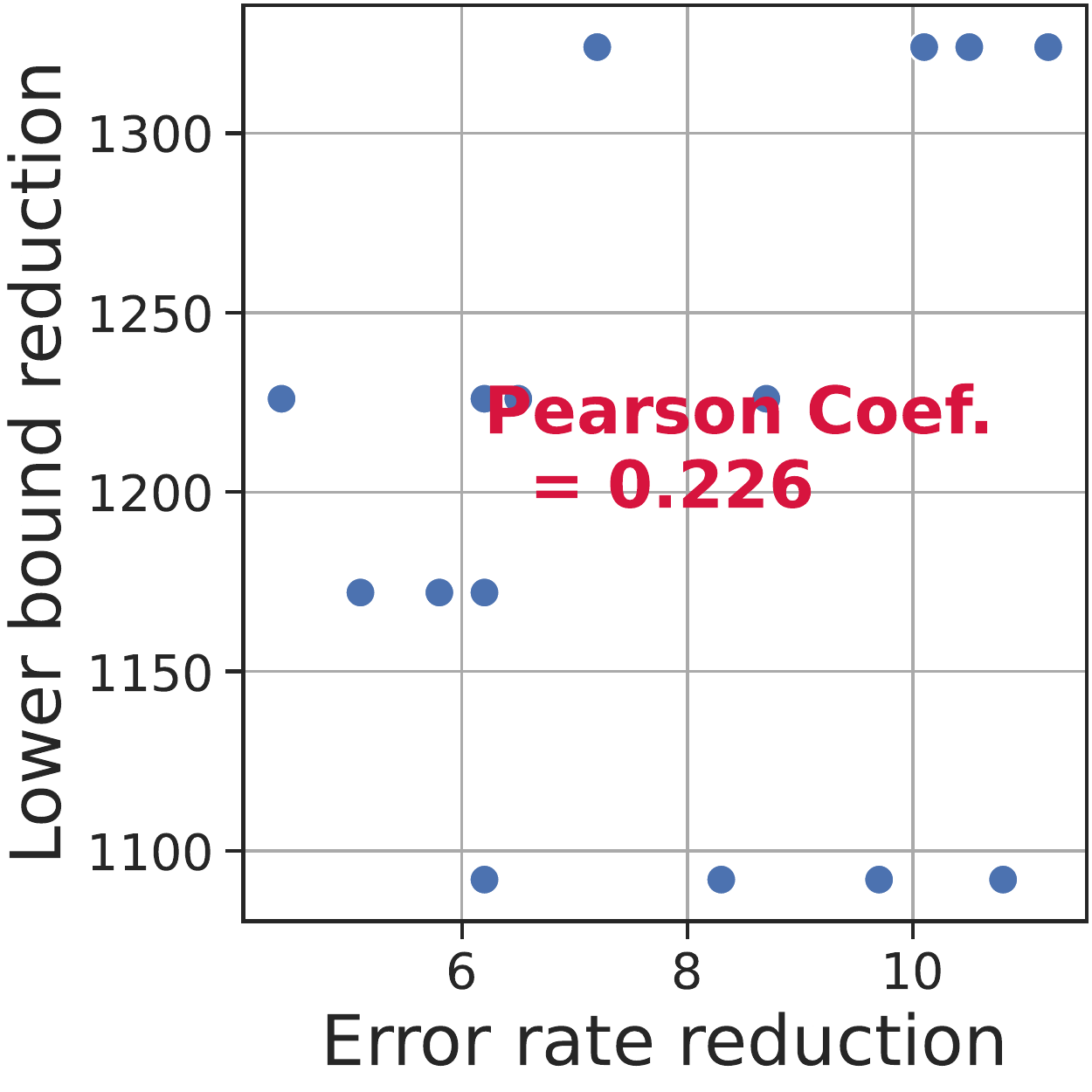}
        \caption{
        \Cref{lemma:ensemble_bound_loose} $\BoundLoose$.
        \label{fig:ERR_LBR_scatter_plot_loose_MNLI}
        }
    \end{subfigure}
    \hfill
    \begin{subfigure}[t]{0.32\linewidth}
        \vskip 0pt
        \centering
        \includegraphics[width=0.65\linewidth]{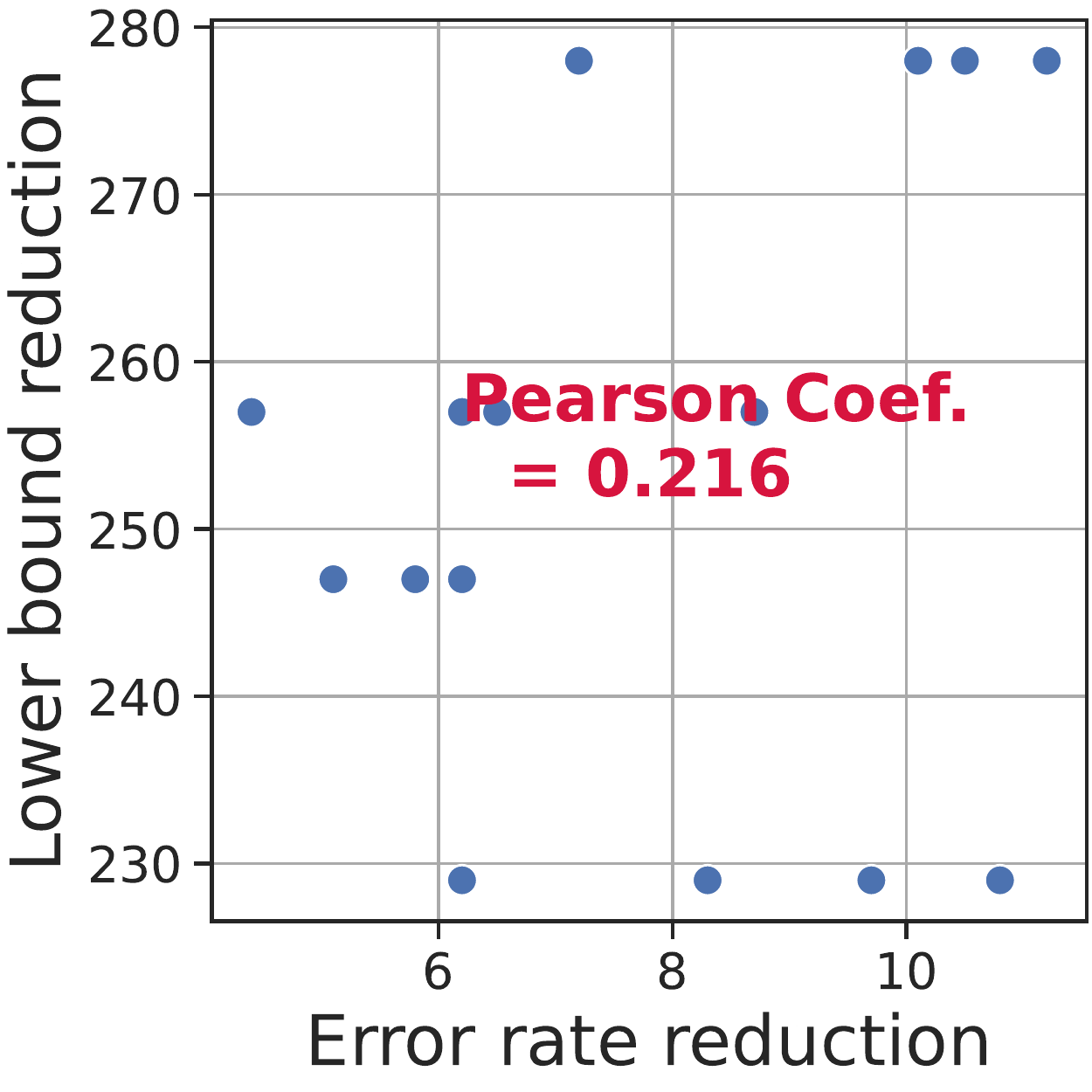}
        \caption{
            $\BoundTightWOCombLoss$.
            \label{fig:ERR_LBR_scatter_plot_tight_wo_combloss_MNLI}
        }
    \end{subfigure}
    \hfill
    \begin{subfigure}[t]{0.32\linewidth}
        \vskip 0pt
        \centering
        \includegraphics[width=0.65\linewidth]{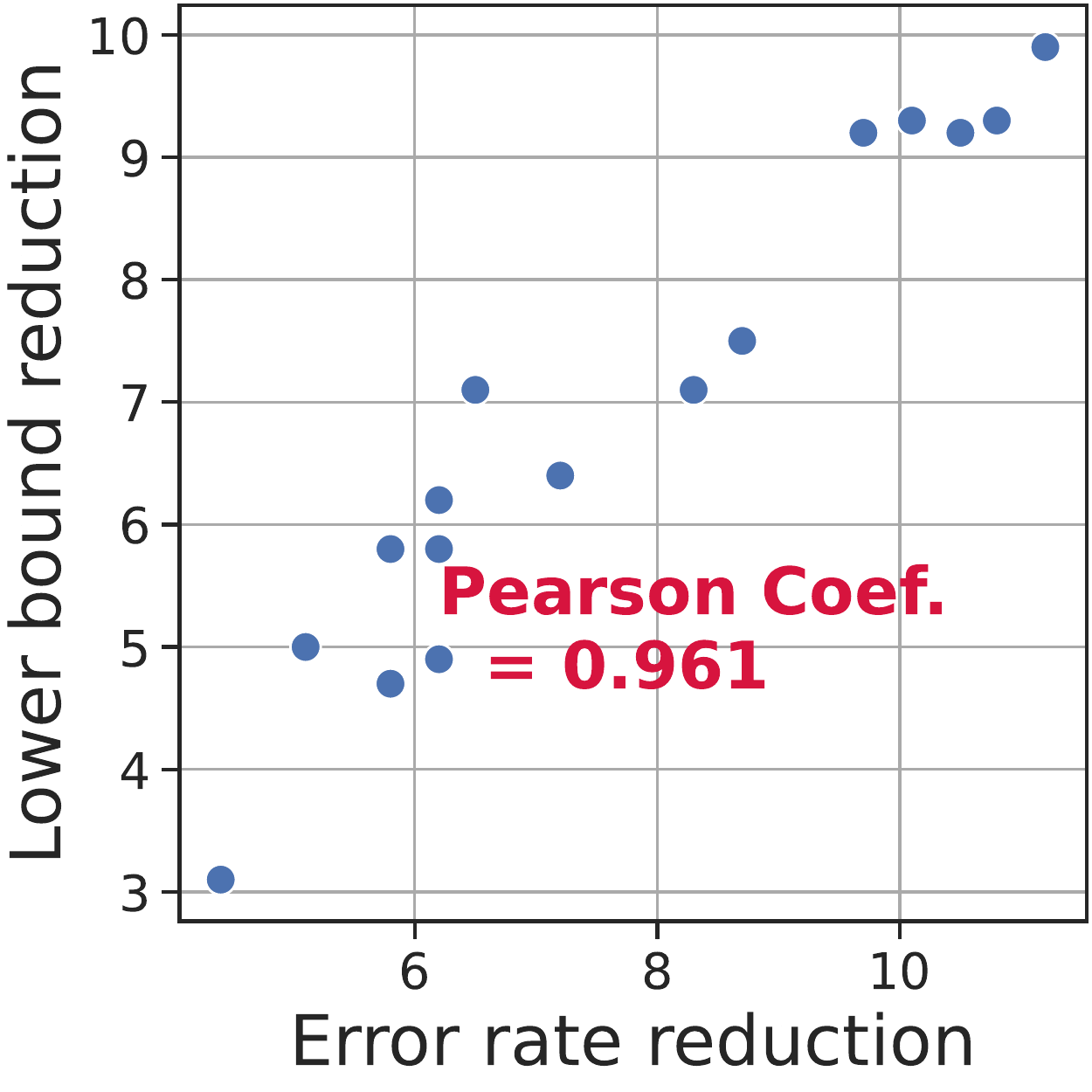}
        \caption{
            \textbf{\Cref{lemma:ensemble_bound_tight}} $\BoundTight$.
            \label{fig:ERR_LBR_scatter_plot_tight_MNLI}
        }
    \end{subfigure}
    \hfill

    \caption{
    \textbf{MNLI task.}
    Correlations between error rate reductions and lower bound reductions.
    Each figure uses different type of lower bound.
    Each point in the figures shows a quantity of a specific ensemble system $s$ and the quantity is the average over the \NumTasks tasks.
    See \Cref{tb:ablation_MNLI} for the real value of each point.
    We used the \NumSystems ensemble systems described in \Cref{sec:ensemble_systems}.
    Each system $s$ used $N=15$ models.
    The baseline values in \Cref{eq:error_reduction,eq:lower_bound_reduction} were the followings:
    ER($s_0$): \SI{18.6}{\percent}.
    LB($s_0$) by $\BoundFuncTight(\StrengthTriplet)$: \SI{3.7}{\percent}.
    LB($s_0$) by $\BoundFuncTight(\StrengthDoublet)$: \SI{3.7}{\percent}.
    LB($s_0$) by $\BoundFuncLoose(\StrengthDoublet)$: \SI{-1.1}{\percent}.
    \label{fig:ERR_LBR_scatter_plot_MNLI}
    }
\end{figure*}

\begin{figure*}[h!]
    \begin{subfigure}[t]{0.19\linewidth}
        \vskip 0pt
        \includegraphics[width=\linewidth]{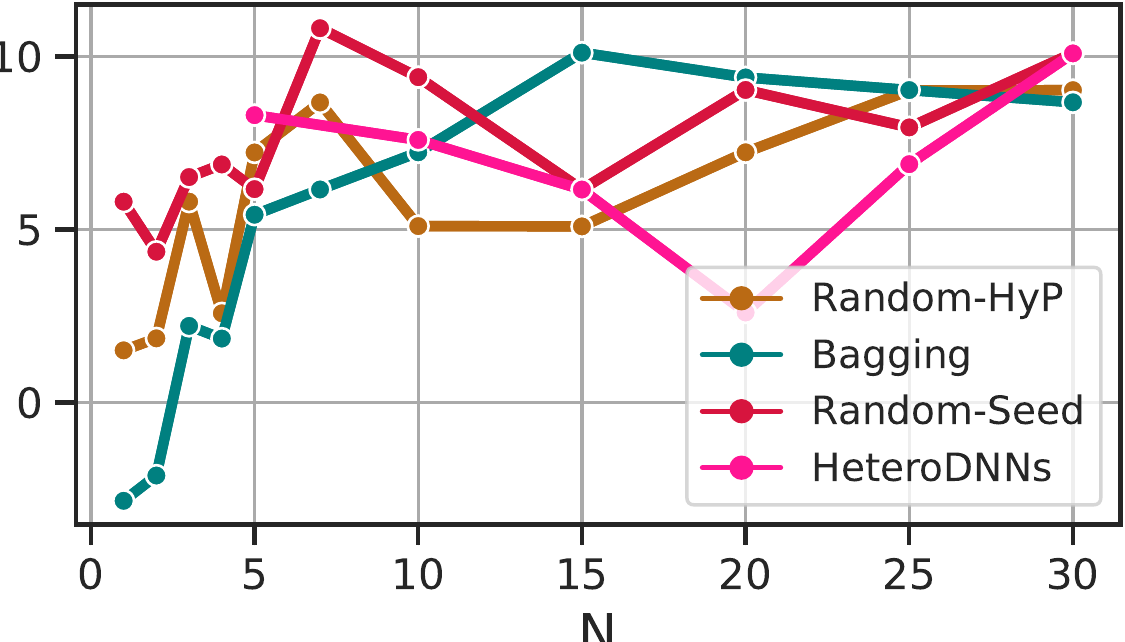}
        \subcaption{Error rate reduction.\label{appendix:fig:scaling.stacking_MNLI}}
    \end{subfigure}
    \hfill
    \begin{subfigure}[t]{0.19\linewidth}
        \vskip 0pt
        \includegraphics[width=\linewidth]{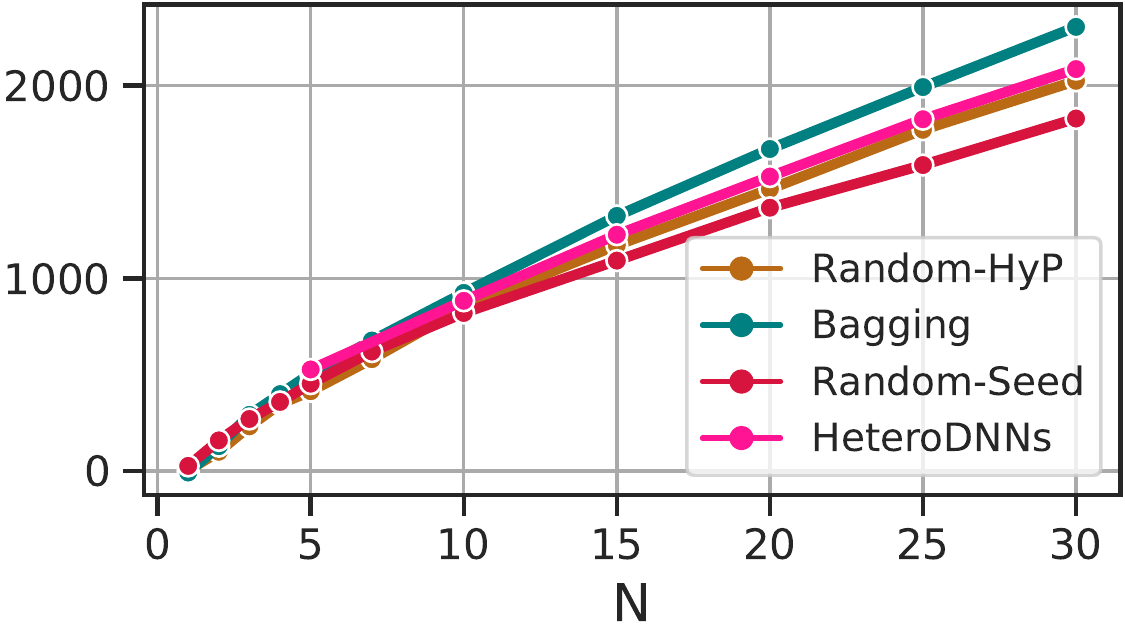}
        \subcaption{Lower bound reduction by \Cref{lemma:ensemble_bound_loose} $\BoundLoose$.\label{appendix:fig:scaling.bound.voting.previous_research_MNLI}}
    \end{subfigure}
    \begin{subfigure}[t]{0.19\linewidth}
        \vskip 0pt
        \includegraphics[width=\linewidth]{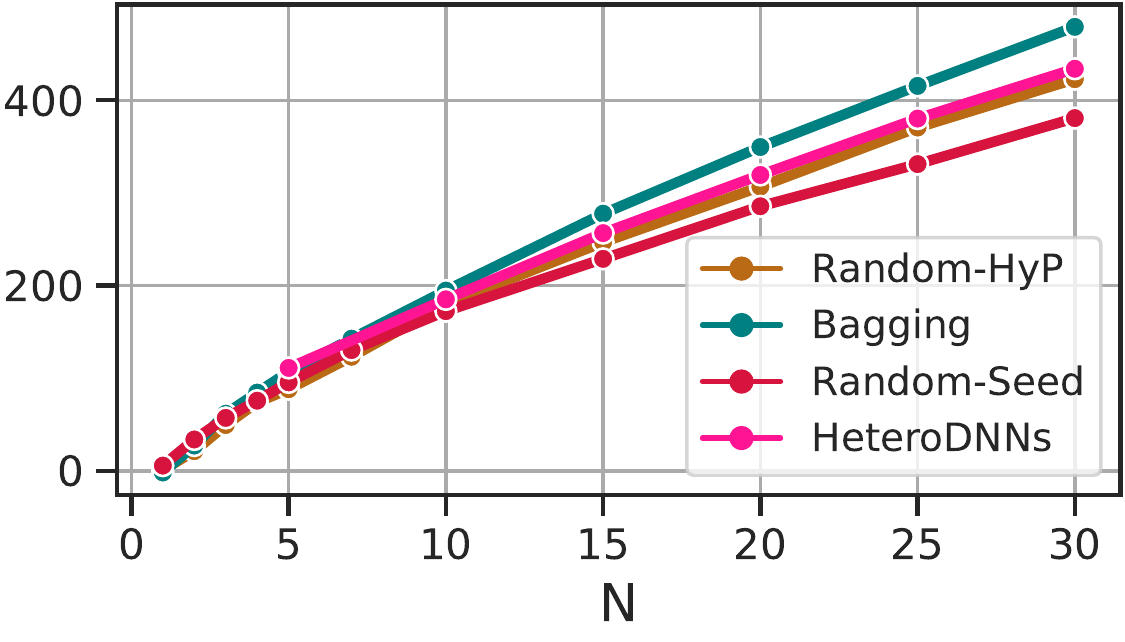}
        \subcaption{Lower bound reduction by $\BoundTightWOCombLoss$. \label{appendix:fig:scaling.bound.voting.ours_wo_combloss_MNLI}}
    \end{subfigure}
    \hfill
     \begin{subfigure}[t]{0.19\linewidth}
        \vskip 0pt
        \includegraphics[width=\linewidth]{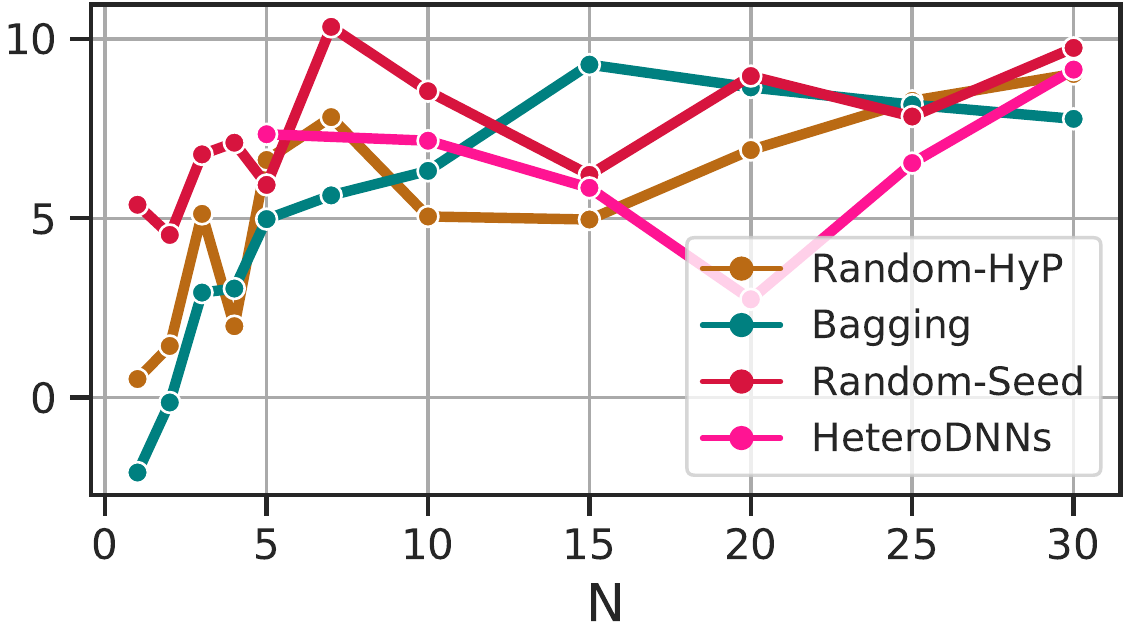}
        \subcaption{Lower bound reduction by \textbf{\Cref{lemma:ensemble_bound_tight}} $\BoundTight$.\label{appendix:fig:scaling.stacking.bound_MNLI}}
    \end{subfigure}
    \hfill
    \begin{subfigure}[t]{0.19\linewidth}
        \vskip 0pt
        \includegraphics[width=\linewidth]{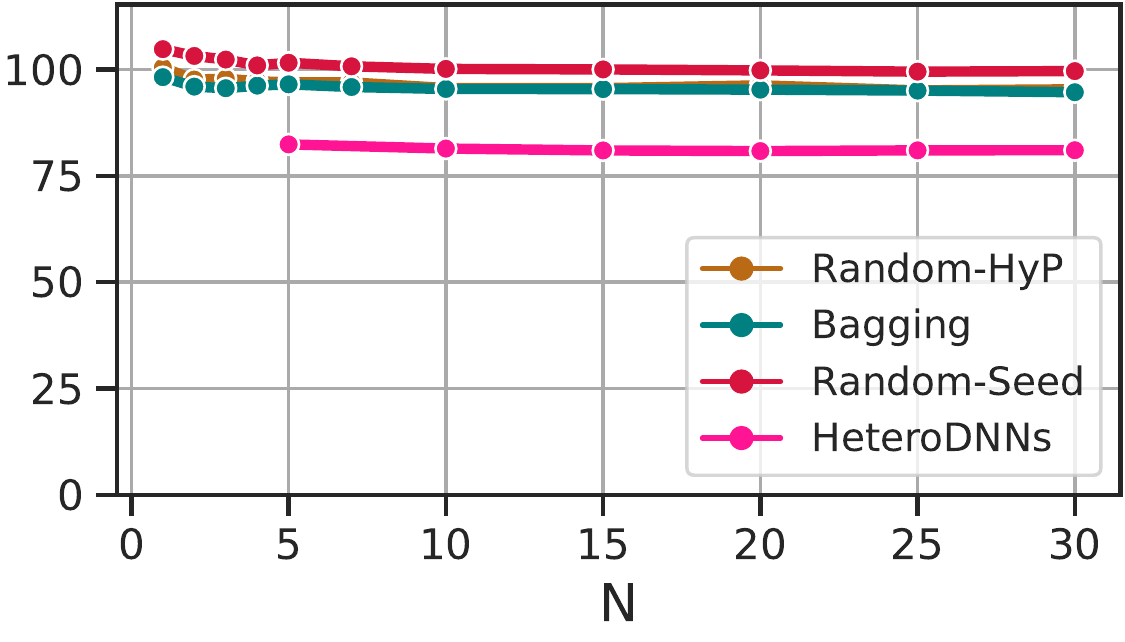}
        \subcaption{$\relev$ \label{appendix:fig:scaling.relevance.per_model_MNLI}}   
    \end{subfigure}   
    \hfill
    \vfill
    \begin{subfigure}[t]{0.19\linewidth}
        \vskip 0pt
        \includegraphics[width=\linewidth]{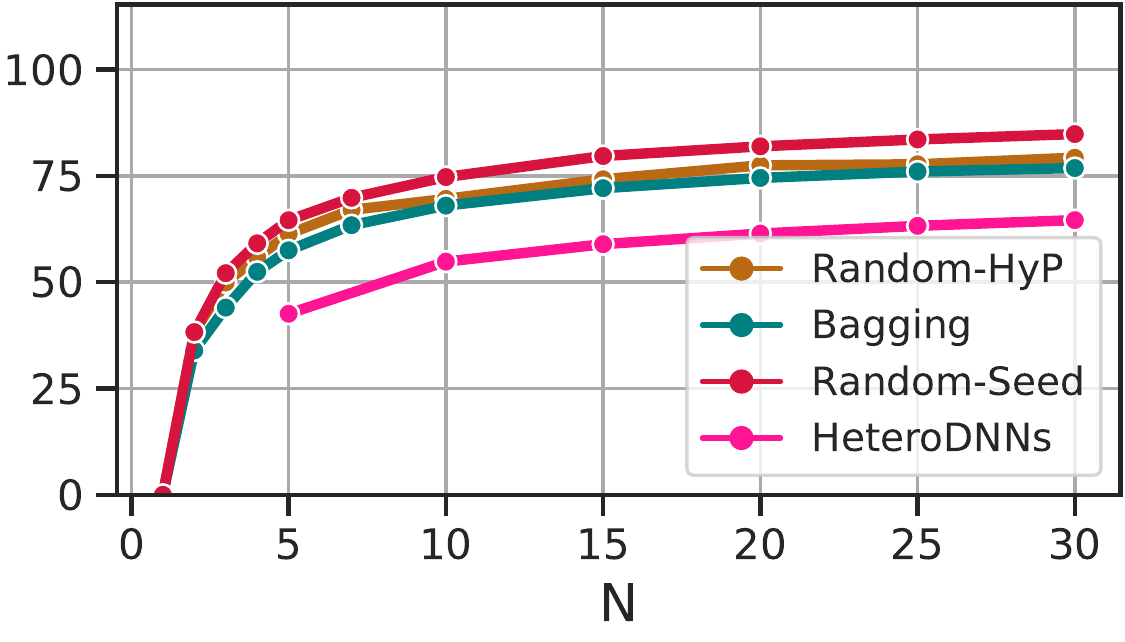}
        \subcaption{$\redun$ \label{appendix:fig:scaling.redundancy.per_model_MNLI}}   
    \end{subfigure}   
    \hfill
    \begin{subfigure}[t]{0.19\linewidth}
        \vskip 0pt
        \includegraphics[width=\linewidth]{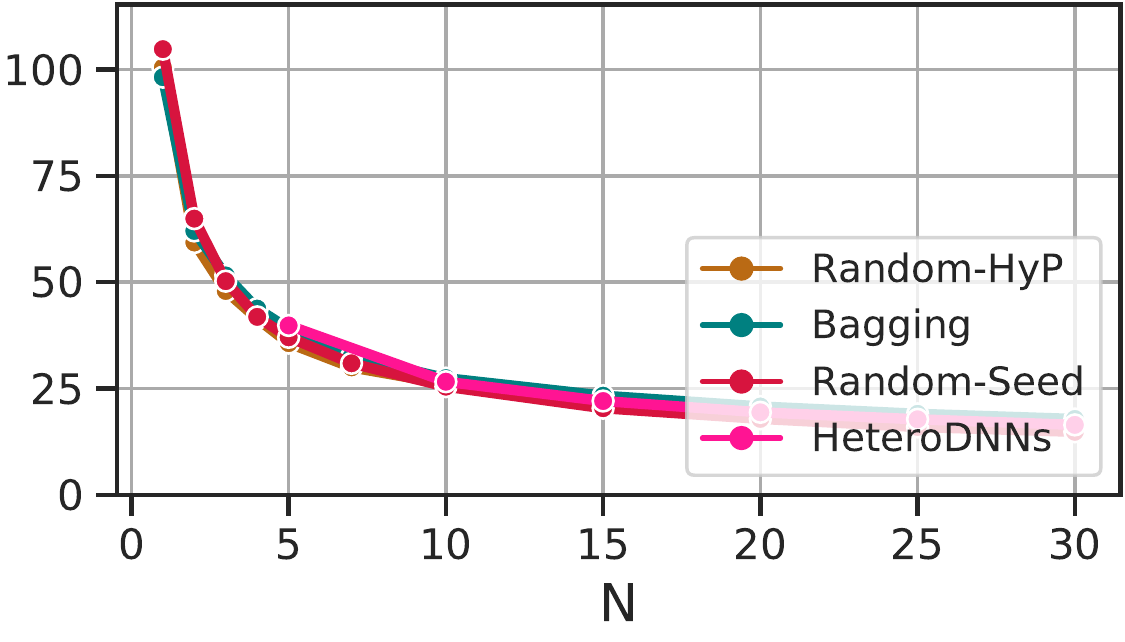}
        \subcaption{$\relev$ $- \redun$ \label{appendix:fig:scaling.novelty.per_model_MNLI}}
    \end{subfigure}   
    \hfill
    \begin{subfigure}[t]{0.19\linewidth}
        \vskip 0pt
        \includegraphics[width=\linewidth]{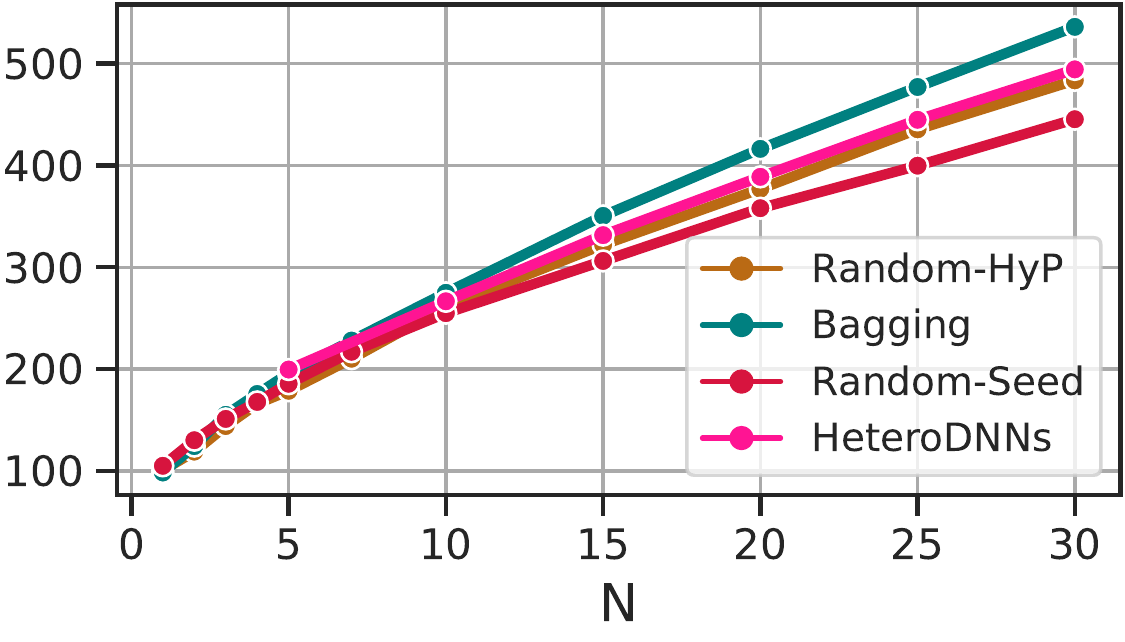}
        \captionsetup{justification=centering}
        \subcaption{$\StrengthDoublet $\newline $= N (\relev - \redun)$ \label{appendix:fig:scaling.E_without_combination_loss_MNLI}}
    \end{subfigure}
    \hfill
    \begin{subfigure}[t]{0.19\linewidth}
        \vskip 0pt
        \includegraphics[width=\linewidth]{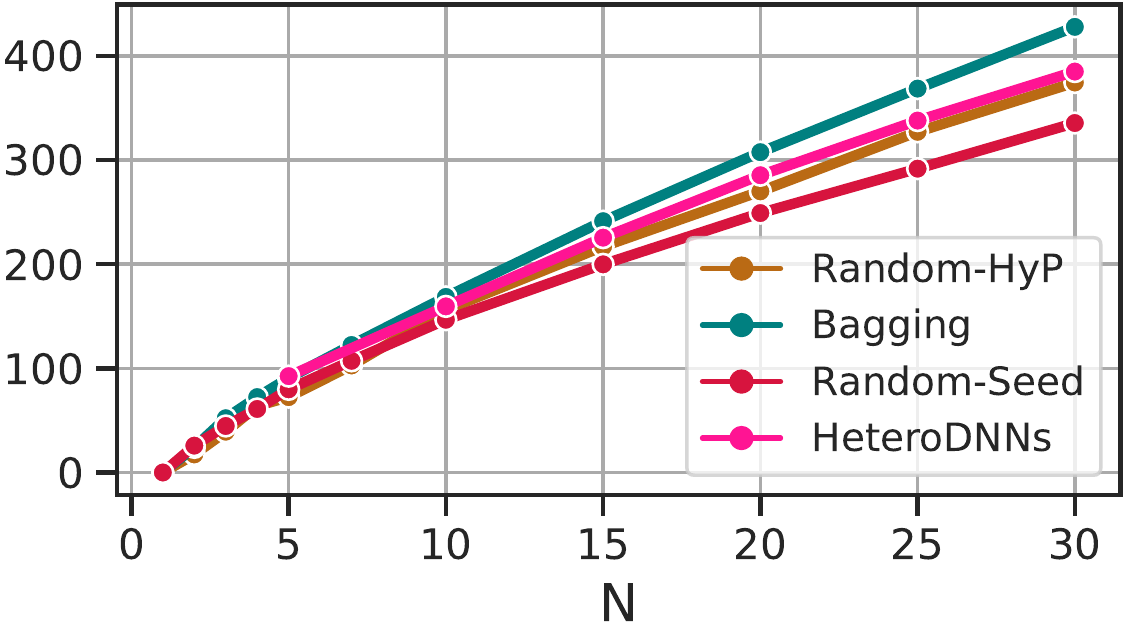}
        \subcaption{$\Combloss$\label{appendix:fig:scaling.combination_loss_MNLI}}
    \end{subfigure}
    \hfill
    \begin{subfigure}[t]{0.19\linewidth}
        \vskip 0pt
        \includegraphics[width=\linewidth]{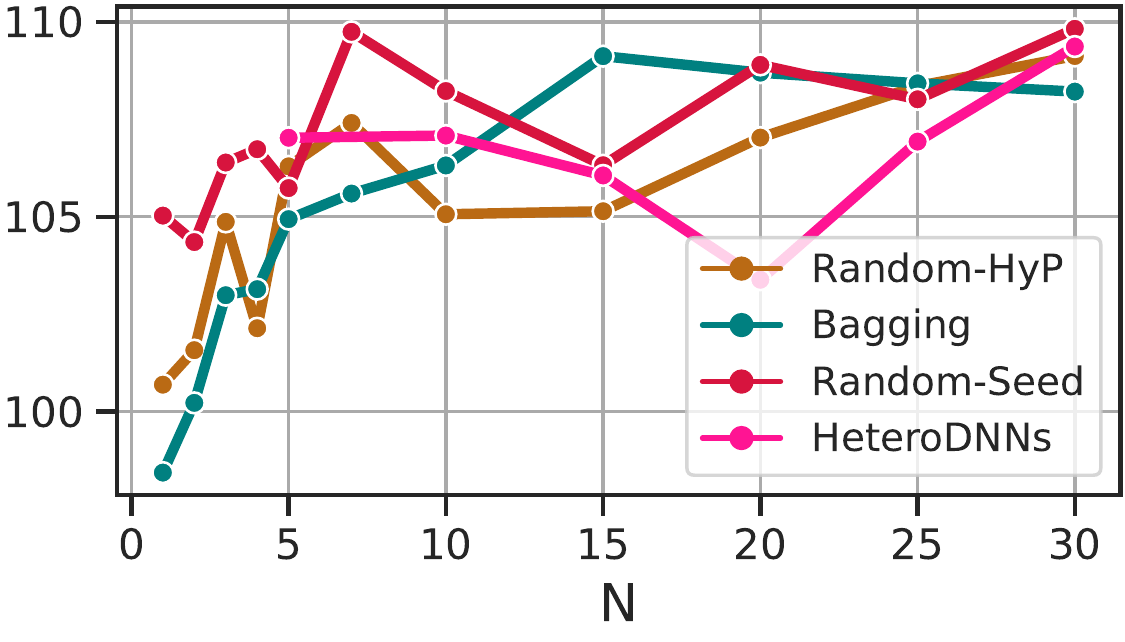}
        \captionsetup{justification=centering}
        \subcaption{$\StrengthTriplet = N (\relev - \redun) - \Combloss $\label{appendix:fig:scaling.E_MNLI}}
    \end{subfigure}
\caption{
\textbf{MNLI task.}
The change in ensemble quantities when the number of models $N$ is changed.
Each figure shows a specific quantity.
The ensemble systems used the SVM model combination.
Each value is an averages of the \NumTasks tasks.
$\perModelMetric$ denotes per-model metric values defined as $\perModelMetricDef$.
\label{appendix:fig:scaling_MNLI}
}
\end{figure*}

\begin{table*}[t]
    \centering
    \caption{
        \textbf{MNLI task}.
        Statistics of ensemble systems described in \Cref{sec:ensemble_systems}.
        The rows and columns list the model generation and combination methods of \Cref{tb:ensemble_methods}, respectively.
        Each cell shows a quantity of a specific system $s$.
        Each quantity is the average over the \NumTasks tasks.
        Each system contains $N=15$ models.
        Color shows the rank within \textit{each column} (brighter is better).
        \label{tb:ablation_MNLI}
    }  
    \begin{subfigure}{\linewidth}
        \centering
        \small
        \tabcolsep 3.0pt
    \subcaption{
        Error rate reductions and lower bound reductions.
        The baseline values used in \Cref{eq:error_reduction,eq:lower_bound_reduction} were the followings.
        ER($s_0$): \SI{18.6}{\percent}.
        LB($s_0$) by $\BoundFuncTight(\StrengthTriplet)$: \SI{3.7}{\percent}.
        LB($s_0$) by $\BoundFuncTight(\StrengthDoublet)$: \SI{3.7}{\percent}.
        LB($s_0$) by $\BoundFuncLoose(\StrengthDoublet)$: \SI{-1.1}{\percent}.
        \label{tb:ablation_errors_MNLI}
    }

\begin{tabular}{lcccccccccccc}
\toprule
& \multicolumn{4}{c}{Error rate reductions \cref{eq:error_reduction}} & &   \multicolumn{7}{c}{Lower bound reductions \Cref{eq:lower_bound_reduction}} \\
\cmidrule(l{\tabcolsep}r{\tabcolsep}){2-5} \cmidrule(l{\tabcolsep}){7-13}
 & \multirow{2}{*}{Voting} & \multirow{2}{*}{LogR} & \multirow{2}{*}{SVM} & \multirow{2}{*}{RForest} & & \multicolumn{4}{c}{\textbf{\Cref{lemma:ensemble_bound_tight}} $\BoundFuncTight(\StrengthTriplet)$} & & \multirow{2}{*}{$\BoundFuncTight(\StrengthDoublet)$} & \multirow{2}{*}{\begin{tabular}{c}\Cref{lemma:ensemble_bound_loose} \\ $\BoundFuncLoose(\StrengthDoublet)$\end{tabular}} \\
 &  &  &  &  & & \multicolumn{1}{c}{Voting} & \multicolumn{1}{c}{LogR} & \multicolumn{1}{c}{SVM} & \multicolumn{1}{c}{RForest} & & &  \\

\midrule
Random-HyP    &   \cThird $6.2_{\pm{\mbox{\tiny 3.0}}}$ &   \cFourth $5.8_{\pm{\mbox{\tiny 3.4}}}$ &   \cThird $5.1_{\pm{\mbox{\tiny 1.8}}}$ &  \cFourth $5.8_{\pm{\mbox{\tiny 1.8}}}$ &        &   \cThird $4.9_{\pm{\mbox{\tiny 2.9}}}$ &  \cFourth $4.7_{\pm{\mbox{\tiny 3.4}}}$ &  \cFourth $5.0_{\pm{\mbox{\tiny 2.3}}}$ &  \cFourth $5.8_{\pm{\mbox{\tiny 2.0}}}$ &        &   \cThird $247_{\pm{\mbox{\tiny 20}}}$ &    \cThird $1172_{\pm{\mbox{\tiny 52}}}$ \\
Bagging       &  \cFirst $11.2_{\pm{\mbox{\tiny 3.8}}}$ &  \cSecond $10.5_{\pm{\mbox{\tiny 3.9}}}$ &  \cFirst $10.1_{\pm{\mbox{\tiny 3.8}}}$ &  \cSecond $7.2_{\pm{\mbox{\tiny 3.0}}}$ &        &   \cFirst $9.9_{\pm{\mbox{\tiny 3.9}}}$ &  \cSecond $9.2_{\pm{\mbox{\tiny 3.7}}}$ &   \cFirst $9.3_{\pm{\mbox{\tiny 3.4}}}$ &   \cThird $6.4_{\pm{\mbox{\tiny 2.6}}}$ &        &   \cFirst $278_{\pm{\mbox{\tiny 21}}}$ &   \cFirst $1324_{\pm{\mbox{\tiny 103}}}$ \\
Random-Seed   &  \cSecond $8.3_{\pm{\mbox{\tiny 1.4}}}$ &   \cFirst $10.8_{\pm{\mbox{\tiny 0.6}}}$ &  \cSecond $6.2_{\pm{\mbox{\tiny 2.2}}}$ &   \cFirst $9.7_{\pm{\mbox{\tiny 4.2}}}$ &        &  \cSecond $7.1_{\pm{\mbox{\tiny 1.3}}}$ &   \cFirst $9.3_{\pm{\mbox{\tiny 0.3}}}$ &  \cSecond $6.2_{\pm{\mbox{\tiny 1.0}}}$ &   \cFirst $9.2_{\pm{\mbox{\tiny 3.9}}}$ &        &  \cFourth $229_{\pm{\mbox{\tiny 22}}}$ &  \cFourth $1092_{\pm{\mbox{\tiny 137}}}$ \\
Hetero-DNNs &  \cFourth $4.4_{\pm{\mbox{\tiny 2.0}}}$ &    \cThird $8.7_{\pm{\mbox{\tiny 2.8}}}$ &  \cSecond $6.2_{\pm{\mbox{\tiny 1.4}}}$ &   \cThird $6.5_{\pm{\mbox{\tiny 3.3}}}$ &        &  \cFourth $3.1_{\pm{\mbox{\tiny 2.1}}}$ &   \cThird $7.5_{\pm{\mbox{\tiny 2.6}}}$ &   \cThird $5.8_{\pm{\mbox{\tiny 1.1}}}$ &  \cSecond $7.1_{\pm{\mbox{\tiny 2.4}}}$ &        &  \cSecond $257_{\pm{\mbox{\tiny 13}}}$ &  \cSecond $1226_{\pm{\mbox{\tiny 116}}}$ \\
\bottomrule

\end{tabular}

    \end{subfigure}
    \vfill
    \begin{subfigure}{\linewidth}
    \centering
    \small
    \tabcolsep 1.0pt
    \subcaption{
    Breakdown of ensemble strength defined in \cref{eq:triplet_decomposition}.
    We show per-model metric values defined as $\perModelMetricDef$. Thus, $\StrengthTriplet = (\relev - \redun - \combloss) \ \times N $ holds.
    For intuitive understanding, all the values are normalized by the ensemble strength of baseline $\StrengthTriplet_{s_0}$, for example, $\Relev = \RelevHat / \StrengthTriplet_{s_0} \times 100$ where $\RelevHat$ is the raw value.
    \label{tb:ablation_triple_MNLI}
    }

\begin{tabular}{lccccccccccccc}
\toprule

{} & \multicolumn{4}{c}{\multirow{1}{*}{$\StrengthTripletWithArgs$}} & & \multicolumn{6}{c}{Per-model metric values} & \\
\cmidrule(l{\tabcolsep}r{\tabcolsep}){2-5} \cmidrule(l{\tabcolsep}r{\tabcolsep}){6-12}
{} &  & &  &  & & \multirow{2}{*}{$\perModelMetric_{\normalfont \text{relev}}$} & \multirow{2}{*}{$\perModelMetric_{\normalfont \text{redun}}$} & \multicolumn{4}{c}{ $\perModelMetric_{\normalfont \text{combloss}}$} & & \multirow{2}{*}{$\perModelMetric_{\normalfont \text{relev}} - \perModelMetric_{\normalfont \text{redun}}$} \\
{} & \multicolumn{1}{c}{Voting} & \multicolumn{1}{c}{LogR} & \multicolumn{1}{c}{SVM} & \multicolumn{1}{c}{RForest} & &  {} &  {} &  \multicolumn{1}{c}{Voting} & \multicolumn{1}{c}{LogR} &  \multicolumn{1}{c}{SVM} & \multicolumn{1}{c}{RForest} & &  {} \\

\midrule
Baseline ($s_0$)                    &    \multicolumn{4}{c}{\cBase 100 (the raw value is 0.681)} &        &   \cBase 100 &                                 \cBase 0 &                                \cBase 0 &                                \cBase 0 &                                \cBase 0 &                                 \cBase 0 &        &   \cBase 100 \\

\midrule

Random-HyP                   &   \cThird $104.3_{\pm{\mbox{\tiny 2.6}}}$ &  \cFourth $104.2_{\pm{\mbox{\tiny 3.0}}}$ &  \cFourth $104.4_{\pm{\mbox{\tiny 2.1}}}$ &  \cFourth $105.1_{\pm{\mbox{\tiny 1.8}}}$ &        &  \cSecond $95.6_{\pm{\mbox{\tiny0.5}}}$ &   \cThird $74.2_{\pm{\mbox{\tiny 1.2}}}$ &                     \cSecond $14.47_{\pm{\mbox{\tiny 1.26}}}$ &                     \cSecond $14.48_{\pm{\mbox{\tiny 1.28}}}$ &                     \cSecond $14.46_{\pm{\mbox{\tiny 1.21}}}$ &                     \cSecond $14.41_{\pm{\mbox{\tiny 1.13}}}$ &        &   \cThird $21.4_{\pm{\mbox{\tiny 1.3}}}$ \\
Bagging                      &   \cFirst $108.8_{\pm{\mbox{\tiny 3.4}}}$ &  \cSecond $108.2_{\pm{\mbox{\tiny 3.3}}}$ &   \cFirst $108.2_{\pm{\mbox{\tiny 3.1}}}$ &   \cThird $105.7_{\pm{\mbox{\tiny 2.3}}}$ &        &   \cThird $95.4_{\pm{\mbox{\tiny0.6}}}$ &  \cSecond $72.1_{\pm{\mbox{\tiny 1.7}}}$ &                     \cFourth $16.05_{\pm{\mbox{\tiny0.97}}}$ &                     \cFourth $16.09_{\pm{\mbox{\tiny 1.05}}}$ &                     \cFourth $16.09_{\pm{\mbox{\tiny 1.01}}}$ &                     \cFourth $16.26_{\pm{\mbox{\tiny 1.24}}}$ &        &   \cFirst $23.3_{\pm{\mbox{\tiny 1.8}}}$ \\
Random-Seed                  &  \cSecond $106.3_{\pm{\mbox{\tiny 1.1}}}$ &   \cFirst $108.3_{\pm{\mbox{\tiny0.2}}}$ &  \cSecond $105.5_{\pm{\mbox{\tiny0.9}}}$ &   \cFirst $108.1_{\pm{\mbox{\tiny 3.5}}}$ &        &  \cFirst $100.0_{\pm{\mbox{\tiny0.0}}}$ &  \cFourth $79.6_{\pm{\mbox{\tiny 1.4}}}$ &                      \cFirst $13.27_{\pm{\mbox{\tiny 1.38}}}$ &                      \cFirst $13.14_{\pm{\mbox{\tiny 1.39}}}$ &                      \cFirst $13.32_{\pm{\mbox{\tiny 1.44}}}$ &                      \cFirst $13.15_{\pm{\mbox{\tiny 1.61}}}$ &        &  \cFourth $20.4_{\pm{\mbox{\tiny 1.4}}}$ \\
Hetero-DNNs                &  \cFourth $102.7_{\pm{\mbox{\tiny 1.9}}}$ &   \cThird $106.7_{\pm{\mbox{\tiny 2.3}}}$ &   \cThird $105.2_{\pm{\mbox{\tiny 1.0}}}$ &  \cSecond $106.3_{\pm{\mbox{\tiny 2.1}}}$ &        &  \cFourth $81.0_{\pm{\mbox{\tiny0.8}}}$ &   \cFirst $58.9_{\pm{\mbox{\tiny 1.0}}}$ &                      \cThird $15.19_{\pm{\mbox{\tiny0.83}}}$ &                      \cThird $14.93_{\pm{\mbox{\tiny0.83}}}$ &                      \cThird $15.03_{\pm{\mbox{\tiny0.89}}}$ &                      \cThird $14.96_{\pm{\mbox{\tiny0.93}}}$ &        &  \cSecond $22.0_{\pm{\mbox{\tiny 1.3}}}$ \\

\bottomrule
\end{tabular}

\end{subfigure}

\end{table*}

\clearpage

\begin{figure*}[t!]
    \begin{subfigure}[t]{0.32\linewidth}
        \vskip 0pt
        \centering
        \includegraphics[width=0.65\linewidth]{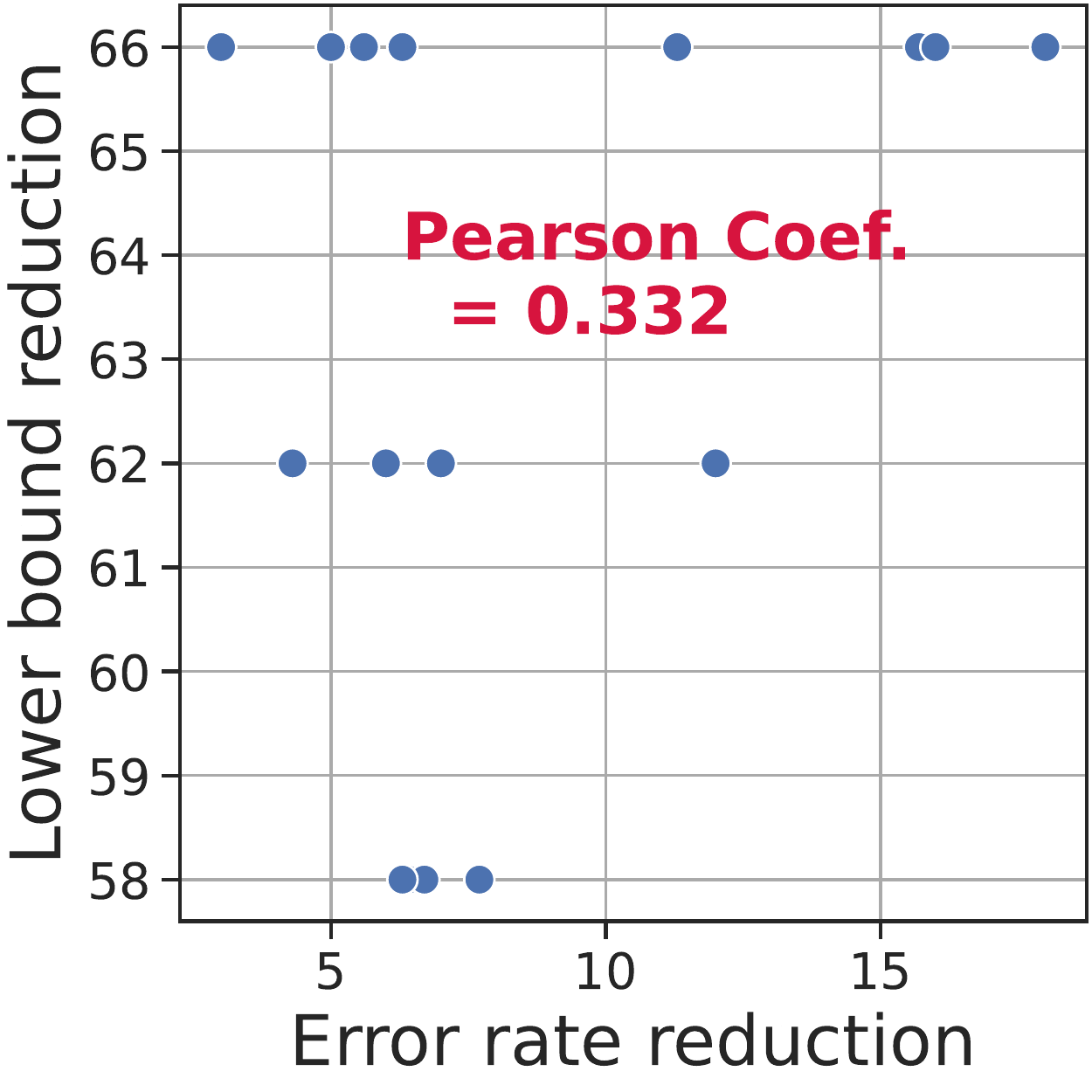}
        \caption{
        \Cref{lemma:ensemble_bound_loose} $\BoundLoose$.
        \label{fig:ERR_LBR_scatter_plot_loose_MRPC}
        }
    \end{subfigure}
    \hfill
    \begin{subfigure}[t]{0.32\linewidth}
        \vskip 0pt
        \centering
        \includegraphics[width=0.65\linewidth]{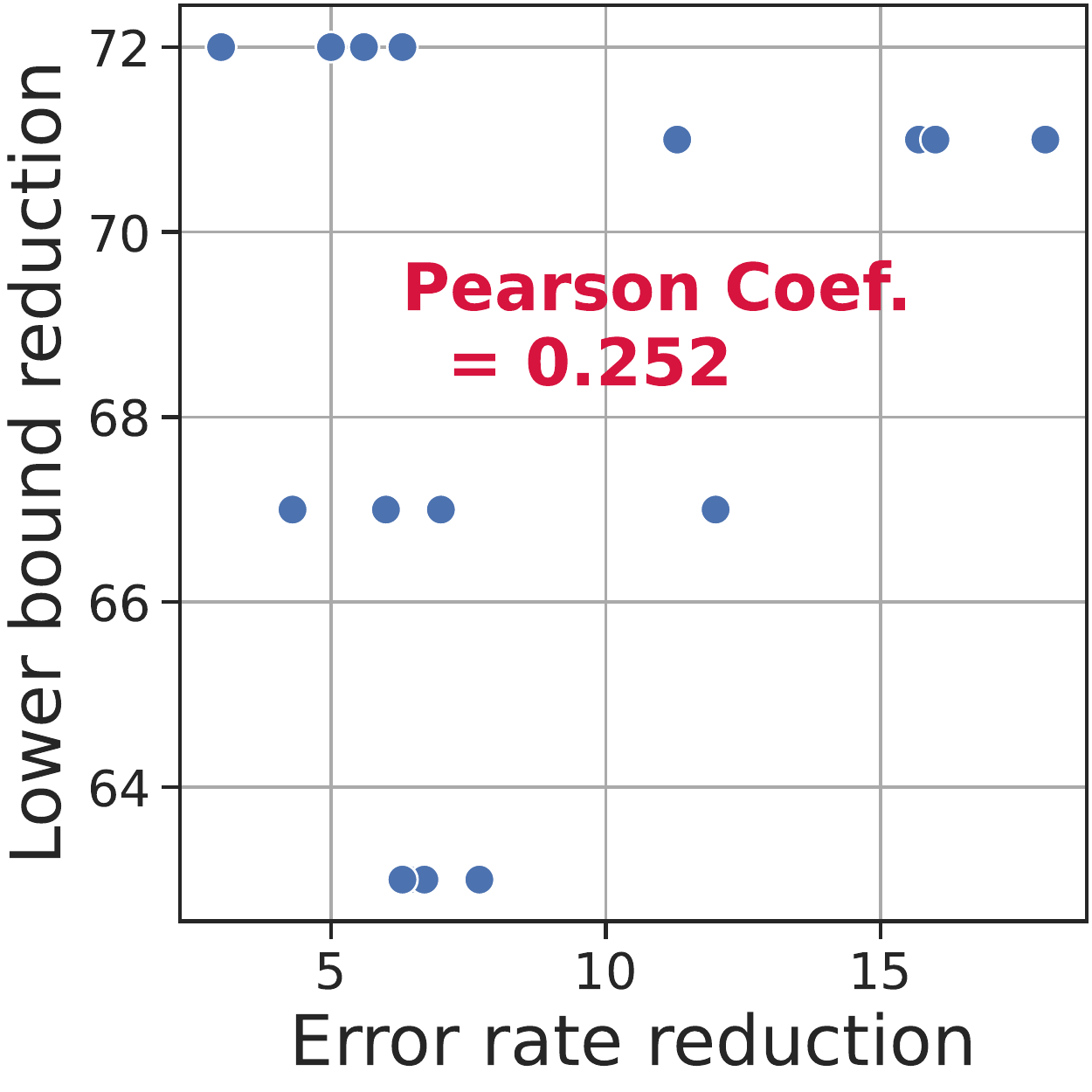}
        \caption{
            $\BoundTightWOCombLoss$.
            \label{fig:ERR_LBR_scatter_plot_tight_wo_combloss_MRPC}
        }
    \end{subfigure}
    \hfill
    \begin{subfigure}[t]{0.32\linewidth}
        \vskip 0pt
        \centering
        \includegraphics[width=0.65\linewidth]{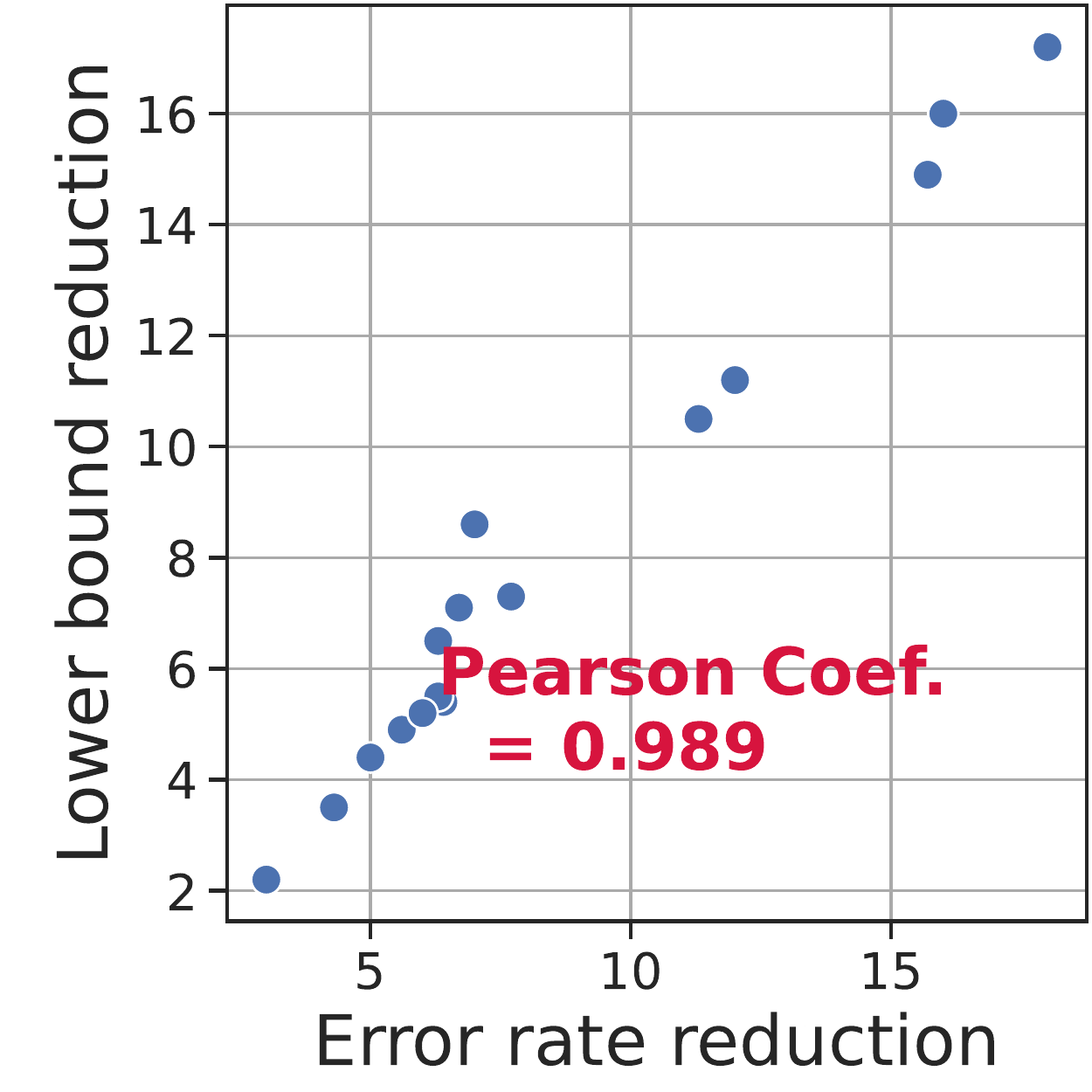}
        \caption{
            \textbf{\Cref{lemma:ensemble_bound_tight}} $\BoundTight$.
            \label{fig:ERR_LBR_scatter_plot_tight_MRPC}
        }
    \end{subfigure}
    \hfill

    \caption{
    \textbf{MRPC task.}
    Correlations between error rate reductions and lower bound reductions.
    Each figure uses different type of lower bound.
    Each point in the figures shows a quantity of a specific ensemble system $s$ and the quantity is the average over the \NumTasks tasks.
    See \Cref{tb:ablation_MRPC} for the real value of each point.
    We used the \NumSystems ensemble systems described in \Cref{sec:ensemble_systems}.
    Each system $s$ used $N=15$ models.
    The baseline values in \Cref{eq:error_reduction,eq:lower_bound_reduction} were the followings:
    ER($s_0$): \SI{13.6}{\percent}.
    LB($s_0$) of $\BoundFuncTight(\StrengthTriplet)$: \SI{2.6}{\percent}.
    LB($s_0$) of $\BoundFuncTight(\StrengthDoublet)$: \SI{2.6}{\percent}.
    LB($s_0$) of $\BoundFuncLoose(\StrengthDoublet)$: \SI{-4.0}{\percent}.
    \label{fig:ERR_LBR_scatter_plot_MRPC}
    }
\end{figure*}

\begin{figure*}[h!]
    \begin{subfigure}[t]{0.19\linewidth}
        \vskip 0pt
        \includegraphics[width=\linewidth]{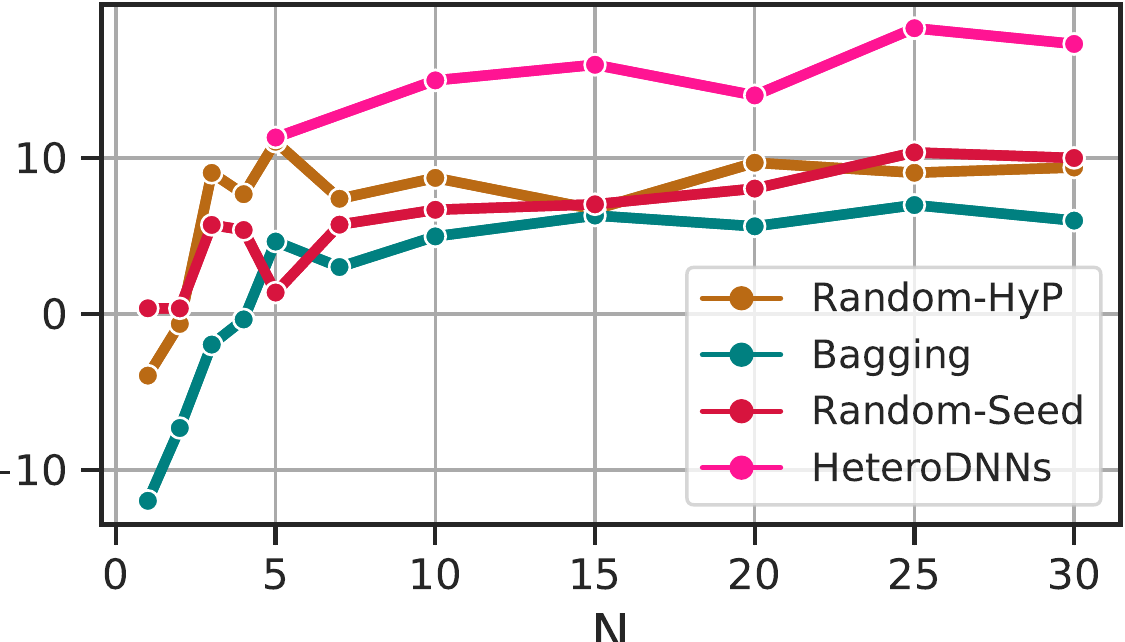}
        \subcaption{Error rate reduction.\label{appendix:fig:scaling.stacking_MRPC}}
    \end{subfigure}
    \hfill
    \begin{subfigure}[t]{0.19\linewidth}
        \vskip 0pt
        \includegraphics[width=\linewidth]{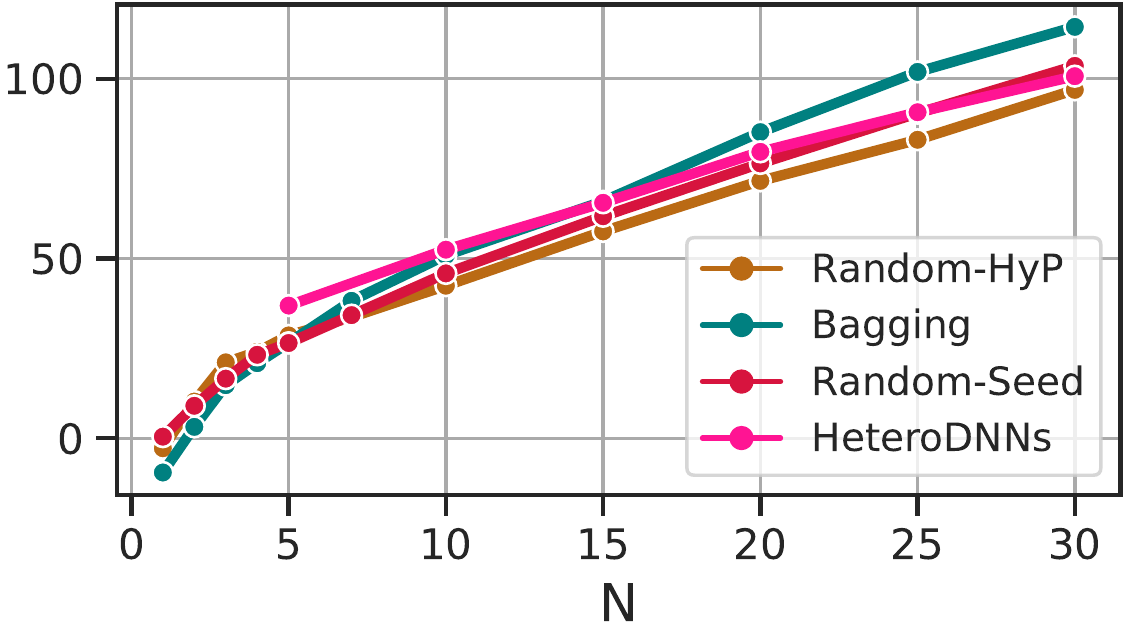}
        \subcaption{Lower bound reduction by \Cref{lemma:ensemble_bound_loose} $\BoundLoose$.\label{appendix:fig:scaling.bound.voting.previous_research_MRPC}}
    \end{subfigure}
    \begin{subfigure}[t]{0.19\linewidth}
        \vskip 0pt
        \includegraphics[width=\linewidth]{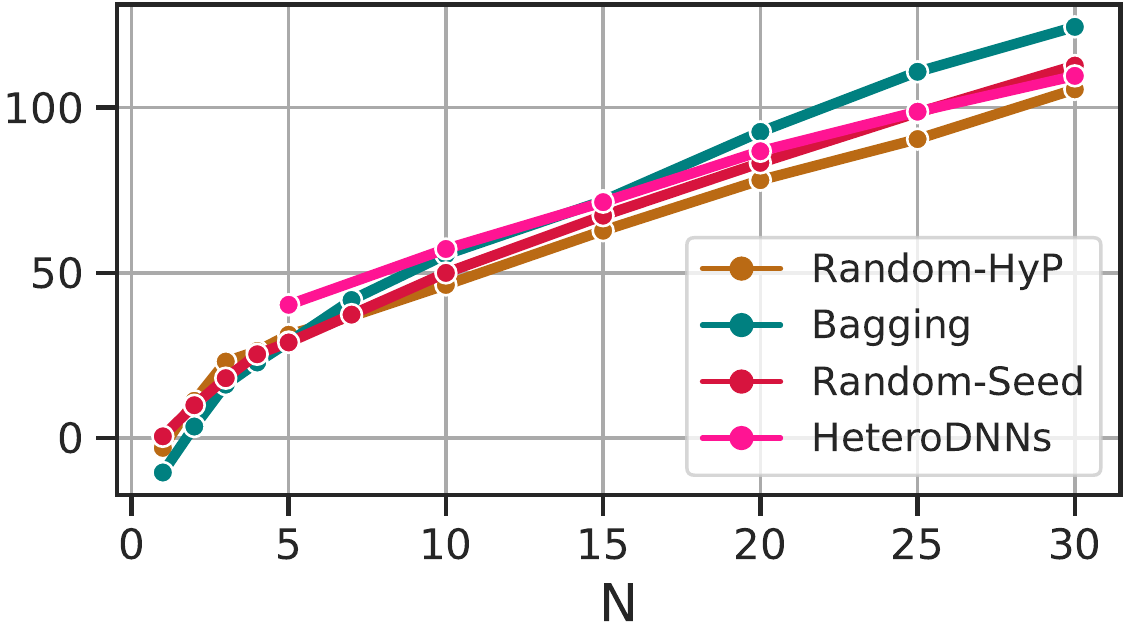}
        \subcaption{Lower bound reduction by $\BoundTightWOCombLoss$. \label{appendix:fig:scaling.bound.voting.ours_wo_combloss_MRPC}}
    \end{subfigure}
    \hfill
     \begin{subfigure}[t]{0.19\linewidth}
        \vskip 0pt
        \includegraphics[width=\linewidth]{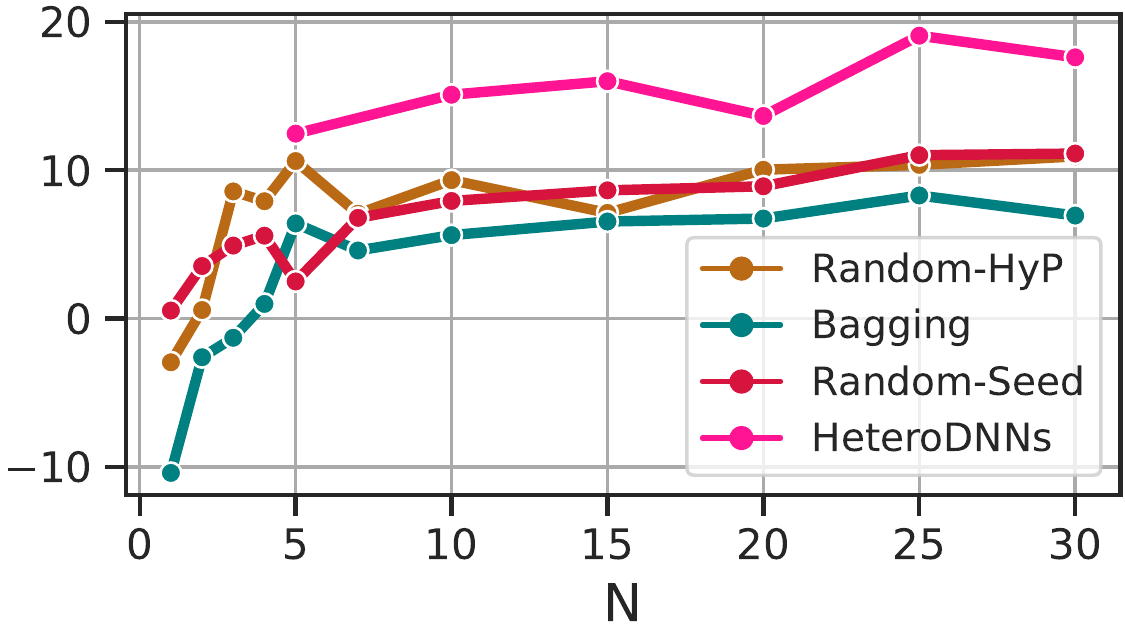}
        \subcaption{Lower bound reduction by \textbf{\Cref{lemma:ensemble_bound_tight}} $\BoundTight$.\label{appendix:fig:scaling.stacking.bound_MRPC}}
    \end{subfigure}
    \hfill
    \begin{subfigure}[t]{0.19\linewidth}
        \vskip 0pt
        \includegraphics[width=\linewidth]{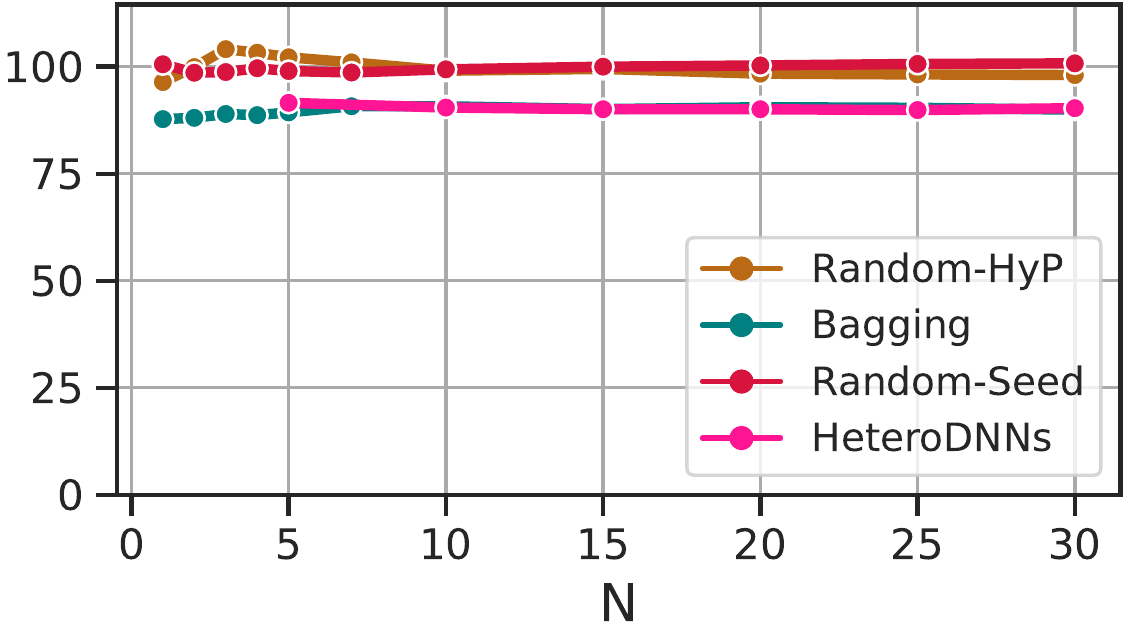}
        \subcaption{$\relev$ \label{appendix:fig:scaling.relevance.per_model_MRPC}}   
    \end{subfigure}   
    \hfill
    \vfill
    \begin{subfigure}[t]{0.19\linewidth}
        \vskip 0pt
        \includegraphics[width=\linewidth]{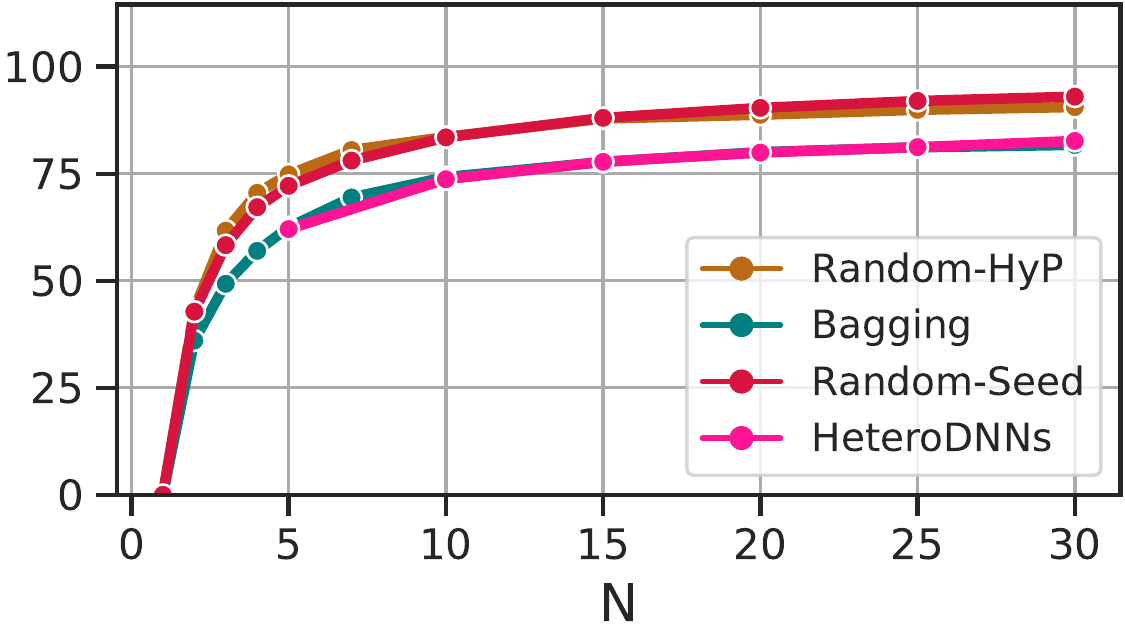}
        \subcaption{$\redun$ \label{appendix:fig:scaling.redundancy.per_model_MRPC}}   
    \end{subfigure}   
    \hfill
    \begin{subfigure}[t]{0.19\linewidth}
        \vskip 0pt
        \includegraphics[width=\linewidth]{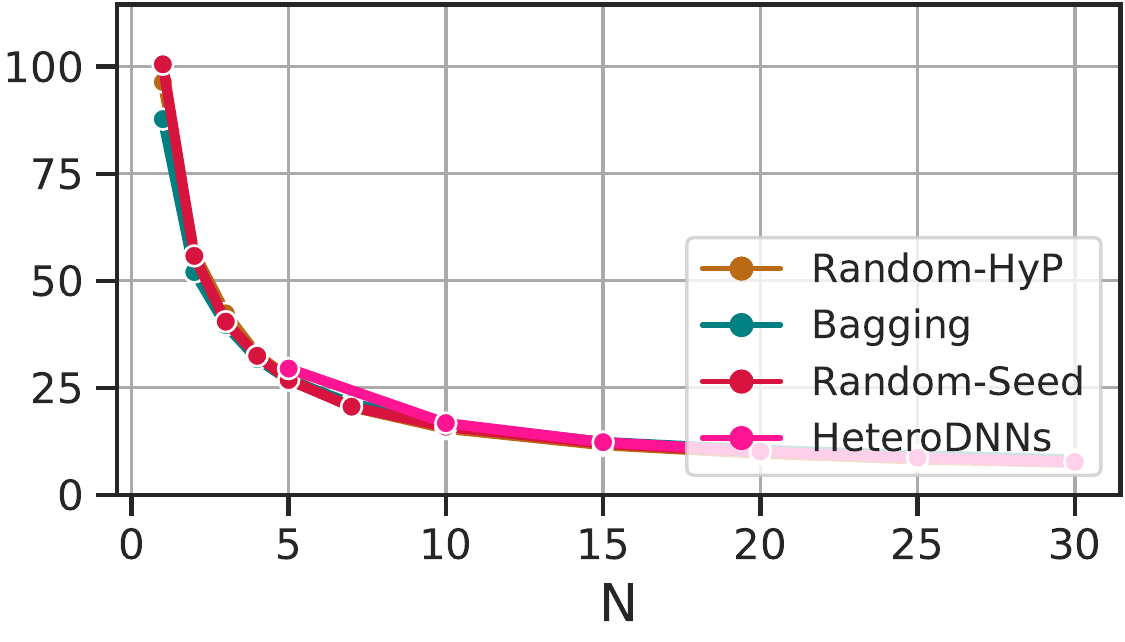}
        \subcaption{$\relev$ $- \redun$ \label{appendix:fig:scaling.novelty.per_model_MRPC}}
    \end{subfigure}   
    \hfill
    \begin{subfigure}[t]{0.19\linewidth}
        \vskip 0pt
        \includegraphics[width=\linewidth]{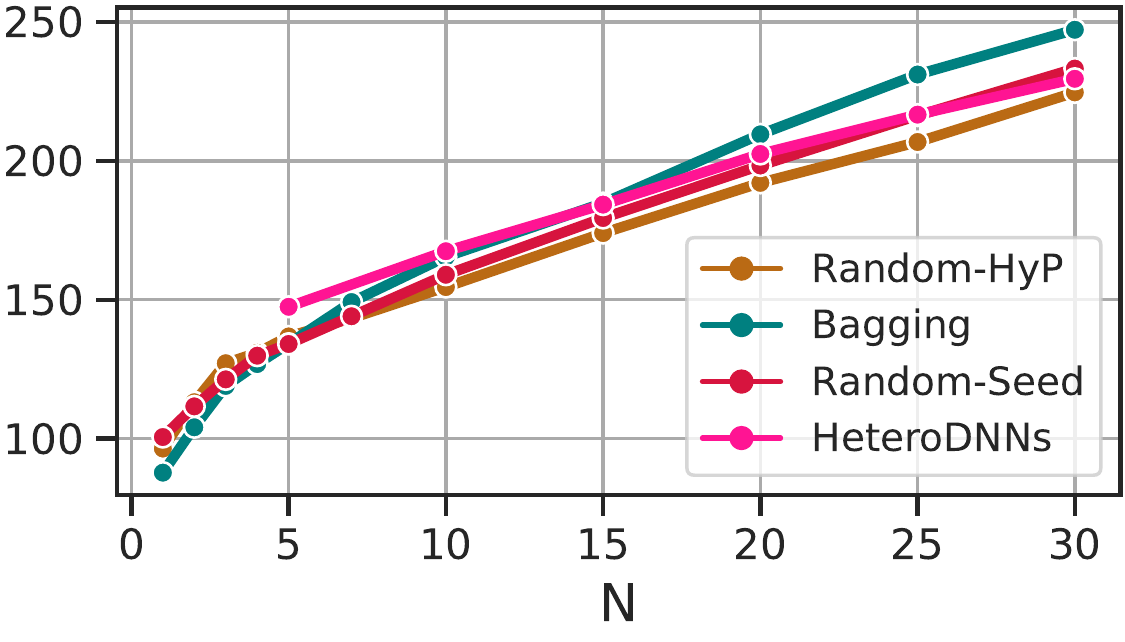}
        \captionsetup{justification=centering}
        \subcaption{$\StrengthDoublet $\newline $= N (\relev - \redun)$ \label{appendix:fig:scaling.E_without_combination_loss_MRPC}}
    \end{subfigure}
    \hfill
    \begin{subfigure}[t]{0.19\linewidth}
        \vskip 0pt
        \includegraphics[width=\linewidth]{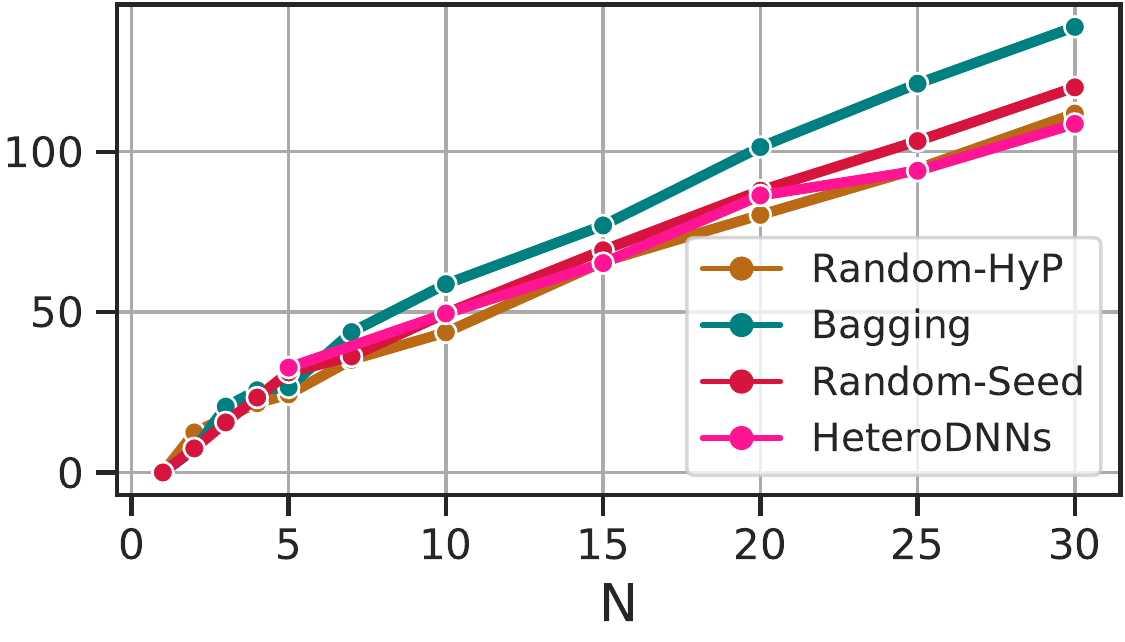}
        \subcaption{$\Combloss$\label{appendix:fig:scaling.combination_loss_MRPC}}
    \end{subfigure}
    \hfill
    \begin{subfigure}[t]{0.19\linewidth}
        \vskip 0pt
        \includegraphics[width=\linewidth]{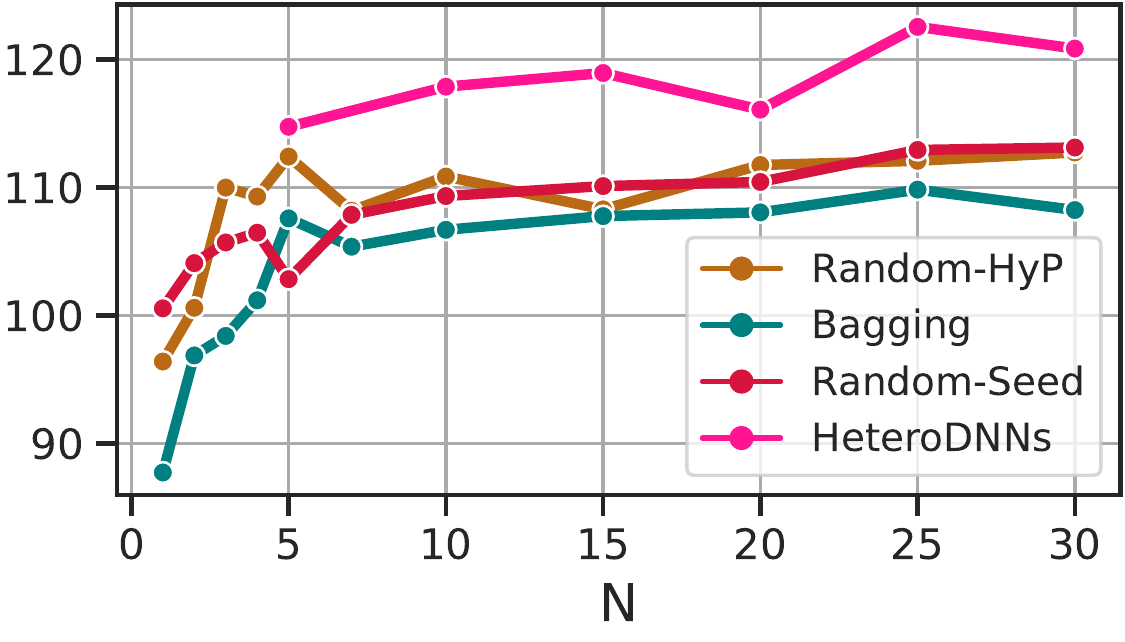}
        \captionsetup{justification=centering}
        \subcaption{$\StrengthTriplet = N (\relev - \redun) - \Combloss $\label{appendix:fig:scaling.E_MRPC}}
    \end{subfigure}
\caption{
\textbf{MRPC task.}
The change in ensemble quantities when the number of models $N$ is changed.
Each figure shows a specific quantity.
The ensemble systems used the SVM model combination.
Each value is an averages of the \NumTasks tasks.
$\perModelMetric$ denotes per-model metric values defined as: $\perModelMetricDef$.
\label{appendix:fig:scaling_MRPC}
}
\end{figure*}

\begin{table*}[t]
    \centering
    \caption{
        \textbf{MRPC task}.
        Statistics of ensemble systems described in \Cref{sec:ensemble_systems}.
        The rows and columns list the model generation and combination methods of \Cref{tb:ensemble_methods}, respectively.
        Each cell shows a quantity of a specific system $s$.
        Each quantity is the average over the \NumTasks tasks.
        Each system contains $N=15$ models.
        Color shows the rank within \textit{each column} (brighter is better).
        \label{tb:ablation_MRPC}
    }  
    \begin{subfigure}{\linewidth}
        \centering
        \small
        \tabcolsep 3.0pt
    \subcaption{
        Error rate reductions and lower bound reductions.
        The baseline values used in \Cref{eq:error_reduction,eq:lower_bound_reduction} were the followings.
        ER($s_0$): \SI{13.6}{\percent}.
        LB($s_0$) of $\BoundFuncTight(\StrengthTriplet)$: \SI{2.6}{\percent}.
        LB($s_0$) of $\BoundFuncTight(\StrengthDoublet)$: \SI{2.6}{\percent}.
        LB($s_0$) of $\BoundFuncLoose(\StrengthDoublet)$: \SI{-4.0}{\percent}.
        \label{tb:ablation_errors_MRPC}
    }

\begin{tabular}{lcccccccccccc}
\toprule
& \multicolumn{4}{c}{Error rate reductions \cref{eq:error_reduction}} & &   \multicolumn{7}{c}{Lower bound reductions \Cref{eq:lower_bound_reduction}} \\
\cmidrule(l{\tabcolsep}r{\tabcolsep}){2-5} \cmidrule(l{\tabcolsep}){7-13}
 & \multirow{2}{*}{Voting} & \multirow{2}{*}{LogR} & \multirow{2}{*}{SVM} & \multirow{2}{*}{RForest} & & \multicolumn{4}{c}{\textbf{\Cref{lemma:ensemble_bound_tight}} $\BoundFuncTight(\StrengthTriplet)$} & & \multirow{2}{*}{$\BoundFuncTight(\StrengthDoublet)$} & \multirow{2}{*}{\begin{tabular}{c}\Cref{lemma:ensemble_bound_loose} \\ $\BoundFuncLoose(\StrengthDoublet)$\end{tabular}} \\
 &  &  &  &  & & \multicolumn{1}{c}{Voting} & \multicolumn{1}{c}{LogR} & \multicolumn{1}{c}{SVM} & \multicolumn{1}{c}{RForest} & & &  \\

\midrule
Random-HyP    &    \cThird $7.7_{\pm{\mbox{\tiny 3.7}}}$ &  \cSecond $6.4_{\pm{\mbox{\tiny 2.3}}}$ &   \cThird $6.7_{\pm{\mbox{\tiny 3.8}}}$ &  \cSecond $6.3_{\pm{\mbox{\tiny 2.5}}}$ &        &    \cThird $7.3_{\pm{\mbox{\tiny 3.7}}}$ &  \cSecond $5.4_{\pm{\mbox{\tiny 2.3}}}$ &   \cThird $7.1_{\pm{\mbox{\tiny 3.7}}}$ &  \cSecond $5.5_{\pm{\mbox{\tiny 2.4}}}$ &        &  \cFourth $63_{\pm{\mbox{\tiny 4}}}$ &   \cThird $58_{\pm{\mbox{\tiny 4}}}$ \\
Bagging       &   \cFourth $5.0_{\pm{\mbox{\tiny 4.5}}}$ &   \cThird $5.6_{\pm{\mbox{\tiny 3.0}}}$ &  \cFourth $6.3_{\pm{\mbox{\tiny 3.2}}}$ &  \cFourth $3.0_{\pm{\mbox{\tiny 2.4}}}$ &        &   \cFourth $4.4_{\pm{\mbox{\tiny 4.2}}}$ &   \cThird $4.9_{\pm{\mbox{\tiny 3.0}}}$ &  \cFourth $6.5_{\pm{\mbox{\tiny 2.6}}}$ &  \cFourth $2.2_{\pm{\mbox{\tiny 2.2}}}$ &        &   \cFirst $72_{\pm{\mbox{\tiny 4}}}$ &   \cFirst $66_{\pm{\mbox{\tiny 5}}}$ \\
Random-Seed   &  \cSecond $12.0_{\pm{\mbox{\tiny 3.2}}}$ &  \cFourth $4.3_{\pm{\mbox{\tiny 2.5}}}$ &  \cSecond $7.0_{\pm{\mbox{\tiny 2.0}}}$ &   \cThird $6.0_{\pm{\mbox{\tiny 2.1}}}$ &        &  \cSecond $11.2_{\pm{\mbox{\tiny 3.0}}}$ &  \cFourth $3.5_{\pm{\mbox{\tiny 2.2}}}$ &  \cSecond $8.6_{\pm{\mbox{\tiny 2.5}}}$ &   \cThird $5.2_{\pm{\mbox{\tiny 1.9}}}$ &        &   \cThird $67_{\pm{\mbox{\tiny 6}}}$ &  \cSecond $62_{\pm{\mbox{\tiny 6}}}$ \\
Hetero-DNNs &   \cFirst $15.7_{\pm{\mbox{\tiny 2.1}}}$ &  \cFirst $18.0_{\pm{\mbox{\tiny 2.2}}}$ &  \cFirst $16.0_{\pm{\mbox{\tiny 4.3}}}$ &  \cFirst $11.3_{\pm{\mbox{\tiny 5.6}}}$ &        &   \cFirst $14.9_{\pm{\mbox{\tiny 2.1}}}$ &  \cFirst $17.2_{\pm{\mbox{\tiny 2.3}}}$ &  \cFirst $16.0_{\pm{\mbox{\tiny 4.7}}}$ &  \cFirst $10.5_{\pm{\mbox{\tiny 6.1}}}$ &        &  \cSecond $71_{\pm{\mbox{\tiny 2}}}$ &   \cFirst $66_{\pm{\mbox{\tiny 4}}}$ \\
\bottomrule

\end{tabular}

    \end{subfigure}
    \vfill
    \begin{subfigure}{\linewidth}
    \centering
    \small
    \tabcolsep 2.0pt
    \subcaption{
    Breakdown of ensemble strength defined in \cref{eq:triplet_decomposition}.
    We show per-model metric values defined as: $\perModelMetricDef$. Thus, $\StrengthTriplet = (\relev - \redun - \combloss) \ \times N $ holds.
    For intuitive understanding, all the values are normalized by the ensemble strength of baseline $\StrengthTriplet_{s_0}$, for example, $\Relev = \RelevHat / \StrengthTriplet_{s_0} \times 100$ where $\RelevHat$ is the raw value.
    \label{tb:ablation_triple_MRPC}
    }

\begin{tabular}{lccccccccccccc}
\toprule

{} & \multicolumn{4}{c}{\multirow{1}{*}{$\StrengthTripletWithArgs$}} & & \multicolumn{6}{c}{Per-model metric values} & \\
\cmidrule(l{\tabcolsep}r{\tabcolsep}){2-5} \cmidrule(l{\tabcolsep}r{\tabcolsep}){6-12}
{} &  & &  &  & & \multirow{2}{*}{$\perModelMetric_{\normalfont \text{relev}}$} & \multirow{2}{*}{$\perModelMetric_{\normalfont \text{redun}}$} & \multicolumn{4}{c}{ $\perModelMetric_{\normalfont \text{combloss}}$} & & \multirow{2}{*}{$\perModelMetric_{\normalfont \text{relev}} - \perModelMetric_{\normalfont \text{redun}}$} \\
{} & \multicolumn{1}{c}{Voting} & \multicolumn{1}{c}{LogR} & \multicolumn{1}{c}{SVM} & \multicolumn{1}{c}{RForest} & &  {} &  {} &  \multicolumn{1}{c}{Voting} & \multicolumn{1}{c}{LogR} &  \multicolumn{1}{c}{SVM} & \multicolumn{1}{c}{RForest} & &  {} \\

\midrule
Baseline ($s_0$)                    &    \multicolumn{4}{c}{\cBase 100 (the raw value is 0.336)} &        &   \cBase 100 &                                 \cBase 0 &                                \cBase 0 &                                \cBase 0 &                                \cBase 0 &                                 \cBase 0 &        &   \cBase 100 \\

\midrule

Random-HyP                   &   \cThird $108.5_{\pm{\mbox{\tiny 4.1}}}$ &  \cSecond $106.3_{\pm{\mbox{\tiny 2.6}}}$ &   \cThird $108.3_{\pm{\mbox{\tiny 4.1}}}$ &  \cSecond $106.5_{\pm{\mbox{\tiny 3.0}}}$ &        &  \cSecond $99.5_{\pm{\mbox{\tiny 1.7}}}$ &  \cSecond $87.9_{\pm{\mbox{\tiny 1.5}}}$ &   \cFirst $4.37_{\pm{\mbox{\tiny 0.50}}}$ &  \cSecond $4.51_{\pm{\mbox{\tiny 0.52}}}$ &  \cSecond $4.38_{\pm{\mbox{\tiny 0.49}}}$ &   \cFirst $4.50_{\pm{\mbox{\tiny 0.41}}}$ &        &   \cThird $11.6_{\pm{\mbox{\tiny 2.3}}}$ \\
Bagging                      &  \cFourth $105.3_{\pm{\mbox{\tiny 5.1}}}$ &   \cThird $105.8_{\pm{\mbox{\tiny 3.6}}}$ &  \cFourth $107.8_{\pm{\mbox{\tiny 3.2}}}$ &  \cFourth $102.6_{\pm{\mbox{\tiny 2.6}}}$ &        &   \cThird $90.1_{\pm{\mbox{\tiny 1.9}}}$ &   \cFirst $77.8_{\pm{\mbox{\tiny 1.5}}}$ &  \cFourth $5.30_{\pm{\mbox{\tiny 0.20}}}$ &  \cFourth $5.26_{\pm{\mbox{\tiny 0.23}}}$ &  \cFourth $5.13_{\pm{\mbox{\tiny 0.27}}}$ &  \cFourth $5.48_{\pm{\mbox{\tiny 0.55}}}$ &        &   \cFirst $12.3_{\pm{\mbox{\tiny 2.4}}}$ \\
Random-Seed                  &  \cSecond $113.1_{\pm{\mbox{\tiny 3.2}}}$ &  \cFourth $104.1_{\pm{\mbox{\tiny 2.6}}}$ &  \cSecond $110.1_{\pm{\mbox{\tiny 2.6}}}$ &   \cThird $106.1_{\pm{\mbox{\tiny 2.2}}}$ &        &  \cFirst $100.0_{\pm{\mbox{\tiny 0.0}}}$ &   \cThird $88.0_{\pm{\mbox{\tiny 0.5}}}$ &  \cSecond $4.42_{\pm{\mbox{\tiny 0.67}}}$ &   \cThird $5.02_{\pm{\mbox{\tiny 0.72}}}$ &   \cThird $4.62_{\pm{\mbox{\tiny 0.70}}}$ &   \cThird $4.89_{\pm{\mbox{\tiny 0.69}}}$ &        &  \cSecond $12.0_{\pm{\mbox{\tiny 0.5}}}$ \\
Hetero-DNNs                &   \cFirst $117.5_{\pm{\mbox{\tiny 2.7}}}$ &   \cFirst $120.3_{\pm{\mbox{\tiny 3.1}}}$ &   \cFirst $118.9_{\pm{\mbox{\tiny 5.9}}}$ &   \cFirst $112.5_{\pm{\mbox{\tiny 7.4}}}$ &        &   \cThird $90.1_{\pm{\mbox{\tiny 1.4}}}$ &   \cFirst $77.8_{\pm{\mbox{\tiny 1.2}}}$ &   \cThird $4.45_{\pm{\mbox{\tiny 0.24}}}$ &   \cFirst $4.26_{\pm{\mbox{\tiny 0.14}}}$ &   \cFirst $4.35_{\pm{\mbox{\tiny 0.12}}}$ &  \cSecond $4.78_{\pm{\mbox{\tiny 0.19}}}$ &        &   \cFirst $12.3_{\pm{\mbox{\tiny 1.9}}}$ \\

\bottomrule
\end{tabular}

\end{subfigure}

\end{table*}

\clearpage

\begin{figure*}[t!]
    \begin{subfigure}[t]{0.32\linewidth}
        \vskip 0pt
        \centering
        \includegraphics[width=0.65\linewidth]{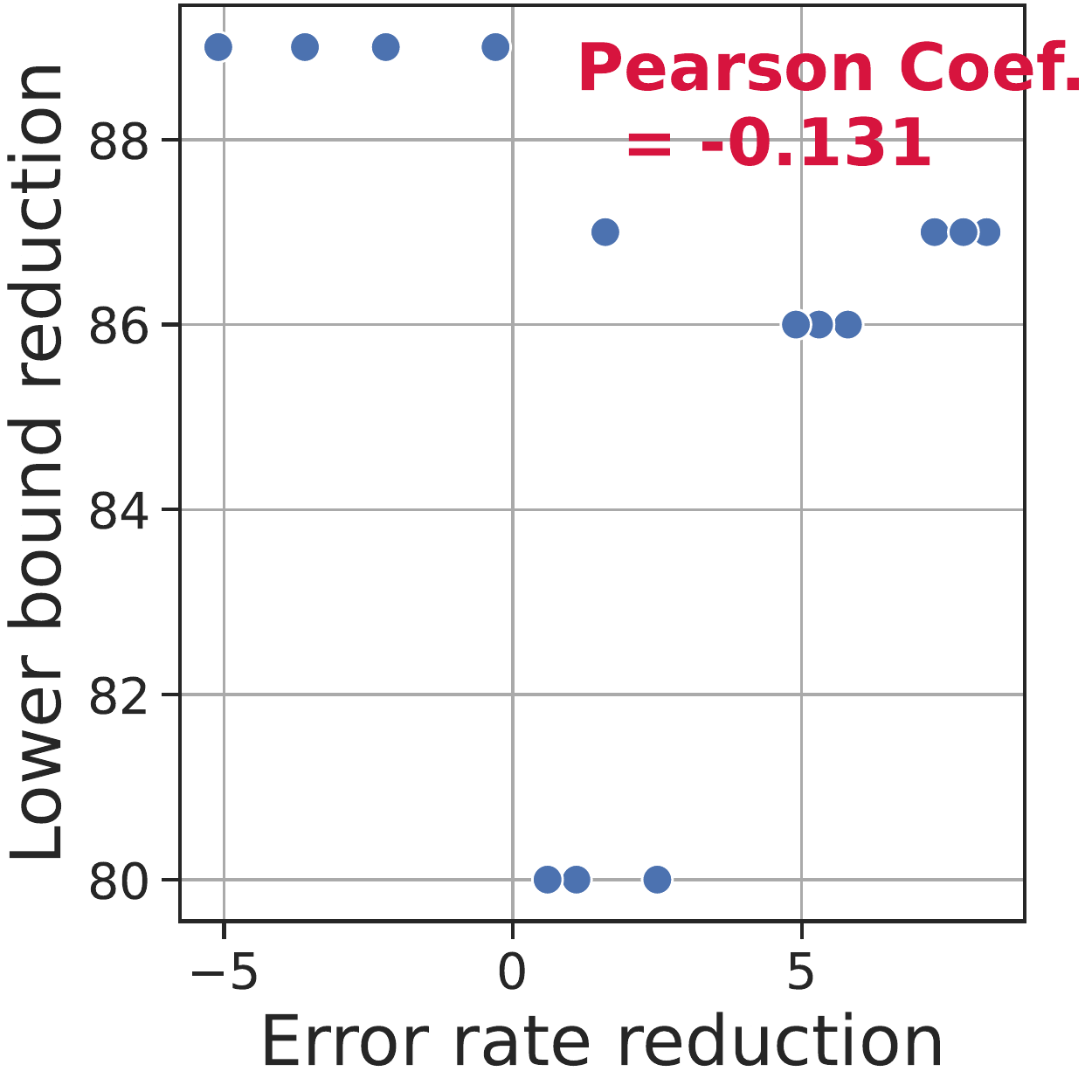}
        \caption{
        \Cref{lemma:ensemble_bound_loose} $\BoundLoose$.
        \label{fig:ERR_LBR_scatter_plot_loose_QQP}
        }
    \end{subfigure}
    \hfill
    \begin{subfigure}[t]{0.32\linewidth}
        \vskip 0pt
        \centering
        \includegraphics[width=0.65\linewidth]{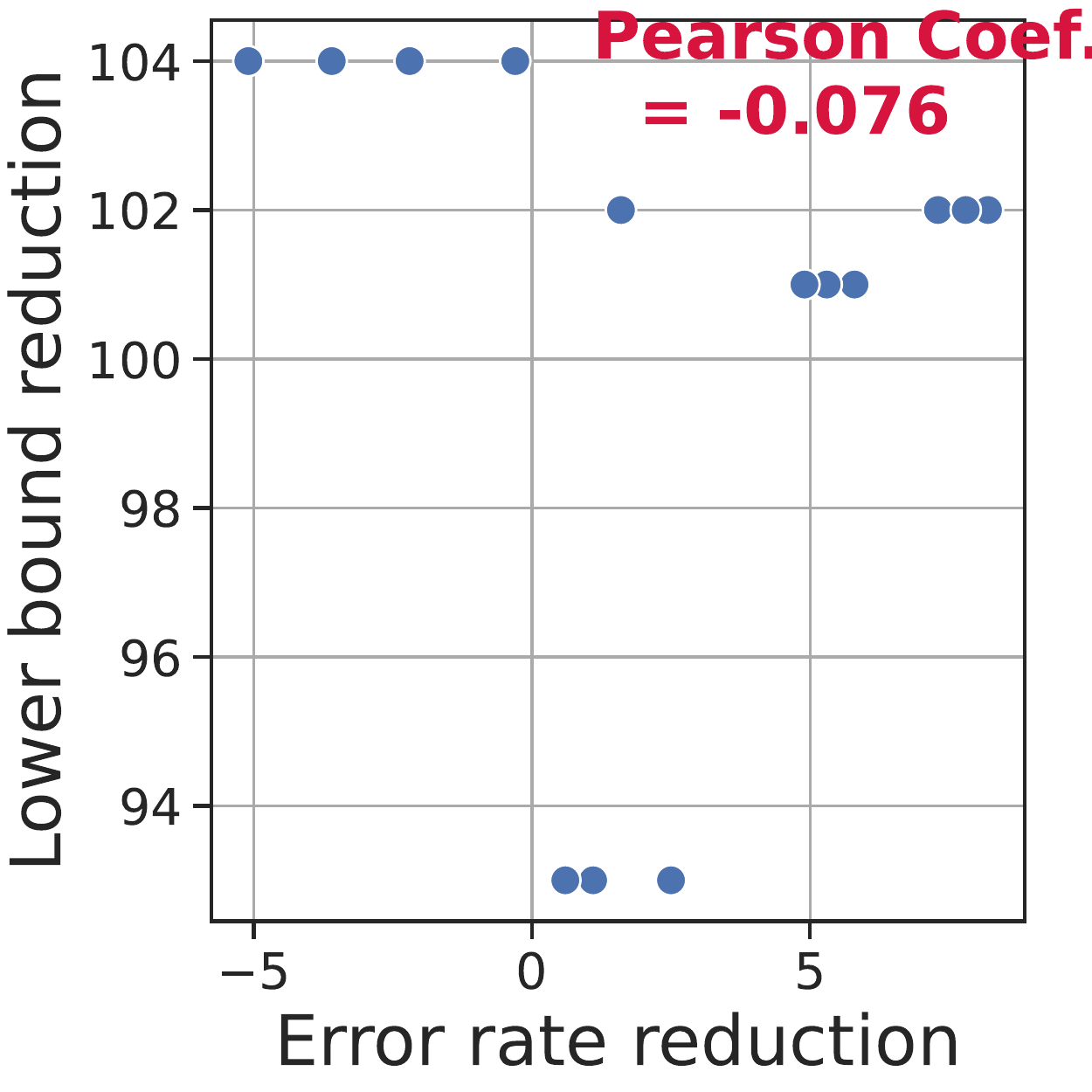}
        \caption{
            $\BoundTightWOCombLoss$.
            \label{fig:ERR_LBR_scatter_plot_tight_wo_combloss_QQP}
        }
    \end{subfigure}
    \hfill
    \begin{subfigure}[t]{0.32\linewidth}
        \vskip 0pt
        \centering
        \includegraphics[width=0.65\linewidth]{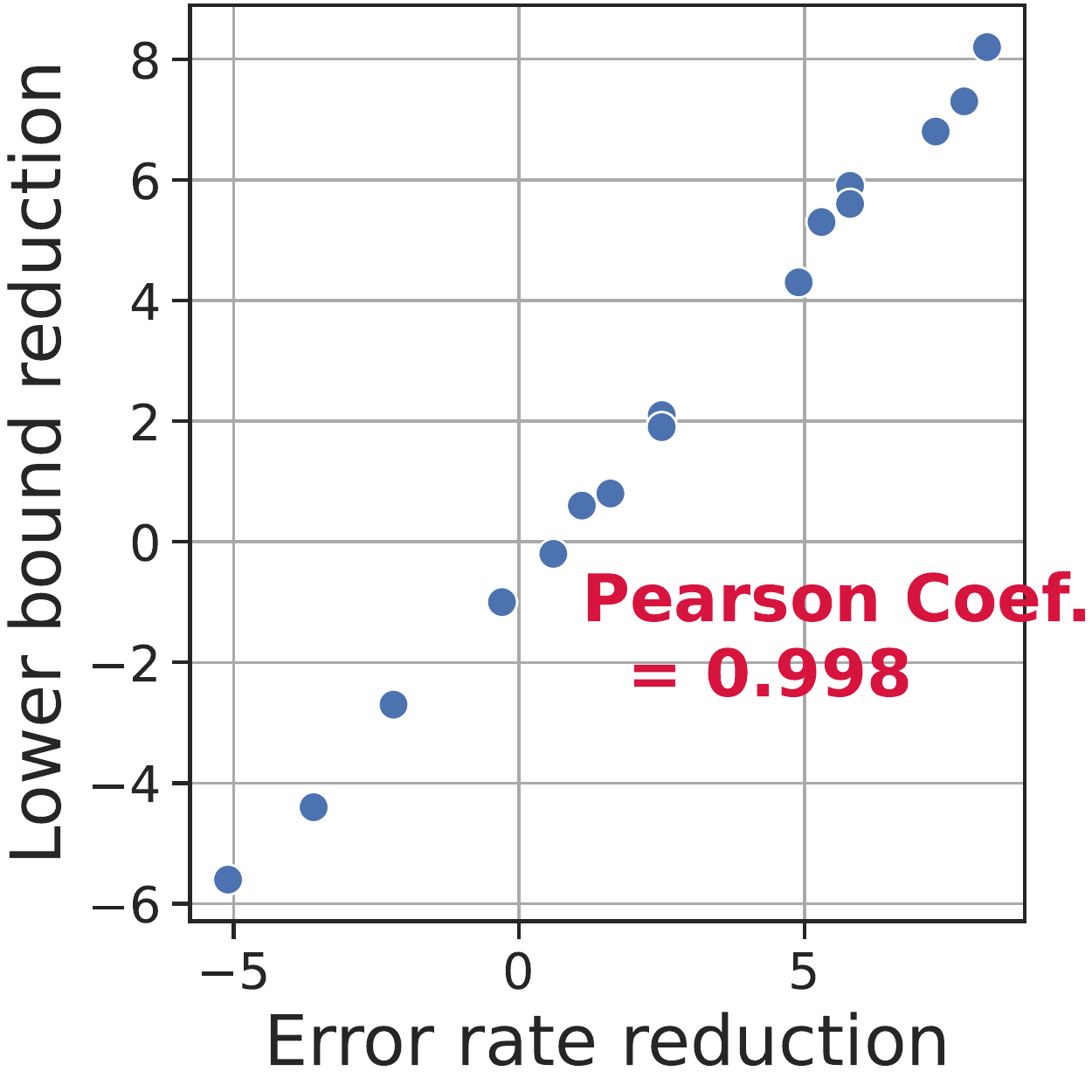}
        \caption{
            \textbf{\Cref{lemma:ensemble_bound_tight}} $\BoundTight$.
            \label{fig:ERR_LBR_scatter_plot_tight_QQP}
        }
    \end{subfigure}
    \hfill

    \caption{
    \textbf{QQP task.}
    Correlations between error rate reductions and lower bound reductions.
    Each figure uses different type of lower bound.
    Each point in the figures shows a quantity of a specific ensemble system $s$ and the quantity is the average over the \NumTasks tasks.
    See \Cref{tb:ablation_QQP} for the real value of each point.
    We used the \NumSystems ensemble systems described in \Cref{sec:ensemble_systems}.
    Each system $s$ used $N=15$ models.
    The baseline values in \Cref{eq:error_reduction,eq:lower_bound_reduction} were the followings:
    ER($s_0$): \SI{14.0}{\percent}.
    LB($s_0$) of $\BoundFuncTight(\StrengthTriplet)$: \SI{2.1}{\percent}.
    LB($s_0$) of $\BoundFuncTight(\StrengthDoublet)$: \SI{2.1}{\percent}.
    LB($s_0$) of $\BoundFuncLoose(\StrengthDoublet)$: \SI{-2.9}{\percent}.
    \label{fig:ERR_LBR_scatter_plot_QQP}
    }
\end{figure*}

\begin{figure*}[h!]
    \begin{subfigure}[t]{0.19\linewidth}
        \vskip 0pt
        \includegraphics[width=\linewidth]{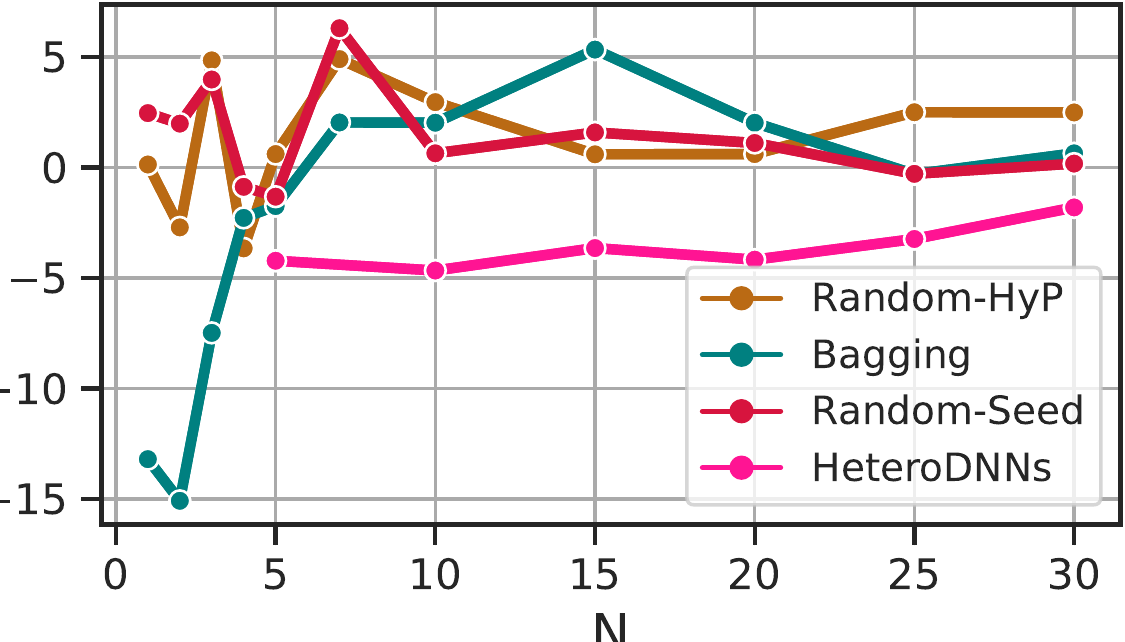}
        \subcaption{Error rate reduction.\label{appendix:fig:scaling.stacking_QQP}}
    \end{subfigure}
    \hfill
    \begin{subfigure}[t]{0.19\linewidth}
        \vskip 0pt
        \includegraphics[width=\linewidth]{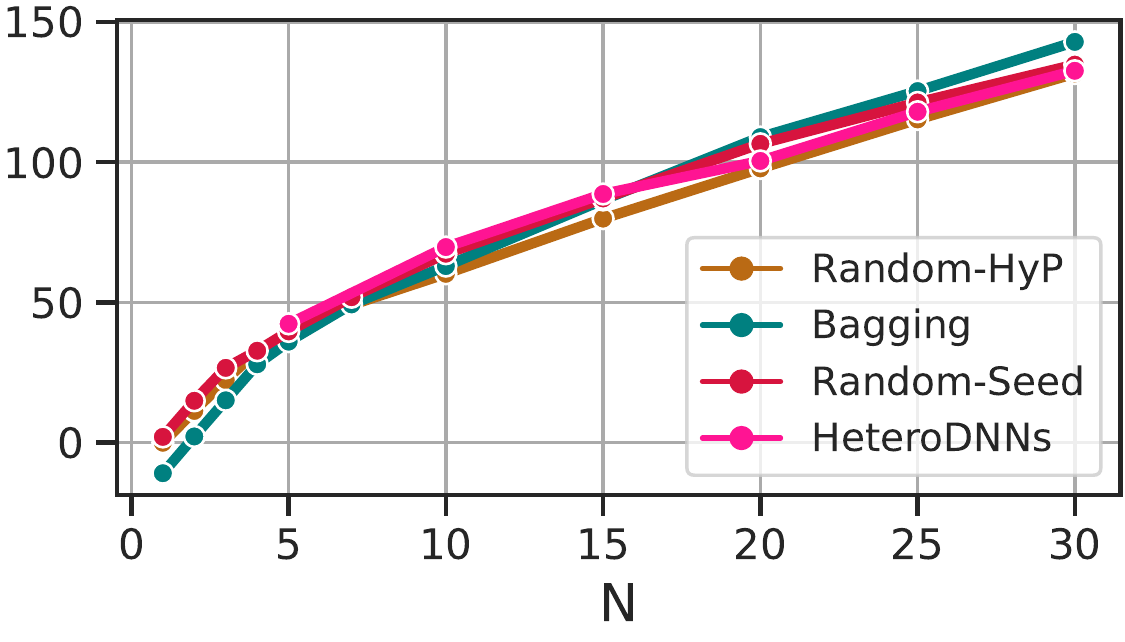}
        \subcaption{Lower bound reduction by \Cref{lemma:ensemble_bound_loose} $\BoundLoose$.\label{appendix:fig:scaling.bound.voting.previous_research_QQP}}
    \end{subfigure}
    \begin{subfigure}[t]{0.19\linewidth}
        \vskip 0pt
        \includegraphics[width=\linewidth]{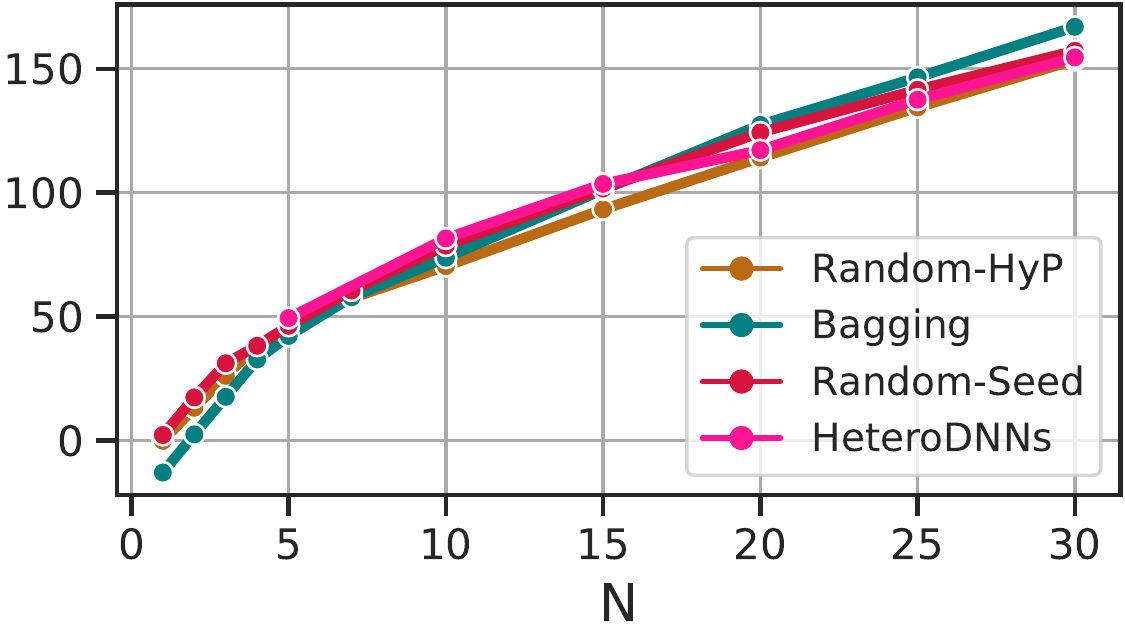}
        \subcaption{Lower bound reduction by $\BoundTightWOCombLoss$. \label{appendix:fig:scaling.bound.voting.ours_wo_combloss_QQP}}
    \end{subfigure}
    \hfill
     \begin{subfigure}[t]{0.19\linewidth}
        \vskip 0pt
        \includegraphics[width=\linewidth]{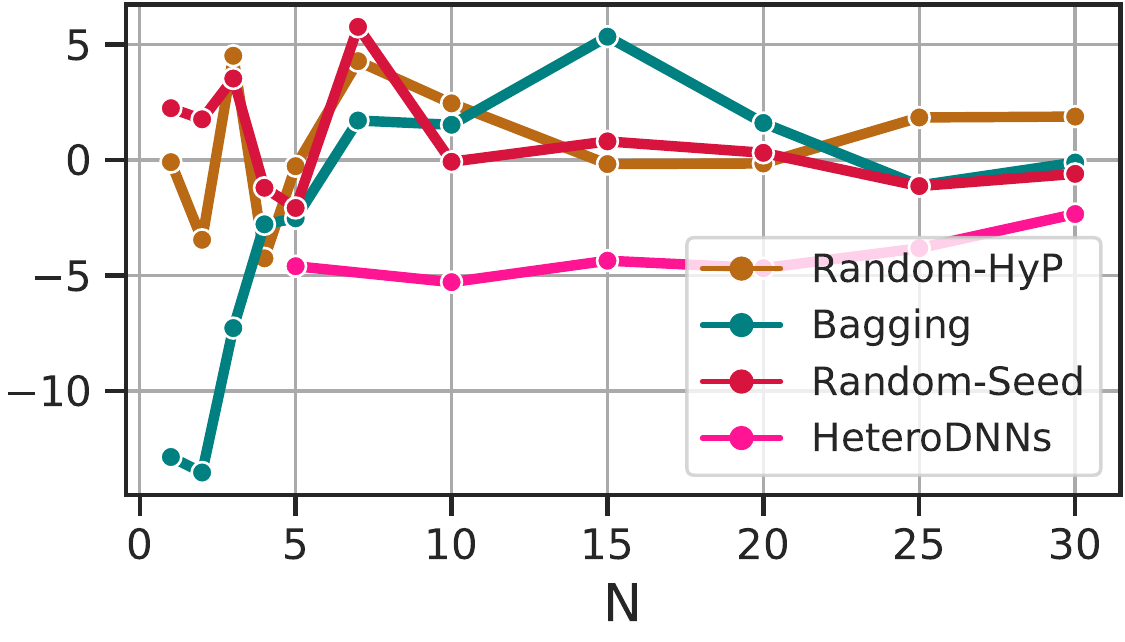}
        \subcaption{Lower bound reduction by \textbf{\Cref{lemma:ensemble_bound_tight}} $\BoundTight$.\label{appendix:fig:scaling.stacking.bound_QQP}}
    \end{subfigure}
    \hfill
    \begin{subfigure}[t]{0.19\linewidth}
        \vskip 0pt
        \includegraphics[width=\linewidth]{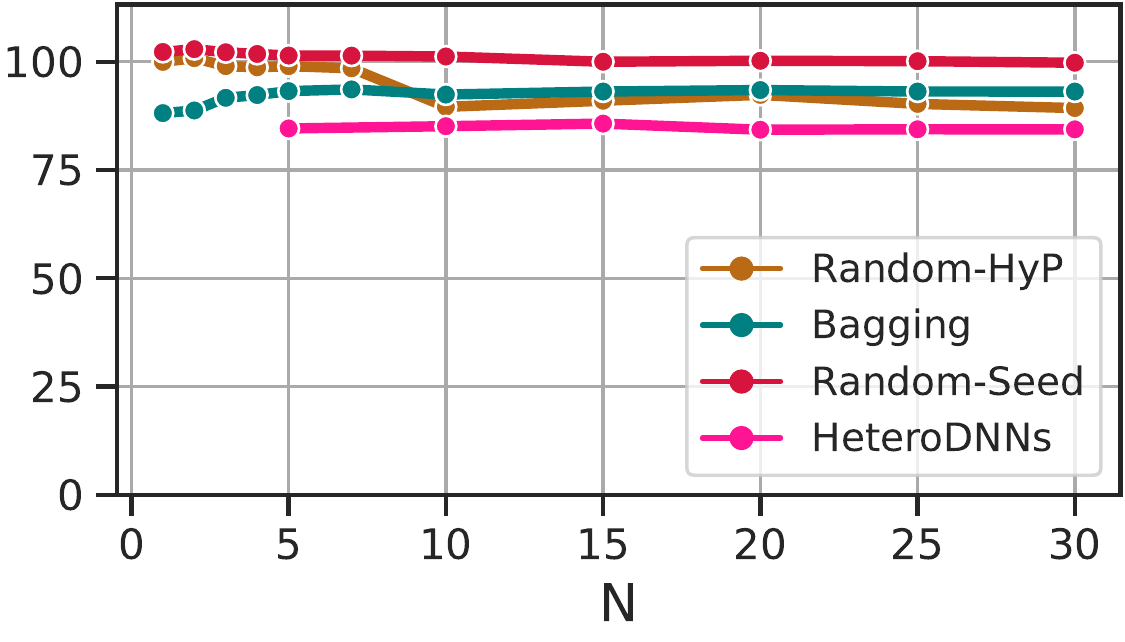}
        \subcaption{$\relev$ \label{appendix:fig:scaling.relevance.per_model_QQP}}   
    \end{subfigure}   
    \hfill
    \vfill
    \begin{subfigure}[t]{0.19\linewidth}
        \vskip 0pt
        \includegraphics[width=\linewidth]{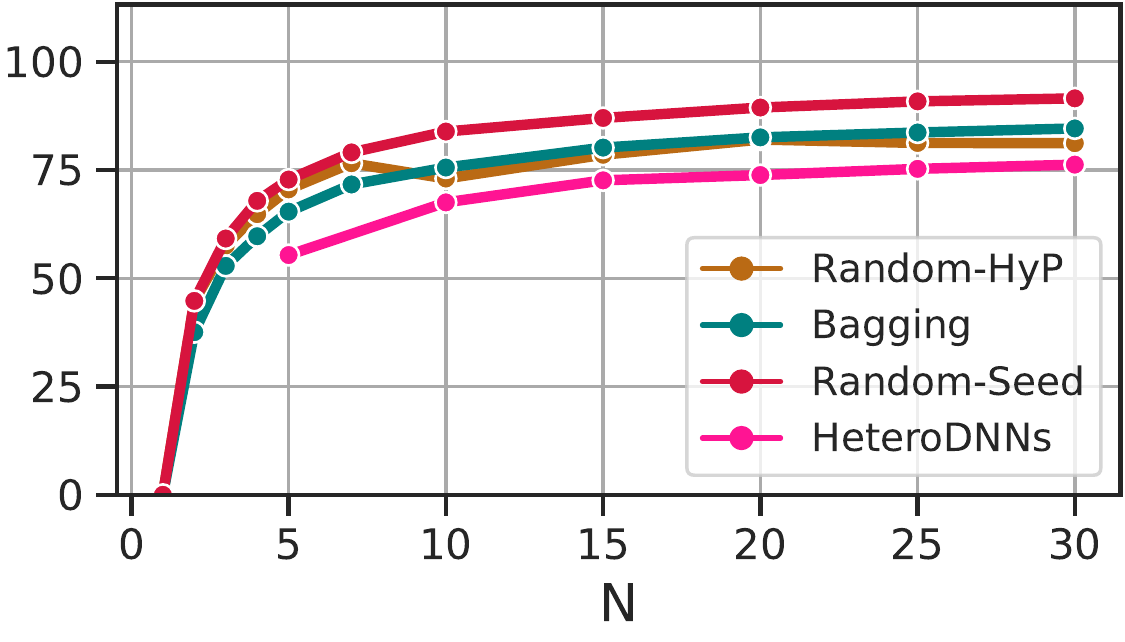}
        \subcaption{$\redun$ \label{appendix:fig:scaling.redundancy.per_model_QQP}}   
    \end{subfigure}   
    \hfill
    \begin{subfigure}[t]{0.19\linewidth}
        \vskip 0pt
        \includegraphics[width=\linewidth]{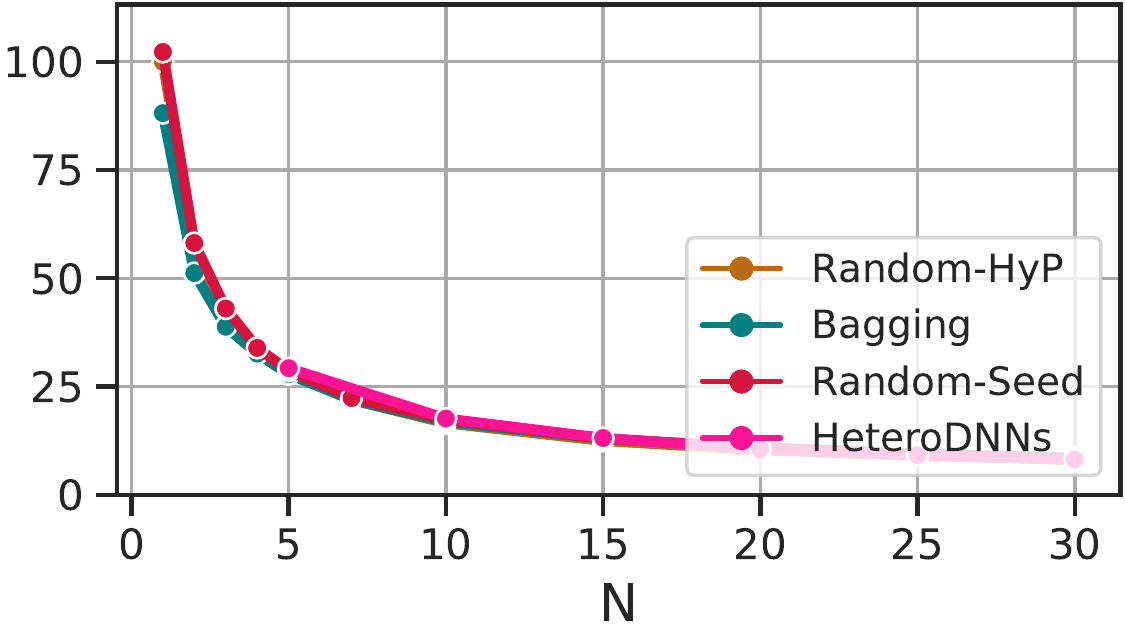}
        \subcaption{$\relev$ $- \redun$ \label{appendix:fig:scaling.novelty.per_model_QQP}}
    \end{subfigure}   
    \hfill
    \begin{subfigure}[t]{0.19\linewidth}
        \vskip 0pt
        \includegraphics[width=\linewidth]{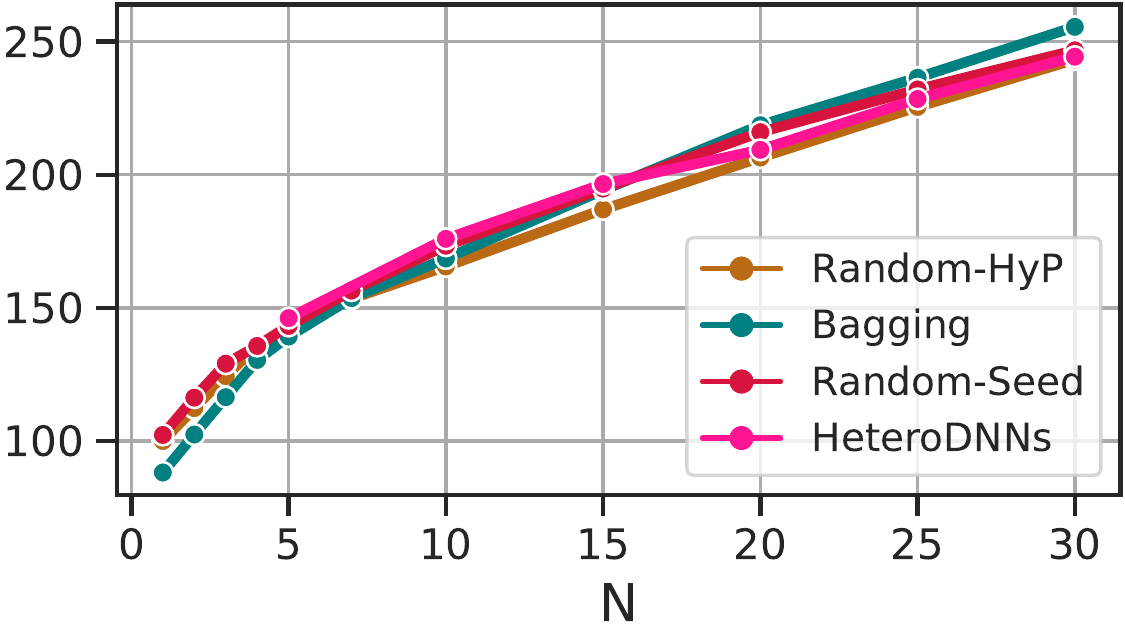}
        \captionsetup{justification=centering}
        \subcaption{$\StrengthDoublet $\newline $= N (\relev - \redun)$ \label{appendix:fig:scaling.E_without_combination_loss_QQP}}
    \end{subfigure}
    \hfill
    \begin{subfigure}[t]{0.19\linewidth}
        \vskip 0pt
        \includegraphics[width=\linewidth]{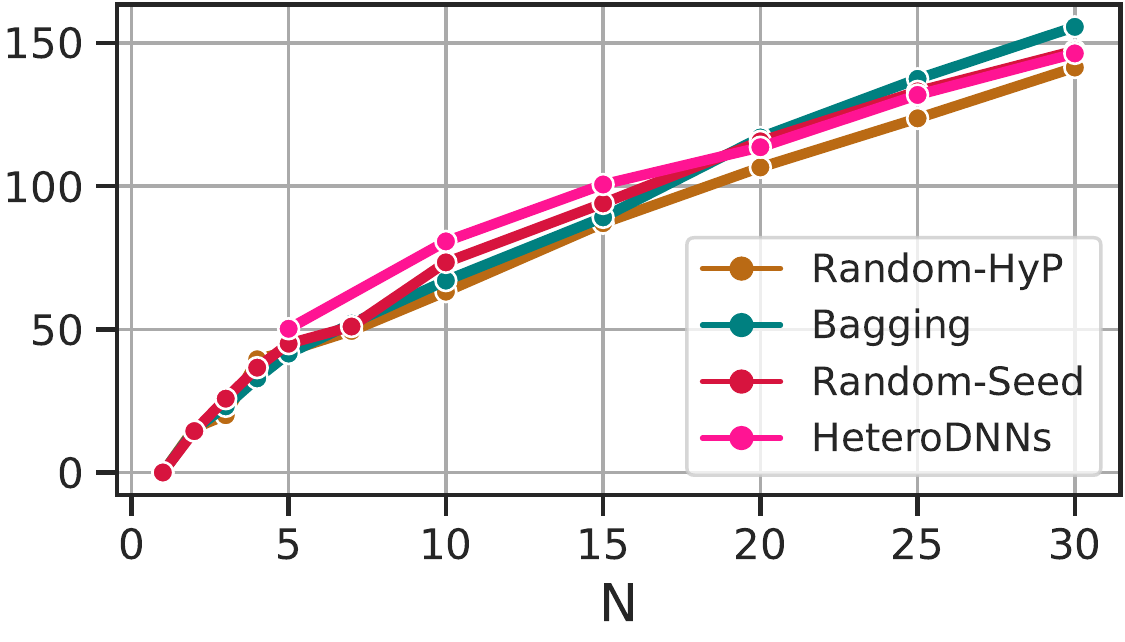}
        \subcaption{$\Combloss$\label{appendix:fig:scaling.combination_loss_QQP}}
    \end{subfigure}
    \hfill
    \begin{subfigure}[t]{0.19\linewidth}
        \vskip 0pt
        \includegraphics[width=\linewidth]{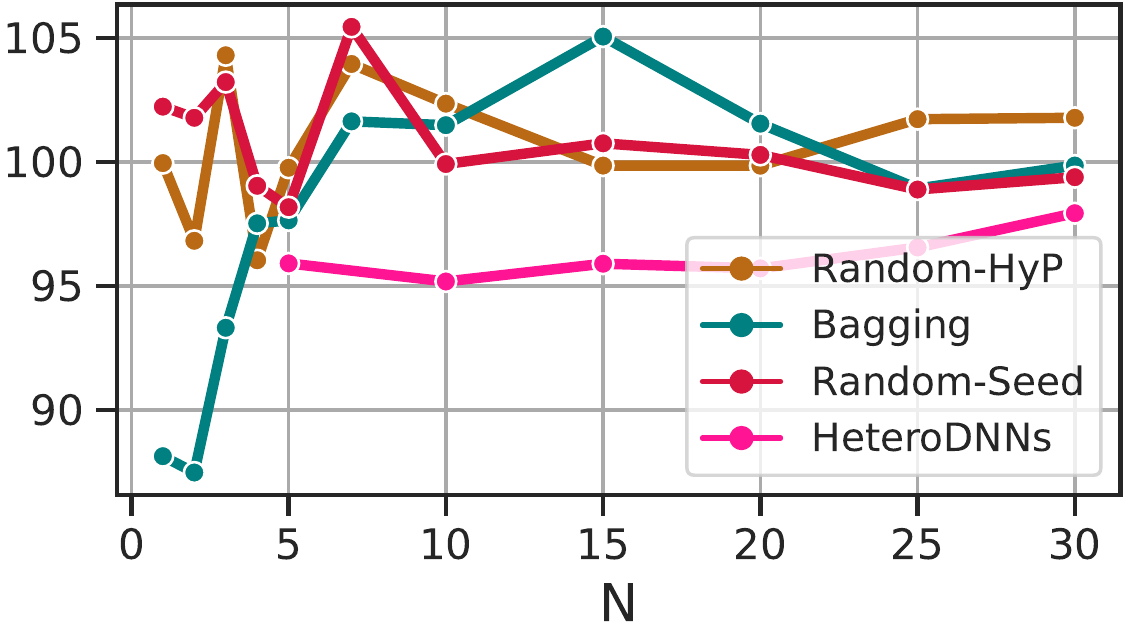}
        \captionsetup{justification=centering}
        \subcaption{$\StrengthTriplet = N (\relev - \redun) - \Combloss $\label{appendix:fig:scaling.E_QQP}}
    \end{subfigure}
\caption{
\textbf{QQP task.}
The change in ensemble quantities when the number of models $N$ is changed.
Each figure shows a specific quantity.
The ensemble systems used the SVM model combination.
Each value is an averages of the \NumTasks tasks.
$\perModelMetric$ denotes per-model metric values defined as: $\perModelMetricDef$.
\label{appendix:fig:scaling_QQP}
}
\end{figure*}

\begin{table*}[h]
    \centering
    \caption{
        \textbf{QQP task}.
        Statistics of ensemble systems described in \Cref{sec:ensemble_systems}.
        The rows and columns list the model generation and combination methods of \Cref{tb:ensemble_methods}, respectively.
        Each cell shows a quantity of a specific system $s$.
        Each quantity is the average over the \NumTasks tasks.
        Each system contains $N=15$ models.
        Color shows the rank within \textit{each column} (brighter is better).
        \label{tb:ablation_QQP}
    }  
    \begin{subfigure}{\linewidth}
        \centering
        \small
        \tabcolsep 3.0pt
    \subcaption{
        Error rate reductions and lower bound reductions.
        The baseline values used in \Cref{eq:error_reduction,eq:lower_bound_reduction} were the followings.
        ER($s_0$): \SI{14.0}{\percent}.
        LB($s_0$) of $\BoundFuncTight(\StrengthTriplet)$: \SI{2.1}{\percent}.
        LB($s_0$) of $\BoundFuncTight(\StrengthDoublet)$: \SI{2.1}{\percent}.
        LB($s_0$) of $\BoundFuncLoose(\StrengthDoublet)$: \SI{-2.9}{\percent}.
        \label{tb:ablation_errors_QQP}
    }

\begin{tabular}{lcccccccccccc}
\toprule
& \multicolumn{4}{c}{Error rate reductions \cref{eq:error_reduction}} & &   \multicolumn{7}{c}{Lower bound reductions \Cref{eq:lower_bound_reduction}} \\
\cmidrule(l{\tabcolsep}r{\tabcolsep}){2-5} \cmidrule(l{\tabcolsep}){7-13}
 & \multirow{2}{*}{Voting} & \multirow{2}{*}{LogR} & \multirow{2}{*}{SVM} & \multirow{2}{*}{RForest} & & \multicolumn{4}{c}{\textbf{\Cref{lemma:ensemble_bound_tight}} $\BoundFuncTight(\StrengthTriplet)$} & & \multirow{2}{*}{$\BoundFuncTight(\StrengthDoublet)$} & \multirow{2}{*}{\begin{tabular}{c}\Cref{lemma:ensemble_bound_loose} \\ $\BoundFuncLoose(\StrengthDoublet)$\end{tabular}} \\
 &  &  &  &  & & \multicolumn{1}{c}{Voting} & \multicolumn{1}{c}{LogR} & \multicolumn{1}{c}{SVM} & \multicolumn{1}{c}{RForest} & & &  \\

\midrule
Random-HyP    &                         \cThird $2.5_{\pm{\mbox{\tiny 2.7}}}$ &                         \cThird $1.1_{\pm{\mbox{\tiny 4.3}}}$ &                         \cThird $0.6_{\pm{\mbox{\tiny 3.7}}}$ &                         \cThird $2.5_{\pm{\mbox{\tiny 1.5}}}$ &        &                         \cThird $2.1_{\pm{\mbox{\tiny 2.9}}}$ &                         \cThird $0.6_{\pm{\mbox{\tiny 4.6}}}$ &   \cThird $\scalebox{1.5}[1.0]{-}0.2_{\pm{\mbox{\tiny 4.0}}}$ &                         \cThird $1.9_{\pm{\mbox{\tiny 1.7}}}$ &        &   \cFourth $93_{\pm{\mbox{\tiny 17}}}$ &  \cFourth $80_{\pm{\mbox{\tiny 16}}}$ \\
Bagging       &                        \cSecond $5.8_{\pm{\mbox{\tiny 6.0}}}$ &                        \cSecond $5.8_{\pm{\mbox{\tiny 7.1}}}$ &                         \cFirst $5.3_{\pm{\mbox{\tiny 5.4}}}$ &                        \cSecond $4.9_{\pm{\mbox{\tiny 6.0}}}$ &        &                        \cSecond $5.9_{\pm{\mbox{\tiny 6.7}}}$ &                        \cSecond $5.6_{\pm{\mbox{\tiny 7.7}}}$ &                         \cFirst $5.3_{\pm{\mbox{\tiny 5.9}}}$ &                        \cSecond $4.3_{\pm{\mbox{\tiny 6.5}}}$ &        &    \cThird $101_{\pm{\mbox{\tiny 3}}}$ &    \cThird $86_{\pm{\mbox{\tiny 2}}}$ \\
Random-Seed   &                         \cFirst $8.2_{\pm{\mbox{\tiny 0.3}}}$ &                         \cFirst $7.3_{\pm{\mbox{\tiny 1.4}}}$ &                        \cSecond $1.6_{\pm{\mbox{\tiny 0.4}}}$ &                         \cFirst $7.8_{\pm{\mbox{\tiny 1.7}}}$ &        &                         \cFirst $8.2_{\pm{\mbox{\tiny 0.3}}}$ &                         \cFirst $6.8_{\pm{\mbox{\tiny 1.4}}}$ &                        \cSecond $0.8_{\pm{\mbox{\tiny 0.4}}}$ &                         \cFirst $7.3_{\pm{\mbox{\tiny 1.8}}}$ &        &  \cSecond $102_{\pm{\mbox{\tiny 15}}}$ &  \cSecond $87_{\pm{\mbox{\tiny 14}}}$ \\
Hetero-DNNs &  \cFourth $\scalebox{1.5}[1.0]{-}5.1_{\pm{\mbox{\tiny 1.8}}}$ &  \cFourth $\scalebox{1.5}[1.0]{-}2.2_{\pm{\mbox{\tiny 1.1}}}$ &  \cFourth $\scalebox{1.5}[1.0]{-}3.6_{\pm{\mbox{\tiny 4.0}}}$ &  \cFourth $\scalebox{1.5}[1.0]{-}0.3_{\pm{\mbox{\tiny 2.5}}}$ &        &  \cFourth $\scalebox{1.5}[1.0]{-}5.6_{\pm{\mbox{\tiny 1.9}}}$ &  \cFourth $\scalebox{1.5}[1.0]{-}2.7_{\pm{\mbox{\tiny 0.6}}}$ &  \cFourth $\scalebox{1.5}[1.0]{-}4.4_{\pm{\mbox{\tiny 4.0}}}$ &  \cFourth $\scalebox{1.5}[1.0]{-}1.0_{\pm{\mbox{\tiny 2.7}}}$ &        &   \cFirst $104_{\pm{\mbox{\tiny 11}}}$ &   \cFirst $89_{\pm{\mbox{\tiny 11}}}$ \\
\bottomrule

\end{tabular}

    \end{subfigure}
    \vfill
    \begin{subfigure}{\linewidth}
    \centering
    \small
    \tabcolsep 2.0pt
    \subcaption{
    Breakdown of ensemble strength defined in \cref{eq:triplet_decomposition}.
    We show per-model metric values defined as: $\perModelMetricDef$. Thus, $\StrengthTriplet = (\relev - \redun - \combloss) \ \times N $ holds.
    For intuitive understanding, all the values are normalized by the ensemble strength of baseline $\StrengthTriplet_{s_0}$, for example, $\Relev = \RelevHat / \StrengthTriplet_{s_0} \times 100$ where $\RelevHat$ is the raw value.
    \label{tb:ablation_triple_QQP}
    }

\begin{tabular}{lccccccccccccc}
\toprule

{} & \multicolumn{4}{c}{\multirow{1}{*}{$\StrengthTripletWithArgs$}} & & \multicolumn{6}{c}{Per-model metric values} & \\
\cmidrule(l{\tabcolsep}r{\tabcolsep}){2-5} \cmidrule(l{\tabcolsep}r{\tabcolsep}){6-12}
{} &  & &  &  & & \multirow{2}{*}{$\perModelMetric_{\normalfont \text{relev}}$} & \multirow{2}{*}{$\perModelMetric_{\normalfont \text{redun}}$} & \multicolumn{4}{c}{ $\perModelMetric_{\normalfont \text{combloss}}$} & & \multirow{2}{*}{$\perModelMetric_{\normalfont \text{relev}} - \perModelMetric_{\normalfont \text{redun}}$} \\
{} & \multicolumn{1}{c}{Voting} & \multicolumn{1}{c}{LogR} & \multicolumn{1}{c}{SVM} & \multicolumn{1}{c}{RForest} & &  {} &  {} &  \multicolumn{1}{c}{Voting} & \multicolumn{1}{c}{LogR} &  \multicolumn{1}{c}{SVM} & \multicolumn{1}{c}{RForest} & &  {} \\

\midrule
Baseline ($s_0$)                    &    \multicolumn{4}{c}{\cBase 100 (the raw value is 0.343)} &        &   \cBase 100 &                                 \cBase 0 &                                \cBase 0 &                                \cBase 0 &                                \cBase 0 &                                 \cBase 0 &        &   \cBase 100 \\

\midrule

Random-HyP                   &   \cThird $102.0_{\pm{\mbox{\tiny 2.8}}}$ &   \cThird $100.6_{\pm{\mbox{\tiny 4.4}}}$ &    \cThird $99.9_{\pm{\mbox{\tiny 3.8}}}$ &   \cThird $101.8_{\pm{\mbox{\tiny 1.6}}}$ &        &   \cThird $91.0_{\pm{\mbox{\tiny 1.6}}}$ &  \cSecond $78.5_{\pm{\mbox{\tiny 1.8}}}$ &   \cFirst $5.66_{\pm{\mbox{\tiny 1.12}}}$ &   \cFirst $5.76_{\pm{\mbox{\tiny 0.97}}}$ &   \cFirst $5.81_{\pm{\mbox{\tiny 0.99}}}$ &   \cFirst $5.68_{\pm{\mbox{\tiny 1.14}}}$ &        &  \cFourth $12.5_{\pm{\mbox{\tiny 2.4}}}$ \\
Bagging                      &  \cSecond $105.6_{\pm{\mbox{\tiny 6.4}}}$ &  \cSecond $105.4_{\pm{\mbox{\tiny 7.3}}}$ &   \cFirst $105.1_{\pm{\mbox{\tiny 5.7}}}$ &  \cSecond $104.1_{\pm{\mbox{\tiny 6.2}}}$ &        &  \cSecond $93.1_{\pm{\mbox{\tiny 2.5}}}$ &   \cThird $80.2_{\pm{\mbox{\tiny 2.4}}}$ &   \cThird $5.91_{\pm{\mbox{\tiny 0.42}}}$ &   \cThird $5.92_{\pm{\mbox{\tiny 0.49}}}$ &  \cSecond $5.94_{\pm{\mbox{\tiny 0.38}}}$ &   \cThird $6.01_{\pm{\mbox{\tiny 0.32}}}$ &        &   \cThird $12.9_{\pm{\mbox{\tiny 3.5}}}$ \\
Random-Seed                  &   \cFirst $107.6_{\pm{\mbox{\tiny 0.5}}}$ &   \cFirst $106.4_{\pm{\mbox{\tiny 1.4}}}$ &  \cSecond $100.7_{\pm{\mbox{\tiny 0.3}}}$ &   \cFirst $106.8_{\pm{\mbox{\tiny 1.7}}}$ &        &  \cFirst $100.0_{\pm{\mbox{\tiny 0.0}}}$ &  \cFourth $87.0_{\pm{\mbox{\tiny 1.0}}}$ &  \cSecond $5.81_{\pm{\mbox{\tiny 0.95}}}$ &  \cSecond $5.89_{\pm{\mbox{\tiny 1.02}}}$ &   \cThird $6.27_{\pm{\mbox{\tiny 0.98}}}$ &  \cSecond $5.86_{\pm{\mbox{\tiny 1.05}}}$ &        &  \cSecond $13.0_{\pm{\mbox{\tiny 1.0}}}$ \\
Hetero-DNNs                &   \cFourth $94.8_{\pm{\mbox{\tiny 1.7}}}$ &   \cFourth $97.5_{\pm{\mbox{\tiny 0.6}}}$ &   \cFourth $95.9_{\pm{\mbox{\tiny 3.6}}}$ &   \cFourth $99.1_{\pm{\mbox{\tiny 2.4}}}$ &        &  \cFourth $85.7_{\pm{\mbox{\tiny 1.3}}}$ &   \cFirst $72.6_{\pm{\mbox{\tiny 0.6}}}$ &  \cFourth $6.78_{\pm{\mbox{\tiny 0.74}}}$ &  \cFourth $6.61_{\pm{\mbox{\tiny 0.81}}}$ &  \cFourth $6.71_{\pm{\mbox{\tiny 0.71}}}$ &  \cFourth $6.50_{\pm{\mbox{\tiny 0.87}}}$ &        &   \cFirst $13.1_{\pm{\mbox{\tiny 1.5}}}$ \\

\bottomrule
\end{tabular}

\end{subfigure}

\end{table*}

\clearpage

\begin{figure*}[t!]
    \begin{subfigure}[t]{0.32\linewidth}
        \vskip 0pt
        \centering
        \includegraphics[width=0.65\linewidth]{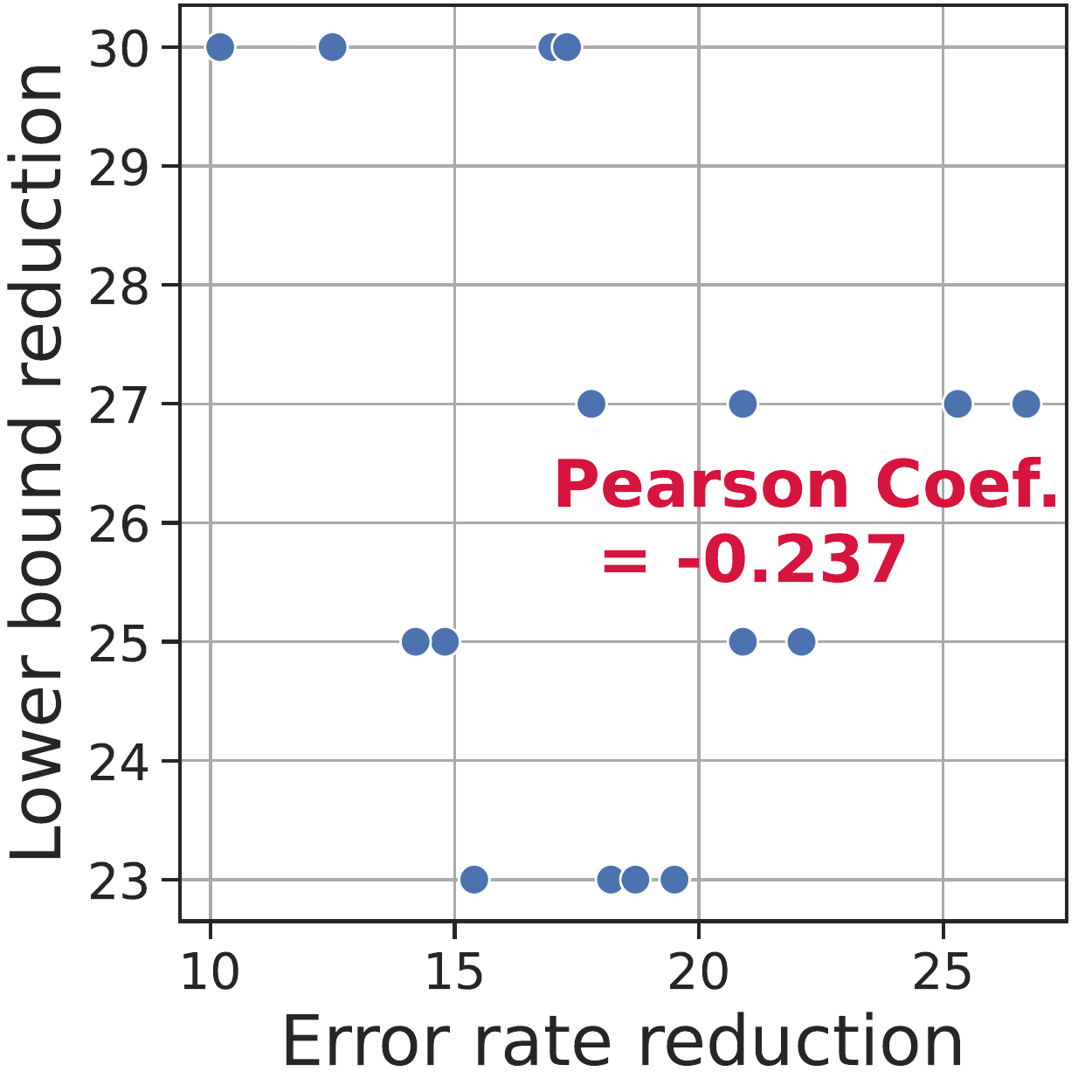}
        \caption{
        \Cref{lemma:ensemble_bound_loose} $\BoundLoose$.
        \label{fig:ERR_LBR_scatter_plot_loose_SciTail}
        }
    \end{subfigure}
    \hfill
    \begin{subfigure}[t]{0.32\linewidth}
        \vskip 0pt
        \centering
        \includegraphics[width=0.65\linewidth]{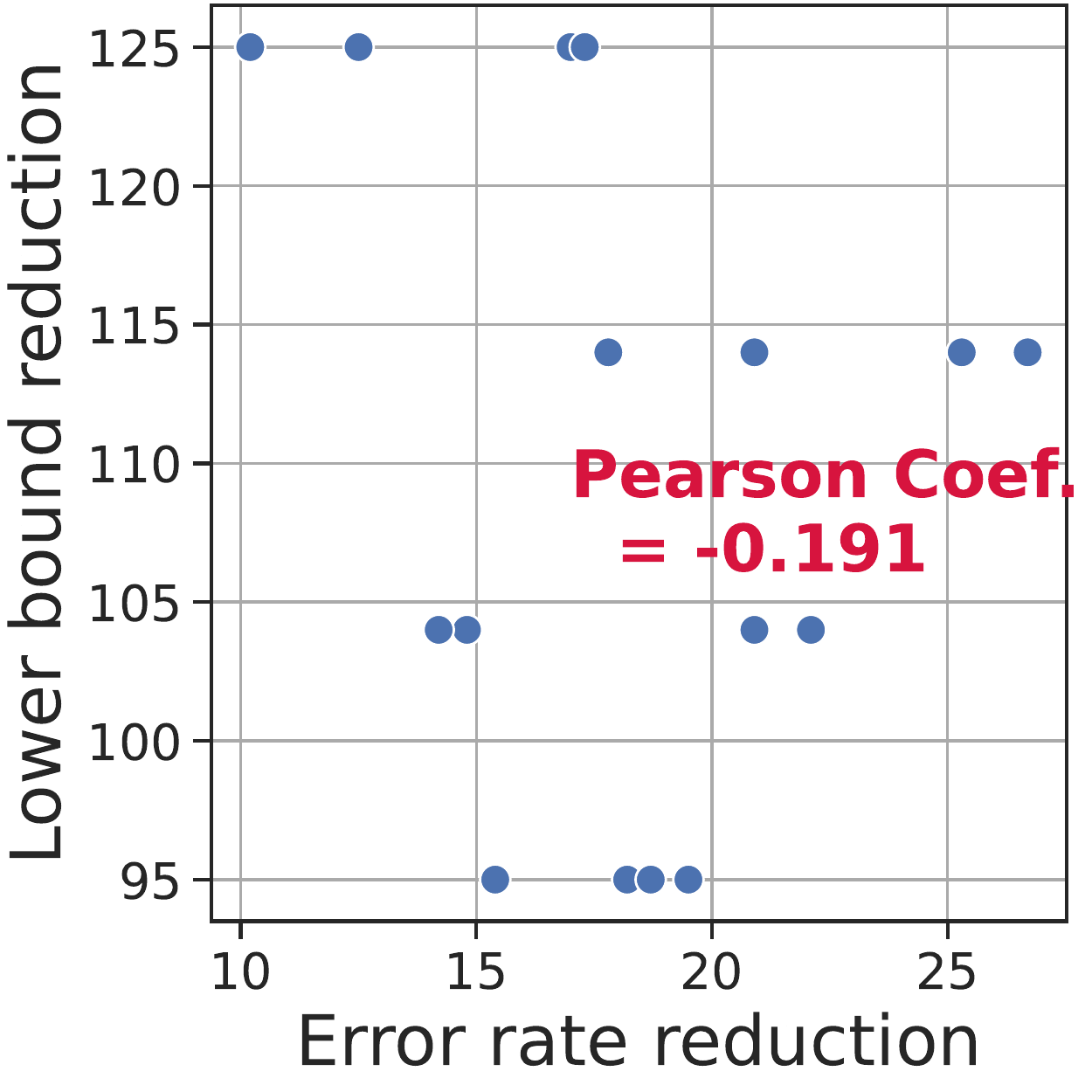}
        \caption{
            $\BoundTightWOCombLoss$.
            \label{fig:ERR_LBR_scatter_plot_tight_wo_combloss_SciTail}
        }
    \end{subfigure}
    \hfill
    \begin{subfigure}[t]{0.32\linewidth}
        \vskip 0pt
        \centering
        \includegraphics[width=0.65\linewidth]{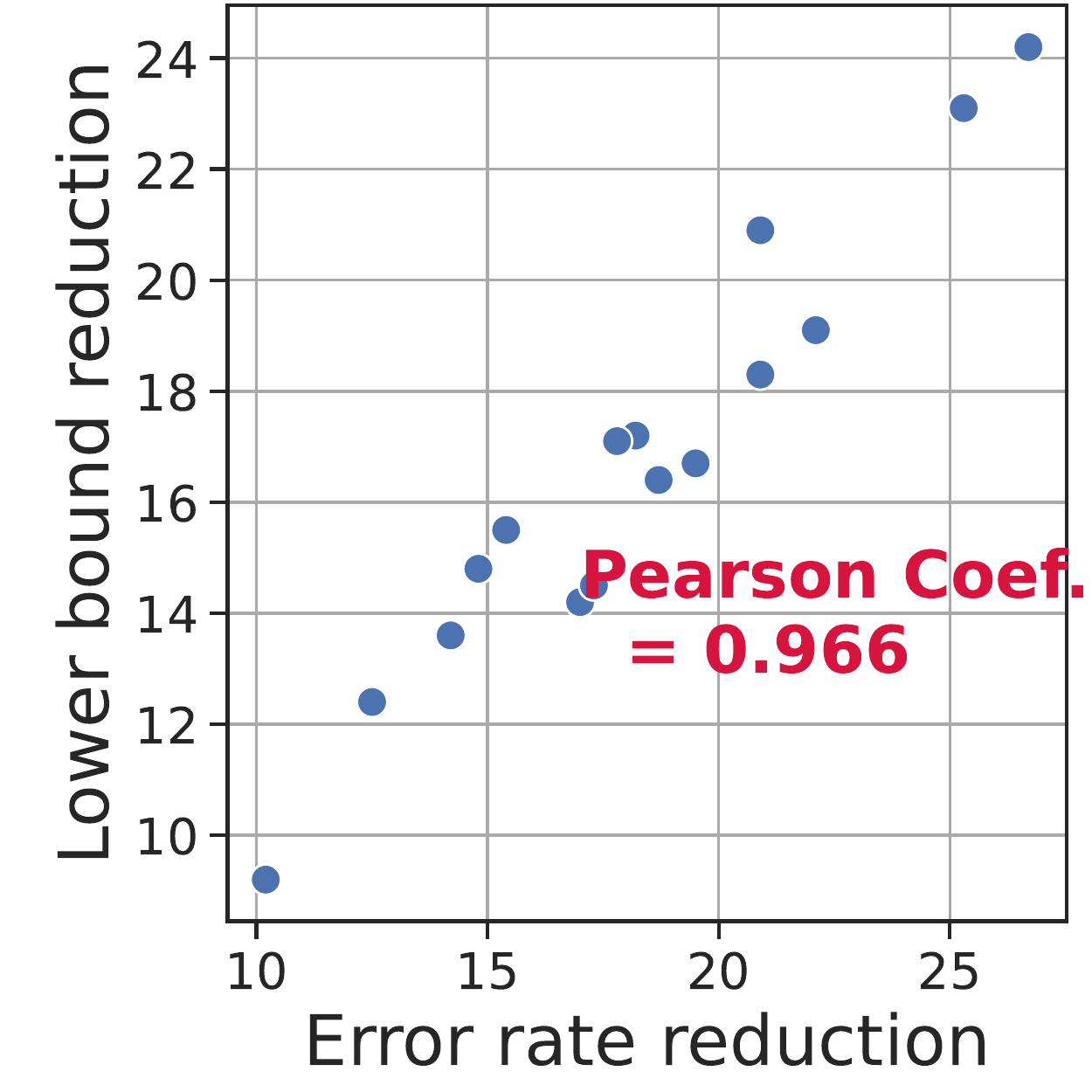}
        \caption{
            \textbf{\Cref{lemma:ensemble_bound_tight}} $\BoundTight$.
            \label{fig:ERR_LBR_scatter_plot_tight_SciTail}
        }
    \end{subfigure}
    \hfill

    \caption{
    \textbf{SciTail task.}
    Correlations between error rate reductions and lower bound reductions.
    Each figure uses different type of lower bound.
    Each point in the figures shows a quantity of a specific ensemble system $s$ and the quantity is the average over the \NumTasks tasks.
    See \Cref{tb:ablation_SciTail} for the real value of each point.
    We used the \NumSystems ensemble systems described in \Cref{sec:ensemble_systems}.
    Each system $s$ used $N=15$ models.
    The baseline values in \Cref{eq:error_reduction,eq:lower_bound_reduction} were the followings:
    ER($s_0$): \SI{5.7}{\percent}.
    LB($s_0$) of $\BoundFuncTight(\StrengthTriplet)$: \SI{1.2}{\percent}.
    LB($s_0$) of $\BoundFuncTight(\StrengthDoublet)$: \SI{1.2}{\percent}.
    LB($s_0$) of $\BoundFuncLoose(\StrengthDoublet)$: \SI{-5.2}{\percent}.
    \label{fig:ERR_LBR_scatter_plot_SciTail}
    }
\end{figure*}

\begin{figure*}[h!]
    \begin{subfigure}[t]{0.19\linewidth}
        \vskip 0pt
        \includegraphics[width=\linewidth]{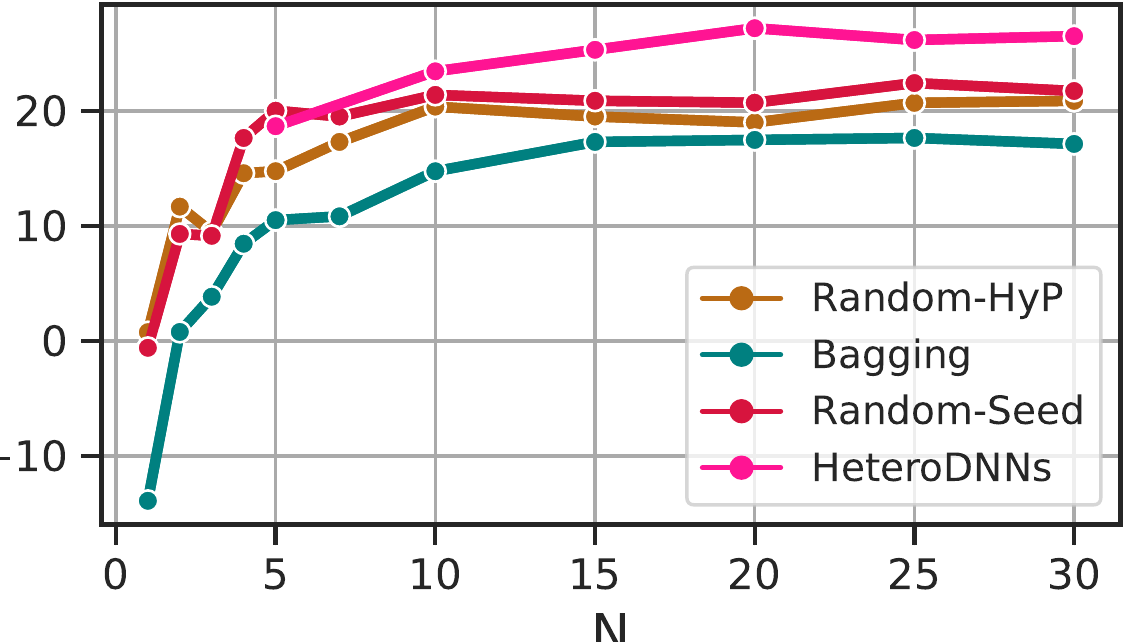}
        \subcaption{Error rate reduction.\label{appendix:fig:scaling.stacking_SciTail}}
    \end{subfigure}
    \hfill
    \begin{subfigure}[t]{0.19\linewidth}
        \vskip 0pt
        \includegraphics[width=\linewidth]{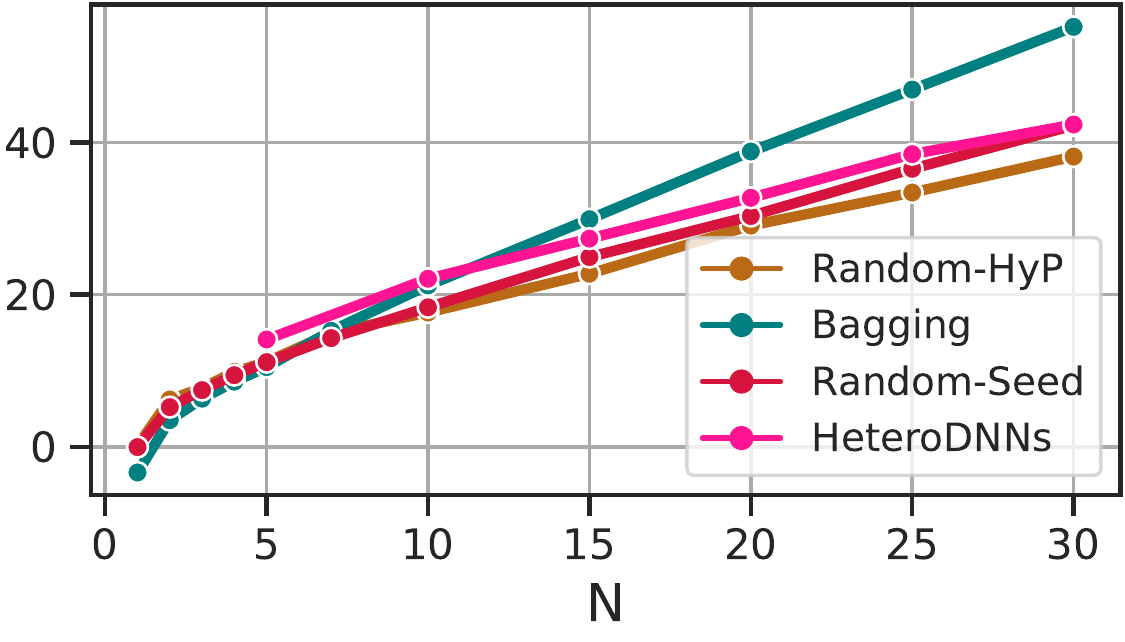}
        \subcaption{Lower bound reduction by \Cref{lemma:ensemble_bound_loose} $\BoundLoose$.\label{appendix:fig:scaling.bound.voting.previous_research_SciTail}}
    \end{subfigure}
    \begin{subfigure}[t]{0.19\linewidth}
        \vskip 0pt
        \includegraphics[width=\linewidth]{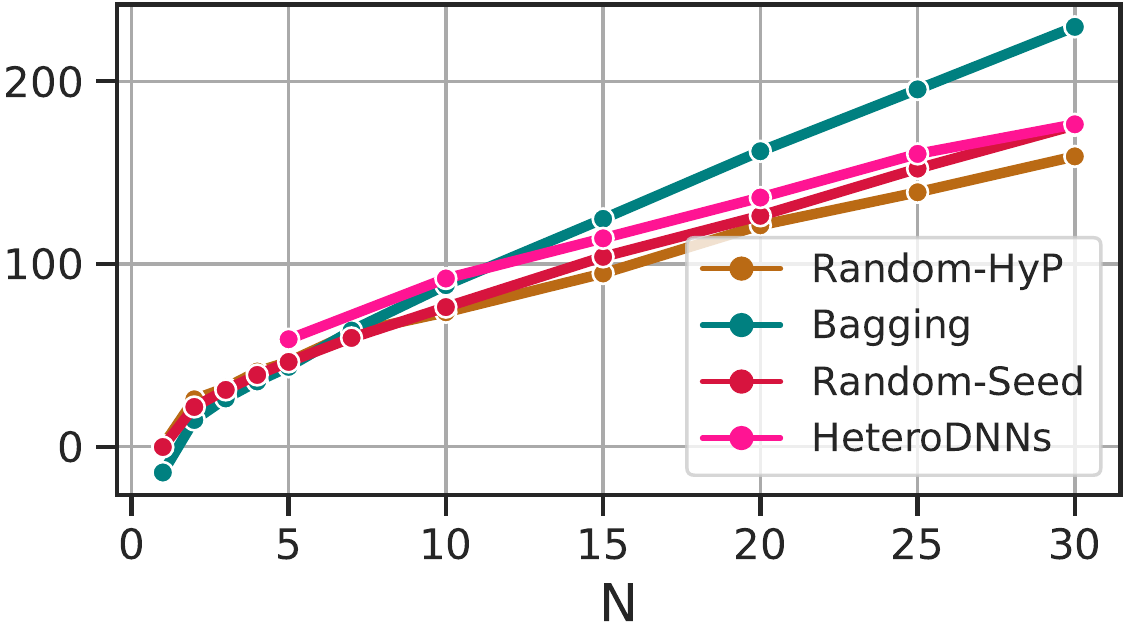}
        \subcaption{Lower bound reduction by $\BoundTightWOCombLoss$. \label{appendix:fig:scaling.bound.voting.ours_wo_combloss_SciTail}}
    \end{subfigure}
    \hfill
     \begin{subfigure}[t]{0.19\linewidth}
        \vskip 0pt
        \includegraphics[width=\linewidth]{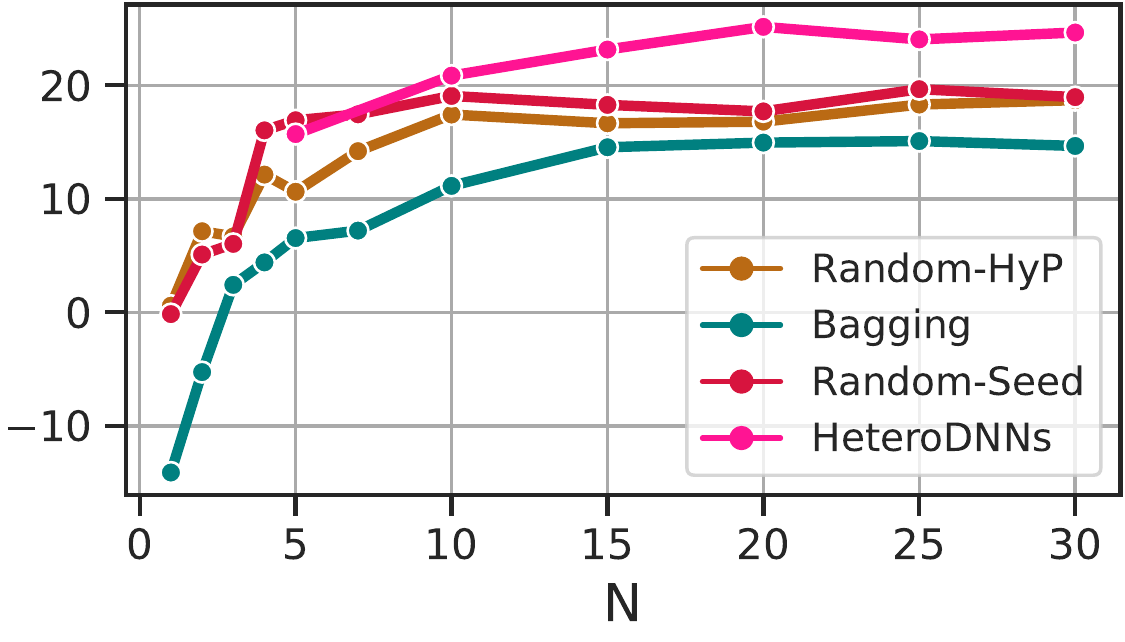}
        \subcaption{Lower bound reduction by \textbf{\Cref{lemma:ensemble_bound_tight}} $\BoundTight$.\label{appendix:fig:scaling.stacking.bound_SciTail}}
    \end{subfigure}
    \hfill
    \begin{subfigure}[t]{0.19\linewidth}
        \vskip 0pt
        \includegraphics[width=\linewidth]{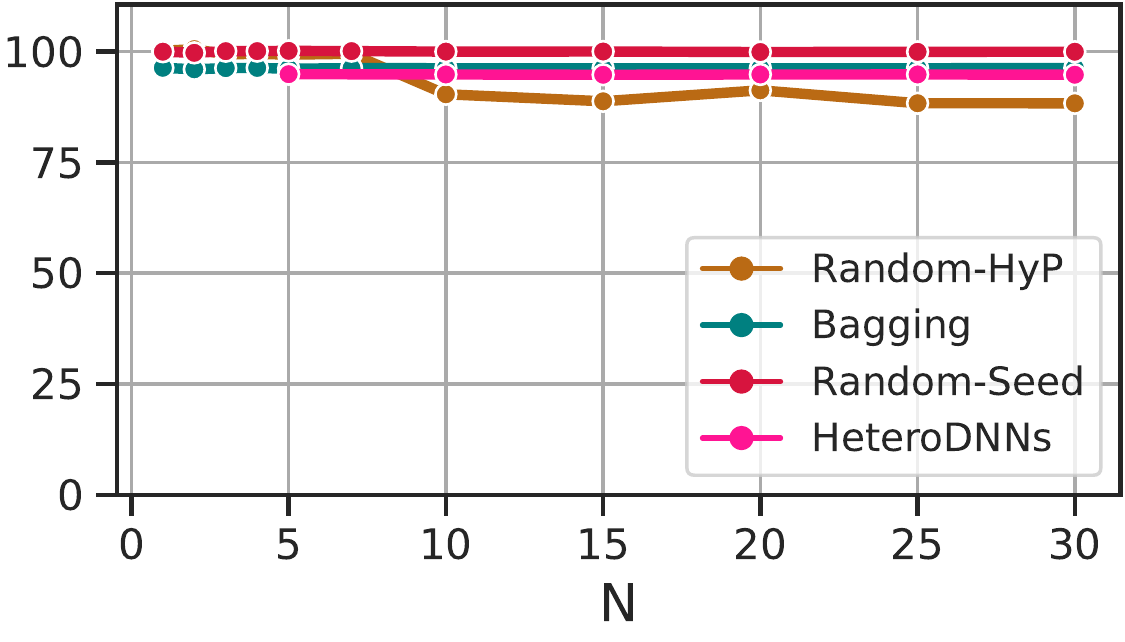}
        \subcaption{$\relev$ \label{appendix:fig:scaling.relevance.per_model_SciTail}}   
    \end{subfigure}   
    \hfill
    \vfill
    \begin{subfigure}[t]{0.19\linewidth}
        \vskip 0pt
        \includegraphics[width=\linewidth]{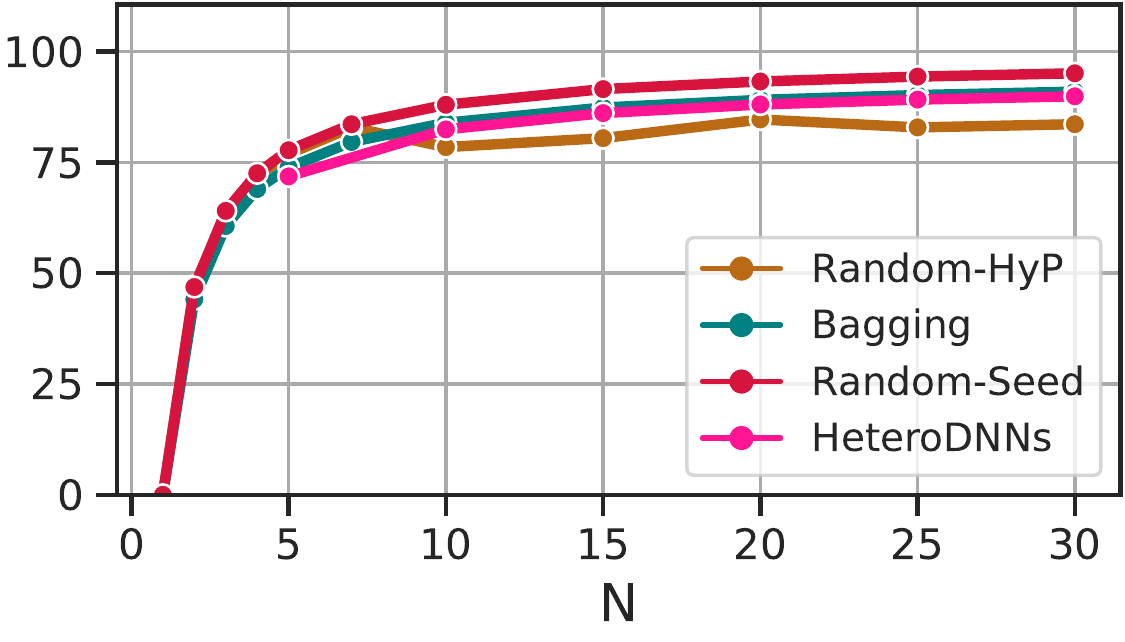}
        \subcaption{$\redun$ \label{appendix:fig:scaling.redundancy.per_model_SciTail}}   
    \end{subfigure}   
    \hfill
    \begin{subfigure}[t]{0.19\linewidth}
        \vskip 0pt
        \includegraphics[width=\linewidth]{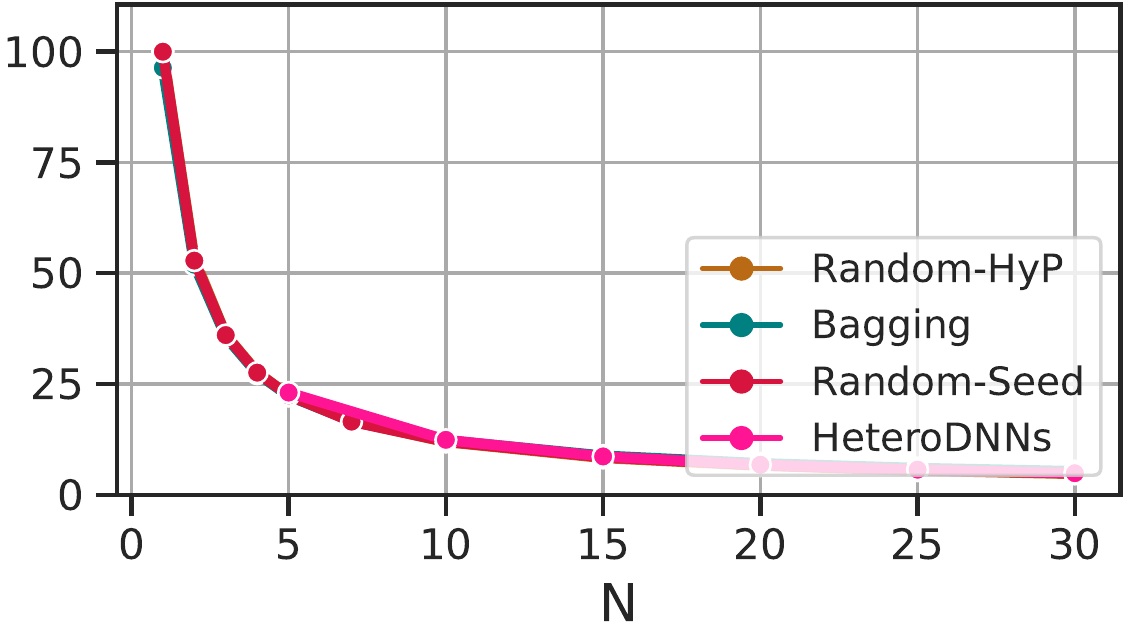}
        \subcaption{$\relev$ $- \redun$ \label{appendix:fig:scaling.novelty.per_model_SciTail}}
    \end{subfigure}   
    \hfill
    \begin{subfigure}[t]{0.19\linewidth}
        \vskip 0pt
        \includegraphics[width=\linewidth]{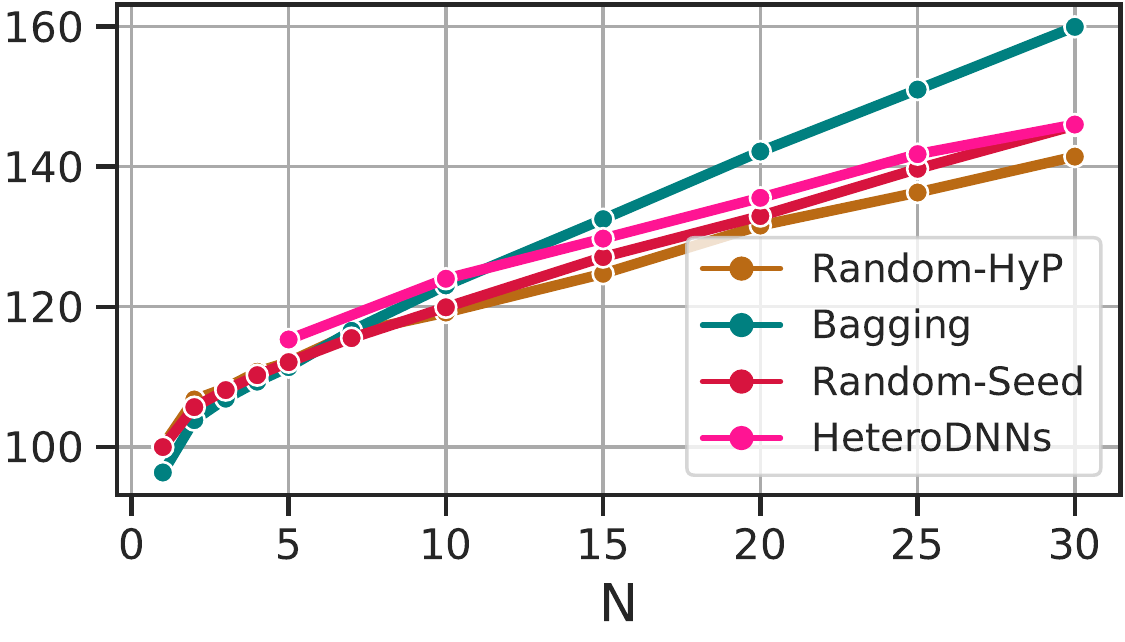}
        \captionsetup{justification=centering}
        \subcaption{$\StrengthDoublet $\newline $= N (\relev - \redun)$ \label{appendix:fig:scaling.E_without_combination_loss_SciTail}}
    \end{subfigure}
    \hfill
    \begin{subfigure}[t]{0.19\linewidth}
        \vskip 0pt
        \includegraphics[width=\linewidth]{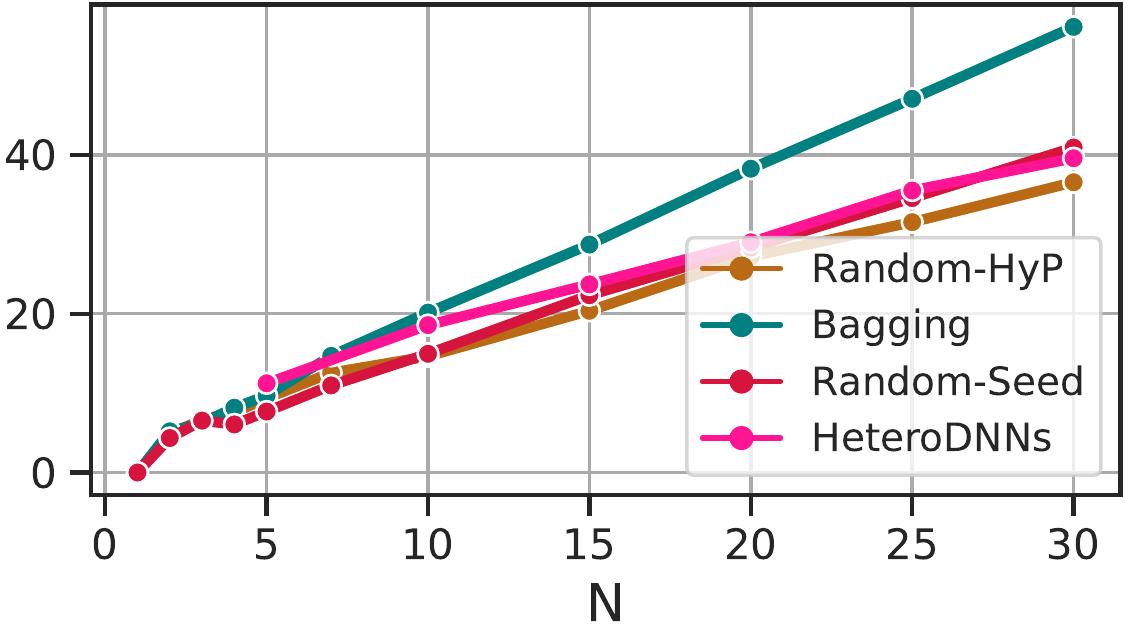}
        \subcaption{$\Combloss$\label{appendix:fig:scaling.combination_loss_SciTail}}
    \end{subfigure}
    \hfill
    \begin{subfigure}[t]{0.19\linewidth}
        \vskip 0pt
        \includegraphics[width=\linewidth]{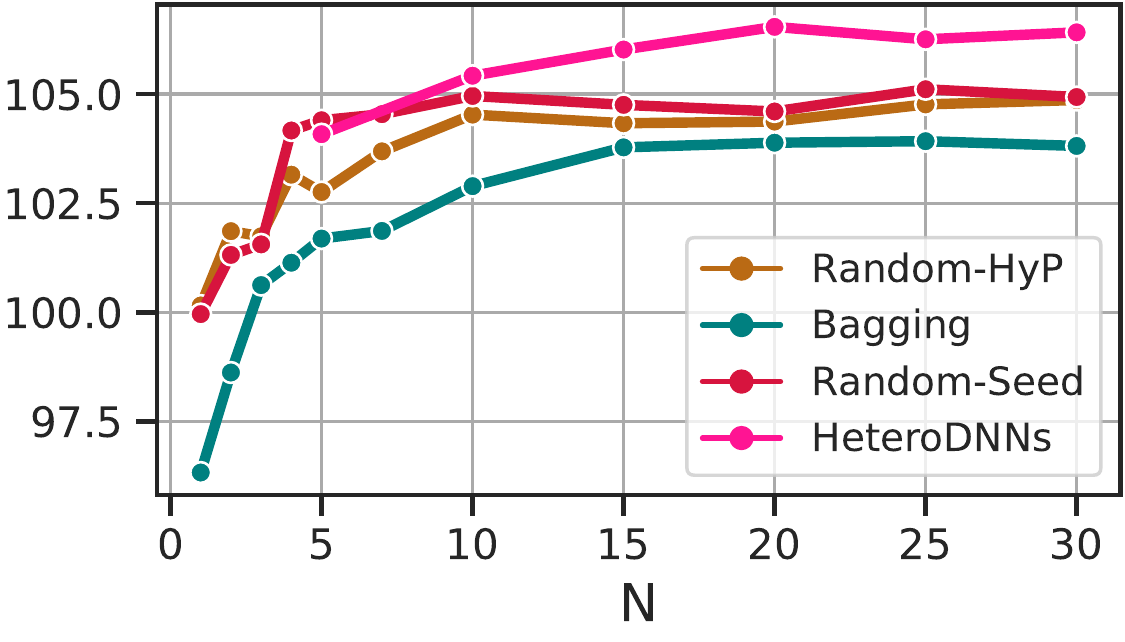}
        \captionsetup{justification=centering}
        \subcaption{$\StrengthTriplet = N (\relev - \redun) - \Combloss $\label{appendix:fig:scaling.E_SciTail}}
    \end{subfigure}
\caption{
\textbf{SciTail task.}
The change in ensemble quantities when the number of models $N$ is changed.
Each figure shows a specific quantity.
The ensemble systems used the SVM model combination.
Each value is an averages of the \NumTasks tasks.
$\perModelMetric$ denotes per-model metric values defined as: $\perModelMetricDef$.
\label{appendix:fig:scaling_SciTail}
}
\end{figure*}

\begin{table*}[h]
    \centering
    \caption{
        \textbf{SciTail task}.
        Statistics of ensemble systems described in \Cref{sec:ensemble_systems}.
        The rows and columns list the model generation and combination methods of \Cref{tb:ensemble_methods}, respectively.
        Each cell shows a quantity of a specific system $s$.
        Each quantity is the average over the \NumTasks tasks.
        Each system contains $N=15$ models.
        Color shows the rank within \textit{each column} (brighter is better).
        \label{tb:ablation_SciTail}
    }  
    \begin{subfigure}{\linewidth}
        \centering
        \small
        \tabcolsep 3.0pt
    \subcaption{
        Error rate reductions and lower bound reductions.
        The baseline values used in \Cref{eq:error_reduction,eq:lower_bound_reduction} were the followings.
        ER($s_0$): \SI{5.7}{\percent}.
        LB($s_0$) of $\BoundFuncTight(\StrengthTriplet)$: \SI{1.2}{\percent}.
        LB($s_0$) of $\BoundFuncTight(\StrengthDoublet)$: \SI{1.2}{\percent}.
        LB($s_0$) of $\BoundFuncLoose(\StrengthDoublet)$: \SI{-5.2}{\percent}.
        \label{tb:ablation_errors_SciTail}
    }

\begin{tabular}{lcccccccccccc}
\toprule
& \multicolumn{4}{c}{Error rate reductions \cref{eq:error_reduction}} & &   \multicolumn{7}{c}{Lower bound reductions \Cref{eq:lower_bound_reduction}} \\
\cmidrule(l{\tabcolsep}r{\tabcolsep}){2-5} \cmidrule(l{\tabcolsep}){7-13}
 & \multirow{2}{*}{Voting} & \multirow{2}{*}{LogR} & \multirow{2}{*}{SVM} & \multirow{2}{*}{RForest} & & \multicolumn{4}{c}{\textbf{\Cref{lemma:ensemble_bound_tight}} $\BoundFuncTight(\StrengthTriplet)$} & & \multirow{2}{*}{$\BoundFuncTight(\StrengthDoublet)$} & \multirow{2}{*}{\begin{tabular}{c}\Cref{lemma:ensemble_bound_loose} \\ $\BoundFuncLoose(\StrengthDoublet)$\end{tabular}} \\
 &  &  &  &  & & \multicolumn{1}{c}{Voting} & \multicolumn{1}{c}{LogR} & \multicolumn{1}{c}{SVM} & \multicolumn{1}{c}{RForest} & & &  \\

\midrule
Random-HyP    &  \cSecond $18.2_{\pm{\mbox{\tiny 1.1}}}$ &   \cThird $18.7_{\pm{\mbox{\tiny 1.1}}}$ &   \cThird $19.5_{\pm{\mbox{\tiny 2.3}}}$ &  \cSecond $15.4_{\pm{\mbox{\tiny 1.5}}}$ &        &  \cSecond $17.2_{\pm{\mbox{\tiny 1.3}}}$ &   \cThird $16.4_{\pm{\mbox{\tiny 1.8}}}$ &   \cThird $16.7_{\pm{\mbox{\tiny 2.6}}}$ &  \cSecond $15.5_{\pm{\mbox{\tiny 1.5}}}$ &        &   \cFourth $95_{\pm{\mbox{\tiny 2}}}$ &  \cFourth $23_{\pm{\mbox{\tiny 0}}}$ \\
Bagging       &  \cFourth $12.5_{\pm{\mbox{\tiny 1.5}}}$ &  \cFourth $17.0_{\pm{\mbox{\tiny 2.5}}}$ &  \cFourth $17.3_{\pm{\mbox{\tiny 2.7}}}$ &  \cFourth $10.2_{\pm{\mbox{\tiny 2.6}}}$ &        &  \cFourth $12.4_{\pm{\mbox{\tiny 1.4}}}$ &  \cFourth $14.2_{\pm{\mbox{\tiny 2.9}}}$ &  \cFourth $14.5_{\pm{\mbox{\tiny 3.1}}}$ &   \cFourth $9.2_{\pm{\mbox{\tiny 2.6}}}$ &        &   \cFirst $125_{\pm{\mbox{\tiny 2}}}$ &   \cFirst $30_{\pm{\mbox{\tiny 0}}}$ \\
Random-Seed   &   \cThird $14.8_{\pm{\mbox{\tiny 1.4}}}$ &  \cSecond $22.1_{\pm{\mbox{\tiny 0.5}}}$ &  \cSecond $20.9_{\pm{\mbox{\tiny 0.9}}}$ &   \cThird $14.2_{\pm{\mbox{\tiny 0.7}}}$ &        &   \cThird $14.8_{\pm{\mbox{\tiny 1.6}}}$ &  \cSecond $19.1_{\pm{\mbox{\tiny 0.5}}}$ &  \cSecond $18.3_{\pm{\mbox{\tiny 0.7}}}$ &   \cThird $13.6_{\pm{\mbox{\tiny 0.4}}}$ &        &   \cThird $104_{\pm{\mbox{\tiny 4}}}$ &   \cThird $25_{\pm{\mbox{\tiny 1}}}$ \\
Hetero-DNNs &   \cFirst $20.9_{\pm{\mbox{\tiny 2.0}}}$ &   \cFirst $26.7_{\pm{\mbox{\tiny 1.0}}}$ &   \cFirst $25.3_{\pm{\mbox{\tiny 2.2}}}$ &   \cFirst $17.8_{\pm{\mbox{\tiny 4.5}}}$ &        &   \cFirst $20.9_{\pm{\mbox{\tiny 2.0}}}$ &   \cFirst $24.2_{\pm{\mbox{\tiny 1.1}}}$ &   \cFirst $23.1_{\pm{\mbox{\tiny 2.2}}}$ &   \cFirst $17.1_{\pm{\mbox{\tiny 4.3}}}$ &        &  \cSecond $114_{\pm{\mbox{\tiny 2}}}$ &  \cSecond $27_{\pm{\mbox{\tiny 1}}}$ \\
\bottomrule

\end{tabular}

    \end{subfigure}
    \vfill
    \begin{subfigure}{\linewidth}
    \centering
    \small
    \tabcolsep 2.0pt
    \subcaption{
    Breakdown of ensemble strength defined in \cref{eq:triplet_decomposition}.
    We show per-model metric values defined as: $\perModelMetricDef$. Thus, $\StrengthTriplet = (\relev - \redun - \combloss) \ \times N $ holds.
    For intuitive understanding, all the values are normalized by the ensemble strength of baseline $\StrengthTriplet_{s_0}$, for example, $\Relev = \RelevHat / \StrengthTriplet_{s_0} \times 100$ where $\RelevHat$ is the raw value.
    \label{tb:ablation_triple_SciTail}
    }

\begin{tabular}{lccccccccccccc}
\toprule

{} & \multicolumn{4}{c}{\multirow{1}{*}{$\StrengthTripletWithArgs$}} & & \multicolumn{6}{c}{Per-model metric values} & \\
\cmidrule(l{\tabcolsep}r{\tabcolsep}){2-5} \cmidrule(l{\tabcolsep}r{\tabcolsep}){6-12}
{} &  & &  &  & & \multirow{2}{*}{$\perModelMetric_{\normalfont \text{relev}}$} & \multirow{2}{*}{$\perModelMetric_{\normalfont \text{redun}}$} & \multicolumn{4}{c}{ $\perModelMetric_{\normalfont \text{combloss}}$} & & \multirow{2}{*}{$\perModelMetric_{\normalfont \text{relev}} - \perModelMetric_{\normalfont \text{redun}}$} \\
{} & \multicolumn{1}{c}{Voting} & \multicolumn{1}{c}{LogR} & \multicolumn{1}{c}{SVM} & \multicolumn{1}{c}{RForest} & &  {} &  {} &  \multicolumn{1}{c}{Voting} & \multicolumn{1}{c}{LogR} &  \multicolumn{1}{c}{SVM} & \multicolumn{1}{c}{RForest} & &  {} \\

\midrule
Baseline ($s_0$)                    &    \multicolumn{4}{c}{\cBase 100 (the raw value is 0.641)} &        &   \cBase 100 &                                 \cBase 0 &                                \cBase 0 &                                \cBase 0 &                                \cBase 0 &                                 \cBase 0 &        &   \cBase 100 \\

\midrule

Random-HyP                   &  \cSecond $104.5_{\pm{\mbox{\tiny 0.4}}}$ &   \cThird $104.3_{\pm{\mbox{\tiny 0.5}}}$ &   \cThird $104.3_{\pm{\mbox{\tiny 0.7}}}$ &  \cSecond $104.0_{\pm{\mbox{\tiny 0.4}}}$ &        &  \cFourth $88.8_{\pm{\mbox{\tiny 1.8}}}$ &   \cFirst $80.5_{\pm{\mbox{\tiny 1.8}}}$ &   \cFirst $1.35_{\pm{\mbox{\tiny 0.03}}}$ &   \cFirst $1.36_{\pm{\mbox{\tiny 0.01}}}$ &   \cFirst $1.36_{\pm{\mbox{\tiny 0.02}}}$ &   \cFirst $1.38_{\pm{\mbox{\tiny 0.03}}}$ &        &  \cFourth $8.3_{\pm{\mbox{\tiny 2.5}}}$ \\
Bagging                      &  \cFourth $103.2_{\pm{\mbox{\tiny 0.4}}}$ &  \cFourth $103.7_{\pm{\mbox{\tiny 0.7}}}$ &  \cFourth $103.8_{\pm{\mbox{\tiny 0.8}}}$ &  \cFourth $102.4_{\pm{\mbox{\tiny 0.7}}}$ &        &  \cSecond $96.2_{\pm{\mbox{\tiny 0.3}}}$ &   \cThird $87.4_{\pm{\mbox{\tiny 0.3}}}$ &  \cFourth $1.95_{\pm{\mbox{\tiny 0.01}}}$ &  \cFourth $1.92_{\pm{\mbox{\tiny 0.03}}}$ &  \cFourth $1.91_{\pm{\mbox{\tiny 0.04}}}$ &  \cFourth $2.01_{\pm{\mbox{\tiny 0.04}}}$ &        &   \cFirst $8.8_{\pm{\mbox{\tiny 0.5}}}$ \\
Random-Seed                  &   \cThird $103.8_{\pm{\mbox{\tiny 0.4}}}$ &  \cSecond $105.0_{\pm{\mbox{\tiny 0.2}}}$ &  \cSecond $104.8_{\pm{\mbox{\tiny 0.2}}}$ &   \cThird $103.6_{\pm{\mbox{\tiny 0.1}}}$ &        &  \cFirst $100.0_{\pm{\mbox{\tiny 0.0}}}$ &  \cFourth $91.5_{\pm{\mbox{\tiny 0.1}}}$ &  \cSecond $1.55_{\pm{\mbox{\tiny 0.07}}}$ &  \cSecond $1.47_{\pm{\mbox{\tiny 0.08}}}$ &  \cSecond $1.49_{\pm{\mbox{\tiny 0.08}}}$ &  \cSecond $1.57_{\pm{\mbox{\tiny 0.08}}}$ &        &   \cThird $8.5_{\pm{\mbox{\tiny 0.1}}}$ \\
Hetero-DNNs                &   \cFirst $105.4_{\pm{\mbox{\tiny 0.5}}}$ &   \cFirst $106.3_{\pm{\mbox{\tiny 0.3}}}$ &   \cFirst $106.0_{\pm{\mbox{\tiny 0.6}}}$ &   \cFirst $104.5_{\pm{\mbox{\tiny 1.1}}}$ &        &   \cThird $94.8_{\pm{\mbox{\tiny 0.5}}}$ &  \cSecond $86.1_{\pm{\mbox{\tiny 0.5}}}$ &   \cThird $1.62_{\pm{\mbox{\tiny 0.08}}}$ &   \cThird $1.56_{\pm{\mbox{\tiny 0.07}}}$ &   \cThird $1.58_{\pm{\mbox{\tiny 0.09}}}$ &   \cThird $1.68_{\pm{\mbox{\tiny 0.12}}}$ &        &  \cSecond $8.6_{\pm{\mbox{\tiny 0.7}}}$ \\

\bottomrule
\end{tabular}

\end{subfigure}

\end{table*}

\clearpage

\begin{figure*}[t!]
    \begin{subfigure}[t]{0.32\linewidth}
        \vskip 0pt
        \centering
        \includegraphics[width=0.65\linewidth]{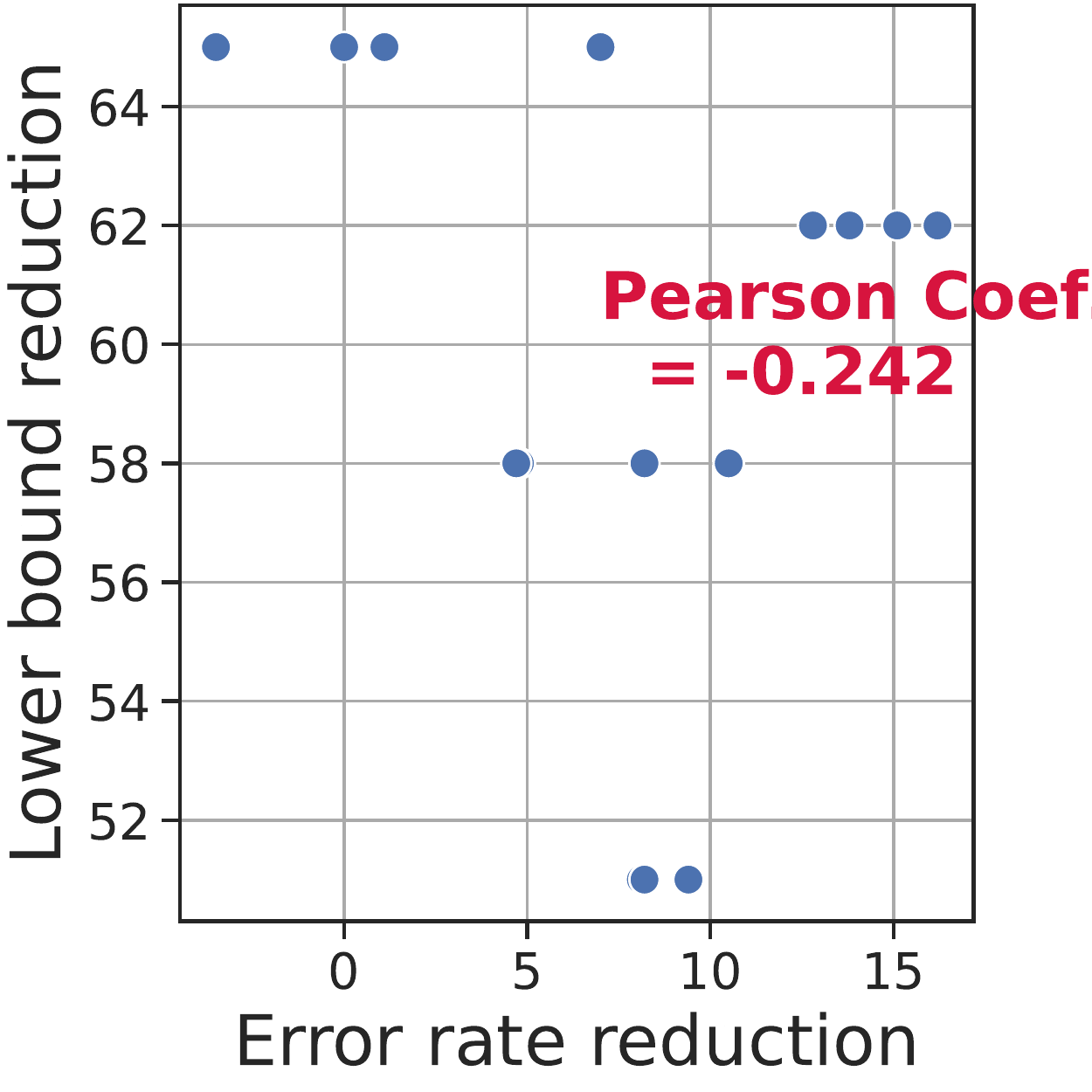}
        \caption{
        \Cref{lemma:ensemble_bound_loose} $\BoundLoose$.
        \label{fig:ERR_LBR_scatter_plot_loose_SST}
        }
    \end{subfigure}
    \hfill
    \begin{subfigure}[t]{0.32\linewidth}
        \vskip 0pt
        \centering
        \includegraphics[width=0.65\linewidth]{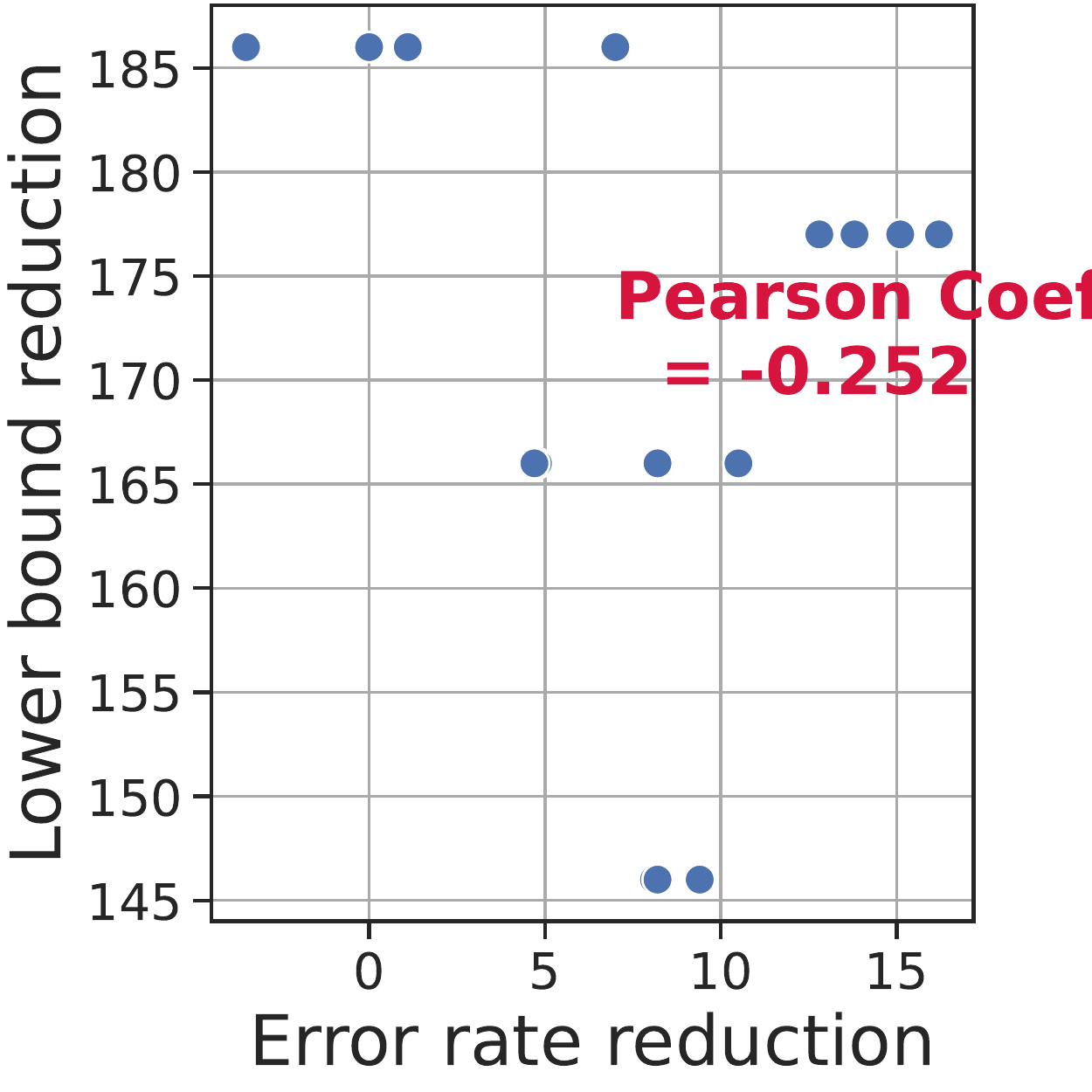}
        \caption{
            $\BoundTightWOCombLoss$.
            \label{fig:ERR_LBR_scatter_plot_tight_wo_combloss_SST}
        }
    \end{subfigure}
    \hfill
    \begin{subfigure}[t]{0.32\linewidth}
        \vskip 0pt
        \centering
        \includegraphics[width=0.65\linewidth]{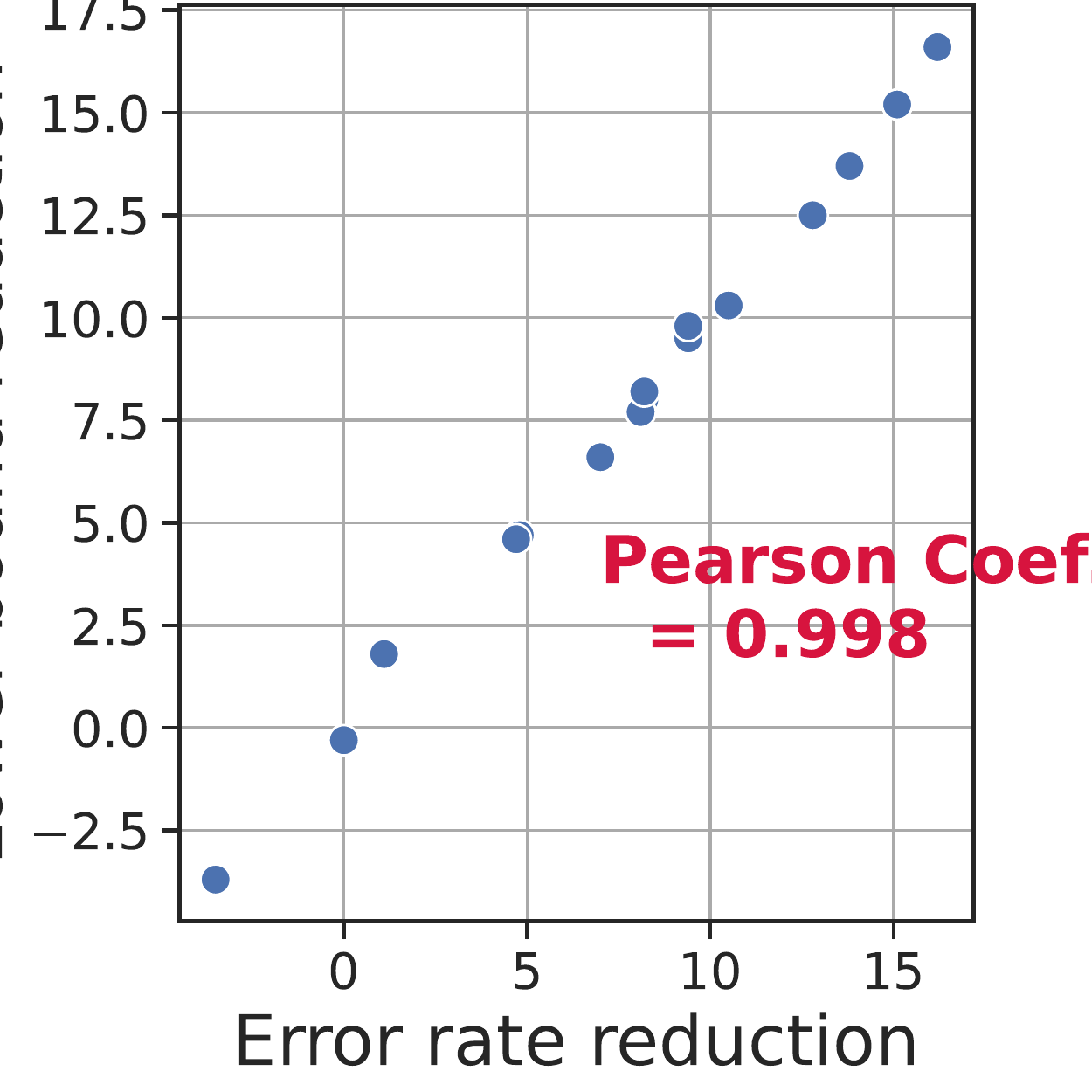}
        \caption{
            \textbf{\Cref{lemma:ensemble_bound_tight}} $\BoundTight$.
            \label{fig:ERR_LBR_scatter_plot_tight_SST}
        }
    \end{subfigure}
    \hfill

    \caption{
    \textbf{SST task.}
    Correlations between error rate reductions and lower bound reductions.
    Each figure uses different type of lower bound.
    Each point in the figures shows a quantity of a specific ensemble system $s$ and the quantity is the average over the \NumTasks tasks.
    See \Cref{tb:ablation_SST} for the real value of each point.
    We used the \NumSystems ensemble systems described in \Cref{sec:ensemble_systems}.
    Each system $s$ used $N=15$ models.
    The baseline values in \Cref{eq:error_reduction,eq:lower_bound_reduction} were the followings:
    ER($s_0$): \SI{15.7}{\percent}.
    LB($s_0$) of $\BoundFuncTight(\StrengthTriplet)$: \SI{2.3}{\percent}.
    LB($s_0$) of $\BoundFuncTight(\StrengthDoublet)$: \SI{2.3}{\percent}.
    LB($s_0$) of $\BoundFuncLoose(\StrengthDoublet)$: \SI{-3.0}{\percent}.
    \label{fig:ERR_LBR_scatter_plot_SST}
    }
\end{figure*}

\begin{figure*}[h!]
    \begin{subfigure}[t]{0.19\linewidth}
        \vskip 0pt
        \includegraphics[width=\linewidth]{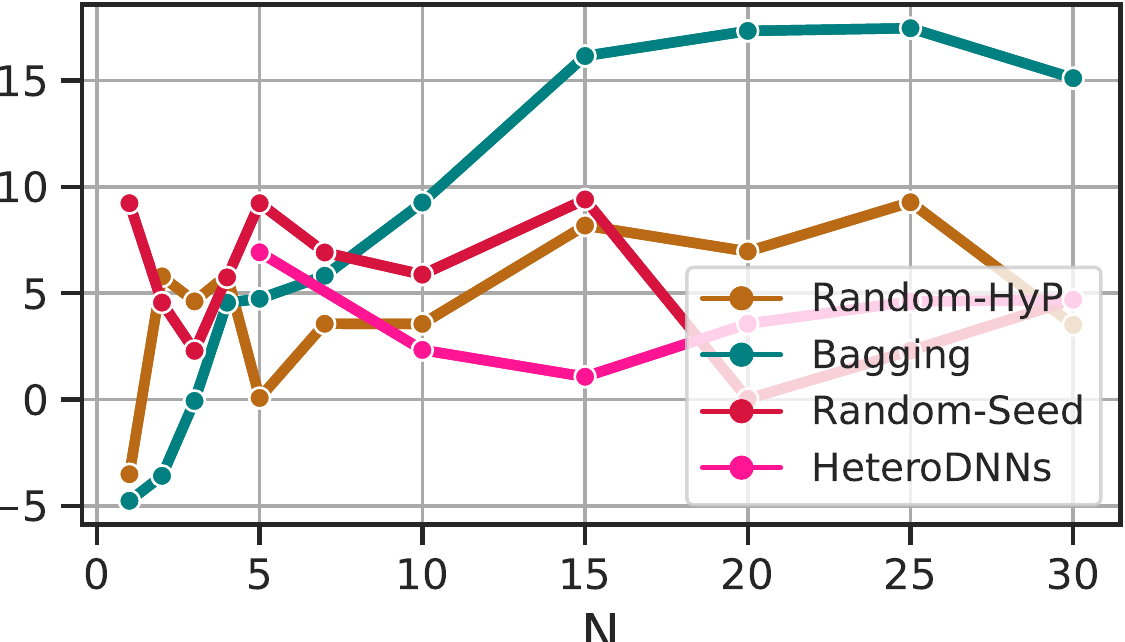}
        \subcaption{Error rate reduction.\label{appendix:fig:scaling.stacking_SST}}
    \end{subfigure}
    \hfill
    \begin{subfigure}[t]{0.19\linewidth}
        \vskip 0pt
        \includegraphics[width=\linewidth]{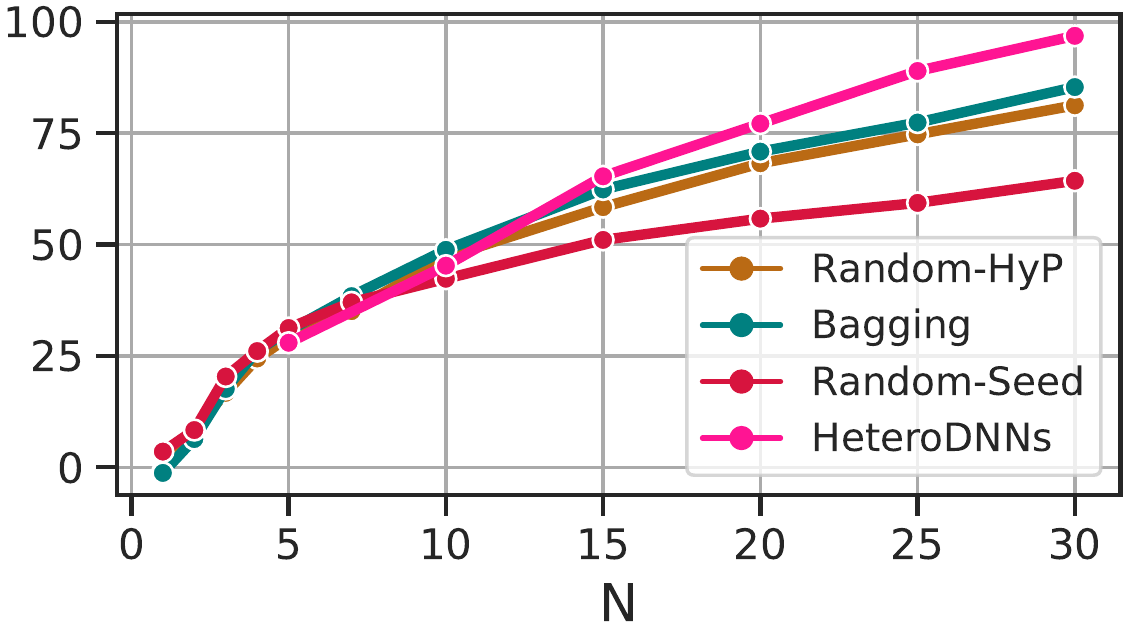}
        \subcaption{Lower bound reduction by \Cref{lemma:ensemble_bound_loose} $\BoundLoose$.\label{appendix:fig:scaling.bound.voting.previous_research_SST}}
    \end{subfigure}
    \begin{subfigure}[t]{0.19\linewidth}
        \vskip 0pt
        \includegraphics[width=\linewidth]{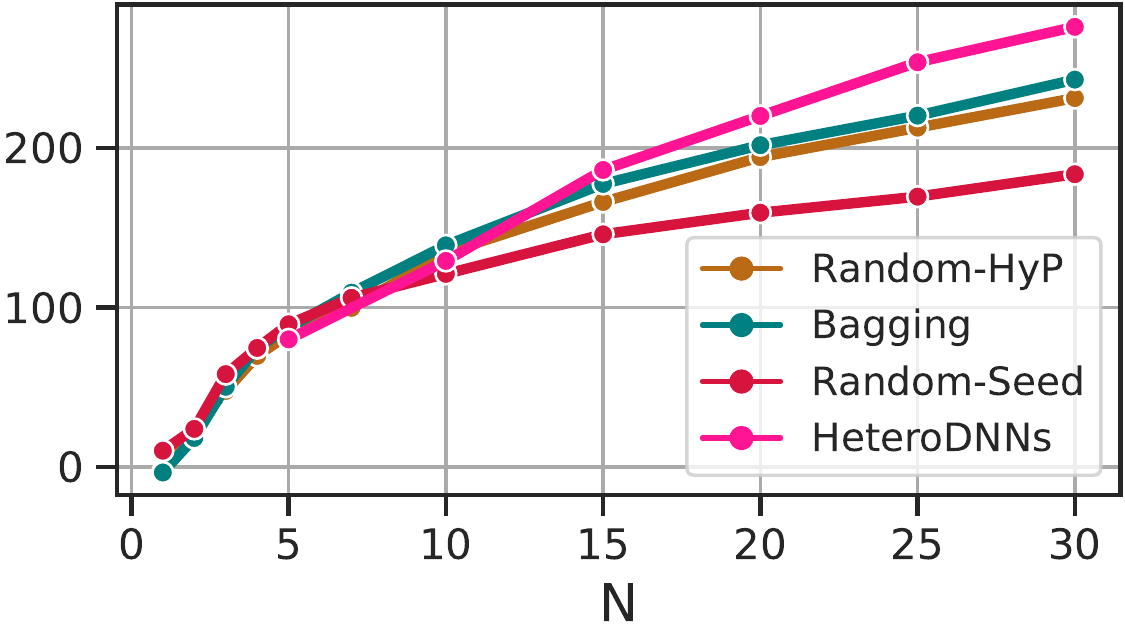}
        \subcaption{Lower bound reduction by $\BoundTightWOCombLoss$. \label{appendix:fig:scaling.bound.voting.ours_wo_combloss_SST}}
    \end{subfigure}
    \hfill
     \begin{subfigure}[t]{0.19\linewidth}
        \vskip 0pt
        \includegraphics[width=\linewidth]{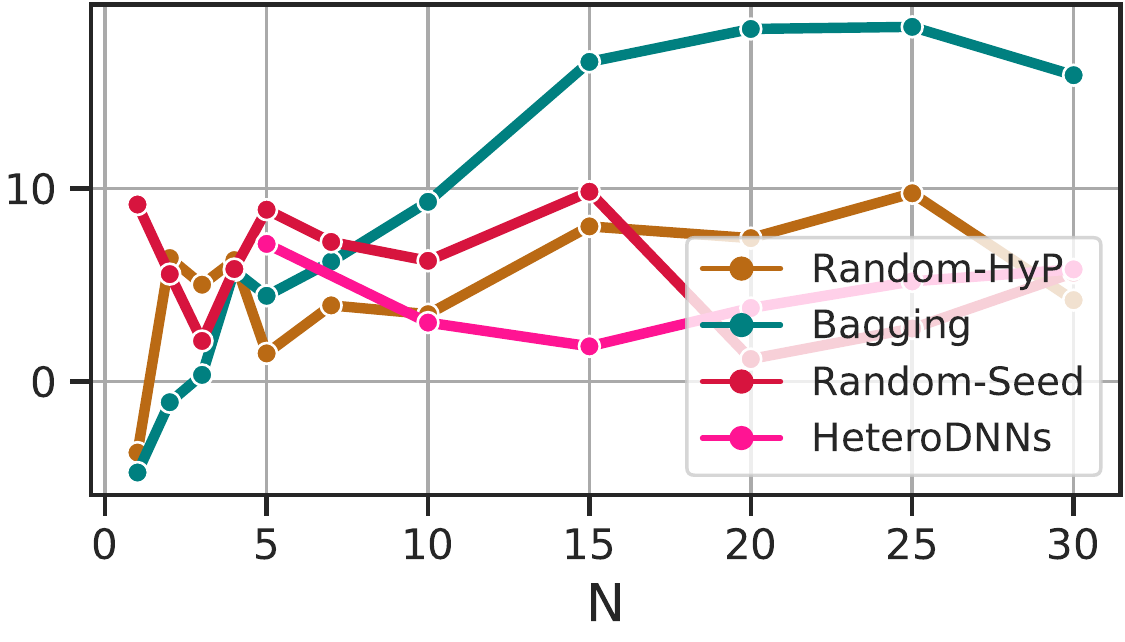}
        \subcaption{Lower bound reduction by \textbf{\Cref{lemma:ensemble_bound_tight}} $\BoundTight$.\label{appendix:fig:scaling.stacking.bound_SST}}
    \end{subfigure}
    \hfill
    \begin{subfigure}[t]{0.19\linewidth}
        \vskip 0pt
        \includegraphics[width=\linewidth]{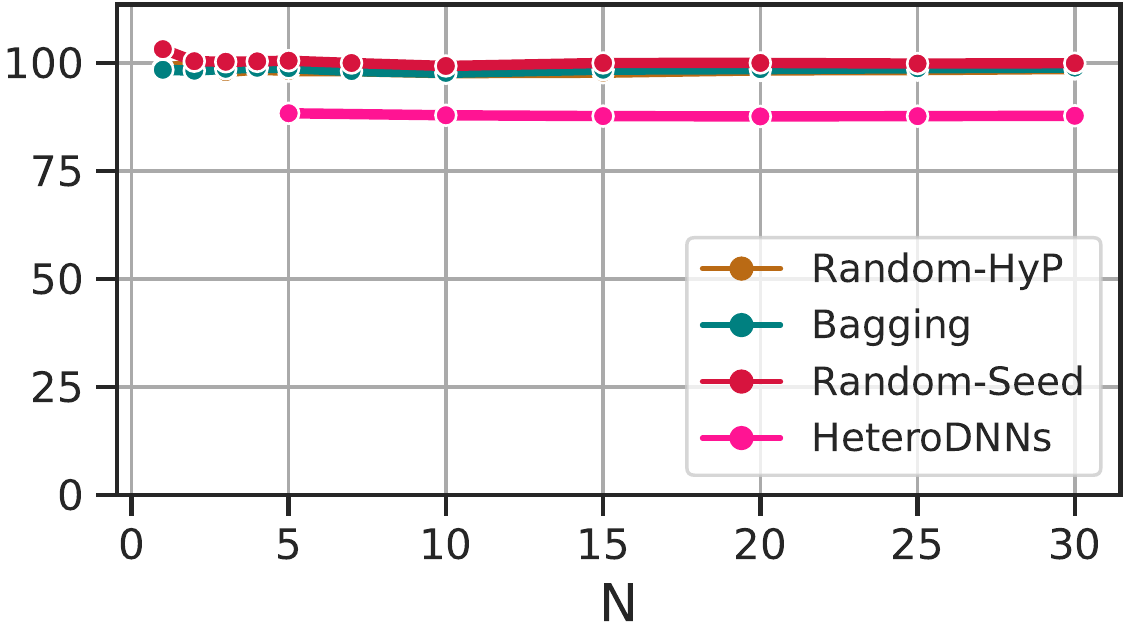}
        \subcaption{$\relev$ \label{appendix:fig:scaling.relevance.per_model_SST}}   
    \end{subfigure}   
    \hfill
    \vfill
    \begin{subfigure}[t]{0.19\linewidth}
        \vskip 0pt
        \includegraphics[width=\linewidth]{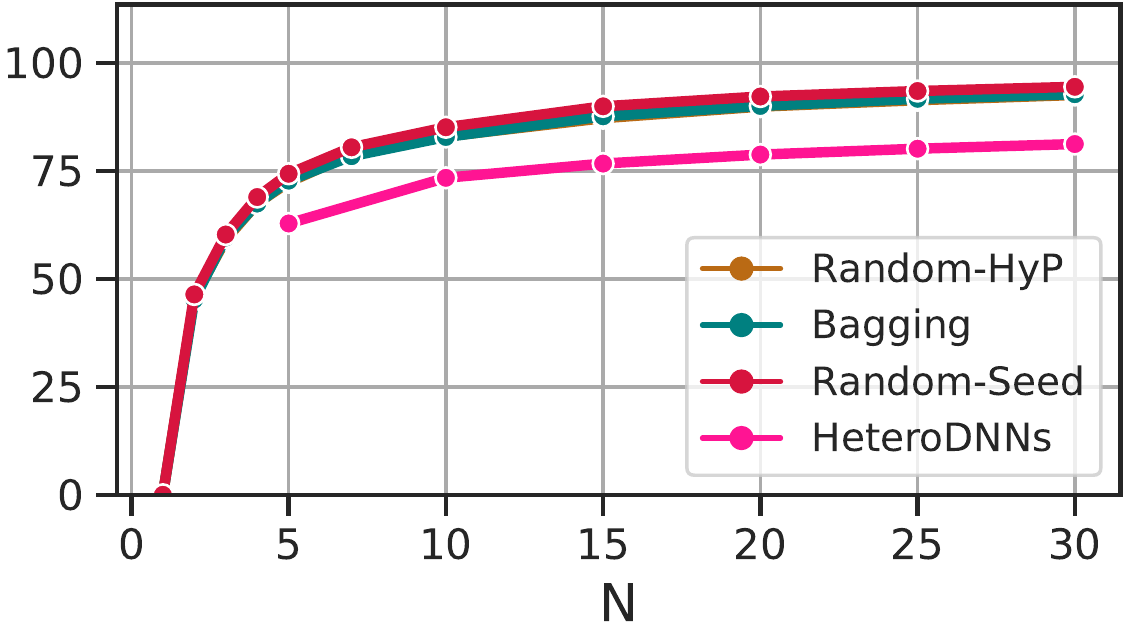}
        \subcaption{$\redun$ \label{appendix:fig:scaling.redundancy.per_model_SST}}   
    \end{subfigure}   
    \hfill
    \begin{subfigure}[t]{0.19\linewidth}
        \vskip 0pt
        \includegraphics[width=\linewidth]{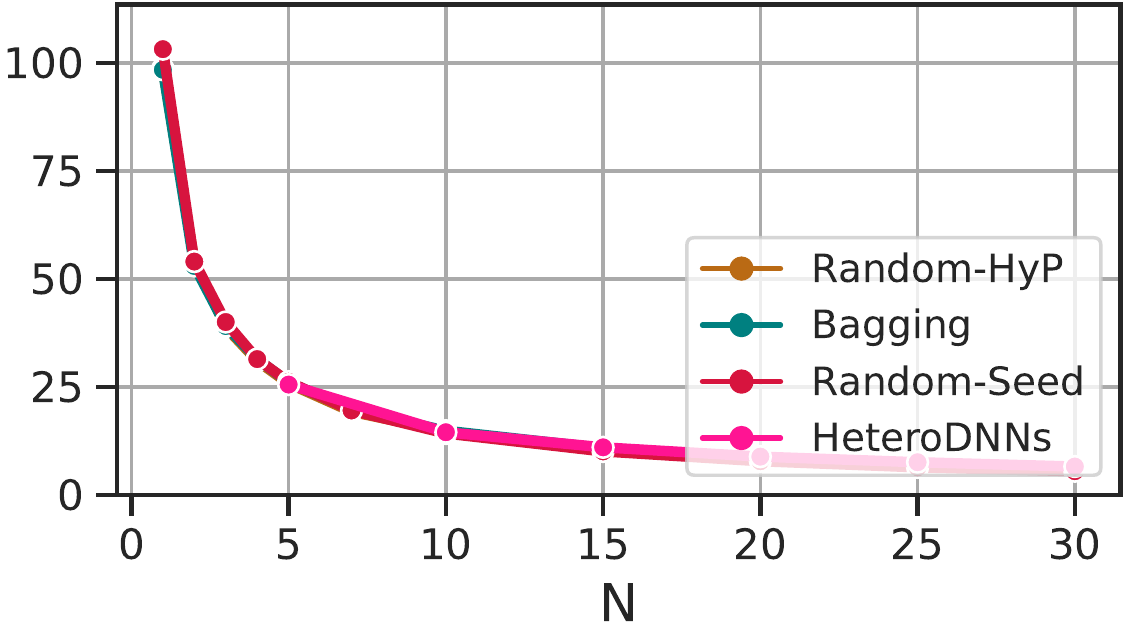}
        \subcaption{$\relev$ $- \redun$ \label{appendix:fig:scaling.novelty.per_model_SST}}
    \end{subfigure}   
    \hfill
    \begin{subfigure}[t]{0.19\linewidth}
        \vskip 0pt
        \includegraphics[width=\linewidth]{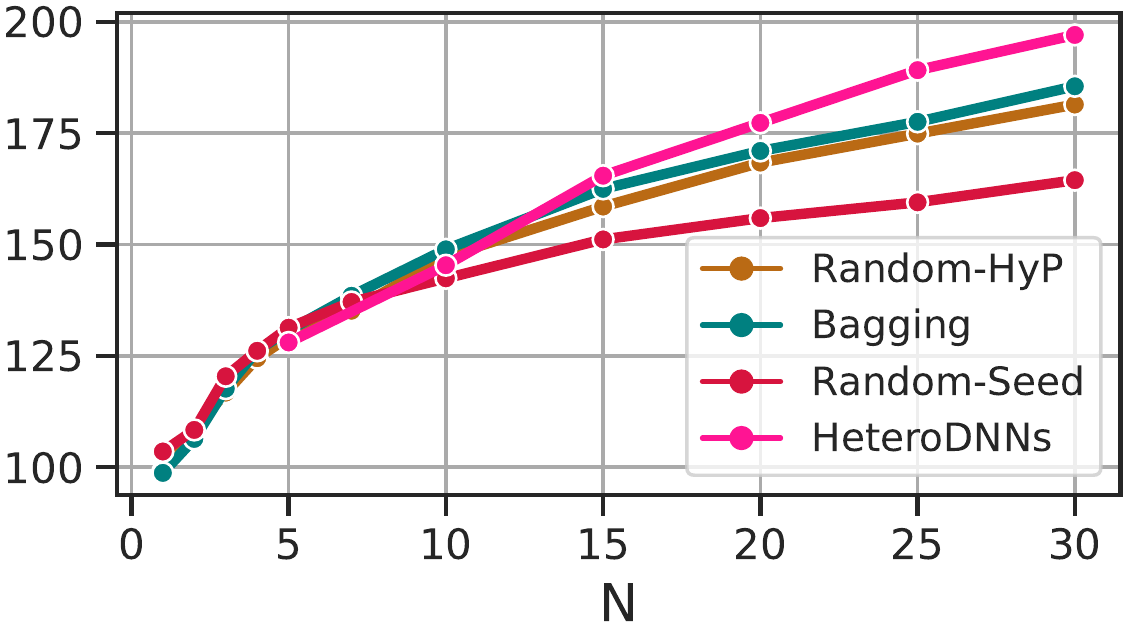}
        \captionsetup{justification=centering}
        \subcaption{$\StrengthDoublet $\newline $= N (\relev - \redun)$ \label{appendix:fig:scaling.E_without_combination_loss_SST}}
    \end{subfigure}
    \hfill
    \begin{subfigure}[t]{0.19\linewidth}
        \vskip 0pt
        \includegraphics[width=\linewidth]{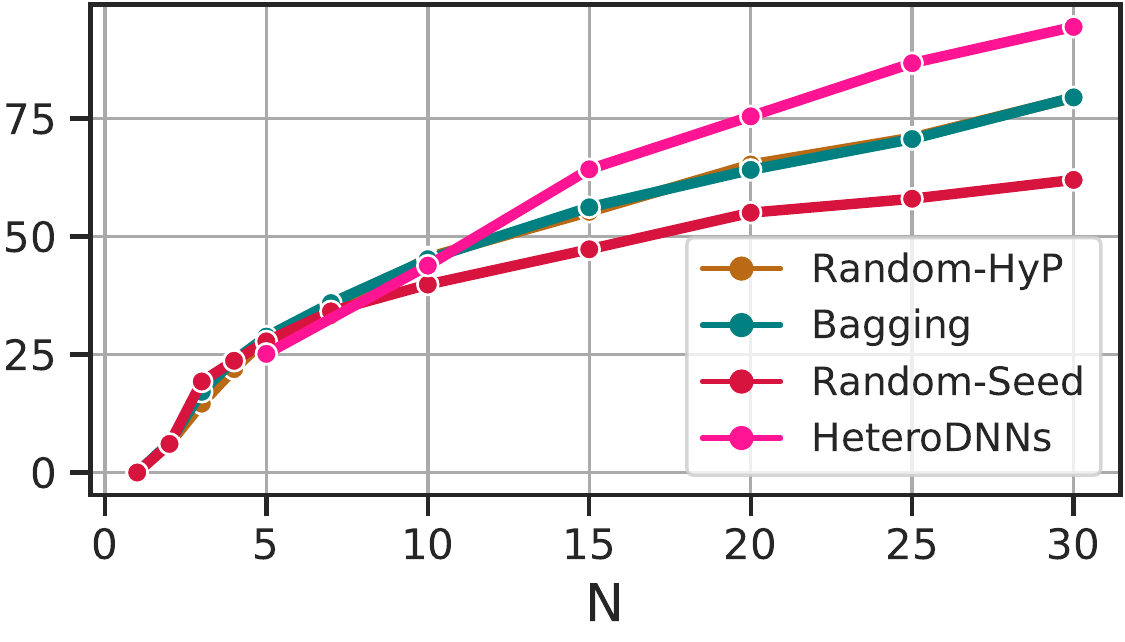}
        \subcaption{$\Combloss$\label{appendix:fig:scaling.combination_loss_SST}}
    \end{subfigure}
    \hfill
    \begin{subfigure}[t]{0.19\linewidth}
        \vskip 0pt
        \includegraphics[width=\linewidth]{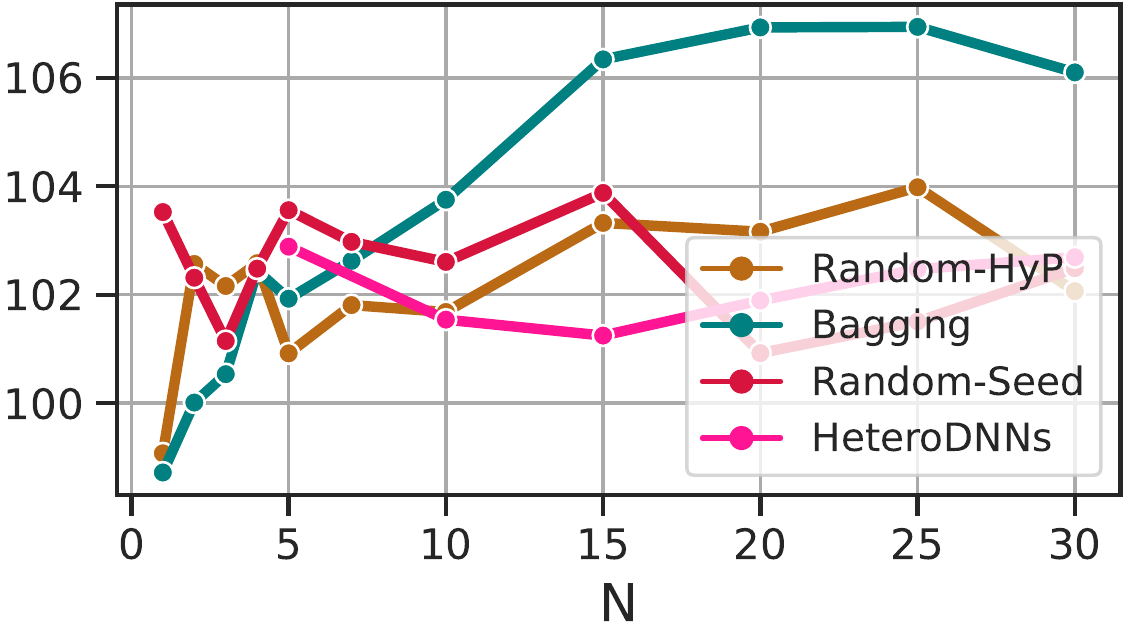}
        \captionsetup{justification=centering}
        \subcaption{$\StrengthTriplet = N (\relev - \redun) - \Combloss $\label{appendix:fig:scaling.E_SST}}
    \end{subfigure}
\caption{
\textbf{SST task.}
The change in ensemble quantities when the number of models $N$ is changed.
Each figure shows a specific quantity.
The ensemble systems used the SVM model combination.
Each value is an averages of the \NumTasks tasks.
$\perModelMetric$ denotes per-model metric values defined as: $\perModelMetricDef$.
\label{appendix:fig:scaling_SST}
}
\end{figure*}

\begin{table*}[h]
    \centering
    \caption{
        \textbf{SST task}.
        Statistics of ensemble systems described in \Cref{sec:ensemble_systems}.
        The rows and columns list the model generation and combination methods of \Cref{tb:ensemble_methods}, respectively.
        Each cell shows a quantity of a specific system $s$.
        Each quantity is the average over the \NumTasks tasks.
        Each system contains $N=15$ models.
        Color shows the rank within \textit{each column} (brighter is better).
        \label{tb:ablation_SST}
    }  
    \begin{subfigure}{\linewidth}
        \centering
        \small
        \tabcolsep 3.0pt
    \subcaption{
        Error rate reductions and lower bound reductions.
        The baseline values used in \Cref{eq:error_reduction,eq:lower_bound_reduction} were the followings.
        ER($s_0$): \SI{15.7}{\percent}.
        LB($s_0$) of $\BoundFuncTight(\StrengthTriplet)$: \SI{2.3}{\percent}.
        LB($s_0$) of $\BoundFuncTight(\StrengthDoublet)$: \SI{2.3}{\percent}.
        LB($s_0$) of $\BoundFuncLoose(\StrengthDoublet)$: \SI{-3.0}{\percent}.
        \label{tb:ablation_errors_SST}
    }

\begin{tabular}{lcccccccccccc}
\toprule
& \multicolumn{4}{c}{Error rate reductions \cref{eq:error_reduction}} & &   \multicolumn{7}{c}{Lower bound reductions \Cref{eq:lower_bound_reduction}} \\
\cmidrule(l{\tabcolsep}r{\tabcolsep}){2-5} \cmidrule(l{\tabcolsep}){7-13}
 & \multirow{2}{*}{Voting} & \multirow{2}{*}{LogR} & \multirow{2}{*}{SVM} & \multirow{2}{*}{RForest} & & \multicolumn{4}{c}{\textbf{\Cref{lemma:ensemble_bound_tight}} $\BoundFuncTight(\StrengthTriplet)$} & & \multirow{2}{*}{$\BoundFuncTight(\StrengthDoublet)$} & \multirow{2}{*}{\begin{tabular}{c}\Cref{lemma:ensemble_bound_loose} \\ $\BoundFuncLoose(\StrengthDoublet)$\end{tabular}} \\
 &  &  &  &  & & \multicolumn{1}{c}{Voting} & \multicolumn{1}{c}{LogR} & \multicolumn{1}{c}{SVM} & \multicolumn{1}{c}{RForest} & & &  \\

\midrule
Random-HyP    &                         \cThird $4.8_{\pm{\mbox{\tiny 7.4}}}$ &  \cSecond $10.5_{\pm{\mbox{\tiny 3.4}}}$ &    \cThird $8.2_{\pm{\mbox{\tiny 2.9}}}$ &   \cThird $4.7_{\pm{\mbox{\tiny 7.6}}}$ &        &                         \cThird $4.7_{\pm{\mbox{\tiny 7.3}}}$ &  \cSecond $10.3_{\pm{\mbox{\tiny 3.5}}}$ &    \cThird $8.0_{\pm{\mbox{\tiny 2.8}}}$ &                         \cThird $4.6_{\pm{\mbox{\tiny 7.5}}}$ &        &   \cThird $166_{\pm{\mbox{\tiny 28}}}$ &   \cThird $58_{\pm{\mbox{\tiny 11}}}$ \\
Bagging       &                        \cFirst $15.1_{\pm{\mbox{\tiny 7.7}}}$ &   \cFirst $12.8_{\pm{\mbox{\tiny 6.1}}}$ &  \cFirst $16.2_{\pm{\mbox{\tiny 10.8}}}$ &  \cFirst $13.8_{\pm{\mbox{\tiny 9.9}}}$ &        &                        \cFirst $15.2_{\pm{\mbox{\tiny 8.1}}}$ &   \cFirst $12.5_{\pm{\mbox{\tiny 6.3}}}$ &  \cFirst $16.6_{\pm{\mbox{\tiny 11.0}}}$ &                       \cFirst $13.7_{\pm{\mbox{\tiny 10.2}}}$ &        &  \cSecond $177_{\pm{\mbox{\tiny 28}}}$ &  \cSecond $62_{\pm{\mbox{\tiny 11}}}$ \\
Random-Seed   &                        \cSecond $9.4_{\pm{\mbox{\tiny 5.8}}}$ &    \cThird $8.1_{\pm{\mbox{\tiny 2.8}}}$ &   \cSecond $9.4_{\pm{\mbox{\tiny 5.8}}}$ &  \cSecond $8.2_{\pm{\mbox{\tiny 2.8}}}$ &        &                        \cSecond $9.5_{\pm{\mbox{\tiny 6.1}}}$ &    \cThird $7.7_{\pm{\mbox{\tiny 2.8}}}$ &   \cSecond $9.8_{\pm{\mbox{\tiny 5.6}}}$ &                        \cSecond $8.2_{\pm{\mbox{\tiny 2.9}}}$ &        &  \cFourth $146_{\pm{\mbox{\tiny 16}}}$ &   \cFourth $51_{\pm{\mbox{\tiny 6}}}$ \\
Hetero-DNNs &  \cFourth $\scalebox{1.5}[1.0]{-}3.5_{\pm{\mbox{\tiny 4.5}}}$ &   \cFourth $7.0_{\pm{\mbox{\tiny 1.8}}}$ &   \cFourth $1.1_{\pm{\mbox{\tiny 5.8}}}$ &                              \cFourth 0 &        &  \cFourth $\scalebox{1.5}[1.0]{-}3.7_{\pm{\mbox{\tiny 4.5}}}$ &   \cFourth $6.6_{\pm{\mbox{\tiny 1.8}}}$ &   \cFourth $1.8_{\pm{\mbox{\tiny 6.1}}}$ &  \cFourth $\scalebox{1.5}[1.0]{-}0.3_{\pm{\mbox{\tiny 1.9}}}$ &        &   \cFirst $186_{\pm{\mbox{\tiny 14}}}$ &    \cFirst $65_{\pm{\mbox{\tiny 7}}}$ \\
\bottomrule

\end{tabular}

    \end{subfigure}
    \vfill
    \begin{subfigure}{\linewidth}
    \centering
    \small
    \tabcolsep 2.0pt
    \subcaption{
    Breakdown of ensemble strength defined in \cref{eq:triplet_decomposition}.
    We show per-model metric values defined as: $\perModelMetricDef$. Thus, $\StrengthTriplet = (\relev - \redun - \combloss) \ \times N $ holds.
    For intuitive understanding, all the values are normalized by the ensemble strength of baseline $\StrengthTriplet_{s_0}$, for example, $\Relev = \RelevHat / \StrengthTriplet_{s_0} \times 100$ where $\RelevHat$ is the raw value.
    \label{tb:ablation_triple_SST}
    }

\begin{tabular}{lccccccccccccc}
\toprule

{} & \multicolumn{4}{c}{\multirow{1}{*}{$\StrengthTripletWithArgs$}} & & \multicolumn{6}{c}{Per-model metric values} & \\
\cmidrule(l{\tabcolsep}r{\tabcolsep}){2-5} \cmidrule(l{\tabcolsep}r{\tabcolsep}){6-12}
{} &  & &  &  & & \multirow{2}{*}{$\perModelMetric_{\normalfont \text{relev}}$} & \multirow{2}{*}{$\perModelMetric_{\normalfont \text{redun}}$} & \multicolumn{4}{c}{ $\perModelMetric_{\normalfont \text{combloss}}$} & & \multirow{2}{*}{$\perModelMetric_{\normalfont \text{relev}} - \perModelMetric_{\normalfont \text{redun}}$} \\
{} & \multicolumn{1}{c}{Voting} & \multicolumn{1}{c}{LogR} & \multicolumn{1}{c}{SVM} & \multicolumn{1}{c}{RForest} & &  {} &  {} &  \multicolumn{1}{c}{Voting} & \multicolumn{1}{c}{LogR} &  \multicolumn{1}{c}{SVM} & \multicolumn{1}{c}{RForest} & &  {} \\

\midrule
Baseline ($s_0$)                    &    \multicolumn{4}{c}{\cBase 100 (the raw value is 0.705)} &        &   \cBase 100 &                                 \cBase 0 &                                \cBase 0 &                                \cBase 0 &                                \cBase 0 &                                 \cBase 0 &        &   \cBase 100 \\

\midrule
Random-HyP                   &   \cThird $101.6_{\pm{\mbox{\tiny 2.5}}}$ &  \cSecond $103.5_{\pm{\mbox{\tiny 1.2}}}$ &   \cThird $102.8_{\pm{\mbox{\tiny 0.9}}}$ &   \cThird $101.6_{\pm{\mbox{\tiny 2.6}}}$ &        &   \cThird $97.7_{\pm{\mbox{\tiny 0.5}}}$ &  \cSecond $87.2_{\pm{\mbox{\tiny 0.2}}}$ &                      \cSecond $3.76_{\pm{\mbox{\tiny 0.72}}}$ &                      \cSecond $3.63_{\pm{\mbox{\tiny 0.63}}}$ &                      \cSecond $3.68_{\pm{\mbox{\tiny 0.67}}}$ &                      \cSecond $3.76_{\pm{\mbox{\tiny 0.60}}}$ &        &   \cThird $10.5_{\pm{\mbox{\tiny 0.6}}}$ \\
Bagging                      &   \cFirst $105.3_{\pm{\mbox{\tiny 2.8}}}$ &   \cFirst $104.3_{\pm{\mbox{\tiny 2.2}}}$ &   \cFirst $105.8_{\pm{\mbox{\tiny 3.9}}}$ &   \cFirst $104.8_{\pm{\mbox{\tiny 3.6}}}$ &        &  \cSecond $98.4_{\pm{\mbox{\tiny 1.2}}}$ &   \cThird $87.6_{\pm{\mbox{\tiny 0.6}}}$ &                       \cThird $3.78_{\pm{\mbox{\tiny 0.53}}}$ &                       \cThird $3.84_{\pm{\mbox{\tiny 0.58}}}$ &                       \cThird $3.74_{\pm{\mbox{\tiny 0.45}}}$ &                       \cThird $3.81_{\pm{\mbox{\tiny 0.50}}}$ &        &  \cSecond $10.8_{\pm{\mbox{\tiny 1.4}}}$ \\
Random-Seed                  &  \cSecond $103.2_{\pm{\mbox{\tiny 2.1}}}$ &   \cThird $102.7_{\pm{\mbox{\tiny 1.0}}}$ &  \cSecond $103.4_{\pm{\mbox{\tiny 1.9}}}$ &  \cSecond $102.8_{\pm{\mbox{\tiny 0.9}}}$ &        &  \cFirst $100.0_{\pm{\mbox{\tiny 0.0}}}$ &  \cFourth $90.0_{\pm{\mbox{\tiny 0.3}}}$ &                       \cFirst $3.16_{\pm{\mbox{\tiny 0.23}}}$ &                       \cFirst $3.19_{\pm{\mbox{\tiny 0.39}}}$ &                       \cFirst $3.15_{\pm{\mbox{\tiny 0.24}}}$ &                       \cFirst $3.19_{\pm{\mbox{\tiny 0.29}}}$ &        &  \cFourth $10.0_{\pm{\mbox{\tiny 0.3}}}$ \\
Hetero-DNNs                &   \cFourth $98.7_{\pm{\mbox{\tiny 1.5}}}$ &  \cFourth $102.3_{\pm{\mbox{\tiny 0.7}}}$ &  \cFourth $100.7_{\pm{\mbox{\tiny 2.1}}}$ &   \cFourth $99.9_{\pm{\mbox{\tiny 0.7}}}$ &        &  \cFourth $87.7_{\pm{\mbox{\tiny 0.6}}}$ &   \cFirst $76.7_{\pm{\mbox{\tiny 0.3}}}$ &                      \cFourth $4.41_{\pm{\mbox{\tiny 0.27}}}$ &                      \cFourth $4.17_{\pm{\mbox{\tiny 0.34}}}$ &                      \cFourth $4.28_{\pm{\mbox{\tiny 0.31}}}$ &                      \cFourth $4.33_{\pm{\mbox{\tiny 0.33}}}$ &        &   \cFirst $11.0_{\pm{\mbox{\tiny 0.6}}}$ \\
\bottomrule

\end{tabular}

\end{subfigure}

\end{table*}

\end{document}